\documentclass[times,twocolumn,final]{elsarticle}

\ifdefined\showrevision

\usepackage{xcolor}
\usepackage[normalem]{ulem}
\usepackage{marginnote}
\usepackage{xspace}

\newcommand{\revref}[2]{%
\marginnote{$R_{#1}C_{#2}$}
}
\newcommand{\secondrevref}[2]{%
\marginnote{$R_{#1}C_{#2}$}[0.3cm]
}
\DeclareRobustCommand{\robustrevref}[2]{%
\revref{#1}{#2}
}

\newcommand{\revmod}[1]{%
{\color{blue}#1}\xspace
}

\newcommand{\revnew}[1]{%
{\color{orange}#1}\xspace
}

\newcommand{\revdel}[1]{%
{\color{darkgray}\sout{#1}}\xspace
}

\else

\usepackage{xspace}

\newcommand{\marginnote}[1]{\ignorespaces}
\newcommand{\revref}[2]{\ignorespaces}
\newcommand{\secondrevref}[2]{\ignorespaces}
\newcommand{\robustrevref}[2]{\ignorespaces}
\newcommand{\revmod}[1]{#1\xspace}
\newcommand{\revnew}[1]{#1\xspace}
\newcommand{\revdel}[1]{\ignorespaces}

\fi

\usepackage{xcolor}

\usepackage{medima}
\usepackage{framed,multirow}
\usepackage{booktabs}  

\usepackage{amssymb}
\usepackage{amsmath}
\usepackage{latexsym}
\usepackage{bm}

\usepackage{url}
\usepackage[export]{adjustbox}
\usepackage{makecell}
\usepackage{placeins}

\usepackage{tikz}
\newcommand{\coloredcircleRGB}[4][1ex]{%
  \tikz[baseline=-0.6ex]{
    \definecolor{tmpcolor}{RGB}{#2,#3,#4}%
    \draw[draw=tmpcolor, fill=tmpcolor, fill opacity=0.3, line width=0.8pt]
      (0,0) circle (#1);
  }%
}

\newcommand*{\nolink}[1]{%
  \begin{NoHyper}#1\end{NoHyper}%
}
\newcommand{\NA}{---}
\newcommand{\figref}[1]{\figurename~\ref{#1}}
\newcommand{\tabref}[1]{\tablename~\ref{#1}}
\newcommand{\secref}[1]{Section~\ref{#1}}

\DeclareMathOperator{\softmax}{softmax}
\DeclareMathOperator{\Conv}{Conv}
\DeclareMathOperator{\BN}{BN}
\DeclareMathOperator{\ID}{ID}
\DeclareMathOperator{\OOD}{OOD}
\DeclareMathOperator{\Metric}{Metric}
\DeclareMathOperator{\CP}{CP}
\DeclareMathOperator{\SP}{SP}
\DeclareMathOperator{\TP}{TP}
\DeclareMathOperator{\FP}{FP}
\DeclareMathOperator{\TN}{TN}
\DeclareMathOperator{\FN}{FN}
\DeclareMathOperator{\TPR}{TPR}
\DeclareMathOperator{\FPR}{FPR}
\DeclareMathOperator{\TNR}{TNR}
\DeclareMathOperator{\Precision}{Precision}
\DeclareMathOperator{\BACC}{BACC}
\DeclareMathOperator{\Fone}{F1}
\DeclareMathOperator{\OVR}{OVR}
\DeclareMathOperator{\AUROC}{AUROC}
\DeclareMathOperator{\AUPR}{AUPR}

\journal{Preprint}

\usepackage{hyperref}
\hypersetup{pdfdisplaydoctitle=true,pdfauthor=Junwen Wang et al.}

\begin{document}

\verso{Junwen Wang \textit{et~al.}}
\begin{frontmatter}

%
%
%

\title{
OOD-SEG: Exploiting out-of-distribution detection techniques for learning image segmentation from sparse multi-class positive-only annotations
}

\author[1]{Junwen \snm{Wang}\corref{cor1}}
\cortext[cor1]{Corresponding author: junwen.wang@kcl.ac.uk}
\author[1]{Zhonghao \snm{Wang}}
\author[1,2]{Oscar \snm{MacCormac}}
\author[1,2]{Jonathan \snm{Shapey}}
\author[1]{Tom \snm{Vercauteren}}

\address[1]{\small\textit{School of Biomedical Engineering \& Imaging Sciences, King’s College London, UK}}
\address[2]{\small\textit{Department of Neurosurgery, King's College Hospital, London,  UK}}


\begin{abstract}
Despite significant advancements, segmentation based on deep neural networks in medical and surgical imaging faces several challenges, two of which we aim to address in this work.
First, acquiring complete pixel-level segmentation labels for medical images is time-consuming and requires domain expertise.
Second, typical segmentation pipelines cannot detect out-of-distribution (OOD) pixels, leaving them prone to spurious outputs during deployment.
In this work, we propose a novel segmentation approach which broadly falls within the positive-unlabelled (PU) learning paradigm and exploits tools from OOD detection techniques. Our framework learns only from sparsely annotated pixels from multiple positive-only classes and does not use any annotation for the background class.
These multi-class positive annotations naturally fall within the in-distribution (ID) set.
Unlabelled pixels may contain positive classes but also negative ones, including what is typically referred to as \emph{background} in standard segmentation formulations.
\revref{1}{2}
\revnew{To the best of our knowledge, this work is the first to formulate multi-class segmentation with sparse positive-only annotations as a pixel-wise PU learning problem and to address it using OOD detection techniques.}
Here, we forgo the need for background annotation and consider these together with any other unseen classes as part of the OOD set.
Our framework can integrate, at a pixel-level, any OOD detection approaches designed for classification tasks.
To address the lack of existing OOD datasets and established evaluation metric for medical image segmentation, we propose a cross-validation strategy that treats held-out labelled classes as OOD.
Extensive experiments on both multi-class hyperspectral and RGB surgical imaging datasets demonstrate the robustness and generalisation capability of our proposed framework.
\end{abstract}


\end{frontmatter}


\section{Introduction}
\label{sec:introduction}

Despite significant progress of deep neural networks based segmentation methods for medical and surgical image analysis in recent years, the efficacy of these methods is highly dependent on the quality and quantity of pixel-level annotations. Specifically, the manual annotation of medical images necessitates professional expertise from experienced domain experts, making the process both costly and time-consuming, thereby leading to a shortage of labelled data in clinical settings. 

To reduce the intensive workload in acquiring pixel-level dense annotation from clinical experts, many efforts have been made to advance Weakly Supervised Learning (WSL)~\citep{tajbakhshEmbracingImperfectDatasets2020,canLearningSegmentMedical2018,xuWeaklySupervisedHistopathology2014} and train learning-based algorithms from coarse-grained annotations instead of precise segmentation masks.
WSL could allow domain experts to annotate regions that they are confident of, leaving intricate details unlabelled.
Such annotations are already starting to be adopted in some open-access datasets.
For example, a recent hyperspectral imaging (HSI) dataset adopted a sparse annotation protocol by annotating representative image regions, omitting marginal areas, superficial blood vessels, adipose tissue and other artefacts~\citep{studier-fischerHeiPorSPECTRALHeidelbergPorcine2023}.
Similarly, the Dresden Surgical Anatomy Dataset (DSAD) offers sparse positive-only annotations for RGB surgical imaging~\citep{carstensDresdenSurgicalAnatomy2023}.
Yet, the proper application of WSL approaches to such cases lacking background class annotations remains an open question.
WSL also preserves less information compared to dense annotations, losing supervisory signal for some object structures.
Such difficulties make the training process from sparse positive-only labels challenging.

Furthermore, to deploy a fully automated system in a safety-critical environment, the system should not only able to produce reliable results in a known context, but should also be able to flag situations in which it may fail~\citep{amodeiConcreteProblemsAI2016,unionArtificialIntelligenceAct2024}.
Conventional segmentation frameworks
follow an assumption that all training data and testing data are drawn from the same distribution and are thus considered in-distribution (ID).
Under this assumption, at inference, the model should only be used in a similar context, which may imply limiting the acquisition hardware and the presence of unexpected classes such as a new model of surgical instrument.
This poses a safety issue when trying to deploy the model for real-world clinical use.
Out-of-distribution (OOD) detection may thus be considered a mandatory feature in many clinical applications.
It is an active research topic in many
classification tasks~\citep{hendrycksBaselineDetectingMisclassified2017,liangEnhancingReliabilityOutofdistribution2018, leeSimpleUnifiedFramework2018,hsuGeneralizedODINDetecting2020}, but has rarely been exploited in medical imaging~\citep{lambertTrustworthyClinicalAI2024}.

We argue that these two challenges outlined above: sparse annotations and the need for OOD detection, share enough similarities to address them under a single methodological approach.
In medical image segmentation with sparse annotations, the absence of an annotation does not necessarily imply that a region is identified as negative.
Two other possibilities could explain why a positive pixel remains unlabelled:
1) it may be deemed ambiguous by the annotator; or 
2) it may simply be skipped due to time-constraints.
The most straightforward, albeit wrong approach to handle unlabelled data would be to assume that all such data belongs to the negative or background class.
In contrast, positive-unlabelled (PU) learning~\citep{bekkerLearningPositiveUnlabeled2020} assumes that an unlabelled example could belong to either the positive or negative class.
Most existing work in PU learning focuses on binary classification problems rather than multi-class ones.
PU learning can be seen as a specific case within the broader domain of OOD detection.
Given the absence of the negative class, traditional PU learning methods are frequently formulated as one-class semi-supervised learning problems~\citep{yangGeneralizedOutDistributionDetection2024}.
However, research on segmentation within the frameworks of both PU learning and OOD detection is limited.
Image segmentation problems often require multi-class learning for which little PU-learning approaches have been proposed.
This scarcity of published work is also partly due to the lack of OOD-based evaluation protocols and publicly available benchmark datasets. One potential solution could be to use a different dataset as OOD data during testing~\citep{karimiImprovingCalibrationOutofDistribution2023}. However, this approach poses significant challenges as annotating multiple medical datasets is labor-intensive and requires domain-specific expertise for each dataset.
In this work, we propose a simple but effective medical image segmentation framework which broadly falls within the PU learning paradigm and exploits tools from OOD detection techniques.
Our framework effectively learns feature representations using sparsely annotated labels, enabling reliable detection of OOD pixels 
with classical OOD approaches~\citep{yangGeneralizedOutDistributionDetection2024} designed for classification purposes.
This allows for state-of-the-art OOD detection performance
without compromising the classification accuracy for ID classes.

To address the challenges in evaluation, we propose a novel dedicated cross-validation strategy that enables systematic assessment of background/OOD performance within a single dataset. By isolating part of the labelled positive classes during training, they do not contribute to updating the model weights during training but are grouped as an additional \emph{background/outlier} class used only for validation purposes.
To effectively evaluate OOD performance for segmentation tasks, we propose using two threshold-independent metrics to measure model performance. Building on these metrics, we further design a threshold selection strategy to visualise OOD segmentation results.

Based on our framework, we compare four different classical OOD detection methods integrated in a common U-Net based backbone segmentation model. Our cross-validation results show that combining a model calibration method with the proposed framework achieves the best overall performance.

Our contributions are mostly along three folds: 
\begin{itemize}
    \item We introduce a novel framework for medical image segmentation learning from sparse positive-only annotations. Our approach is technically straightforward and effectively segments negative/OOD data without compromising performance for multi-class positive/ID data.
    \item To assess model performance in both ID and OOD scenarios, we propose a two-level cross-validation method and metrics for evaluation. The cross-validation is based on both subjects/patients and classes present in the dataset. Our evaluation approach eliminates the need for an additional OOD testing set.
    \item The proposed framework can seamlessly incorporate any given OOD detection method or backbone architecture. In particular, we introduce a novel convolutional adaptation of the GODIN method, extending its applicability to segmentation tasks within our framework.
\end{itemize}

To the best of our knowledge, this represents the first work to address the setting of positive-only learning for multi-class medical image segmentation.
\revref{1}{2}
\revnew{While the proposed solution intentionally builds upon established components, its primary novelty thus lies in this recasting of the sparse multi-class positive-only segmentation problem as a PU learning one and showing that pixel-level OOD detection methods offer a practical set of tools to address it in practice.}

\section{Related works}
\label{sec:related_works}
\subsection{Medical image segmentation with sparse annotation}
Existing WSL methods utilise sparse annotation at different level, including image-level annotation~\citep{kuangWeaklySupervisedLearning2024}, bounding box~\citep{wangBoundingBoxTightness2021,wangInteractiveMedicalImage2018,xuWeaklySupervisedHistopathology2014}, scribbles~\citep{canLearningSegmentMedical2018, wangDeepIGeoSDeepInteractive2019}, points~\citep{glockerVertebraeLocalizationPathological2013,quWeaklySupervisedDeep2019,dorentInterExtremePoints2021} and 2D slices within a 3D structure~\citep{bitarafan3DImageSegmentation2021, cai3DMedicalImage2023}.
These methods use weak labels as supervision signals to train the model and produce full segmentation mask for test image.
Specifically, \citet{glockerVertebraeLocalizationPathological2013} introduced a semi-automatic labeling strategy that transforms sparse
point-wise
annotations into dense probabilistic labels for vertebrae localisation and identification; \citet{xuWeaklySupervisedHistopathology2014} propose to segment both healthy and cancerous tissue from colorectal histopathological biopsies using bounding boxes; and  
\citet{wangInteractiveMedicalImage2018} reported improved CNN performance on sparse annotated input through image-specific fine-tuning; and \citet{wangDeepIGeoSDeepInteractive2019} combined sparsely annotated input with a CNN through geodesic distance transforms, followed by a resolution-preserving network resulting in better dense prediction. However, all of these methods primarily focussed on addressing partial or incomplete annotations, thereby overlooking the context in which no background annotations are present.

Within a related research domain, partial-label learning seeks to train a unified model on multiple partially annotated datasets. This annotation protocol is often adopted in multi-class segmentation tasks, where each dataset contains fully annotated labels for only a subset of classes.
As a way of examples, \citet{liuCOSSTMultiOrganSegmentation2024} proposed a two-stage pipeline that first train on available ground truth label, then self-train with pseudo-labels while discarding outliers in the latent space.
\citet{fidonLabelSetLossFunctions2021} proposed label-set loss functions for training fully segmented images from annotations that may be a a higher-level of granularity than the target task. 
\revdel{\mbox{\citet{shiMarginalLossExclusion2021}} re-engineer the loss to treat hidden classes as marginal probabilities of and forbid overlap between organs.}\revref{2}{2}
\revmod{\citet{shiMarginalLossExclusion2021} re-engineer the loss function to model hidden classes as marginal probabilities while enforcing mutual exclusivity between organ labels.}
\citet{lianLearningMultiorganSegmentation2023} incorporate anatomical organ specific or mutual priors to generate pseudo-labels, then retrain a single unified network on these refined labels. 
\citet{jiangLabeledtounlabeledDistributionAlignment2025} proposed a distribution alignment framework to close the distribution gap between labelled and unlabelled pixels in partially-supervised multi-organ segmentation. 

Compared to partial-label learning, our method addresses a different problem setting where only parts of foreground object regions have annotations. 
In this setting it is erroneous to assume that the classes present in unlabelled pixels are disjoint from those in the annotated regions, or to treat unlabelled regions as background or negative examples.

\subsection{Learning from positive-only data}
\label{sec:learning_from_positive_only_data}
Positive and unlabelled (PU) learning considers a scenario where only a subset of positive data are labelled, while the unlabelled set contains both positive and negative data~\citep{bekkerLearningPositiveUnlabeled2020}. 
It is closely related to semi-supervised learning and positive-only learning.

Positive-only or one-class learning, illustrated in \figref{fig:boundary_illustration}, is a supervised method which involves learning a decision boundary that corresponds to a desired density level of the positive data distribution~\citep{pereraOneClassClassificationSurvey2021}.
Early approaches utilised statistical features to build one-class classifiers. For instance, Principal Component Analysis (PCA)~\citep{bishopPatternRecognitionMachine2006} or Kernel PCA identifies a lower-dimensional subspace that best represents the training data distribution. Leveraging robust feature extraction capabilities, some studies have integrated deep learning models into one-class learning methods.
Another method, Deep Support Vector Data Descriptor (DeepSVDD)~\citep{ruffDeepOneClassClassification2018} learns a representation that encloses embedding of all positively labelled data with the smallest possible hyper-sphere. One-class CNN~\citep{ozaOneClassConvolutionalNeural2019} use a zero-centered Gaussian noise in the latent space as the pseudo-negative class and trains a CNN to learn a decision boundary for the given class. 
In medical imaging domain, Han~\textit{et al.} recast COVID-19 image screening as a positive-unlabelled problem and proposed a variant of PU learning method to robust against overfitting given limited positive data \citep{hanSemiSupervisedScreeningCOVID192021}.

Positive-only learning extends the binary classification in one-class methods by learning decision boundaries for multiple classes of positive labelled data.
However, very few studies have examined the multi-class setup in detail. 
In this work, we frame positive-only learning for image segmentation as a multi-class problem with pixel-level OOD detection.

\subsection{Out-of-distribution detection}
Several studies have explored OOD detection within the context of image classification~\citep{hendrycksBaselineDetectingMisclassified2017,liangEnhancingReliabilityOutofdistribution2018,leeSimpleUnifiedFramework2018, hsuGeneralizedODINDetecting2020}. 
As an early example exploiting deep learning,
\citet{hendrycksBaselineDetectingMisclassified2017} proposed using the maximum softmax score as a baseline for OOD detection based on an observation that correctly classified images tend to have higher softmax probabilities than erroneously classified examples. \citet{liangEnhancingReliabilityOutofdistribution2018} found that applying confidence calibration through temperature scaling~\citep{guoCalibrationModernNeural2017a} effectively separates ID and OOD images. \citet{leeSimpleUnifiedFramework2018} suggested measuring the Mahalanobis distance between test image features and the training distribution from the penultimate convolutional layer of the model. \citet{hsuGeneralizedODINDetecting2020} proposed decomposing the confidence score to learn temperature parameters during training.

Despite methodological advances and positive demonstration for image classification purposes, usage of OOD detection in medical image segmentation is uncommon.
Some studies hypothesize that this may be due to the lack of OOD-based evaluation protocols and the difficulty in gathering relevant data for it~\citep{lambertTrustworthyClinicalAI2024, bulusuAnomalousExampleDetection2020}.
Recent research has attempted to address this issue by using other datasets as OOD examples. \citet{karimiImprovingCalibrationOutofDistribution2023} used two separate datasets: one for training the neural network and evaluating its performance on ID data, and another for testing specifically for OOD detection. \citet{gonzalezDistancebasedDetectionOutofdistribution2022} collected four types of OOD datasets to account for different distribution shifts from ID data for COVID-19 lung lesion segmentation task. However, acquiring an additional dataset that can be considered OOD is a difficult and time-consuming process. Therefore, a more scalable approach would be to establish both training and evaluation within a single dataset.

\subsection{Uncertainty estimation in medical image segmentation}
As illustrated in the previous section, several typical OOD detection approaches rely on estimating the uncertainty of a deep learning prediction~\citep{lambertTrustworthyClinicalAI2024}.
Better uncertainty modelling could thus benefit OOD detection.
Several uncertainty estimation approaches rely on measuring the empirical variance of the network predictions under a set of perturbations.
%
Strategies to generate ensembles of predictions include using several deep learning models with:
differences in model hyperparameters~\citep{wenzelHyperparameterEnsemblesRobustness2020}; random initialization of the network parameters; random shuffling of the data points~\citep{lakshminarayananSimpleScalablePredictive2017}; and applying dropout during test time~\citep{galDropoutBayesianApproximation2016a}. 
In medical image segmentation, uncertainty estimation has mostly been applied with binary classes.
As way of examples, \citet{wangAleatoricUncertaintyEstimation2019} apply test time augmentation to estimate aleatoric uncertainty for fetal brains and brain tumours segmentation from 2D and 3D Magnetic Resonance Images (MRI); \citet{wangAutomaticBrainTumor2019} propose a CNN-based cascaded framework with test-time augmentation for brain tumour segmentation.
Beyond prediction ensembling, recent studies have focused on providing better uncertainty prediction out of the box by calibrating the model uncertainty using dedicated loss functions. In particular, \citet{liangImprovedTrainableCalibration2020} proposed
an auxiliary loss term
based on the difference between accuracy and confidence.
\citet{barfootAverageCalibrationError2024} extend the expected calibration error~\citep{guoCalibrationModernNeural2017a} to 
a differentiable loss function to train a segmentation model.
However, none of the works in the medical imaging field have demonstrated the benefits of improved uncertainty calibration in the context of
unlabelled or OOD data.

\subsection{Segmentation of surgical spectral imaging data}
Having looked at related work in the key methodological areas of interest, we now turn to the related work in the main clinical application of interest in this work, namely hyperspectral imaging for surgical guidance.
Early works on segmentation of surgical HSI
data are based on traditional machine learning techniques \citep{raviManifoldEmbeddingSemantic2017,fabeloSpatiospectralClassificationHyperspectral2018,mocciaUncertaintyAwareOrganClassification2018a}. For example, \citet{raviManifoldEmbeddingSemantic2017} trained a Semantic Texton Forest \citep{shottonSemanticTextonForests2008} on HSI embedding which generated by using an adapted version of \textit{t}-distributed stochastic neighbour approach (\textit{t}-SNE) \citep{maatenVisualizingDataUsing2008}; \citet{fabeloSpatiospectralClassificationHyperspectral2018} proposed a hybrid framework utilising supervised learning and unsupervised learning techniques. The supervised classification map is obtained by using a pixel-wise Support Vector Machine (SVM) classifier that was spatially homogenized through k-nearest neighbours filtering. The authors then combined it with a segmentation map obtained via unsupervised clustering using a hierarchical k-means algorithm. However, the experiment is conducted on $5$ HSI datasets and the separation between training, validation and testing is unclear.

The use of deep learning for biomedical segmentation using spectral imaging data is increasing~\citep{khanTrendsDeepLearning2021}.
Most studies adopt standard U-Net and similar architectures~\citep{ronnebergerUNetConvolutionalNetworks2015, jegouOneHundredLayers2017} and train their model with patch-based or pixel-based input. 
Some works have looked at the impact of training models with different types of input spanning different levels of granularity such as pixel, patches and images~\citep{seidlitzRobustDeepLearningbased2022,garciaperazaherreraHyperspectralImageSegmentation2023}. 
In \citep{seidlitzRobustDeepLearningbased2022}, the authors segmented $20$ types of organs from $506$ HSI hypercubes taken from $20$ pigs. They compared the segmentation performance by training the model with single pixels (no spatial context), patches and full HSI images with the same hyperparameter setup.
They reported that the best performance was achieved with full HSI image input~\citep{seidlitzRobustDeepLearningbased2022}. 
Similarly, \citet{garciaperazaherreraHyperspectralImageSegmentation2023}
used the ODSI-DB dataset~\citep{hyttinenOralDentalSpectral2020}
segmenting $35$ dental tissues from $30$ human subjects after data preprocessing and partitioning training and testing set.
They trained a deep learning model on full HSI and hyperspectral pixels with spatial context removed, reporting a baseline segmentation result.
Recently, work by \citet{martin-perezMachineLearningPerformance2024} compared various pixel-level classification algorithms for brain tissue differentiation. The study evaluated conventional algorithms, deep learning methods, and advanced classification models. Their findings highlighted that reducing the number of training pixels could improve performance, regardless of the dataset and classifiers.

Overall, available surgical HSI data remains limited in size, and the inherent complexity and variability of the surgical environment further complicate its analysis. Furthermore, the available annotations are sparse, as the data often consists of annotations on isolated pixels or small regions rather than comprehensive labelling of entire images \citep{zhuSpectralSpatialDependentGlobalLearning2022}. 
While relevant, none of the previous works have demonstrated effective methods for leveraging sparse, positive-only annotations.

\section{Material and methods}
\label{sec:material_and_methods}
The section starts by describing the HSI and RGB imaging datasets and associated annotations that serve as a foundation and motivation for this work (\secref{sec:dataset}), followed by the problem formulation, where we clearly distinguish between positive-only learning, PU learning, and OOD detection (\secref{sec:relation_OOD_PU}).
We then describe our proposed learning framework for sparse multi-class positive-only medical image segmentation (\secref{sec:positive_only_learning_for_seg}).
Lastly, we introduce our proposed OOD-focused evaluation framework (\secref{sec:evaluation_framework}), evaluation metrics (\secref{sec:evaluation_metrics}), and threshold selection method for negative / OOD detection (\secref{sec:threshold_selection}).

\subsection{Datasets}
\label{sec:dataset}
Hyperspectral imaging (HSI) and multispectral imaging are emerging optical imaging techniques that collect and process spectral data distributed across number of wavelengths~\citep{shapeyIntraoperativeMultispectralHyperspectral2019}. 
HSI splits light into multiple narrow bands beyond the standard red, green, and blue spectral bands, allowing examination of image details that are invisible to human sight.
This technique gathers diagnostic data about tissue properties, allowing for objective characterization of tissues without the use of any external contrast agents.
Recent studies~\cite{liSpatialGradientConsistency2023, liDeepLearningApproach2022, ayalaSpectralImagingEnables2023} have demonstrated the feasibility and clinical potential of real-time HSI in surgery using snapshot mosaic sensors, typically configured with 16 (4×4) or 25 (5×5) spectral bands. 
Real-time performance is key to accelerate clinical adoption of HSI. 
However, only a limited number of annotated surgical datasets for snapshot mosaic HSI are currently available. In this work, we leverage publicly accessible annotated HSI datasets~\citep{studier-fischerHeiPorSPECTRALHeidelbergPorcine2023, hyttinenOralDentalSpectral2020} acquired with scanning-based systems and simulate consistent 16-band snapshot mosaic data from them, thus essentially providing realistic annotated mosaic snapshot HSI data for our study.

The \emph{Heidelberg Porcine HyperSPECTRAL Imaging} (\textbf{Heiporspectral}) dataset~\citep{studier-fischerHeiPorSPECTRALHeidelbergPorcine2023} comprise $5758$ hyperspectral images with resolution of $480 \times 640$ acquired over the $500$-$1000$nm wavelength range.
Hyperspectral images were captured using the TIVITA tissue hyperspectral camera system, which provides 100 spectral bands for each image.
\revdel{For consistency with all hyperspectral datasets used in this study, for each dataset we sample 16 bands in the available wavelength range at equal intervals.}
The background-free, sparse annotations include $20$ physiological porcine organs, which are obtained from a total $11$ pigs. For each organ, annotations are distributed across $8$ pigs. In each acquired organ image series, representative image regions of the $20$ structures depending on the respective organ image series were annotated.

The \emph{Oral and Dental Spectral Image Database} (\textbf{ODSI-DB})~\citep{hyttinenOralDentalSpectral2020} contains $316$ hyperspectral images of $30$ human subjects of which $215$ have annotations. Images have a varied resolution and wavelength range due to two different cameras being used in the study. 
$59$ annotated images were taken with a Nuance EX (CRI, PerkinElmer, Inc., Waltham, MA, USA) and $156$ were obtained with a Specim IQ (Specim, Spectral Imaging Ltd., Oulu, Finland). The pictures taken by the Nuance EX contain $51$ spectral bands ($450$–$950$ nm with $10$ nm bands) and special resolution $1392 \times 1040$;
Those captured by the Specim IQ have $204$ bands ($400$–$1000$ nm with approximately $3$ nm steps) and spatial resolution $512 \times 512$. Some images are further cropped to ensure the anonymity of the testing subject. 
\revdel{To alleviate the discrepancy from the camera setup, we sample $16$ bands at equal intervals in the available range.} 
We resize all images to a spatial size $512\times512$ by either centrally cropping or padding the image.
Annotations from these $215$ images are sparse and background-free. The number of annotated pixels per image varies from image to image. The annotated pixels can belong to $35$ possible dental tissues, which do not contain the background class.
Inspection of this dataset shows that the majority of classes are underrepresented.
We select classes with at least 1 million pixel samples and discard the rest of the classes for further analysis as done in previous work by~\citet{garciaperazaherreraHyperspectralImageSegmentation2023}, resulting total $9$ classes selected in this study.
While the application of our proposed  methodological approach to spectral imaging data represents our main focus,
we also demonstrate the capability of our method in a non-spectral imaging dataset.
More specifically, we make use of an RGB laparoscopic dataset that shares many similarities in terms of anatomical content and annotation style.

The \emph{Dresden Surgical Anatomy Dataset}
(\textbf{DSAD})~\citep{carstensDresdenSurgicalAnatomy2023} comprises $13195$ laparoscopic images from $32$ patients of robot-assisted anterior rectal resections or rectal extirpation surgeries. Images provided in the dataset were extracted from the video and were stored in PNG format at a resolution of $1920 \times 1080$.
The annotation of $11$ abdominal organs provided pixel-wise segmentation with multiple inclusion criteria for anatomical structures, resulting in sparsely annotated images across the dataset. The majority of annotations in this dataset only account for a single organ per image. 
However, a subset of the data is associated with multi-organ segmentation for all $11$ anatomical structures. This includes a total of $1430$ images in $32$ patients. 

\begin{figure}[t]
    \centering\footnotesize
    \setlength\tabcolsep{1.5pt} 
    \begin{tabular}{ccc}
        \includegraphics[width=0.32\linewidth, height=0.32\linewidth]{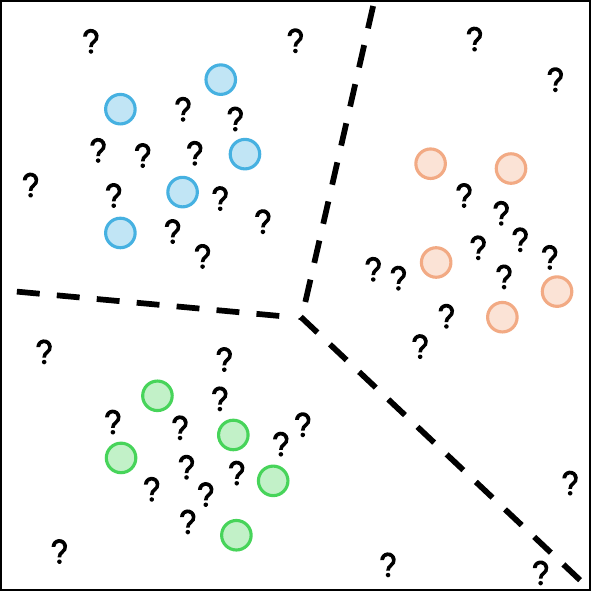} &  
        \includegraphics[width=0.32\linewidth, height=0.32\linewidth]{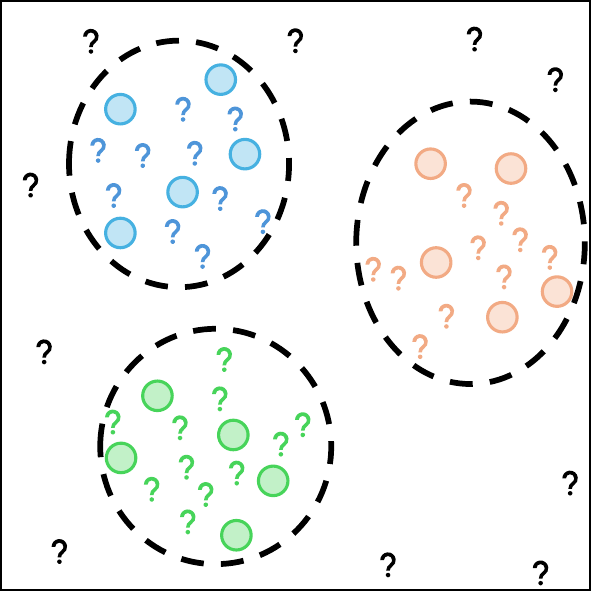} &
        \includegraphics[width=0.32\linewidth, height=0.32\linewidth]{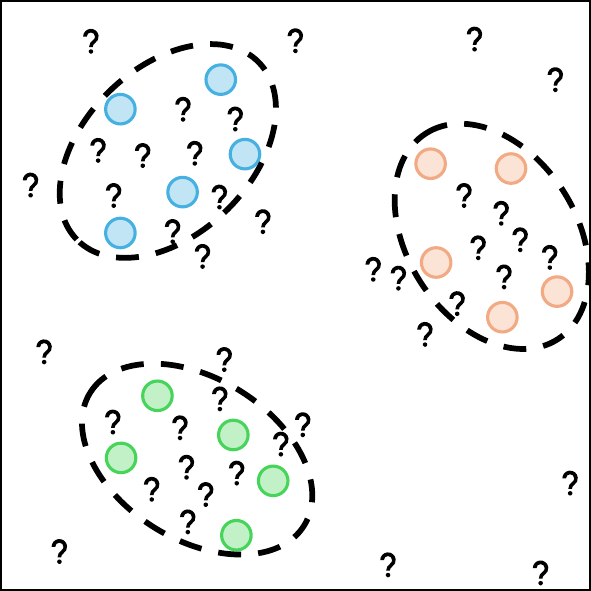} \\
        \makecell[c] {\textbf{Multi-class} \textbf{learning}} &
        \makecell[c] {\textbf{PU learning}} &
        \makecell[c] {\textbf{Positive-only learning}} \\
        
    \end{tabular}
    \caption{Decision boundaries under different learning settings for a multi-class problem with annotations from only positive classes. 
    Coloured circle markers (\coloredcircleRGB[0.5ex]{93}{187}{228}/\coloredcircleRGB[0.5ex]{244}{180}{146}/\coloredcircleRGB[0.5ex]{105}{220}{120}) denote positively labelled samples for each class. Black question marks (\textcolor{black}{\textsf{?}}) indicate unused unlabelled training data, while coloured question marks (\textcolor[RGB]{93,187,228}{\textsf{?}}/\textcolor[RGB]{244,180,146}{\textsf{?}}/\textcolor[RGB]{105,220,120}{\textsf{?}}) represent used unlabelled training data. 
    Unlabelled data could be stemming from a positive class or from the background class.
    Our proposed framework adapts a multi-class positive-only learning setting (as shown on the right hand side), forming distinct decision boundaries that enclose positively labelled data for each class. 
    Unlabelled data points outside these boundaries are detected by OOD techniques and flagged as background.}
    \label{fig:boundary_illustration}
\end{figure}

\subsection{\revdel{Problem statement and relationship to positive-only learning, PU learning, and OOD detection}\revmod{Problem formulation: Positive-only learning and its relation to PU learning and OOD detection}}\revref{1}{1}\label{sec:relation_OOD_PU}

\revdel{Our method aims to learn image segmentation from sparse multi-class positive-only annotations.
Our training datasets are composed of images where only a subset of the pixels are annotated for positive (foreground) classes (e.g. organs), while the remaining pixels are unlabelled.
We do not have any negative annotations that would correspond to the background class. 
Unlabelled pixels in the training set can be either from a positive class or be negative (background).}

\revdel{Given this problem formulation, PU learning approaches would appear as a good fit.
Traditional PU learning exploits unlabelled samples in a semi-supervised manner to tease out positive and negative class distributions. 
However, it has been studied primarily for binary classification and there exist no trivial extension to multi-class PU learning \mbox{\cite{bekkerLearningPositiveUnlabeled2020}}.}

\revref{1}{1}
\revmod{We consider the problem of image segmentation under sparse, multi-class positive-only annotation, motivated by practical annotation constraints in surgical imaging where dense pixel-level labelling is infeasible and expert annotations are intentionally sparse. In this setting, training images contain pixel-level annotations only for a subset of foreground classes (e.g.\ organs or anatomical structures), while the remaining pixels are deliberately left unlabelled. Such unlabelled pixels should not be interpreted as confidently belonging to a background class: due to ambiguity, uncertainty, or overlapping anatomical structures, they may correspond either to foreground classes or to true background. Consequently, no reliable negative (background) annotations are available during training.}

\revmod{PU learning paradigm \cite{bekkerLearningPositiveUnlabeled2020} explicitly captures this ambiguity by acknowledging that unlabelled data may contain both positive and negative samples. In practice, however, existing PU learning methods are largely designed for binary classification and do not extend straightforwardly to multi-class, pixel-wise segmentation. For this reason, our method adopts a positive-only learning formulation, in which only positively annotated pixels are used during training, and no assumptions are made about the class membership of unannotated pixels. This contrasts with conventional fully supervised segmentation, where unannotated pixels are implicitly treated as negative or background, introducing systematic label noise when these regions contain target or clinically relevant tissues. PU learning thus provides the conceptual justification for our setting, while positive-only learning provides practical training strategy that avoids making assumptions about unannotated pixels.}

\revdel{For simplicity, we propose to discard the unlabelled data for training purposes.
We thus revert to a multi-class positive-only learning problem (rather than a multi-class PU one).
Learning from only the positively annotated pixels allows us to straightforwardly learn the distribution of features within the positive classes.
However, this does not directly tackle the problem of detecting and segmenting background pixels.}

\revdel{To address the background class, our method takes advantage, at test time, of OOD detection approaches.
Having learned the distribution of positive classes, background pixels should indeed appear as outliers falling outside of the positive class distributions.
At inference time, a model trained from positive-only data suggests a positive class assignment for each pixel and feeds it to an OOD detection module which either accepts the suggestion or flags the pixel as a background one.}

\revref{1}{1}
\revmod{Our framework repurposes OOD detection as a principled mechanism for addressing medical image segmentation problem with positive-only annotations. Conceptually, all PU learning, positive-only learning and OOD detection rely on modelling the positive data distribution and identifying samples that deviate from it. In PU learning, such deviations are interpreted as reliable negatives; in OOD detection, they correspond to out-of-distribution samples. Our method builds on this shared conceptual structure by repurposing OOD detection techniques as a principled solution for positive-only segmentation.}

\figref{fig:boundary_illustration} illustrates different decision boundaries in the presence of unlabelled data. 
Conventional multi-class classifiers trained on positive-only annotations cannot detect background points within the unlabelled set and instead assign them to one of the known positive classes. 
In contrast, PU learning defines distinct decision boundaries by enclosing both the labelled positive data and part of the unlabelled data. 
Our proposed framework offers a simple positive-only learning approach by combining standard multi-class learning with tools from OOD detection. 
It effectively identifies background data points as outliers with respect to the distributions of the positive classes by establishing clear decision boundaries around the labelled positive data for each class.

\subsection{Positive-only learning for multi-class segmentation}
\label{sec:positive_only_learning_for_seg}
\begin{figure*}[t]
    \centering
    \includegraphics[width=\textwidth]{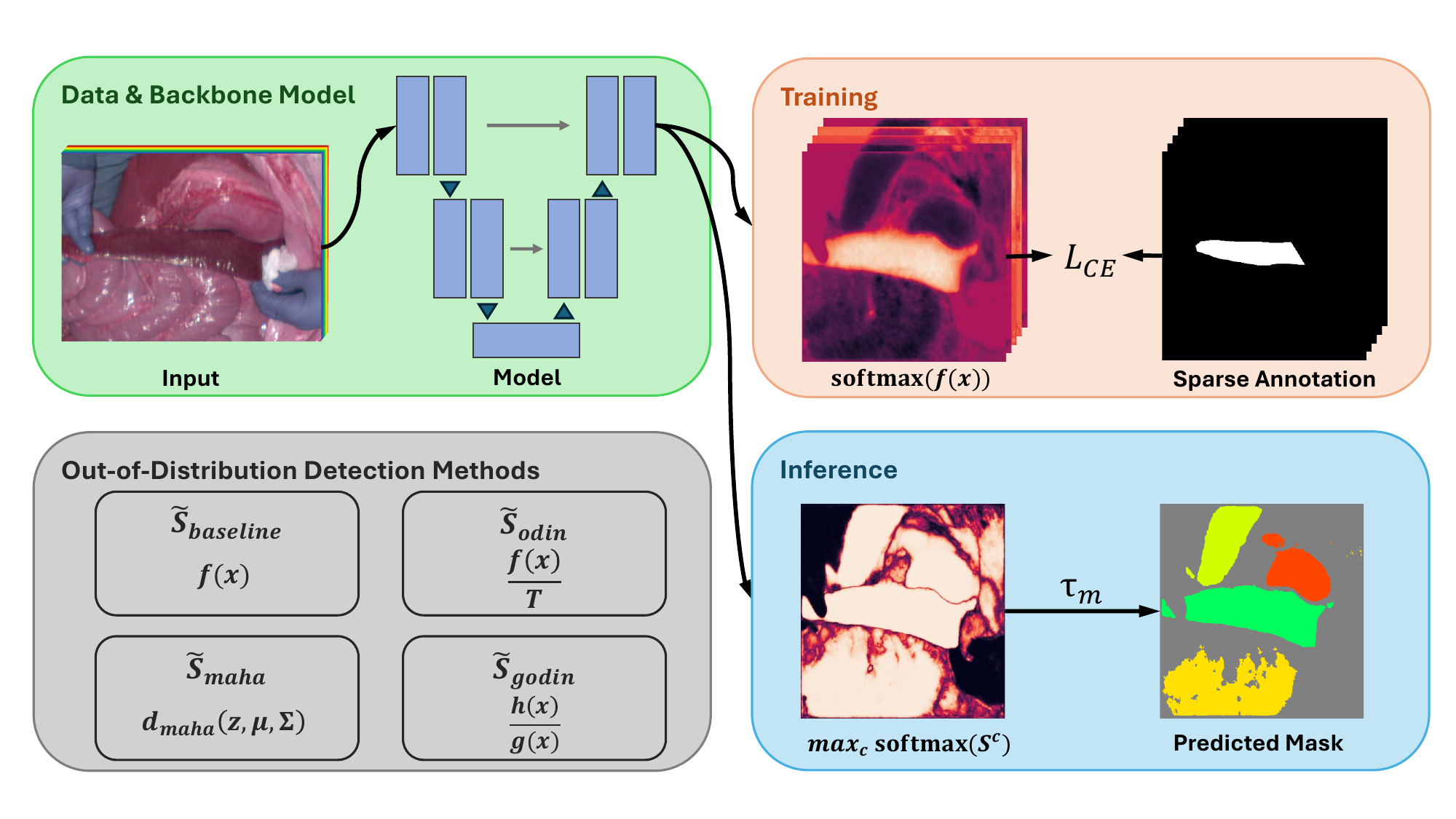}
    \caption{Overview of the proposed OOD-SEG framework. During the training stage, only annotated pixels for the multiple positive classes are used to update the model weights. We define a confidence score $\boldsymbol{S}$ to correlate probability distribution for ID classes. $\boldsymbol{S}$ can be replaced by multiple OOD detection methods (See bottom left block). At the inference stage, we compute the maximum probability of $\boldsymbol{S}$ from $c$ classes followed by thresholding from a pre-selected threshold $\tau_m$ to obtain the predicted mask.}
    \label{fig:overview}
\end{figure*}
In this study, we propose addressing
the positive-only learning scenario
by leveraging concepts from OOD detection. 
\figref{fig:overview} shows an overview of our proposed framework for image segmentation. Given a 2D image $x$, each annotated spatial location $(i,j)$ from $x$ has a corresponding annotation $y_{ij}$, where $y_{ij} \in\{c\}=\{1,2,\dots,C\}$ and $C$ is the number of class that marked as in-distribution.
We purposely refrain from starting numbering positive classes from 0 to retain that 0 index for negative data such as background and OOD samples. 

Our approach entails training a standard multi-class
semantic segmentation network using a loss computed from the sparse positive-only annotations as seen in \figref{fig:overview} (Training).
In practice, we restrict the output of the network to a $C$-dimensional output per pixel.
That is, if no further post-processing were to be applied, the network would not predict any background class.

To incorporate background predictions, as illustrated in \figref{fig:overview} (Inference), during inference, we introduce a pixel-wise scoring function $\boldsymbol{S}_{ij}^c$ which 
aims to capture the probability of that pixel $(i,j)$ belonging to the ID class $c$ while acknowledging the possibility of it being OOD.
If a pixel-wise score is high, we maintain the assignment of that pixel to the best ID class.
In contrast, if the ID class scores are all low, the pixel is considered as OOD.
Our proposed framework utilises a confidence threshold $\tau$ to detect OOD samples at a pixel-level:
\begin{equation}
    \hat{y}_{ij}= 
    \begin{cases}
        \operatorname*{arg\,max}_c \boldsymbol{S}_{ij}^c,& \text{if } \max_c \boldsymbol{S}_{ij}^c > \tau \\
        0,     & \text{otherwise}
    \end{cases}
    \label{eqa:ood_decision}
\end{equation}

\paragraph{Baseline OOD scoring}
Let $f(x)$ denote the logit output of the segmentation network trained using the positive-only ID sparse annotation.
In its simplest implementation, the score function can be the
softmax output from the network
as shown in Equation~\eqref{eqa:baseline} below where dependence on pixel location is omitted for brevity:
\begin{equation}
    \boldsymbol{S}_{baseline}
    = [\boldsymbol{S}_{baseline}^1, \ldots, \boldsymbol{S}_{baseline}^C]
    =\softmax\big(f(x)\big)
    \label{eqa:baseline}
\end{equation}
The resulting OOD approach is a commonly used baseline method for OOD detection in classification tasks~\citep{hendrycksBaselineDetectingMisclassified2017}.

Beyond this baseline, our framework allows integrating state-of-the-art OOD detection methods by changing the predefined pixel-level scoring function $\boldsymbol{S}$.
In this work, we investigate methods related to confidence calibration and Mahalanobis distance as they have demonstrated effectiveness in many OOD detection for classification tasks \citep{liangEnhancingReliabilityOutofdistribution2018,leeSimpleUnifiedFramework2018,hsuGeneralizedODINDetecting2020}.

\paragraph{ODIN}
\citet{liangEnhancingReliabilityOutofdistribution2018} adds temperature scaling and adversarial perturbation to the logit output from the pretrained network to improve the OOD performance: 
\begin{equation}
    \boldsymbol{S}_{odin} = \softmax\left(\frac{f(x)}{T}\right)
\end{equation}
It has been shown that using a large temperature $T$ is generally preferred for OOD classification tasks~\citep{liangEnhancingReliabilityOutofdistribution2018}.
However, from our experiments, we find that using a relatively small $T$ (albeit still much larger than 1) is beneficial for segmentation. We chose a fixed value of $T=10$ across our experiments. 
Furthermore, \citet{liangEnhancingReliabilityOutofdistribution2018} employed adversarial perturbation to further enhance OOD performance by optimising the value of $\sigma$ using a validation set composed equally of ID and OOD data.
In our study, we did not incorporate adversarial perturbation for two primary reasons.
First, we aimed to simplify the training to allow fairer and more reliable comparisons across OOD approaches.
Second, the original paper reported only minor improvements from applying adversarial perturbations, and these came at a significant computational cost \citep{liangEnhancingReliabilityOutofdistribution2018}.

\paragraph{Mahalanobis}
\citet{leeSimpleUnifiedFramework2018} propose an OOD mechanism 
based on a statistical analysis of features observed in each ID class.
Let $\varphi(x)$ be some pixel-level features obtained from intermediate layers of the network where, as before, the dependence on pixel location is dropped for brevity. 
We chose the features before segmentation head as our intermediate feature in the study. 
The class-conditioned distributions of the features are modelled as Gaussians with a class-specific mean $\mu_c$ and a tied, i.e. class independent, covariance matrix $\Sigma$.
A first
scoring $\tilde{\boldsymbol{S}}_{maha}$ is obtained by computing the negative Mahalanobis distance between a prediction feature and each class Gaussian:
\begin{equation}\label{eq:rawmaha}
    \tilde{\boldsymbol{S}}_{maha}^c 
    = -(\varphi(x)-\mu_c)^T \Sigma^{-1} (\varphi(x)-\mu_c)
\end{equation}
To make a head-to-head comparison fairer and easier across OOD methods,
we apply a softmax operator to the
$\tilde{\boldsymbol{S}}_{maha}$ Mahalanobis scores and obtain normalised final scores:
\begin{equation}
    \boldsymbol{S}_{maha} =
    \softmax{ \big( [
    \tilde{\boldsymbol{S}}_{maha}^1,
    \ldots,
    \tilde{\boldsymbol{S}}_{maha}^C
    ] \big)}
\end{equation}
We note that this use of the softmax is not advocated by \citet{leeSimpleUnifiedFramework2018} nor is it strictly necessary.  
We however found it to have no measurable impact on the performance while it helped
provide more consistency in evaluation and mask visualisation.
We thus use it in our subsequent experiments.
Furthermore, as with our use of ODIN, to ensure a fair comparison and to reduce computational burden, we did not incorporate the adversarial perturbation and feature ensembling calibration techniques initially proposed in \citep{leeSimpleUnifiedFramework2018}.



The mean vectors and covariance matrix in Equation~\eqref{eq:rawmaha} are dataset-wide parameters.
To alleviate the computational burden associated with estimating $\mu_c$ and $\Sigma$ at once from all pixel-level features extracted across the entire training dataset, we first compute the per-class mean and a shared covariance for each image in the training set through a spatial averaging procedure.
These image-level estimates are then aggregated using standard reduction to produce the dataset-level estimates of $\mu_c$ and $\Sigma$.

%


\paragraph{Generalised ODIN (GODIN)}
\citet{hsuGeneralizedODINDetecting2020} proposed a dividend and divisor structure for OOD detection that learns a temperature scaling function $g(x)$ during training. 
Assuming a trivial extension for pixel-wise operation and dropping the dependence on pixel location from the equation for brevity,
the un-normalised scoring is expressed per class as:
\begin{equation}
    \tilde{\boldsymbol{S}}^c_{godin} = \frac{h^c(x)}{g(x)}
\end{equation}
A softmax operator is then applied to get the final score:
\begin{equation}
    \boldsymbol{S}_{godin} =
    \softmax{ \big( [
    \tilde{\boldsymbol{S}}_{godin}^1,
    \ldots,
    \tilde{\boldsymbol{S}}_{godin}^C
    ] \big)}
\end{equation}
Both $h^c(x)$ and $g(x)$ are chosen to take features from the penultimate layer $\varphi(x)$ of the backbone model $f(x)$.
For the temperature $g(x)$, these features are fed through an extra pixel-wise linear layer with trainable weights $\bm{w}_{g}$ and bias $b_{g}$, the batch norm (BN), and the sigmoid~($\sigma$) function:
\begin{equation}\label{eq:godin_g}
    g(x) = \sigma(\BN(\bm{w}_{g}\varphi(x)+b_{g}))
\end{equation}
For $h^c(x)$, an extra layer with trainable per-class weights $\bm{w}_{c}$ and bias $b_{c}$ is used to extract a class similarity.
In the context of classification,
\citet{hsuGeneralizedODINDetecting2020} investigated three similarity measures.
The default one in the original work is the inner product between the penultimate features and the per-class parameters:
\begin{equation}\label{eq:godin_hip}
    h^c(x)=\bm{w}_{c}^{T}\varphi(x)+b_{c}
\end{equation}
The other proposed options consisted in the Euclidean distance ($||\varphi(x)-\bm{w}_{c}||^2$) and the cosine similarity ($\frac{\bm{w}_{c}^{T}\varphi(x)}{||\bm{w}_{c}|| \, ||\varphi(x)||}$) between the penultimate features and the per-class parameters. 
A trivial pixel-wise extension of these temperature scaling and three similarity measures for segmentation purposes is achieved by training spatially-invariant weights and bias terms.
Of particular interest is the fact that the pixel-wise linear layer in Equation~\eqref{eq:godin_g} and the inner product operation in Equation~\eqref{eq:godin_hip} can efficiently be implemented with a $1\times1$ convolution layer.

In this work, to capture some additional spatial context for both $g$ and $h^c$, we extend the pixel-wise operations in Equation~\eqref{eq:godin_g} and Equation~\eqref{eq:godin_hip} by introducing convolutional layers with $3\times3$ kernels:
\begin{equation}
\begin{aligned}
    g(x) &= \sigma(\BN(\Conv_{g}(\varphi(x)))) \\
    h^c(x) &= \Conv_{h}(\varphi(x)).
\end{aligned}
\end{equation}
where it should be understood that our notation makes dependence on spatial location implicit.

\begin{figure}[t]
    \includegraphics[width=\linewidth]{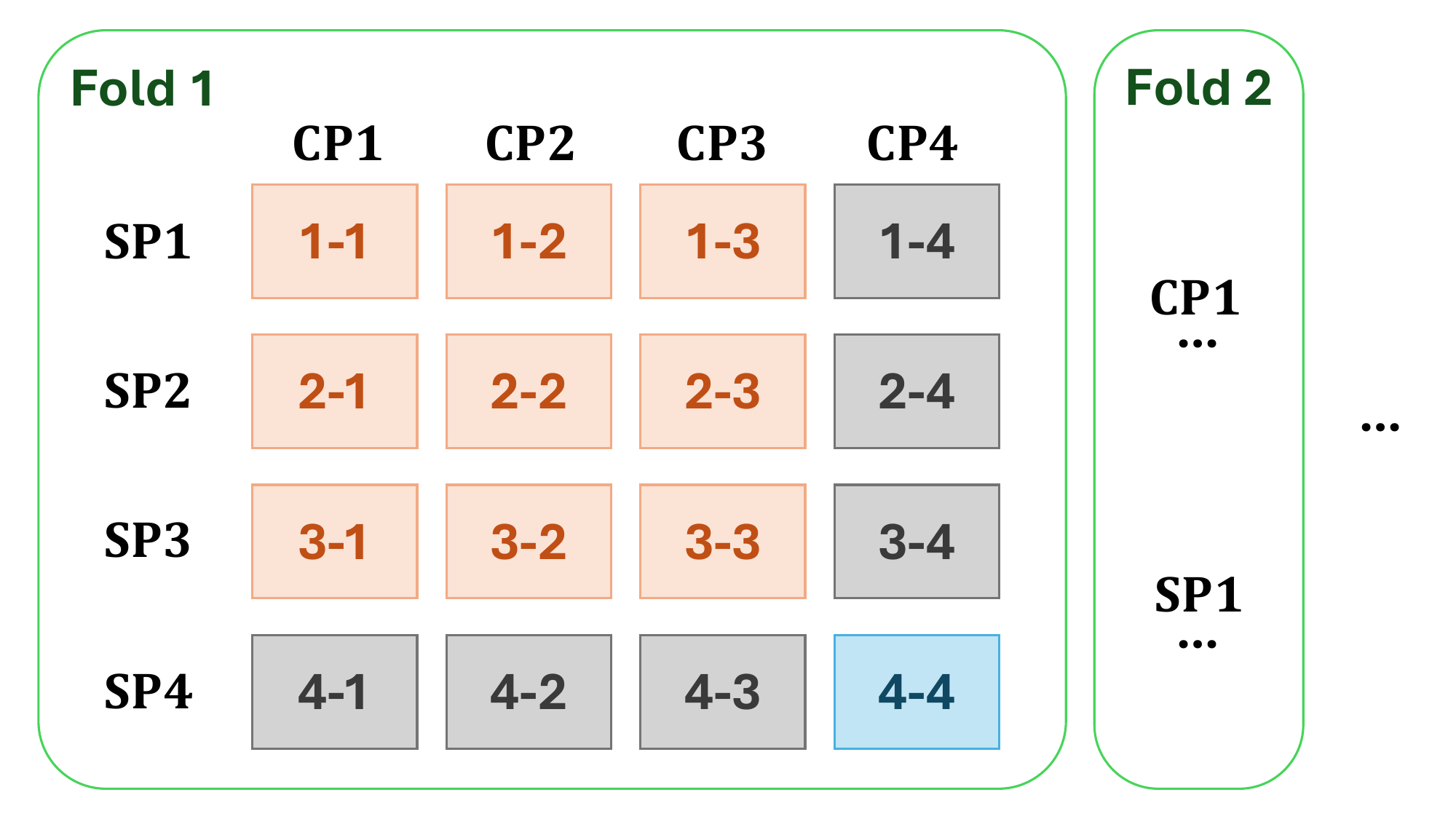}
    \caption{Graphical representation of the proposed OOD-focused two-level cross-validation strategy. For simplicity, only the first fold is shown in detail.
    In this example, the number of subject partitions (SP) and class partitions (CP) are set to $4$, resulting in a total of $16$ partitions.
    Subject-Class Partitions (SCP) marked in red, blue and grey respectively highlight training, testing or untouched data for a particular cross-validation fold.}
    \label{fig:nested_cv}
\end{figure}

\subsection{Two-level OOD-focused Cross-validation Evaluation}
\label{sec:evaluation_framework}
To evaluate the model performance in detecting OOD samples, existing methods utilise other datasets as OOD test set~\citep{liangEnhancingReliabilityOutofdistribution2018,hsuGeneralizedODINDetecting2020}.
A distinctive aspect of these OOD datasets
is the presence of class categories that were not encountered during training.
Such discrepancy is named semantic shift in the original OOD research, which remains an active research topic~\citep{yangGeneralizedOutDistributionDetection2024}.
This approach requires additional annotations beyond the target use case and thus pose an additional burden on the clinical experts.

In our context of sparse multi-class positive-only image segmentation, a semantic shift can already occur at the pixel-level.
This allows us to propose an evaluation framework established without using extra annotated medical image datasets. 
Since there is limited research on benchmarking in positive-only learning, it is interesting to consider common approaches in benchmarking PU learning. 
Benchmarking in PU learning typically assumes a test set with exhaustive class annotations. 
However, meeting this requirement would entail dense pixel-level labelling in a segmentation context, which is impractical for the dataset used in this study.
\figref{fig:nested_cv} shows a simplified view of our proposed two-level cross-validation pipeline for pixel-level OOD detection.
Our two-level cross-validation is built as a combination of two types of data partitions based on subjects and classes: \emph{Subject Partitions} (SP) and \emph{Class Partitions}~(CP).
The subject-level grouping in the SPs ensures that there is no patient overlap bias within our cross-validation experiments.
The class-level groupings CPs allow us to hold out some annotated classes from the training in a specific fold. These classes can thus be considered as OOD for this fold.
For clarity, we note that the number $\text{N}_{\SP}$ of subject partitions (respectively $\text{N}_{\CP}$ the number of class partitions) is upper bounded by the number of subjects (respectively positive classes) in the training data.
By combining these partitions, we obtain $\text{N}_{\SP} \times \text{N}_{\CP}$ two-level folds for cross-validation purposes.

While this approach is effective is establishing an OOD-focused evaluation with no need for OOD-specific annotations, it should be clear that none of the models trained for a particular two-level fold would be trained to recognise all ID classes in the training set.
As such, a complete model for inference purposes should still be trained with all ID classes.

In this work, we chose $\text{N}_{\SP}=4$ and $\text{N}_{\CP}=4$ by default for our OOD-focused evaluations.
To provide some insight on the performance of models trained with all ID classes, we also used a more standard subject-level only cross-validation strategy by setting $\text{N}_{\SP}=4$ and $\text{N}_{\CP}=1$.
This scenario only allowed us to evaluate the capability to recognise ID classes but could not be used for OOD evaluation.

\subsection{Evaluation metric}
\label{sec:evaluation_metrics}

\begin{figure}[t]
    \centering\setlength\tabcolsep{1.5pt}
    \begin{tabular}{cc}
      \includegraphics[width=0.49\linewidth]{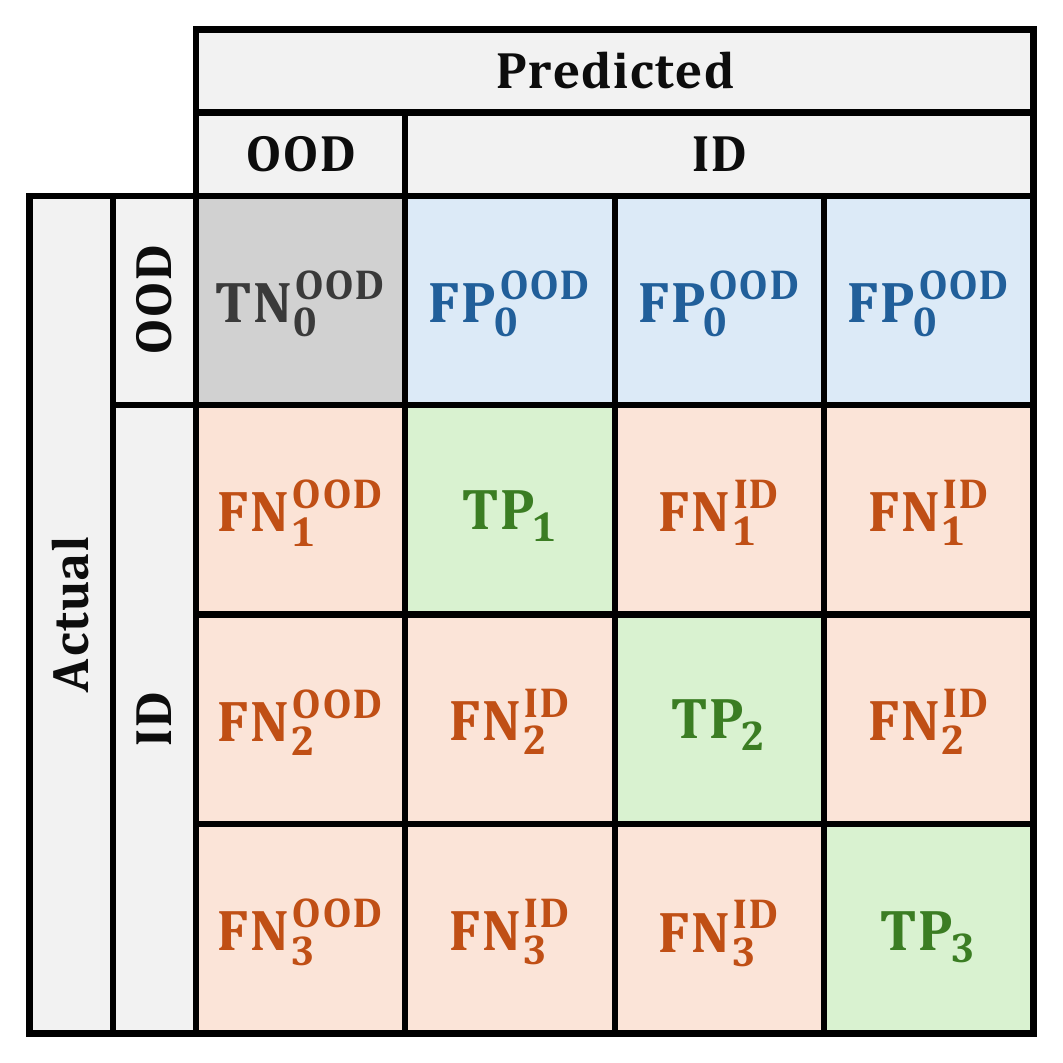} & 
      \includegraphics[width=0.49\linewidth]{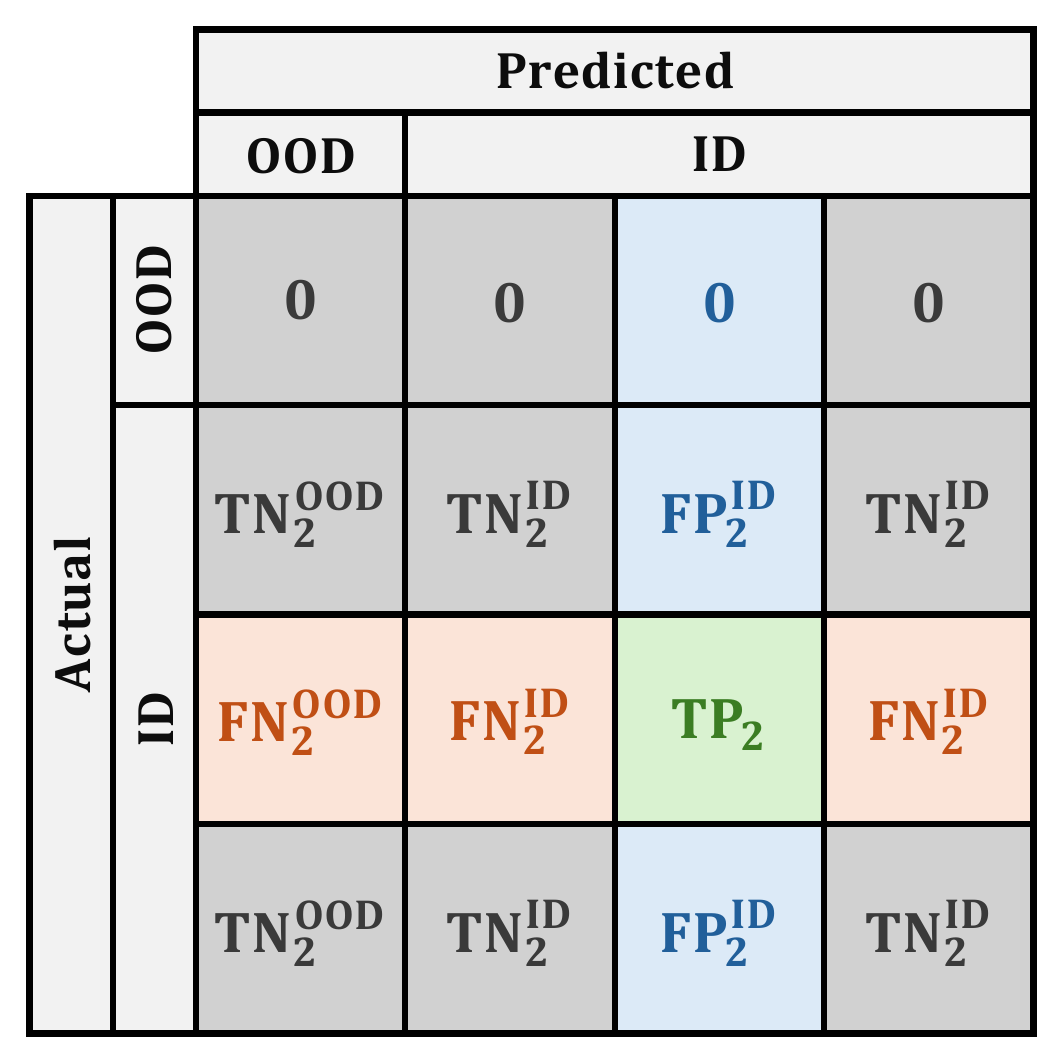}
    \end{tabular}
    \caption{Graphical illustration of confusion matrix incorporating multi-class ID and OOD data. Left: with actual OOD data as negative class.
    Right: In the case without actual OOD data and with class $2$ considered as positive while others are negative in a one-vs-rest approach.}
    \label{fig:ood_cmf}
\end{figure}

\paragraph{OOD-focused metrics}
Within our two-level cross-validation approach,
assuming $\CP_k$ is the current held-out class partition and noting
$\overline{\CP}_k$ its class complement,
we start by building a multi-class confusion matrix that includes all ID classes in $\overline{\CP}_k$ and uses a single outlier class for classes in $\CP_k$.
This is illustrated in \figref{fig:ood_cmf}-left.
Specifically, the outlier class is obtained by aggregating classes in the $\CP_k$ class partition.
Every annotated class is excluded $\text{N}_{\SP}$ times
in our cross-validation approach, as it becomes part of the aggregated outlier class. 
We categorised pixels belonging to the outlier class as negative OOD examples and all other classes as positive ID examples for this particular two-level fold.

Subsequently, we define the true positive rate ($\TPR_{\ID}$) from multi-class positive ID examples and the false negative rate ($\FPR_{\OOD}$) from negative OOD examples as follows:
\begin{equation}
    \TPR_{\ID} = \frac{\sum_{c=1}^{C} \TP_c}{\sum_{c=1}^{C} \big( \TP_c + \FN_c \big)},
    \quad
    \FPR_{\OOD} = \frac{\FP_0^{\OOD}}{\TN_0^{\OOD} + \FP_0^{\OOD}}
    \label{eqa:tpr_and_fpr}
\end{equation}
where $\FN_c = \FN_c^{\OOD} + \FN_c^{\ID}$. It should be clear that since our annotations are sparse, unlabelled data is omitted from these statistics.

By computing $\TPR_{\ID}$ and $\FPR_{\OOD}$ under multiple threshold $\tau$, we obtain a Receiver Operating Characteristic (ROC) curve.
For clarity, we emphasize that this definition of the ROC curve specifically takes advantage of the distinction between the positive classes and the negative/OOD class to provide a single well-posed binarisation of the multi-class problem that doesn't rely on a one-vs-rest strategy. 
The area under the ROC curve (\textbf{AUROC}) is a threshold independent metric which is
commonly used by many image-level OOD detection methods~\citep{hendrycksBaselineDetectingMisclassified2017, liangEnhancingReliabilityOutofdistribution2018, leeSimpleUnifiedFramework2018, hsuGeneralizedODINDetecting2020}.
We thus use the AUROC metric (with our definition of $\TPR_{\ID}$ and $\FPR_{\OOD}$) for quantitative evaluation. 

Additionally, we propose to measure the Area Under the Precision-Recall curve (\textbf{AUPR})~\citep{saitoPrecisionRecallPlotMore2015} as our second metric. 
Again, we define the precision within our multi-class setting by taking advantage of the distinction between the positive classes and the negative/OOD one:
\begin{equation} 
    \Precision = \frac{\sum_{c=1}^{C} \TP_c}{\FP_0^{\OOD} + \sum_{c=1}^{C} \TP_c + \FN_c^{\ID}}
    \label{eqa:precision}
\end{equation}
Recall being a synonym for TPR, we use Equation~\eqref{eqa:tpr_and_fpr} to define it.
Finally, we measure AUPR by evaluating recall and precision under multiple $\tau$ thresholds.

\paragraph{All-classes metrics}
For our experiment using all labelled classes, we do not have any ground-truth pixels associated with the OOD class.
As illustrated in \figref{fig:ood_cmf}-right, $\FP_0^{\OOD}$ is thus 0 by construction and this would skew the previous metrics. 
In this context, we thus choose to compute TPR, TNR, balanced accuracy (BACC) and F1 score based on a one-vs-rest strategy ~\citep{tahaMetricsEvaluating3D2015}. 
To distinguish these one-vs-rest metrics used in the all-classes setting from the OOD-focused ones, we use a superscript $^{\OVR}$ when referring to them.
These expression for each individual positive class $c$ is given by:
\begin{equation}
\begin{aligned}
    \TPR_c^{\OVR} &= \frac{\TP_c}{\TP_c + \FN_c}, \\
    \TNR_c^{\OVR} &= \frac{\TN_c}{\TN_c + \FP_c}, \\
    \BACC_c^{\OVR} &= \frac{1}{2} (\TPR_c^{\OVR} + \TNR_c^{\OVR}), \\
    \Fone_c^{\OVR} &= \frac{2\TP_c}{2\TP_C + \FP_c + \FN_c}
\end{aligned}
\label{eqa:four_scores}
\end{equation}
where $\TN_c = \TN_c^{\ID} + \TN_c^{\OOD}$ and in OVR setting $\FP_c = \FP_c^{\ID}$. These class-specific OVR metrics are then averaged across the positive classes to provide mean scores: $\TPR^{\OVR}$, $\TNR^{\OVR}$, $\BACC^{\OVR}$ and $\Fone^{\OVR}$.
Furthermore, we compute the $\AUROC^{OVR}$ and $\AUPR^{OVR}$ metric in OVR setting by computing $\TPR_c^{\OVR}$, $\FPR_c^{\OVR} = 1 - \TNR_c^{\OVR}$ and $\Precision_c^{\OVR} = \frac{\TP_c}{\TP_c + \FP_c}$ under multiple thresholds.

\subsection{OOD confidence threshold selection}
\label{sec:threshold_selection}
As detailed in Equation~\eqref{eqa:ood_decision}, our approach relies on a OOD confidence threshold $\tau$ to generate the final segmentation masks.
This threshold should be chosen to
1) accurately classify pixels belonging to an ID class, and 2) detect background / OOD test pixels.
For comparison purposes, we can also define a baseline with $\tau_0 = 0$ to represent the method without outlier detection.
To fulfill the two criteria above within our two-level cross-validation setup, we propose to find the optimal threshold $\tau_m$ which maximises a weighted sum of ID and OOD performance across 
the two-level folds using a pair of threshold-sensitive metrics:
\begin{equation}
    \tau_m = \underset{\tau}{\max}\ \frac{1}{\text{N}} \sum_{k=1}^\text{N} w_{\ID} \Metric^{k}_{\ID}(\tau) + w_{\OOD} \Metric^{k}_{\OOD}(\tau)
    \label{eqa:threshold_selection}
\end{equation}
where $\text{N} = \text{N}_{\SP} \times \text{N}_{\CP}$ is total number of cross-validation folds;
and $\Metric^{k}_{\ID}(\tau)$ (respectively $\Metric^{k}_{\OOD}(\tau)$) represents the
ID (respectively OOD) performance of the model on the $k^{\textrm{th}}$ cross-validation fold when using threshold $\tau$.
In this work,
we choose $\TPR_{\ID}$ and $\TNR_{\OOD} = 1 - \FPR_{\OOD}$ as our ID and OOD metric respectively. Computation of $\TPR_{\ID}$ and $\FPR_{\OOD}$ can be found in Equation \eqref{eqa:tpr_and_fpr}.
For the weighting parameters in Equation~\eqref{eqa:threshold_selection}, we choose $w_{\ID}=w_{\OOD}=0.5$.

When used outside of our two-level cross-validation approach, the OOD performance metrics are skewed by the absence of negative/OOD annotations, in which case our threshold selection approach can be extended to only account for ID performance, essentially setting $w_{\ID}=1$ and $w_{\OOD}=0$.
An alternative is to use the optimal threshold from the two-level cross-validation experiments. 
We empirically found this threshold to offer a good trade-off between ID classification and OOD detection performance.
When the ID data distribution is similar to that of the validation set used during cross-validation, applying this threshold can be beneficial for generalisation purposes.

\section{Experimental setup}
\label{sec:experimental_setup}
We start by describing relevant details on our models and training details in \secref{sec:dl_model_and_training_setup} and followed by data preprocessing pipeline in \secref{sec:data_preprocessing}.

\subsection{Deep learning model and training setup}
\label{sec:dl_model_and_training_setup}
For all experiments, we use a U-Net architecture with an efficientnet-b4 encoder~\citep{tanEfficientNetRethinkingModel2019} pretrained on the ImageNet dataset~\citep{dengImageNetLargescaleHierarchical2009}.
Our implementation relies on the \emph{Segmentation models} PyTorch library\footnote{\href{https://github.com/qubvel/segmentation_models.pytorch}{https://github.com/qubvel/segmentation\_models.pytorch}}.
The choice of the encoder is based on good performance reported by previous study~\citep{seidlitzRobustDeepLearningbased2022} and graphical memory limits in the hardware used for this work.
The model inputs are either a pre-processed hyperspectral imaging (HSI) hypercube or an RGB image.
The number of input channels and weights of the first convolutional layer are re-initialised and set to match the number of channels of our input data.
The output of the network is passed on to a segmentation head to calculate the output logits.
The number of output classes is set to be equal to the number of positive classes (i.e. marked as ID) for a given experimental setup.
Note that during our two-level cross-validation, this number will be lower than the number of positive classes in the training dataset as some classes are being held out.

\begin{table}[tbh]
    \centering
    \caption{Hyperparameter setup.
    \label{tab:choice_of_hyperparameters}
    }
    \begin{tabular}{cccc}
    \hline
    \textbf{Dataset} & \textbf{Init. LR} & \textbf{Batch size} & \textbf{Epoch} \\
    \hline
    Heiporspectal & 1e-4 & 8 & 20 \\
    ODSI-DB & 1e-3 & 4 & 80 \\
    DSAD & 1e-4 & 4 & 10 \\
    \hline
    \end{tabular}
\end{table}

To perform model training, we minimise the cross-entropy loss between the softmax output and the one-hot encoded sparse annotation mask for pixels marked as ID (\figref{fig:overview} training stage). 
This approach is used for the Baseline and GODIN models.
We use Adam~\citep{kingmaAdamMethodStochastic2017} optimizer ($\beta_1$: 0.9 and $\beta_2$: 0.999) and exponential learning rate scheme with decay rate $\gamma=0.999$. The initial learning rate, mini-batch size and total number of epochs are varied across datasets. We show the choice of these hyperparameters in \tabref{tab:choice_of_hyperparameters}. For a fair comparison, we use the same hyperparameters for both methods.

To enable as fair a comparison as possible, we take advantage of the fact that the ODIN and Mahalanobis methods can be applied on a frozen, pre-trained model.
In this case, we re-use the weights from the Baseline model and implement the scoring function as a post-processing step.
We nonetheless allow for tuning the confidence threshold as described in \secref{sec:threshold_selection}

\subsection{Data preprocessing pipeline}
\label{sec:data_preprocessing}
For the DSAD dataset, data are stored in PNG format. We use the Pillow library\footnote{\href{https://pillow.readthedocs.io/en/latest/}{https://pillow.readthedocs.io/en/latest/}} to read the RGB data and convert them into PyTorch tensors. For the two HSI datasets, we first extract the hypercube using the provided Python libraries~\nolink{\citep{studier-fischerHeiPorSPECTRALHeidelbergPorcine2023, hyttinenOralDentalSpectral2020}} and manually select $16$ channels at equal intervals from the total available spectral bands, sorted in ascending order.

After exporting the data, we apply $\ell^1$-normalisation at each spatial location $ij$ to account for the non-uniform illumination of the tissue surface. This is routinely applied in hyperspectral imaging because of the dependency of the signal on the distance between the camera and the tissue~\citep{bahlSyntheticWhiteBalancing2023, studier-fischerHeiPorSPECTRALHeidelbergPorcine2023}. The uneven surface of the tissue can also cause some image areas to have different lighting conditions, which affects the classification accuracy and can be mitigated by data normalisation.
For data augmentation, we adopt similar setup reported in~\citep{seidlitzRobustDeepLearningbased2022}:
random rotation (rotation angle limit: $45^{\circ}$); random flip; random scaling (scaling factor limit: $0.1$); random shift (shift factor limit: $0.0625$).
All transformations are applied with a probability of $0.5$.

\section{Results}
\label{sec:results}
We begin by visualising confusion matrices and ROC curves as illustrated in \secref{sec:visualise_cfm_evaluation_curve}. These measures provide the foundation for both our qualitative and quantitative analysis in the later sections. \secref{sec:cross_valition_result_include_OOD} and \secref{sec:cross_valition_result_without_OOD} shows overall performance of our proposed framework comparing different methods as the scoring function. \secref{sec:qualitative_evaluation_all_class} and \secref{sec:qualitative_evaluation_ood_class} shows qualitative evaluation for all methods plus a scenario in which our proposed OOD segmentation framework has not applied. 
Furthermore, we have tested performance of our method under the scenario that all labelled classes are considered as ID. 
The results are shown in \secref{sec:cross_valition_result_without_OOD}.

\begin{figure}[htb]
    \centering
    \setlength\tabcolsep{1.5pt} 
    \begin{tabular}{cc}
    \includegraphics[width=.49\linewidth,height=.49\linewidth,valign=m]{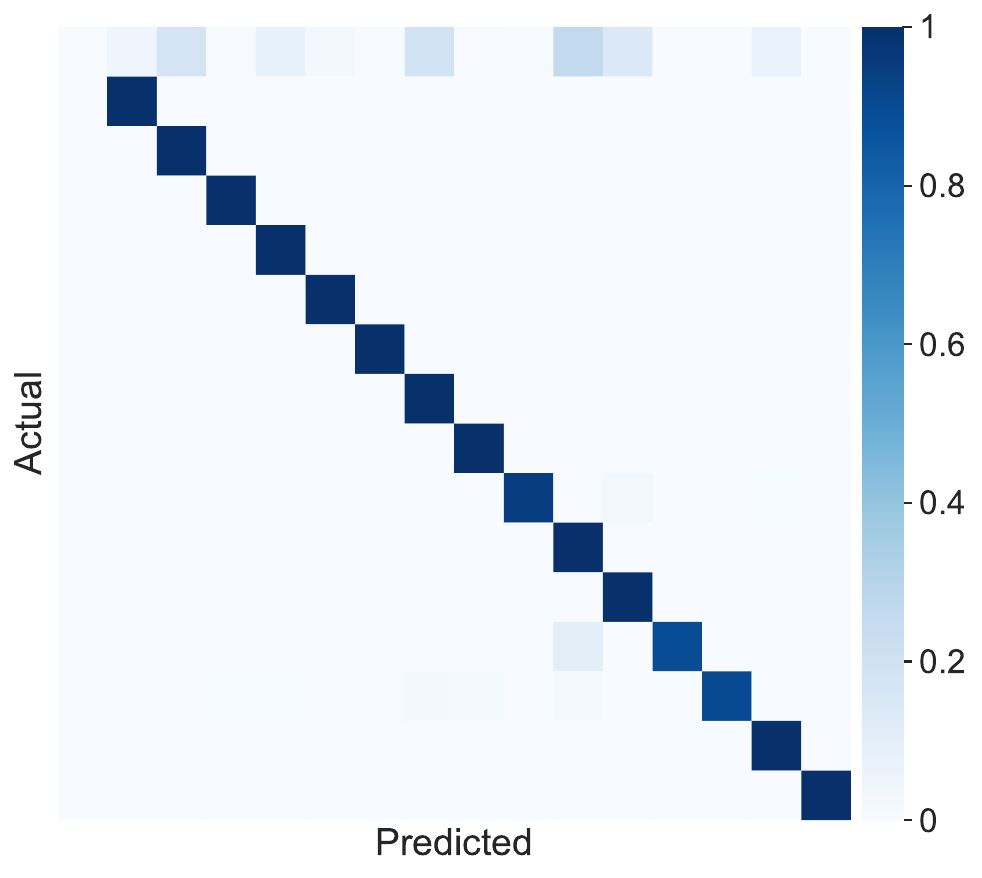} &
    \includegraphics[width=.49\linewidth,height=.49\linewidth,valign=m]{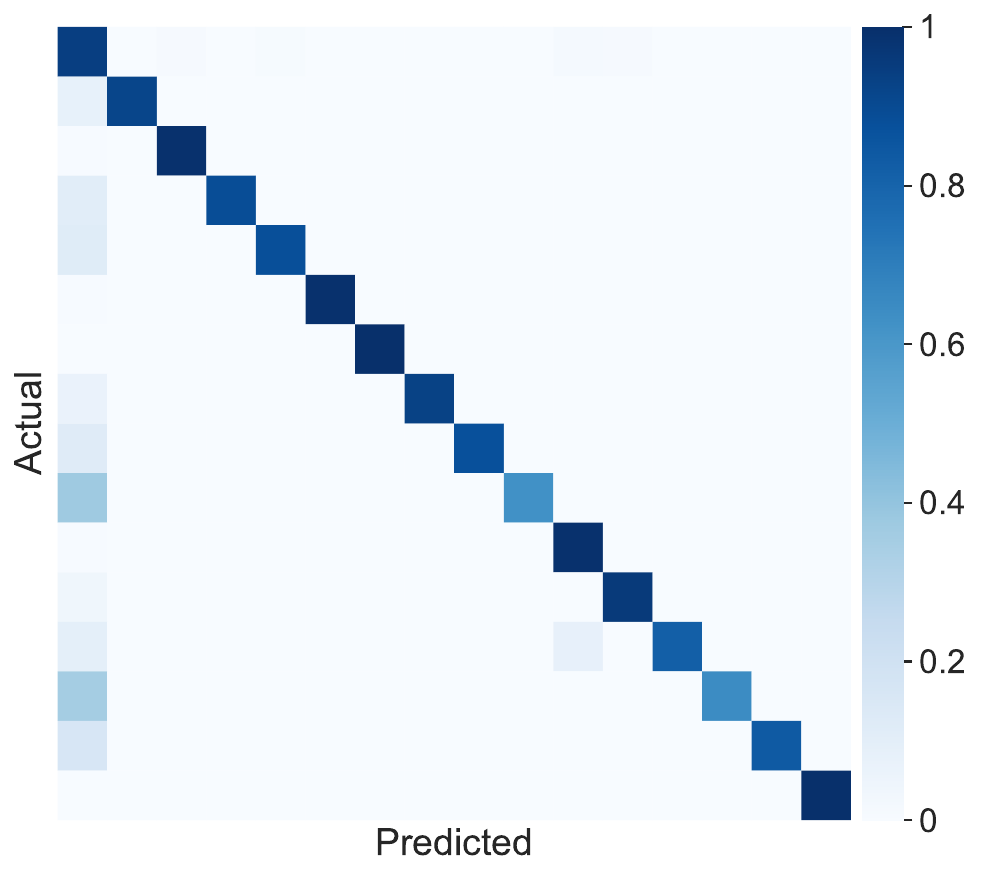}
    \end{tabular}
    \caption{Example confusion matrix at threshold $\tau_0=0$ (left) and $\tau_m$ (right) for the ODIN method on the Heiporspectral dataset with a specific held-out class partition. The first row and column represent the negative / outlier class.}
    \label{fig:confusion_matrix}
\end{figure} 

\begin{figure*}[tbh]
    \centering
    \setlength\tabcolsep{1.5pt} 
    \begin{tabular}{cccc}
    \includegraphics[width=.24\linewidth,height=.24\linewidth,valign=m]{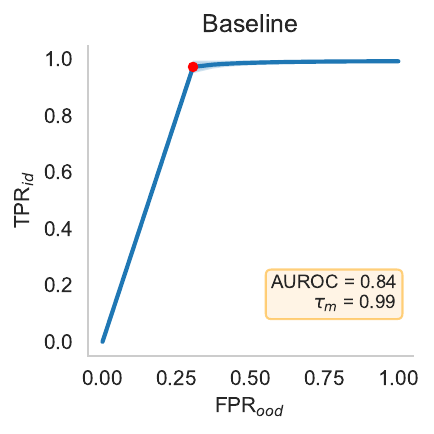} & 
    \includegraphics[width=.24\linewidth,height=.24\linewidth,valign=m]{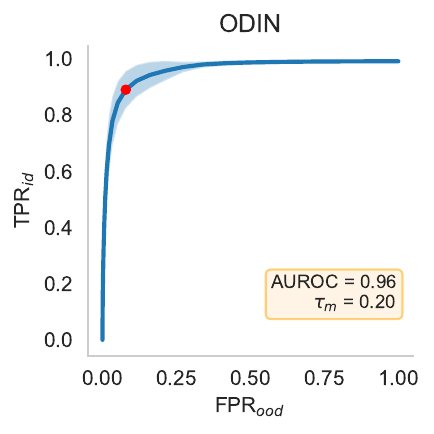} &
    \includegraphics[width=.24\linewidth,height=.24\linewidth,valign=m]{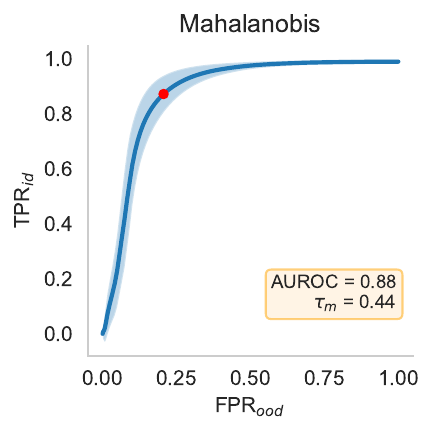} &
    \includegraphics[width=.24\linewidth,height=.24\linewidth,valign=m]{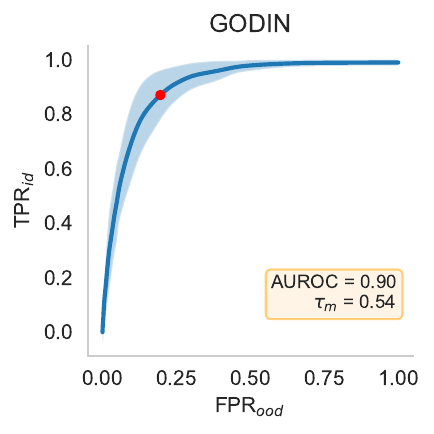} \\
    \end{tabular}
    \caption{ROC curves comparing different OOD detection methods within our framework. The curve shows the average performance of $16$ models over all two-level folds. Shaded area represents standard deviation across the folds. Experiments were conducted on the Heiporspectral dataset.
    }
    \label{fig:evaluation_curve}
\end{figure*}

\subsection{Visualising confusion matrix and ROC curve}
\label{sec:visualise_cfm_evaluation_curve}
\figref{fig:confusion_matrix} demonstrates that our method effectively separates OOD pixels while maintaining classification accuracy for ID pixels.
\figref{fig:evaluation_curve} shows ROC curve with ID class on the y-axis and OOD classes on the x-axis.
As $\tau$ increases, more data will be flagged as OOD.
For the baseline method, fewer pixels are rejected under threshold $0.99$, indicating that the model tends to make overconfident predictions and this decreases the overall AUC.
The ODIN method addresses this issue by providing a calibrated score that aligns with the actual likelihood of correctness, resulting in a less overconfident threshold $\tau_m$ for generating the predicted mask.

\subsection{Cross-validations results for OOD detection}
\label{sec:cross_valition_result_include_OOD}
\tabref{tab:cross_validation_different_CF} shows cross-validation results on both $\text{AUROC}$ and $\text{AUPR}$ metrics.
For each method and metric, we report results across different CP.
For each CP, we aggregate results across all folds sharing the same SP and report mean and standard deviation. Similarly, \tabref{tab:cross_validation_different_SF} shows cross-validation results on both $\text{AUROC}$ and $\text{AUPR}$ metrics. Here for each method and metric, we report results across different SP by aggregating results across all folds sharing the same CP.
Again we report the mean and standard deviation for all results. We observed a consistent trend between two hyperspectral image datasets and one RGB color image dataset, suggesting that our framework is applicable to these medical imaging modalities.

Overall, we found that using the ODIN method as a scoring function yielded the best performance. This suggests that a well-calibrated confidence score is crucial for detecting OOD data. The GODIN method employs a learnable temperature parameter for calibration during training, which may increase training complexity as it requires learning both the model parameters for optimal ID performance and the temperature parameter for effective calibration. 
The underperformance of the Mahalanobis distance based approach could be explained by the relatively high dimension of the feature space on which we use it. Covariance estimation in high dimensional space is indeed prone to ill-conditioned results and thus poor reliability of inverse covariance based features.

\begin{table*}[tb]
    \centering\setlength\tabcolsep{3pt}
    \caption{Cross-validation results for different class partitions (CP). For each CP, we aggregate results across all two-level folds sharing same subject partition (SP) and report mean and standard deviation. We compared different OOD segmentation methods on two threshold independent metrics. Best performances are highlighted in bold.
    \label{tab:cross_validation_different_CF}
    }
    \begin{tabular}{cc ccccc ccccc}
    \toprule
    \textbf{Dataset} & \textbf{Method} & 
    \multicolumn{5}{c}{$\boldsymbol{\AUROC}\uparrow$} &
    \multicolumn{5}{c}{$\boldsymbol{\AUPR}\uparrow$} \\
    \cmidrule(lr){3-7}
    \cmidrule(lr){8-12}
    \multicolumn{2}{c}{} & \boldmath$\CP_1$ & \boldmath$\CP_2$ & \boldmath$\CP_3$ & \boldmath$\CP_4$ & \boldmath$\CP_{\text{mean}}$
                         & \boldmath$\CP_1$ & \boldmath$\CP_2$ & \boldmath$\CP_3$ & \boldmath$\CP_4$ & \boldmath$\CP_{\text{mean}}$ \\
    \midrule
    \multirow{4}{*}{Heiporspectral} & Baseline & $0.85_{\pm0.04}$ & $0.78_{\pm0.08}$ & $0.85_{\pm0.02}$ & $0.85_{\pm0.06}$ & $0.84_{\pm0.03}$ & $0.45_{\pm0.04}$ & $0.45_{\pm0.03}$ & $0.46_{\pm0.01}$ & $0.47_{\pm0.02}$ & $0.46_{\pm0.01}$ \\
     & ODIN & $\boldsymbol{0.97}_{\pm0.02}$ & $\boldsymbol{0.95}_{\pm0.03}$ & $\boldsymbol{0.96}_{\pm0.02}$ & $\boldsymbol{0.97}_{\pm0.01}$ & $\boldsymbol{0.96}_{\pm0.01}$ & $\boldsymbol{0.98}_{\pm0.01}$ & $\boldsymbol{0.98}_{\pm0.02}$ & $\boldsymbol{0.98}_{\pm0.02}$ & $\boldsymbol{0.99}_{\pm0.00}$ & $\boldsymbol{0.98}_{\pm0.00}$ \\
     & Mahalanobis & $0.93_{\pm0.01}$ & $0.87_{\pm0.03}$ & $0.88_{\pm0.07}$ & $0.86_{\pm0.03}$ & $0.88_{\pm0.03}$ & $0.95_{\pm0.01}$ & $0.94_{\pm0.02}$ & $0.93_{\pm0.03}$ & $0.92_{\pm0.02}$ & $0.93_{\pm0.01}$ \\
     & GODIN & $0.91_{\pm0.07}$ & $0.86_{\pm0.08}$ & $\boldsymbol{0.96}_{\pm0.02}$ & $0.92_{\pm0.05}$ & $0.91_{\pm0.04}$ & $0.93_{\pm0.06}$ & $0.93_{\pm0.06}$ & $0.97_{\pm0.02}$ & $0.93_{\pm0.07}$ & $0.94_{\pm0.02}$ \\
    \midrule
    \multirow{4}{*}{ODSI-DB} & Baseline & $0.66_{\pm0.08}$ & $\boldsymbol{0.76}_{\pm0.02}$ & $0.80_{\pm0.05}$ & $0.71_{\pm0.04}$ & $0.73_{\pm0.05}$ & $0.52_{\pm0.05}$ & $0.51_{\pm0.03}$ & $0.45_{\pm0.04}$ & $0.49_{\pm0.02}$ & $0.49_{\pm0.02}$ \\
     & ODIN & $0.66_{\pm0.08}$ & $0.74_{\pm0.05}$ & $\boldsymbol{0.84}_{\pm0.04}$ & $\boldsymbol{0.78}_{\pm0.02}$ & $\boldsymbol{0.75}_{\pm0.07}$ & $\boldsymbol{0.78}_{\pm0.03}$ & $\boldsymbol{0.85}_{\pm0.03}$ & $\boldsymbol{0.85}_{\pm0.05}$ & $\boldsymbol{0.87}_{\pm0.05}$ & $\boldsymbol{0.84}_{\pm0.03}$ \\
     & Mahalanobis & $\boldsymbol{0.67}_{\pm0.07}$ & $\boldsymbol{0.76}_{\pm0.02}$ & $0.78_{\pm0.06}$ & $0.76_{\pm0.04}$ & $0.74_{\pm0.04}$ & $0.60_{\pm0.07}$ & $0.65_{\pm0.02}$ & $0.64_{\pm0.05}$ & $0.58_{\pm0.04}$ & $0.62_{\pm0.03}$ \\
     & GODIN & $0.66_{\pm0.08}$ & $0.65_{\pm0.06}$ & $0.69_{\pm0.05}$ & $0.50_{\pm0.25}$ & $0.62_{\pm0.08}$ & $0.71_{\pm0.07}$ & $0.73_{\pm0.05}$ & $0.66_{\pm0.07}$ & $0.49_{\pm0.32}$ & $0.65_{\pm0.09}$ \\
    \midrule
    \multirow{4}{*}{DSAD} & Baseline & $0.65_{\pm0.11}$ & $0.68_{\pm0.07}$ & $0.68_{\pm0.13}$ & $0.76_{\pm0.05}$ & $0.69_{\pm0.04}$ & $0.54_{\pm0.08}$ & $0.42_{\pm0.05}$ & $0.46_{\pm0.05}$ & $0.60_{\pm0.03}$ & $0.51_{\pm0.07}$ \\
     & ODIN & $0.65_{\pm0.12}$ & $\boldsymbol{0.70}_{\pm0.09}$ & $\boldsymbol{0.70}_{\pm0.13}$ & $\boldsymbol{0.78}_{\pm0.06}$ & $\boldsymbol{0.71}_{\pm0.04}$ & $\boldsymbol{0.80}_{\pm0.11}$ & $\boldsymbol{0.73}_{\pm0.09}$ & $\boldsymbol{0.72}_{\pm0.13}$ & $\boldsymbol{0.85}_{\pm0.03}$ & $\boldsymbol{0.78}_{\pm0.05}$ \\
     & Mahalanobis & $0.64_{\pm0.08}$ & $0.66_{\pm0.05}$ & $0.68_{\pm0.12}$ & $0.70_{\pm0.02}$ & $0.67_{\pm0.02}$ & $0.77_{\pm0.10}$ & $0.64_{\pm0.05}$ & $0.67_{\pm0.14}$ & $0.83_{\pm0.02}$ & $0.73_{\pm0.07}$ \\
     & GODIN & $\boldsymbol{0.68}_{\pm0.04}$ & $0.67_{\pm0.10}$ & $0.68_{\pm0.08}$ & $0.71_{\pm0.10}$ & $0.68_{\pm0.02}$ & $0.76_{\pm0.06}$ & $0.68_{\pm0.15}$ & $0.67_{\pm0.11}$ & $0.80_{\pm0.05}$ & $0.72_{\pm0.05}$ \\
    \bottomrule
    \end{tabular}
\end{table*}

\begin{table*}[tb]
    \centering\setlength\tabcolsep{3pt}
    \caption{Cross-validation results for different subject partitions (SP). For each SP, we aggregate results across all two-level folds sharing the same class partition (CP) and report mean and standard deviation. We compared different OOD segmentation methods on two threshold independent metrics. Best performances are highlighted in bold.
    \label{tab:cross_validation_different_SF}
    }
    \begin{tabular}{cc ccccc ccccc}
    \toprule
    \textbf{Dataset} & \textbf{Method} & 
    \multicolumn{5}{c}{$\boldsymbol{\AUROC}\uparrow$} &
    \multicolumn{5}{c}{$\boldsymbol{\AUPR}\uparrow$} \\
    \cmidrule(lr){3-7}
    \cmidrule(lr){8-12}
    \multicolumn{2}{c}{} & \boldmath$\SP_1$ & \boldmath$\SP_2$ & \boldmath$\SP_3$ & \boldmath$\SP_4$ & \boldmath$\SP_{\text{mean}}$
                         & \boldmath$\SP_1$ & \boldmath$\SP_2$ & \boldmath$\SP_3$ & \boldmath$\SP_4$ & \boldmath$\SP_{\text{mean}}$ \\
    \midrule
    \multirow{4}{*}{Heiporspectral} & Baseline & $0.81_{\pm0.03}$ & $0.84_{\pm0.09}$ & $0.83_{\pm0.06}$ & $0.86_{\pm0.03}$ & $0.84_{\pm0.02}$ & $0.44_{\pm0.02}$ & $0.46_{\pm0.04}$ & $0.47_{\pm0.02}$ & $0.46_{\pm0.02}$ & $0.46_{\pm0.01}$ \\
     & ODIN & $\boldsymbol{0.96}_{\pm0.03}$ & $\boldsymbol{0.97}_{\pm0.01}$ & $\boldsymbol{0.97}_{\pm0.01}$ & $\boldsymbol{0.95}_{\pm0.02}$ & $\boldsymbol{0.96}_{\pm0.01}$ & $\boldsymbol{0.97}_{\pm0.01}$ & $\boldsymbol{0.99}_{\pm0.01}$ & $\boldsymbol{0.99}_{\pm0.01}$ & $\boldsymbol{0.97}_{\pm0.01}$ & $\boldsymbol{0.98}_{\pm0.01}$ \\
     & Mahalanobis & $0.88_{\pm0.05}$ & $0.85_{\pm0.06}$ & $0.90_{\pm0.03}$ & $0.91_{\pm0.03}$ & $0.88_{\pm0.02}$ & $0.93_{\pm0.02}$ & $0.91_{\pm0.03}$ & $0.94_{\pm0.02}$ & $0.95_{\pm0.01}$ & $0.93_{\pm0.01}$ \\
     & GODIN & $0.93_{\pm0.02}$ & $0.91_{\pm0.11}$ & $0.94_{\pm0.02}$ & $0.86_{\pm0.06}$ & $0.91_{\pm0.03}$ & $0.96_{\pm0.02}$ & $0.95_{\pm0.07}$ & $0.96_{\pm0.01}$ & $0.89_{\pm0.06}$ & $0.94_{\pm0.03}$ \\
    \midrule
    \multirow{4}{*}{ODSI-DB} & Baseline & $0.78_{\pm0.03}$ & $0.71_{\pm0.04}$ & $0.72_{\pm0.09}$ & $\boldsymbol{0.72}_{\pm0.11}$ & $0.73_{\pm0.03}$ & $0.50_{\pm0.03}$ & $0.48_{\pm0.08}$ & $0.49_{\pm0.02}$ & $0.49_{\pm0.03}$ & $0.49_{\pm0.01}$ \\
     & ODIN & $\boldsymbol{0.81}_{\pm0.04}$ & $0.72_{\pm0.07}$ & $\boldsymbol{0.76}_{\pm0.09}$ & $\boldsymbol{0.72}_{\pm0.11}$ & $\boldsymbol{0.75}_{\pm0.04}$ & $\boldsymbol{0.87}_{\pm0.05}$ & $\boldsymbol{0.82}_{\pm0.03}$ & $\boldsymbol{0.85}_{\pm0.07}$ & $\boldsymbol{0.81}_{\pm0.04}$ & $\boldsymbol{0.84}_{\pm0.02}$ \\
     & Mahalanobis & $0.80_{\pm0.04}$ & $\boldsymbol{0.76}_{\pm0.06}$ & $0.72_{\pm0.08}$ & $0.69_{\pm0.05}$ & $0.74_{\pm0.04}$ & $0.64_{\pm0.07}$ & $0.63_{\pm0.02}$ & $0.60_{\pm0.07}$ & $0.60_{\pm0.02}$ & $0.62_{\pm0.02}$ \\
     & GODIN & $0.56_{\pm0.23}$ & $0.64_{\pm0.03}$ & $0.59_{\pm0.19}$ & $0.70_{\pm0.04}$ & $0.62_{\pm0.05}$ & $0.54_{\pm0.28}$ & $0.67_{\pm0.08}$ & $0.61_{\pm0.18}$ & $0.77_{\pm0.04}$ & $0.65_{\pm0.08}$ \\
    \midrule
    \multirow{4}{*}{DSAD} & Baseline & $0.73_{\pm0.05}$ & $0.74_{\pm0.11}$ & $0.66_{\pm0.11}$ & $0.63_{\pm0.07}$ & $0.69_{\pm0.05}$ & $0.54_{\pm0.09}$ & $0.50_{\pm0.09}$ & $0.47_{\pm0.07}$ & $0.51_{\pm0.12}$ & $0.51_{\pm0.03}$ \\
     & ODIN & $\boldsymbol{0.76}_{\pm0.05}$ & $0.76_{\pm0.13}$ & $\boldsymbol{0.69}_{\pm0.11}$ & $0.63_{\pm0.10}$ & $\boldsymbol{0.71}_{\pm0.05}$ & $\boldsymbol{0.84}_{\pm0.03}$ & $\boldsymbol{0.81}_{\pm0.13}$ & $\boldsymbol{0.74}_{\pm0.11}$ & $\boldsymbol{0.72}_{\pm0.10}$ & $\boldsymbol{0.78}_{\pm0.05}$ \\
     & Mahalanobis & $0.71_{\pm0.06}$ & $0.70_{\pm0.08}$ & $0.62_{\pm0.07}$ & $\boldsymbol{0.64}_{\pm0.08}$ & $0.67_{\pm0.04}$ & $0.78_{\pm0.10}$ & $0.77_{\pm0.12}$ & $0.67_{\pm0.12}$ & $0.71_{\pm0.10}$ & $0.73_{\pm0.05}$ \\
     & GODIN & $0.67_{\pm0.07}$ & $\boldsymbol{0.77}_{\pm0.04}$ & $0.67_{\pm0.07}$ & $0.63_{\pm0.06}$ & $0.68_{\pm0.05}$ & $0.77_{\pm0.04}$ & $0.80_{\pm0.04}$ & $0.67_{\pm0.12}$ & $0.66_{\pm0.14}$ & $0.72_{\pm0.06}$ \\
    \bottomrule
    \end{tabular}
\end{table*}

\begin{figure*}[!htb]
\centering\footnotesize
\setlength\tabcolsep{10pt} 
\begin{tabular}{cccc}
\vspace{5pt}
& \textbf{Heiporspectral} & \textbf{ODSI-DB} & \textbf{DSAD} \\
\vspace{5pt}
\makecell[c]
{\textbf{Sparsely} \\ \textbf{annotated} \\ \textbf{ground truth}} & 
\includegraphics[width=.22\linewidth,height=.17\linewidth,valign=m]{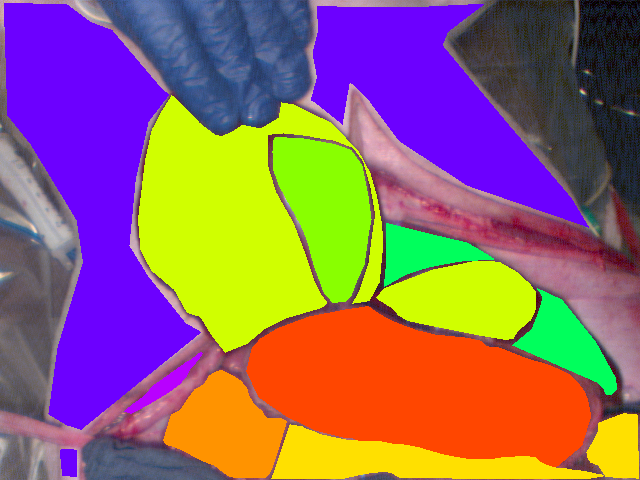} &
\includegraphics[width=.22\linewidth,height=.17\linewidth,valign=m]{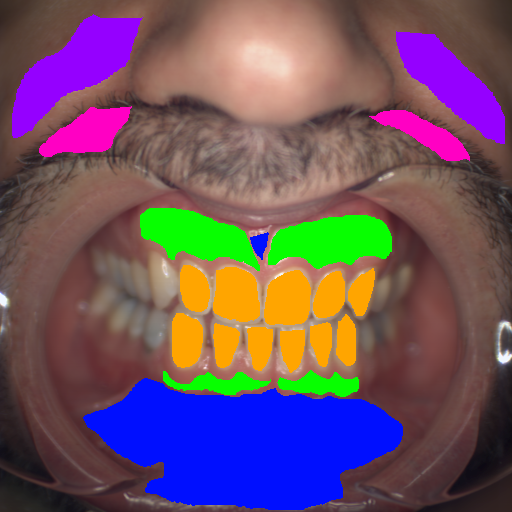} &
\includegraphics[width=.22\linewidth,height=.17\linewidth,valign=m]{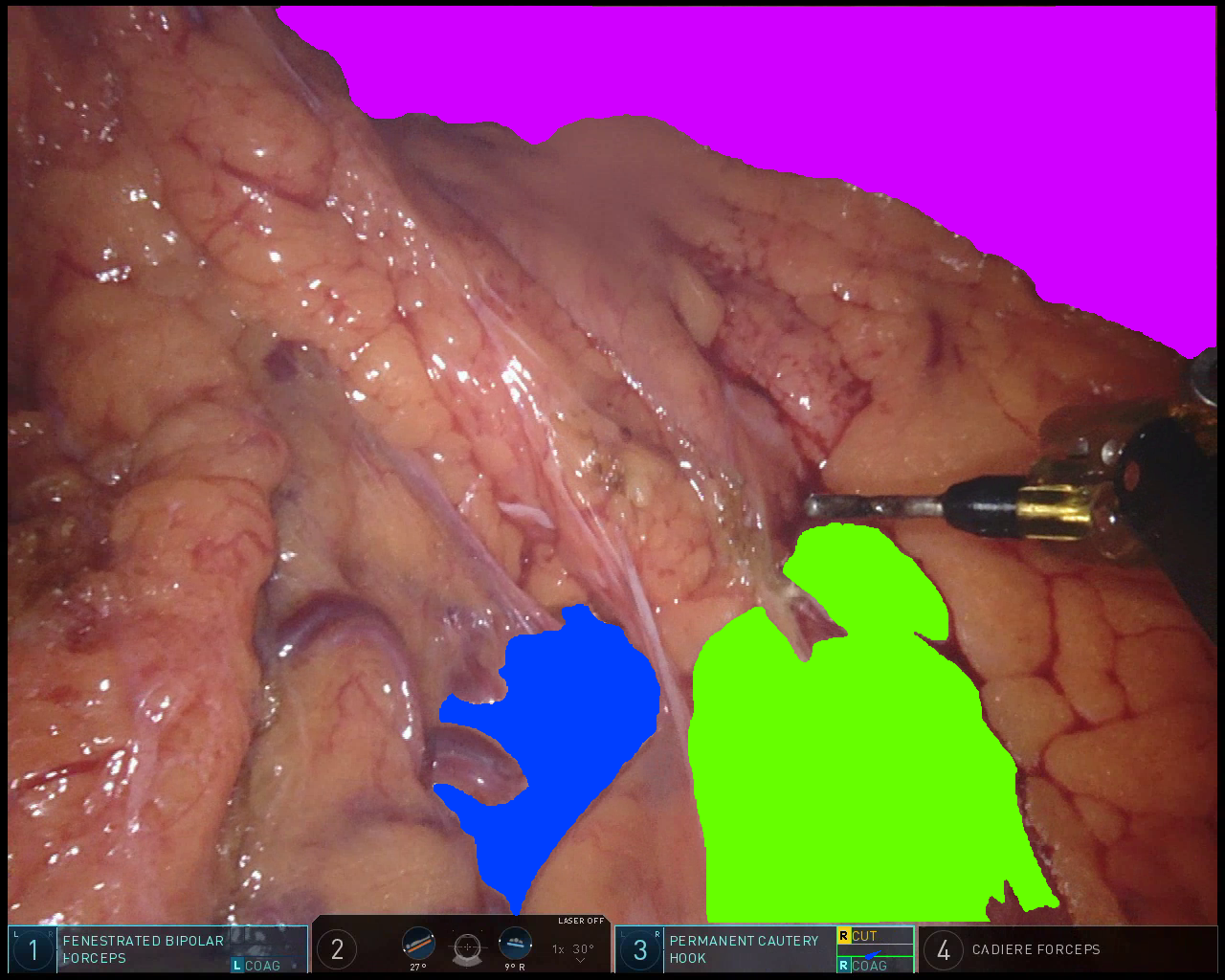}\\
\vspace{5pt}
\textbf{Baseline\ $(\tau_{0}=0$)} & 
\includegraphics[width=.22\linewidth,height=.17\linewidth,valign=m]{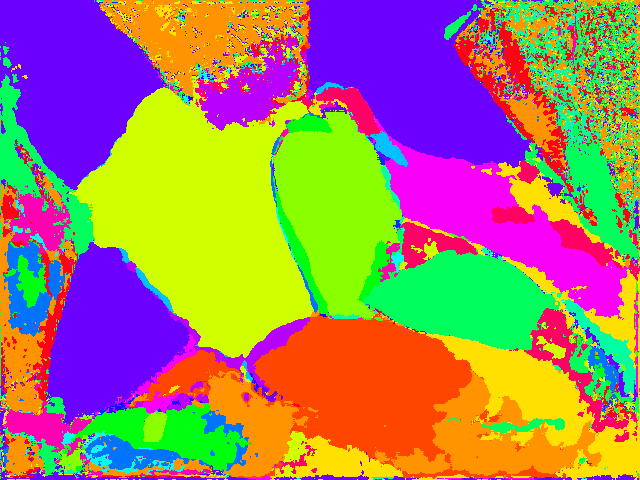} &
\includegraphics[width=.22\linewidth,height=.17\linewidth,valign=m]{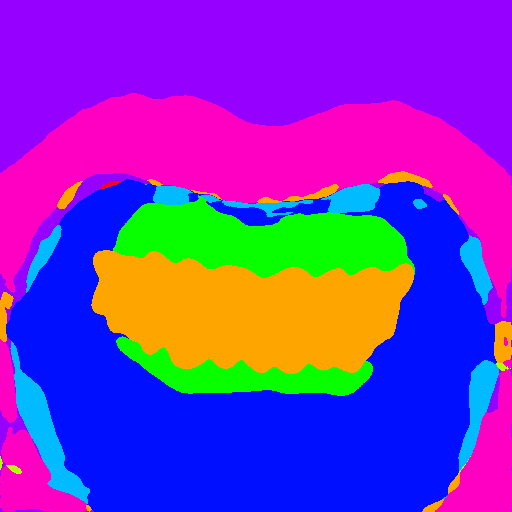} &
\includegraphics[width=.22\linewidth,height=.17\linewidth,valign=m]{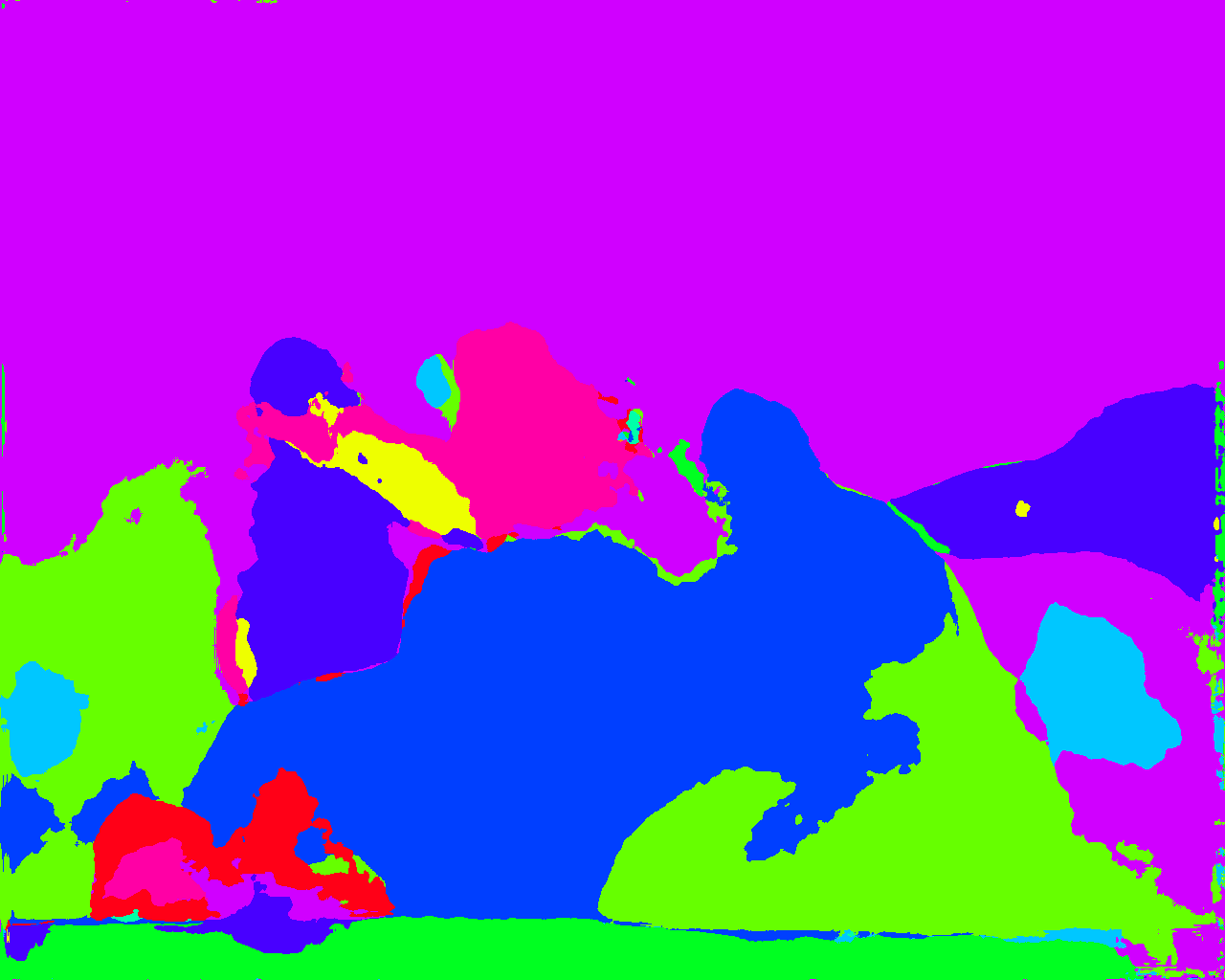}\\
\vspace{5pt}
\textbf{Baseline\ $(\tau_{m}$)} & 
\includegraphics[width=.22\linewidth,height=.17\linewidth,valign=m]{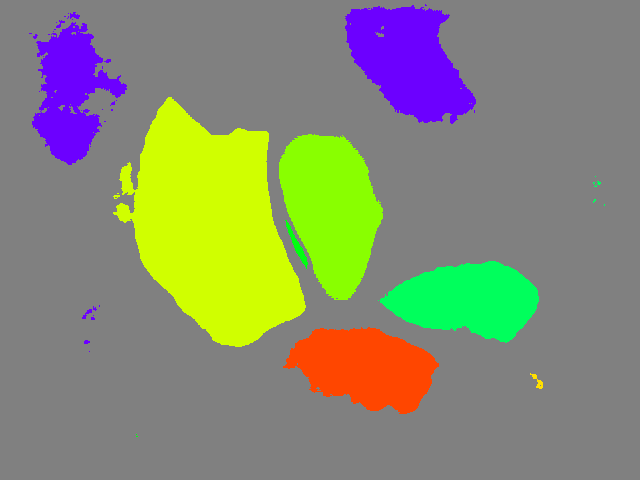} &
\includegraphics[width=.22\linewidth,height=.17\linewidth,valign=m]{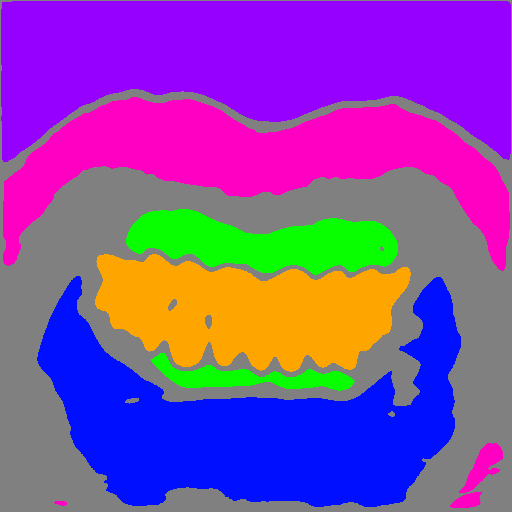} &
\includegraphics[width=.22\linewidth,height=.17\linewidth,valign=m]{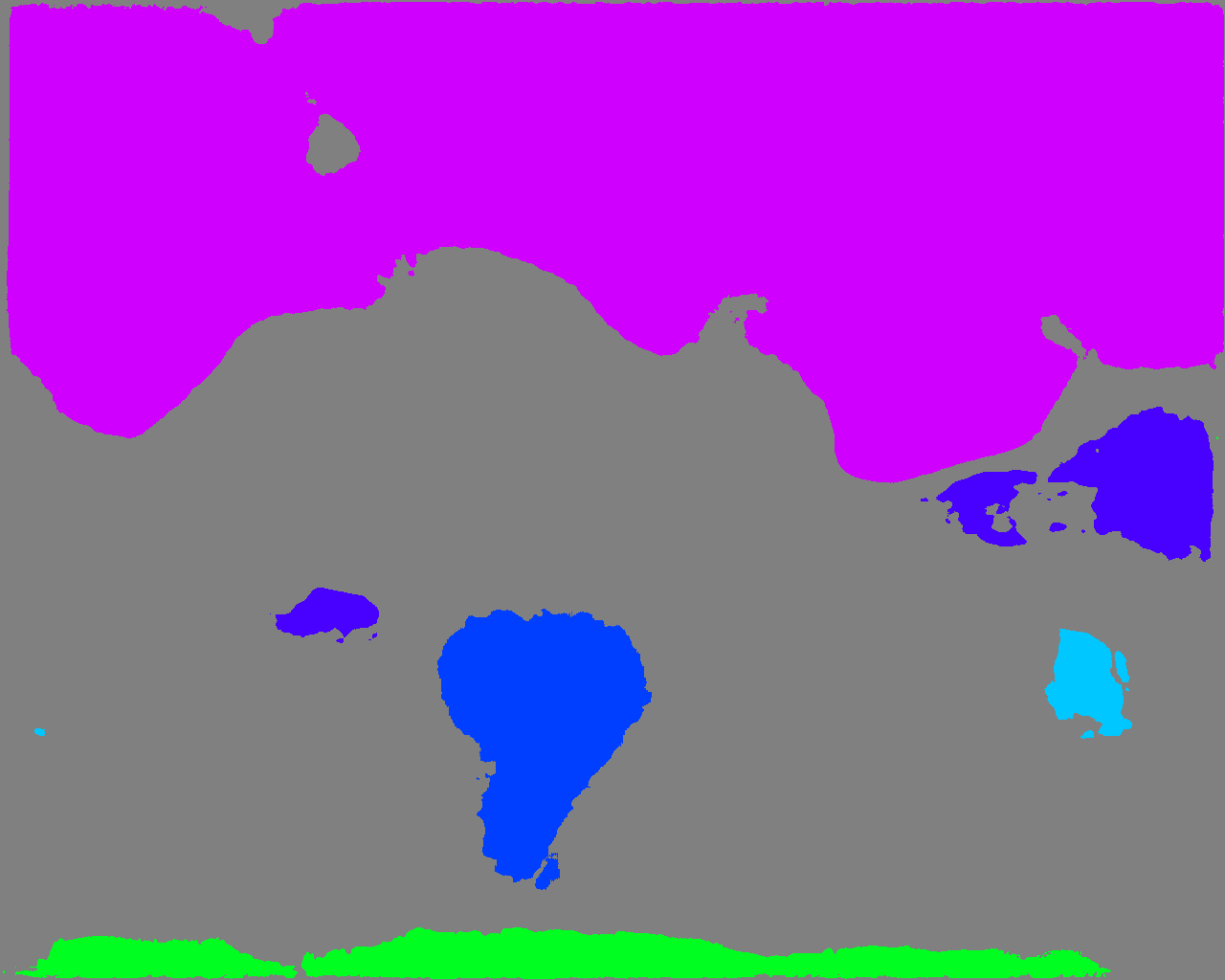}\\
\vspace{5pt}
\textbf{ODIN} & 
\includegraphics[width=.22\linewidth,height=.17\linewidth,valign=m]{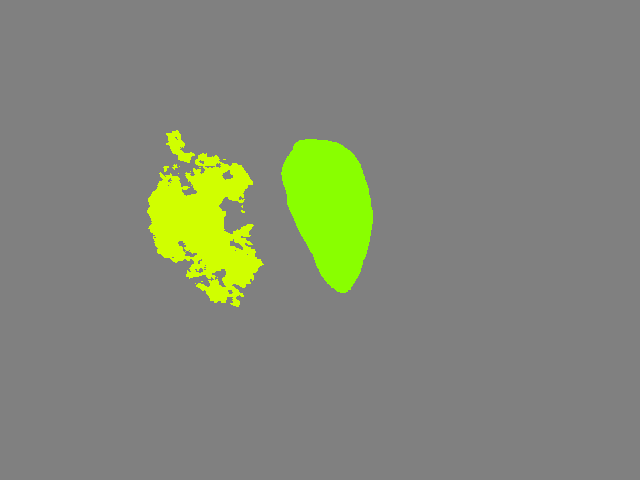} &
\includegraphics[width=.22\linewidth,height=.17\linewidth,valign=m]{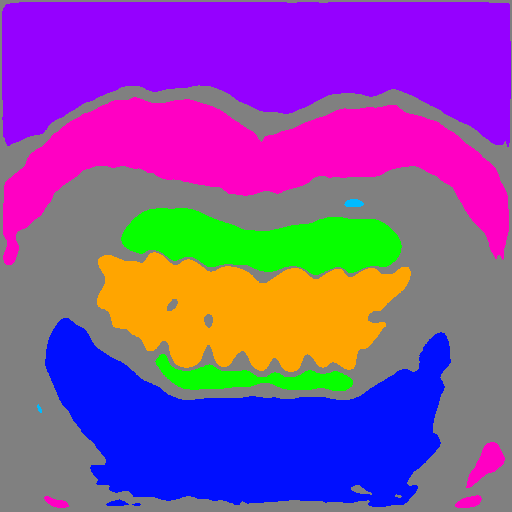} &
\includegraphics[width=.22\linewidth,height=.17\linewidth,valign=m]{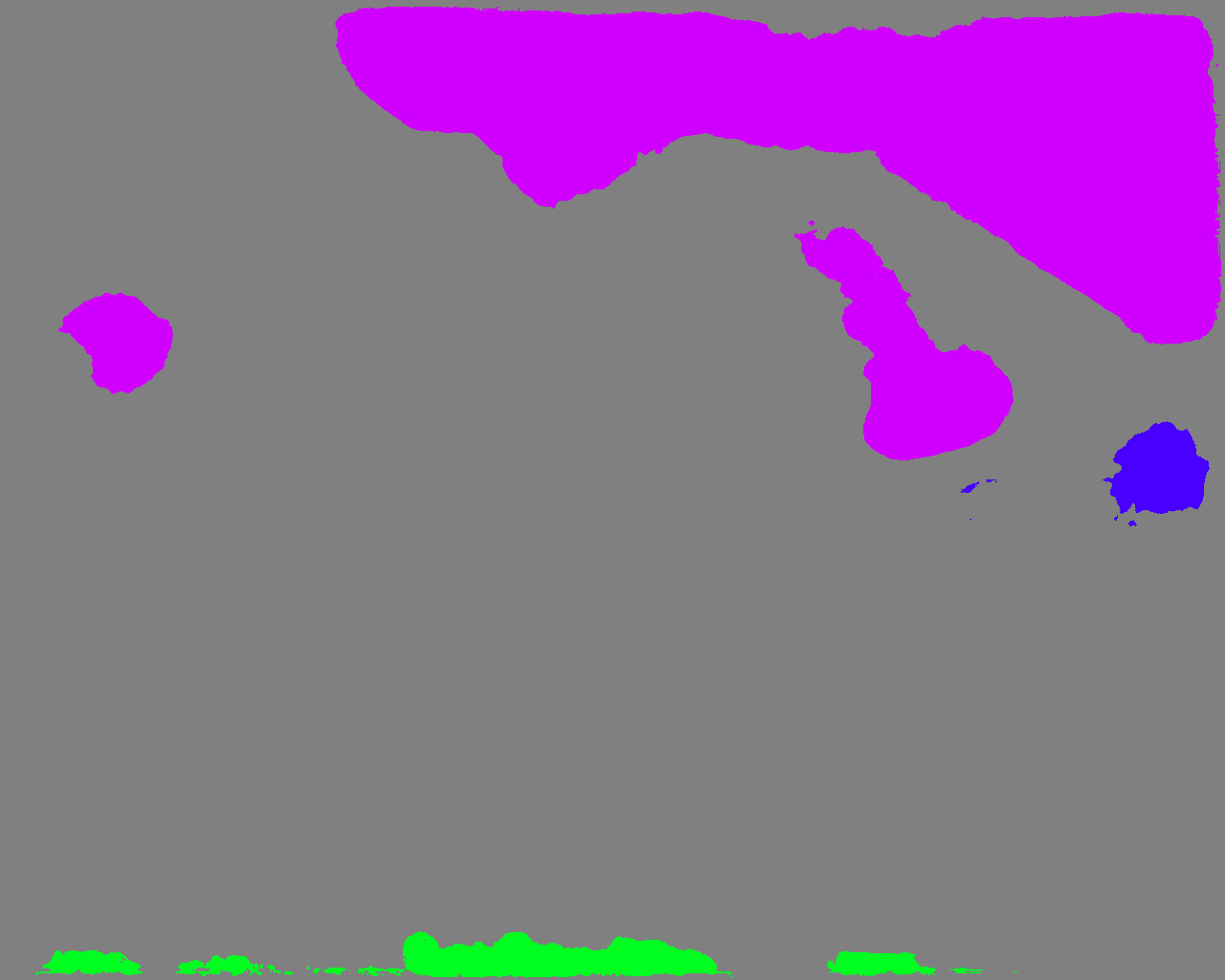}\\
\vspace{5pt}
\textbf{Mahalanobis} & 
\includegraphics[width=.22\linewidth,height=.17\linewidth,valign=m]{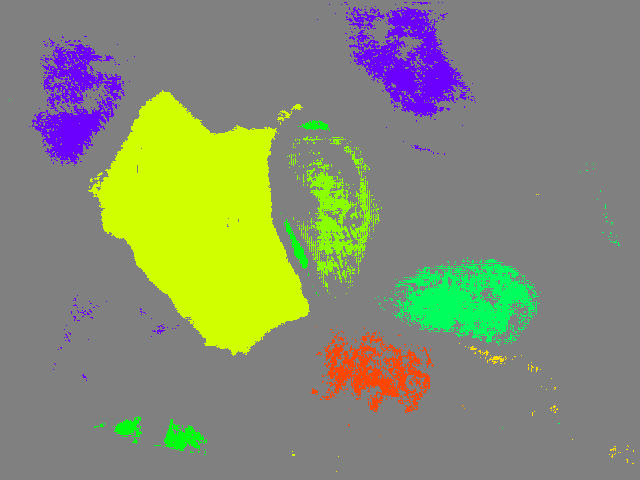} &
\includegraphics[width=.22\linewidth,height=.17\linewidth,valign=m]{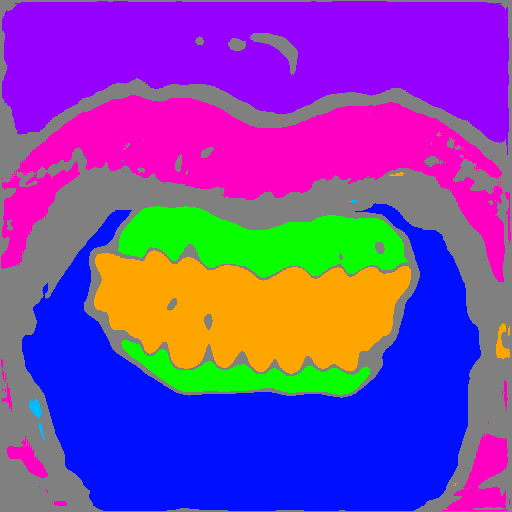} &
\includegraphics[width=.22\linewidth,height=.17\linewidth,valign=m]{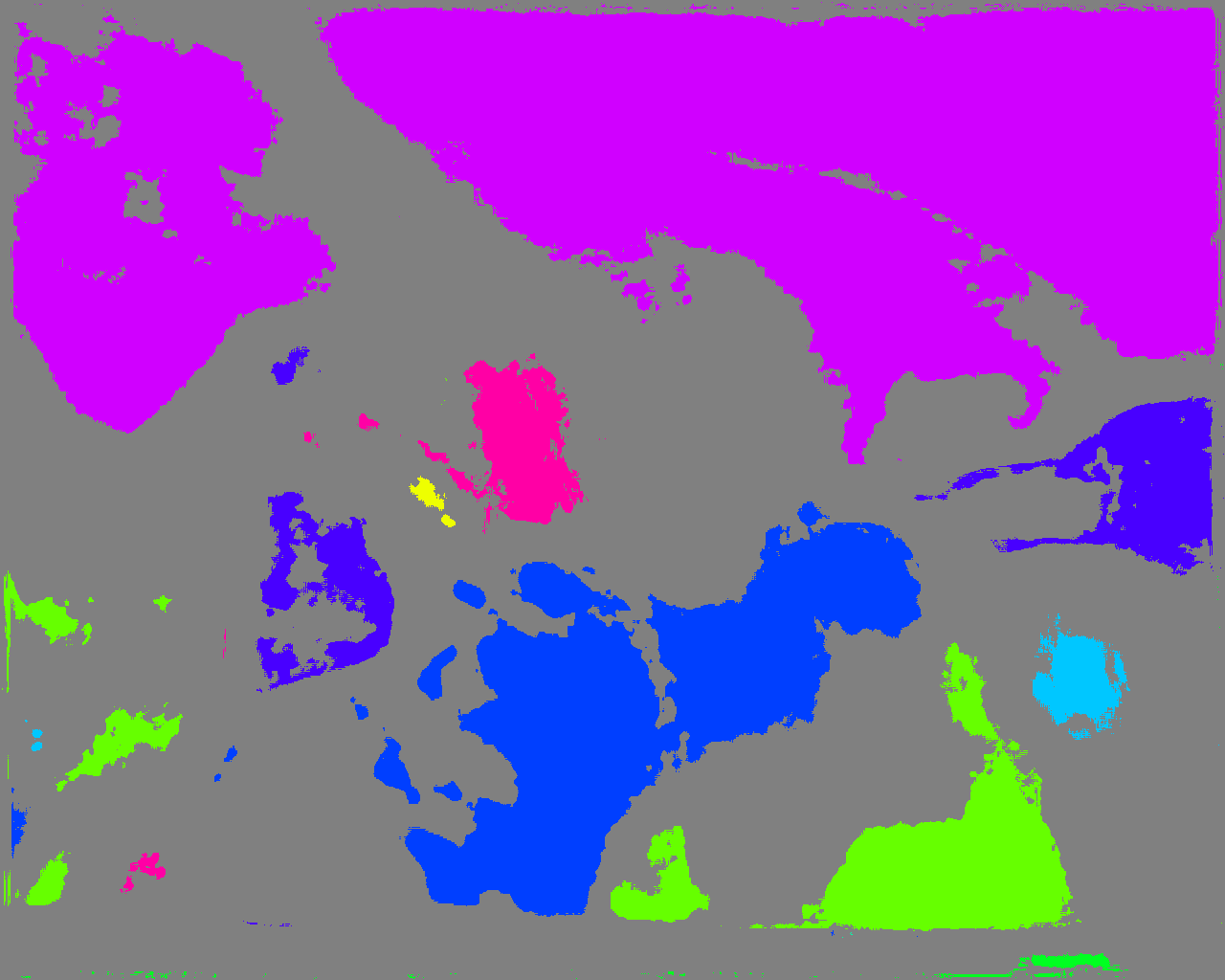}\\
\vspace{5pt}
\textbf{GODIN} & 
\includegraphics[width=.22\linewidth,height=.17\linewidth,valign=m]{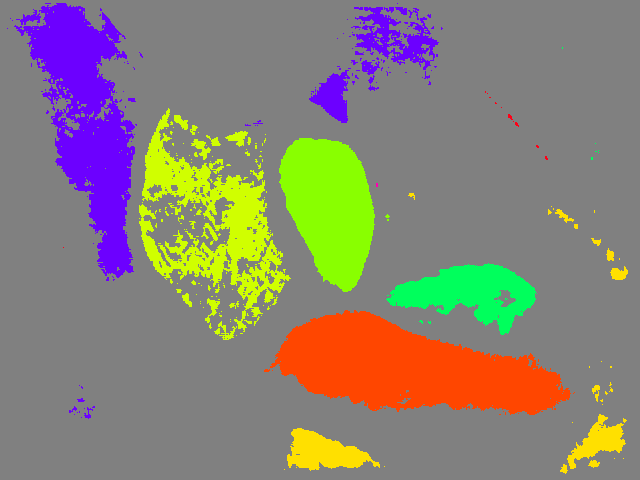} &
\includegraphics[width=.22\linewidth,height=.17\linewidth,valign=m]{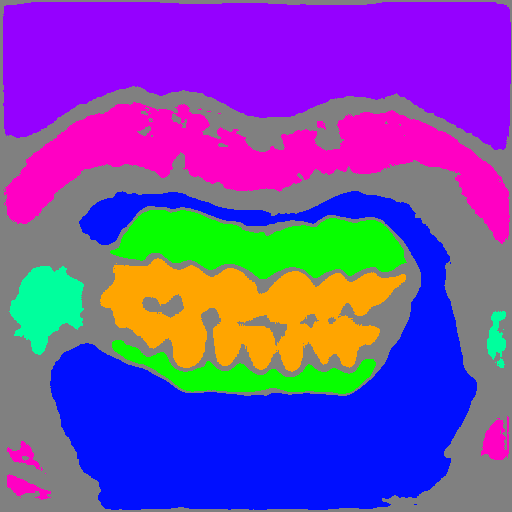} &
\includegraphics[width=.22\linewidth,height=.17\linewidth,valign=m]{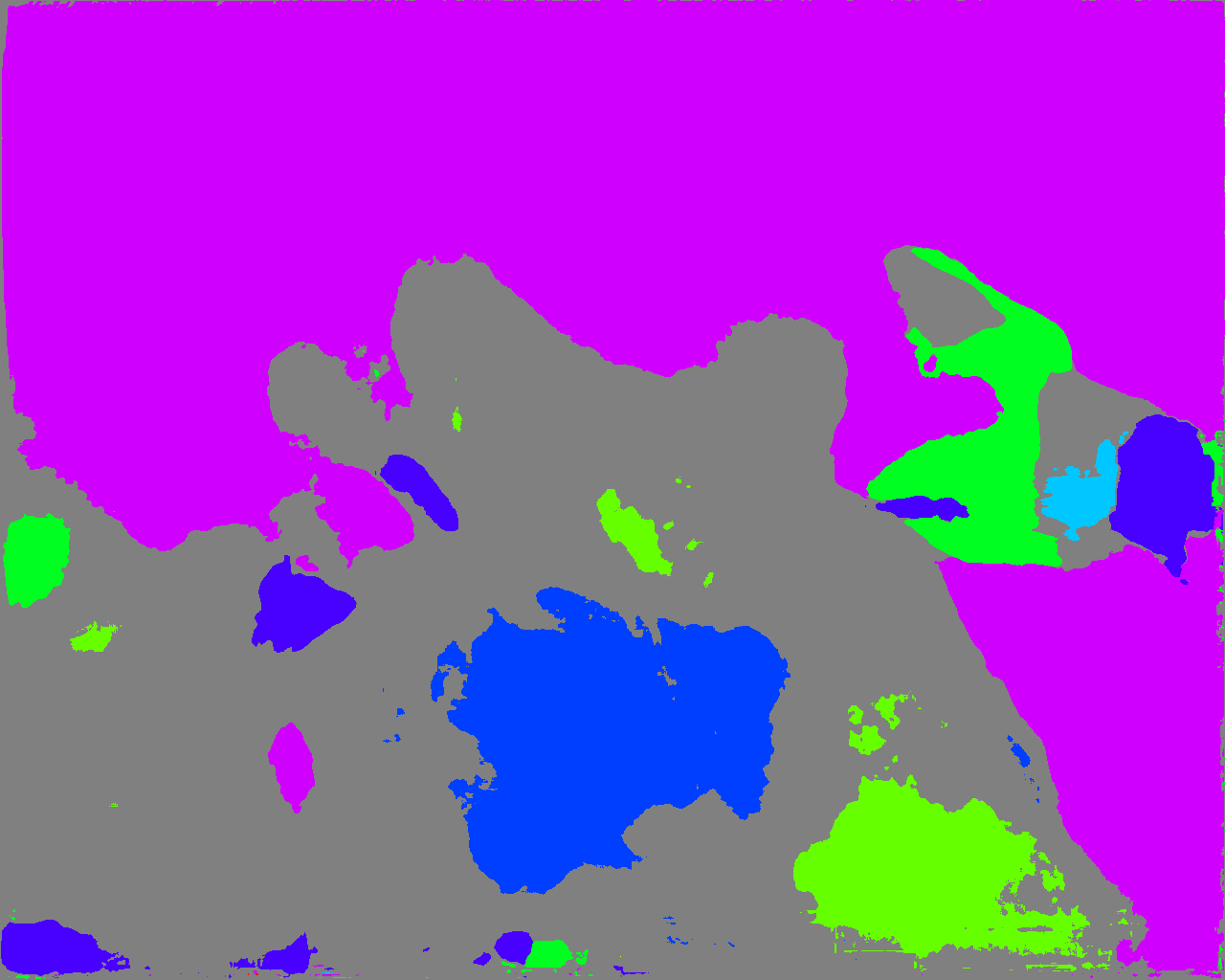}\\
\end{tabular}
\caption{Qualitative result using all labelled classes during training, i.e. with $\text{N}_{\CP}=1$. For each dataset, we show the result of same image using different methods at confidence threshold~$\tau_m$. $\tau_m$ is choose from the optimal threshold from the two-level cross-validation experiments. Baseline results at $\tau_0=0$ are added to represent result without outlier detection.}
\label{fig:qualitative_result_all_class}
\end{figure*}

\subsection{Qualitative results for all labelled classes}
\label{sec:qualitative_evaluation_all_class}
\figref{fig:qualitative_result_all_class} presents qualitative results when training for all labelled classes. For each dataset, we display the results for the same image obtained using different methods at a confidence threshold of $\tau_m$. Additionally, we include baseline results at $\tau_0=0$ to illustrate the outcome without outlier detection. We selected examples with multi-class labels across our three datasets.

For the ODSI-DB dataset, one image with multi-class annotation was randomly selected, given that many images are annotated with more than one class.
In the case of the Heiporspectral dataset, where no multi-class data is available within a single image, an experienced clinical expert in our team manually annotated an image to serve as the ground truth for this qualitative demonstration.
For the DSAD dataset, we chose an image from the additional subset created from $1430$ stomach frames that contain multi-class ground truth data.

\subsection{Qualitative results with held-out OOD classes}
\label{sec:qualitative_evaluation_ood_class}
We also considered qualitative results across all datasets using our CP hold-out approach.
To keep the main content concise for readability, we refer the reader to the appendix for these results.
\figref{fig:segmentation_result_heiporspectral_1} to \figref{fig:segmentation_result_dsad_2} show some cases with segmentation mask overlay. For each case, we show ground truth and predicted segmentation mask from one of the subject partition $\SP_k$ and all class partitions $\CP_1$ to $\CP_4$ within $\SP_k$. Different classes are held-out in each $\CP$ related folds.
Each class from the training set is held-out at least $\text{N}_{\SP}$ times.
For each $\CP$, we visualise and compare masks generated using different methods at threshold $\tau_{m}$. We also show the baseline results at $\tau_0=0$ to represent the baseline method without using our proposed framework. 

Given the sparse nature of our ground-truth annotations, qualitative evaluation reveals insights that quantitative measures alone fail to capture.
In few cases, we observed that while the model performs well on annotated pixels, including those from held-out classes, the overall quality of the segmentation mask is relatively poor. This indicates that quantitative metrics are insufficient to fully represent the model's performance on unlabelled pixels. Our findings emphasize the importance of qualitative analysis in assessing model performance on unlabelled sections.

\subsection{Cross-validation results using all labelled classes}
\label{sec:cross_valition_result_without_OOD}
We further analyse cross-validation results when training using all labelled classes. 
In this experiment, no classes are marked as OOD, therefore there is no negative data for training or evaluation.
Instead, we adopt the one-vs-rest strategy discussed in \secref{sec:evaluation_metrics} and measure $\TPR^{\OVR}$; $\TNR^{\OVR}$; $\BACC^{\OVR}$ and $\Fone^{\OVR}$ across the positive classes. In addition to the threshold-dependent metrics, we further measure the threshold-independent metrics $\AUROC^{\OVR}$ and $\AUPR^{\OVR}$.

For the threshold-dependent metrics, we report model performance with and without OOD detection. This is represented by computing metrics at two different thresholds, $\tau_0 = 0$ and $\tau_m$, respectively. $\tau_m$ is selected from optimal threshold from two-level cross-validation experiments. However, we also experimented with the ID-only threshold selection strategy described in \secref{sec:threshold_selection} for the $\OVR$ setting. We found the difference in performance between the two types of $\tau_m$ to be negligible.
Furthermore, we found there is no drop in ID performance at $\tau_m$ compare to $\tau_0$ highlighting that our OOD rejection does not aggressively mark pixels as OOD.
Since the total positive data remains the same, this indeed indicates that our framework does not compromise the detection of true positives for $\ID$ classes but moves false negatives from misclassified ID classes to the outlier class. Since there is no ground truth for the outlier class, we can discard these pixels and only account for misclassified ID classes. For brevity, these results are shown in the Appendix, \tabref{tab:cross_validation_all_classes} for the threshold-dependent metrics and \tabref{tab:cross_validation_all_classes_auc} for the threshold-independent metrics.



\section{Conclusion}
In this work, we have presented a novel framework named OOD-SEG for detecting negative/out-of-distribution (OOD) pixels while preserving multi-class positive/in-distribution (ID) classification accuracy in medical image segmentation. Our framework is based on a positive-only learning setting, which establishes distinct decision boundaries that enclose positively labelled data for each class. This approach allows for training with sparse annotations under a weakly supervised learning (WSL) setting and demonstrates robustness against various anomalies compared to methods relying on full annotations.

To assess model performance in OOD scenarios, we proposed a novel evaluation protocol based on subjects and classes, facilitating a more comprehensive assessment of OOD detection capabilities in medical imaging. Our framework and evaluation pipeline bridge the gap between OOD detection methods originally designed for image classification and their application in medical image segmentation. Extensive experiments conducted on two hyperspectral and one RGB laparoscopic imaging datasets validate the efficacy of our framework, showing improved OOD detection performance without compromising classification accuracy.

\paragraph{Limitations}
Despite the promising results demonstrated by our OOD-SEG framework, limitations need to be acknowledged to provide a comprehensive understanding of its applicability and areas for improvement.

First, our framework 
does not utilise unlabelled data during the learning process. Future work could explore the use of unlabelled data to expand the diversity of the data seen during training and improve OOD performance.

Second, unlike many OOD detection approaches used in image classification \citep{liangEnhancingReliabilityOutofdistribution2018, leeSimpleUnifiedFramework2018, hsuGeneralizedODINDetecting2020}, we did not include any adversarial perturbations into our experiments.
Although modest and coming at a high computational cost, adversarial perturbations have often demonstrated improved OOD performance. Future studies could investigate the effectiveness of adding perturbations for image segmentation tasks, building on the proposed framework.

Lastly, in our method there is an inherent trade-off between ID classification accuracy and OOD detection performance which inadvertently diminishes one aspect when another is enhanced. Such a trade-off suggests that future work could explore strategies to better balance or synergistically improve both classification accuracy and OOD detection.


\paragraph{Impact}
Our findings suggest that OOD-SEG has the potential to significantly impact downstream medical imaging applications. By enabling reliable OOD detection with sparse positive-only annotations, our framework can enhance the safety and robustness of automated segmentation systems used in clinical settings. This could reduce the risk of misclassification of unknown or anomalous tissue types. Our evaluation protocol may also serve as a benchmark for future research, promoting the development of more advanced OOD detection methods in medical image segmentation.

\section*{Acknowledgments}
TV and JS are co-founders and shareholders of Hypervision Surgical Ltd, London, UK.
The authors have no other relevant interests to declare.

This project received funding by the National Institute for Health and Care Research (NIHR) under its Invention for Innovation (i4i) Programme [NIHR202114].
The views expressed are those of the author(s) and not necessarily those of the NIHR or the Department of Health and Social Care.
This work was supported by core funding from the Wellcome/EPSRC [WT203148/Z/16/Z; NS/A000049/1].
OM is funded by the EPSRC DTP [EP/T517963/1].
For the purpose of open access, the authors have applied a CC BY public copyright license to any Author Accepted Manuscript version arising from this submission.

\bibliographystyle{model2-names.bst}\biboptions{authoryear}
\bibliography{refs}

\appendix
\clearpage
\onecolumn

\section{Additional results}
\vspace{3cm}

\begin{table}[!h]
    \centering
    \caption{Cross-validation results when using all annotated classes for training, i.e. with $\text{N}_{\CP}=1$. For each method and metric, performance is reported at thresholds $\tau_0=0$ and $\tau_m$. $\tau_m$ is choose from the optimal threshold from the two-level cross-validation experiments. Best performance among all methods at each dataset are highlighted in bold.}
    \begin{tabular}{cc cc cc cc cc}
    \toprule
    \textbf{Dataset} & \textbf{Method} & 
    \multicolumn{2}{c}{$\boldsymbol{\TPR}^{\OVR}\uparrow$} &
    \multicolumn{2}{c}{$\boldsymbol{\TNR}^{\OVR}\uparrow$} &
    \multicolumn{2}{c}{$\boldsymbol{\BACC}^{\OVR}\uparrow$} &
    \multicolumn{2}{c}{$\boldsymbol{\Fone}^{\OVR}\uparrow$} \\
    \cmidrule(lr){3-4}
    \cmidrule(lr){5-6}
    \cmidrule(lr){7-8}
    \cmidrule(lr){9-10}
    \multicolumn{2}{c}{} & $\boldsymbol{\tau_0=0}$ & $\boldsymbol{\tau_m}$
                         & $\boldsymbol{\tau_0=0}$ & $\boldsymbol{\tau_m}$
                         & $\boldsymbol{\tau_0=0}$ & $\boldsymbol{\tau_m}$
                         & $\boldsymbol{\tau_0=0}$ & $\boldsymbol{\tau_m}$
                         \\
    \midrule
    \multirow{4}{*}{Heiporspectral} & Baseline & $\boldsymbol{0.98}_{\pm0.02}$ & $\boldsymbol{0.98}_{\pm0.02}$ & $\boldsymbol{1.00}_{\pm0.00}$ & $\boldsymbol{1.00}_{\pm0.00}$ & $\boldsymbol{0.99}_{\pm0.01}$ & $\boldsymbol{0.99}_{\pm0.01}$ & $\boldsymbol{0.98}_{\pm0.02}$ & $\boldsymbol{0.98}_{\pm0.02}$ \\
 & ODIN & $\boldsymbol{0.98}_{\pm0.02}$ & $0.96_{\pm0.02}$ & $\boldsymbol{1.00}_{\pm0.00}$ & $\boldsymbol{1.00}_{\pm0.00}$ & $\boldsymbol{0.99}_{\pm0.01}$ & $0.98_{\pm0.01}$ & $\boldsymbol{0.98}_{\pm0.02}$ & $0.96_{\pm0.02}$ \\
 & Mahalanobis & $0.97_{\pm0.02}$ & $\boldsymbol{0.98}_{\pm0.02}$ & $\boldsymbol{1.00}_{\pm0.00}$ & $\boldsymbol{1.00}_{\pm0.00}$ & $\boldsymbol{0.99}_{\pm0.01}$ & $\boldsymbol{0.99}_{\pm0.01}$ & $\boldsymbol{0.98}_{\pm0.02}$ & $\boldsymbol{0.98}_{\pm0.02}$ \\
 & GODIN & $0.97_{\pm0.02}$ & $0.86_{\pm0.05}$ & $\boldsymbol{1.00}_{\pm0.00}$ & $\boldsymbol{1.00}_{\pm0.00}$ & $0.98_{\pm0.01}$ & $0.93_{\pm0.03}$ & $0.97_{\pm0.02}$ & $0.86_{\pm0.05}$ \\
    \midrule
    \multirow{4}{*}{ODSI-DB} & Baseline & $0.86_{\pm0.03}$ & $0.88_{\pm0.05}$ & $\boldsymbol{0.99}_{\pm0.00}$ & $\boldsymbol{1.00}_{\pm0.00}$ & $0.92_{\pm0.01}$ & $0.94_{\pm0.03}$ & $0.85_{\pm0.02}$ & $0.89_{\pm0.05}$ \\
 & ODIN & $0.86_{\pm0.03}$ & $0.90_{\pm0.05}$ & $\boldsymbol{0.99}_{\pm0.00}$ & $\boldsymbol{1.00}_{\pm0.00}$ & $0.92_{\pm0.01}$ & $0.95_{\pm0.03}$ & $0.85_{\pm0.02}$ & $0.91_{\pm0.05}$ \\
 & Mahalanobis & $\boldsymbol{0.87}_{\pm0.02}$ & $\boldsymbol{0.95}_{\pm0.02}$ & $\boldsymbol{0.99}_{\pm0.00}$ & $\boldsymbol{1.00}_{\pm0.00}$ & $\boldsymbol{0.93}_{\pm0.01}$ & $\boldsymbol{0.98}_{\pm0.01}$ & $0.84_{\pm0.04}$ & $\boldsymbol{0.95}_{\pm0.02}$ \\
 & GODIN & $\boldsymbol{0.87}_{\pm0.02}$ & $0.87_{\pm0.05}$ & $\boldsymbol{0.99}_{\pm0.00}$ & $\boldsymbol{1.00}_{\pm0.00}$ & $\boldsymbol{0.93}_{\pm0.01}$ & $0.93_{\pm0.03}$ & $\boldsymbol{0.86}_{\pm0.04}$ & $0.87_{\pm0.04}$ \\
    \midrule
    \multirow{4}{*}{DSAD} & Baseline & $0.80_{\pm0.04}$ & $0.88_{\pm0.07}$ & $\boldsymbol{0.99}_{\pm0.00}$ & $\boldsymbol{1.00}_{\pm0.00}$ & $\boldsymbol{0.90}_{\pm0.02}$ & $0.94_{\pm0.04}$ & $\boldsymbol{0.80}_{\pm0.07}$ & $0.88_{\pm0.07}$ \\
 & ODIN & $0.80_{\pm0.04}$ & $0.88_{\pm0.08}$ & $\boldsymbol{0.99}_{\pm0.00}$ & $\boldsymbol{1.00}_{\pm0.00}$ & $\boldsymbol{0.90}_{\pm0.02}$ & $0.94_{\pm0.04}$ & $\boldsymbol{0.80}_{\pm0.07}$ & $\boldsymbol{0.90}_{\pm0.08}$ \\
 & Mahalanobis & $\boldsymbol{0.82}_{\pm0.05}$ & $\boldsymbol{0.91}_{\pm0.04}$ & $\boldsymbol{0.99}_{\pm0.00}$ & $\boldsymbol{1.00}_{\pm0.00}$ & $\boldsymbol{0.90}_{\pm0.02}$ & $\boldsymbol{0.95}_{\pm0.02}$ & $0.78_{\pm0.08}$ & $\boldsymbol{0.90}_{\pm0.08}$ \\
 & GODIN & $0.75_{\pm0.10}$ & $0.76_{\pm0.14}$ & $\boldsymbol{0.99}_{\pm0.00}$ & $0.99_{\pm0.00}$ & $0.87_{\pm0.05}$ & $0.88_{\pm0.07}$ & $0.74_{\pm0.11}$ & $0.76_{\pm0.15}$ \\
    \bottomrule
    \end{tabular}
    \label{tab:cross_validation_all_classes}
\end{table}

\begin{table}[h]
    \centering\setlength\tabcolsep{3pt}
    \caption{Cross-validation results when using all annotated classes for training, i.e. with $\text{N}_{\CP}=1$. For each method and metric, average performance among classes is reported with standard deviation. Best performance among all methods at each dataset are highlighted in bold. Some Mahalanobis distance results are not shown (\NA) due to stability issue.}
    \begin{tabular}{cc ccccc ccccc}
    \toprule
    \textbf{Dataset} & \textbf{Method} & 
    \multicolumn{5}{c}{$\boldsymbol{\AUROC}^{\OVR}\uparrow$} &
    \multicolumn{5}{c}{$\boldsymbol{\AUPR}^{\OVR}\uparrow$} \\
    \cmidrule(lr){3-7}
    \cmidrule(lr){8-12}
    \multicolumn{2}{c}{} & \boldmath$\SP_1$ & \boldmath$\SP_2$ & \boldmath$\SP_3$ & \boldmath$\SP_4$ & \boldmath$\SP_{\text{mean}}$
                         & \boldmath$\SP_1$ & \boldmath$\SP_2$ & \boldmath$\SP_3$ & \boldmath$\SP_4$ & \boldmath$\SP_{\text{mean}}$ \\
    \midrule
    \multirow{4}{*}{Heiporspectral} & Baseline & $\boldsymbol{0.99}_{\pm0.04}$ & $\boldsymbol{0.97}_{\pm0.11}$ & $\boldsymbol{1.00}_{\pm0.00}$ & $\boldsymbol{0.99}_{\pm0.03}$ & $0.98_{\pm0.01}$ & $0.84_{\pm0.14}$ & $0.87_{\pm0.11}$ & $0.86_{\pm0.10}$ & $0.82_{\pm0.12}$ & $0.85_{\pm0.02}$ \\
 & ODIN & $\boldsymbol{0.99}_{\pm0.04}$ & $\boldsymbol{0.97}_{\pm0.11}$ & $\boldsymbol{1.00}_{\pm0.00}$ & $\boldsymbol{0.99}_{\pm0.03}$ & $\boldsymbol{0.99}_{\pm0.01}$ & $\boldsymbol{0.99}_{\pm0.04}$ & $\boldsymbol{0.97}_{\pm0.11}$ & $\boldsymbol{1.00}_{\pm0.00}$ & $\boldsymbol{0.99}_{\pm0.03}$ & $\boldsymbol{0.99}_{\pm0.01}$ \\
 & Mahalanobis & $0.96_{\pm0.04}$ & $0.95_{\pm0.11}$ & $0.97_{\pm0.00}$ & $0.96_{\pm0.02}$ & $0.96_{\pm0.01}$ & \NA & \NA & \NA &\NA & \NA \\
 & GODIN & $\boldsymbol{0.99}_{\pm0.04}$ & $\boldsymbol{0.97}_{\pm0.11}$ & $\boldsymbol{1.00}_{\pm0.00}$ & $0.98_{\pm0.06}$ & $0.98_{\pm0.01}$ & $\boldsymbol{0.99}_{\pm0.04}$ & $\boldsymbol{0.97}_{\pm0.11}$ & $\boldsymbol{1.00}_{\pm0.00}$ & $0.97_{\pm0.07}$ & $0.98_{\pm0.01}$ \\
    \midrule
    \multirow{4}{*}{ODSI-DB} & Baseline & $\boldsymbol{0.94}_{\pm0.05}$ & $0.93_{\pm0.09}$ & $0.91_{\pm0.12}$ & $\boldsymbol{0.92}_{\pm0.09}$ & $\boldsymbol{0.93}_{\pm0.01}$ & $\boldsymbol{0.91}_{\pm0.07}$ & $\boldsymbol{0.91}_{\pm0.11}$ & $\boldsymbol{0.90}_{\pm0.12}$ & $0.85_{\pm0.21}$ & $\boldsymbol{0.89}_{\pm0.02}$ \\
 & ODIN & $\boldsymbol{0.94}_{\pm0.05}$ & $0.93_{\pm0.09}$ & $0.91_{\pm0.12}$ & $\boldsymbol{0.92}_{\pm0.09}$ & $\boldsymbol{0.93}_{\pm0.01}$ & $0.90_{\pm0.07}$ & $\boldsymbol{0.91}_{\pm0.11}$ & $0.89_{\pm0.12}$ & $0.84_{\pm0.23}$ & $\boldsymbol{0.89}_{\pm0.02}$ \\
 & Mahalanobis & $0.92_{\pm0.05}$ & $0.91_{\pm0.08}$ & $0.92_{\pm0.06}$ & $0.89_{\pm0.07}$ & $0.91_{\pm0.01}$ & \NA & \NA & \NA & \NA & \NA \\
 & GODIN & $\boldsymbol{0.94}_{\pm0.05}$ & $\boldsymbol{0.94}_{\pm0.05}$ & $\boldsymbol{0.93}_{\pm0.08}$ & $\boldsymbol{0.92}_{\pm0.09}$ & $\boldsymbol{0.93}_{\pm0.01}$ & $0.90_{\pm0.09}$ & $\boldsymbol{0.91}_{\pm0.11}$ & $\boldsymbol{0.90}_{\pm0.12}$ & $\boldsymbol{0.86}_{\pm0.15}$ & $\boldsymbol{0.89}_{\pm0.02}$ \\
    \midrule
    \multirow{4}{*}{DSAD} & Baseline & $\boldsymbol{0.93}_{\pm0.05}$ & $\boldsymbol{0.91}_{\pm0.06}$ & $\boldsymbol{0.89}_{\pm0.08}$ & $\boldsymbol{0.88}_{\pm0.10}$ & $\boldsymbol{0.90}_{\pm0.02}$ & $\boldsymbol{0.90}_{\pm0.07}$ & $\boldsymbol{0.86}_{\pm0.11}$ & $\boldsymbol{0.82}_{\pm0.13}$ & $\boldsymbol{0.78}_{\pm0.16}$ & $\boldsymbol{0.84}_{\pm0.04}$ \\
 & ODIN & $\boldsymbol{0.93}_{\pm0.05}$ & $\boldsymbol{0.91}_{\pm0.06}$ & $\boldsymbol{0.89}_{\pm0.08}$ & $\boldsymbol{0.88}_{\pm0.10}$ & $\boldsymbol{0.90}_{\pm0.02}$ & $\boldsymbol{0.90}_{\pm0.07}$ & $0.85_{\pm0.11}$ & $\boldsymbol{0.82}_{\pm0.13}$ & $\boldsymbol{0.78}_{\pm0.16}$ & $\boldsymbol{0.84}_{\pm0.04}$ \\
 & Mahalanobis & $0.89_{\pm0.04}$ & $0.87_{\pm0.06}$ & $0.86_{\pm0.07}$ & $0.84_{\pm0.09}$ & $0.87_{\pm0.02}$ & \NA & \NA & \NA & \NA & \NA \\
 & GODIN & $\boldsymbol{0.93}_{\pm0.05}$ & $0.90_{\pm0.09}$ & $0.81_{\pm0.13}$ & $0.84_{\pm0.14}$ & $0.87_{\pm0.05}$ & $\boldsymbol{0.90}_{\pm0.07}$ & $0.80_{\pm0.22}$ & $0.63_{\pm0.25}$ & $0.77_{\pm0.18}$ & $0.77_{\pm0.10}$ \\
    \bottomrule
    \end{tabular}
    \label{tab:cross_validation_all_classes_auc}
\end{table}

\begin{table}[htb]
    \centering
    \caption{Subject-level cross-validation results on the ODSI-DB dataset. The model performance on background/OOD classes is evaluated with the 26 under-represented classes being held out during training.
    \label{tab:revision_odsi}}
    \begin{tabular}{cccc}
    \toprule
    \textbf{Dataset} & \textbf{Method} & 
    $\boldsymbol{\AUROC}\uparrow$ &
    $\boldsymbol{\AUPR}\uparrow$ \\
    \midrule
    \multirow{4}{*}{ODSI-DB} & Baseline & $0.60_{\pm0.02}$ & $0.42_{\pm0.02}$ \\
     & ODIN & $\boldsymbol{0.61}_{\pm0.03}$ & $\boldsymbol{0.68}_{\pm0.05}$ \\
     & Mahalanobis & $0.55_{\pm0.04}$ & $0.51_{\pm0.05}$ \\
     & GODIN & $0.56_{\pm0.04}$ & $0.65_{\pm0.07}$ \\
    \bottomrule
    \end{tabular}
\end{table}

\FloatBarrier

\begin{figure}[p]
\centering\footnotesize
\begin{tabular}{ccccc}
& \boldmath$\CP_1$ & \boldmath$\CP_2$ & \boldmath$\CP_3$ & \boldmath$\CP_4$ \\
\vspace{5pt}
\makecell[c] {\textbf{Sparsely} \\ \textbf{annotated} \\ \textbf{ground truth}} &  
\includegraphics[width=.2\linewidth,valign=m]{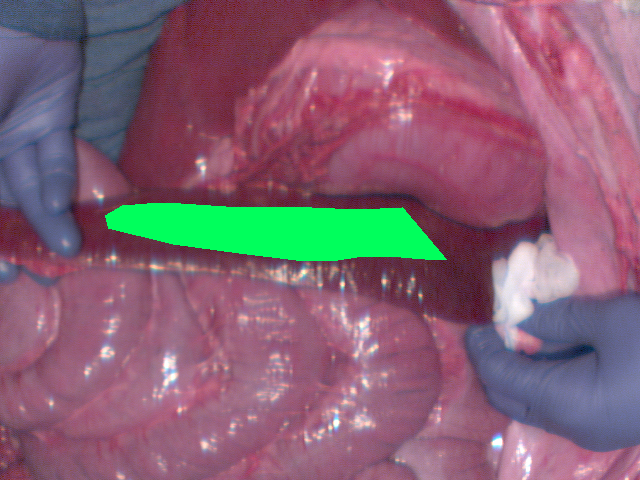} &
\includegraphics[width=.2\linewidth,valign=m]{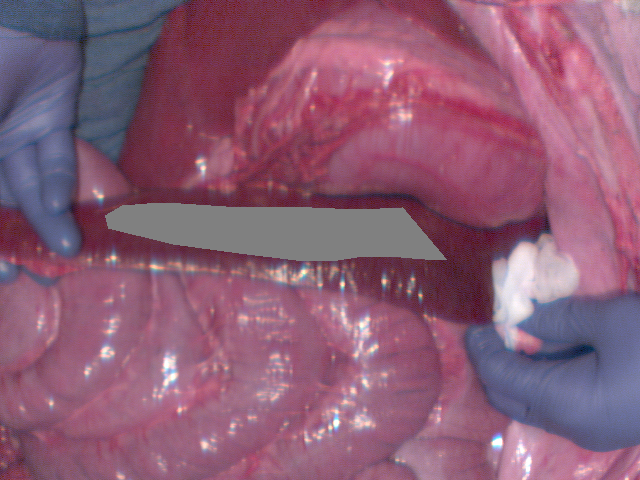} &
\includegraphics[width=.2\linewidth,valign=m]{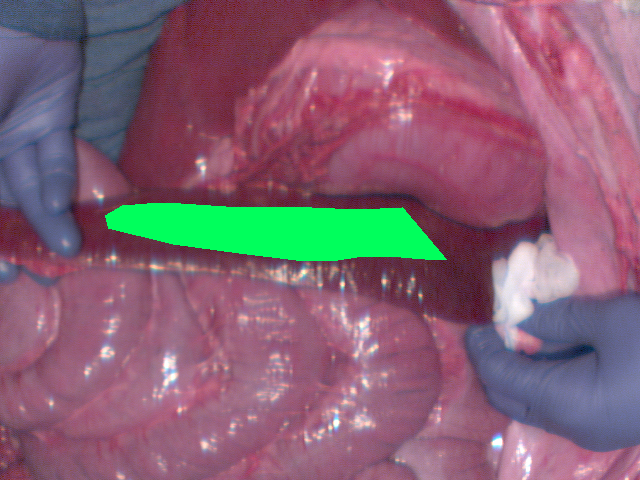} &
\includegraphics[width=.2\linewidth,valign=m]{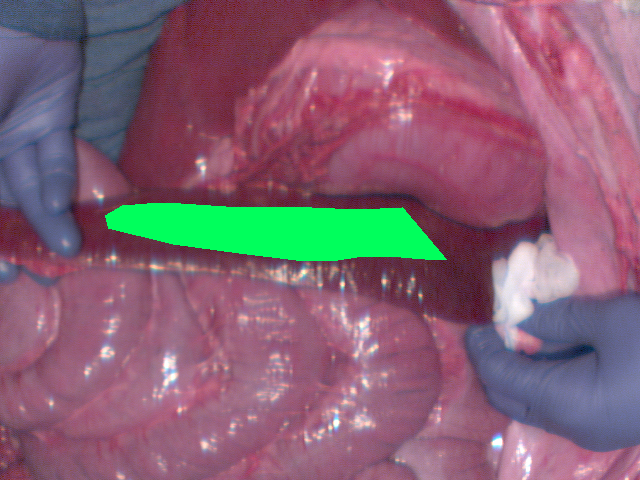}\\
\vspace{5pt}
\textbf{Baseline\ $(\tau_{0}=0$)} & 
\includegraphics[width=.2\linewidth,valign=m]{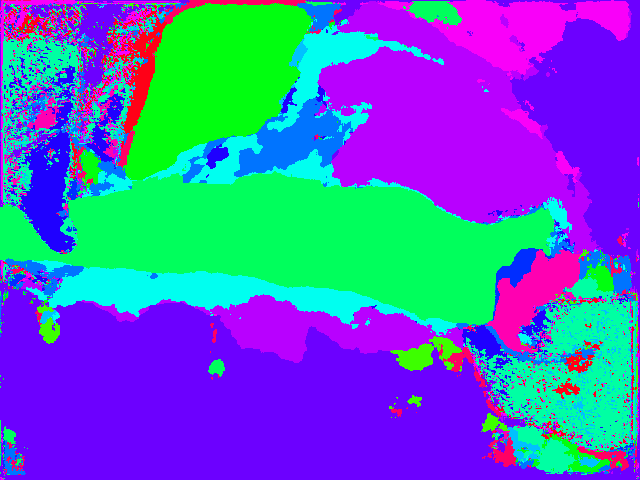} &
\includegraphics[width=.2\linewidth,valign=m]{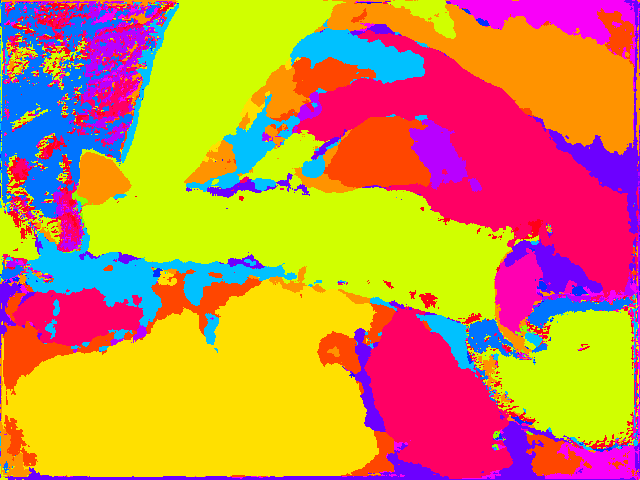} &
\includegraphics[width=.2\linewidth,valign=m]{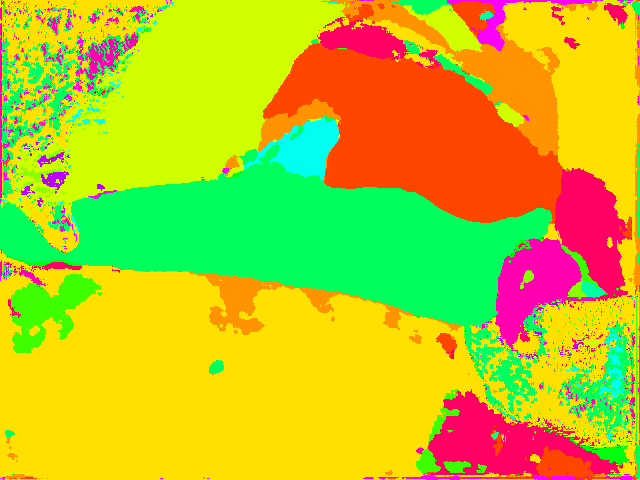} &
\includegraphics[width=.2\linewidth,valign=m]{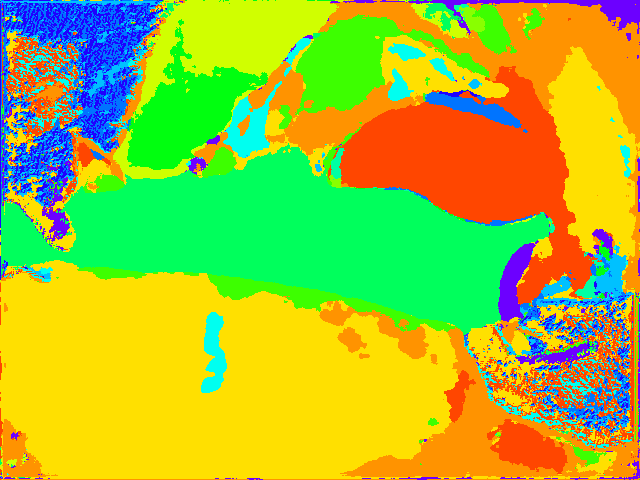}\\
\vspace{5pt}
\textbf{Baseline\ $(\tau_{m}$)} & 
\includegraphics[width=.2\linewidth,valign=m]{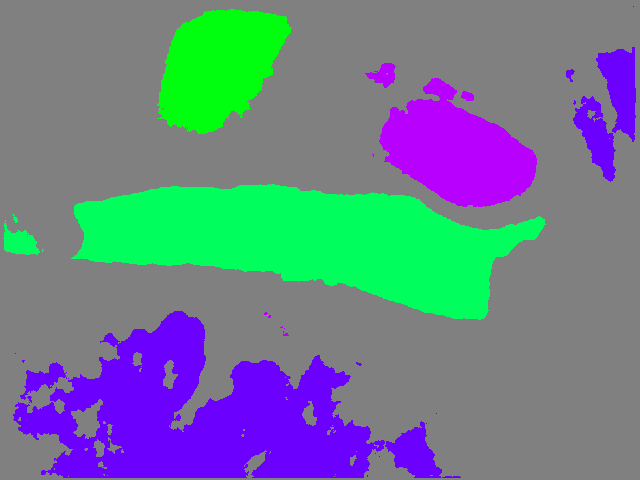} &
\includegraphics[width=.2\linewidth,valign=m]{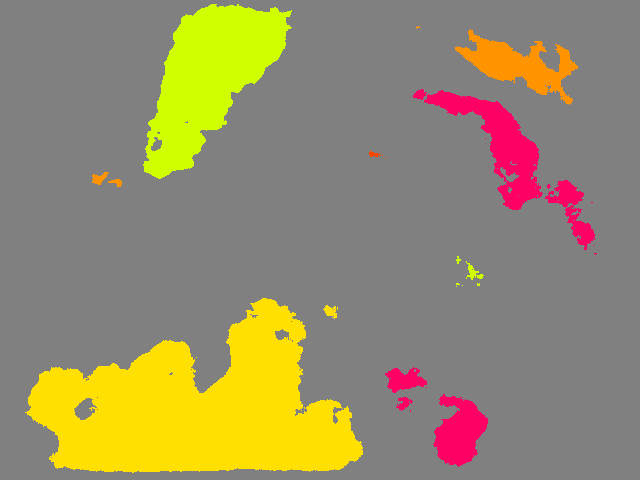} &
\includegraphics[width=.2\linewidth,valign=m]{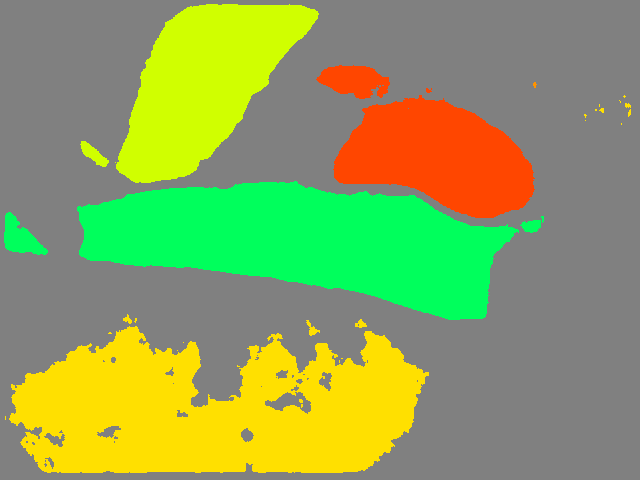} &
\includegraphics[width=.2\linewidth,valign=m]{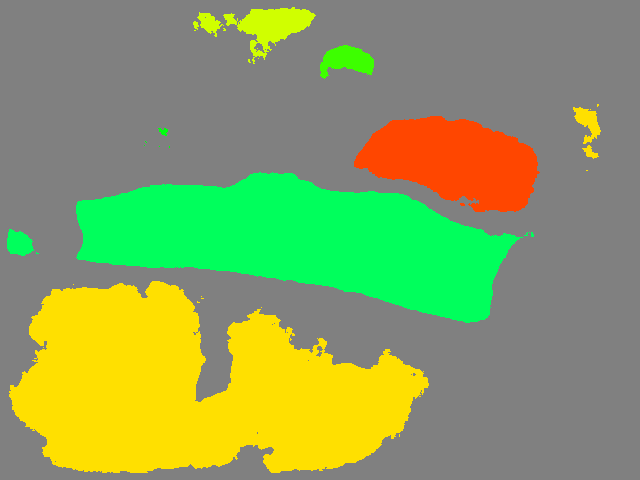}\\
\vspace{5pt}
\textbf{ODIN} & 
\includegraphics[width=.2\linewidth,valign=m]{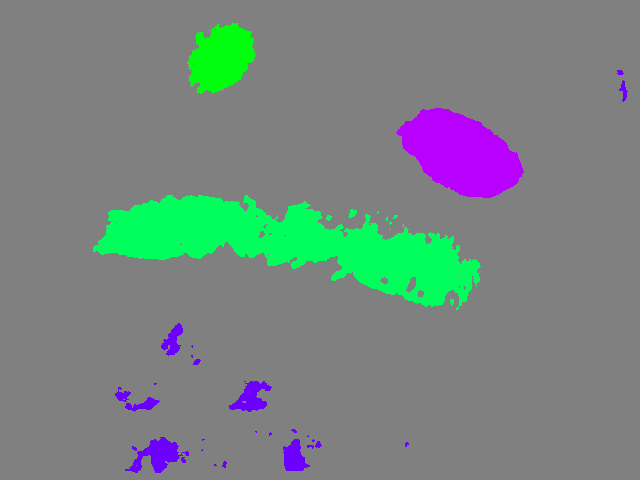} &
\includegraphics[width=.2\linewidth,valign=m]{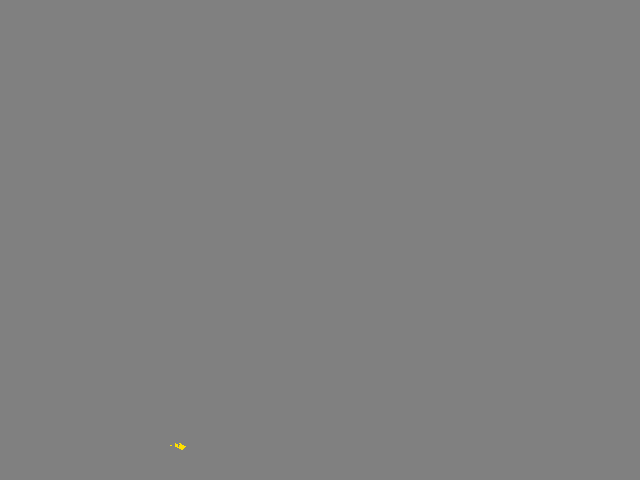} &
\includegraphics[width=.2\linewidth,valign=m]{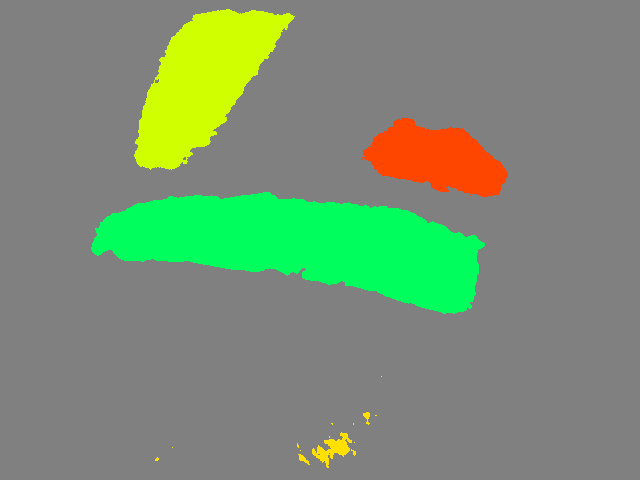} &
\includegraphics[width=.2\linewidth,valign=m]{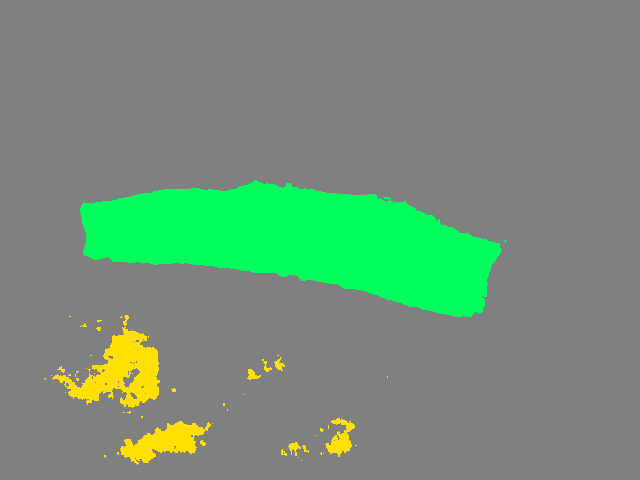}\\
\vspace{5pt}
\textbf{Mahalanobis} & 
\includegraphics[width=.2\linewidth,valign=m]{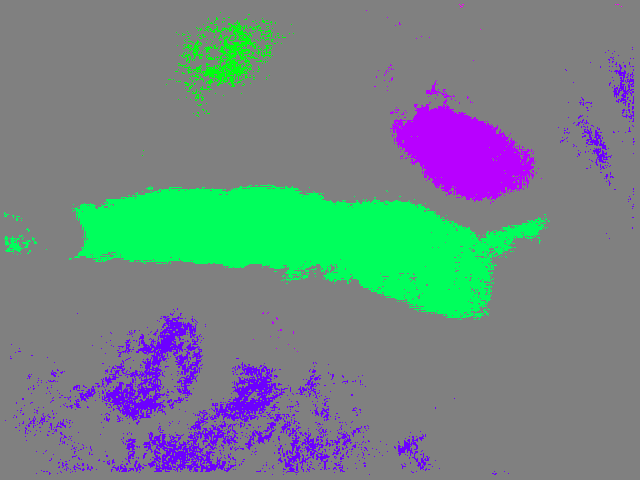} &
\includegraphics[width=.2\linewidth,valign=m]{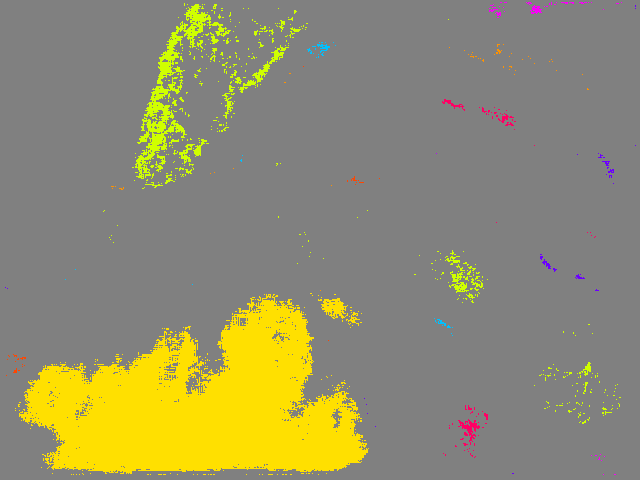} &
\includegraphics[width=.2\linewidth,valign=m]{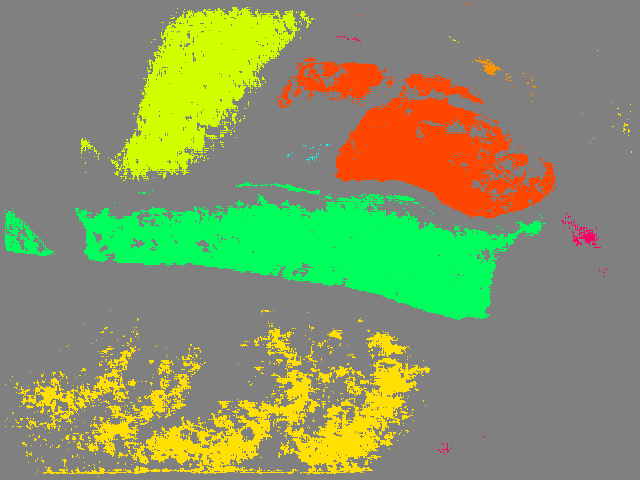} &
\includegraphics[width=.2\linewidth,valign=m]{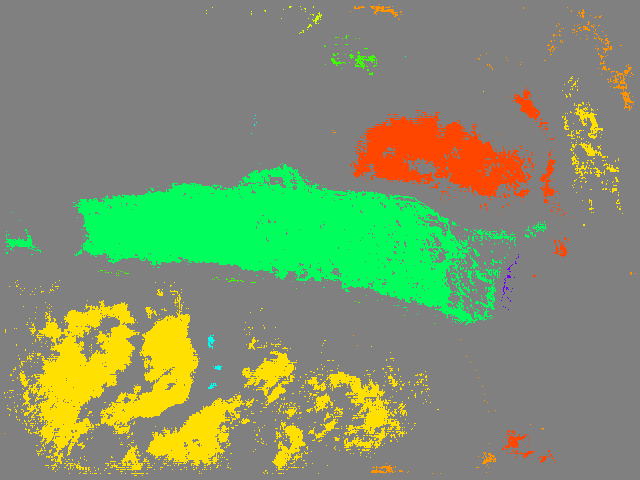}\\
\vspace{5pt}
\textbf{GODIN} & 
\includegraphics[width=.2\linewidth,valign=m]{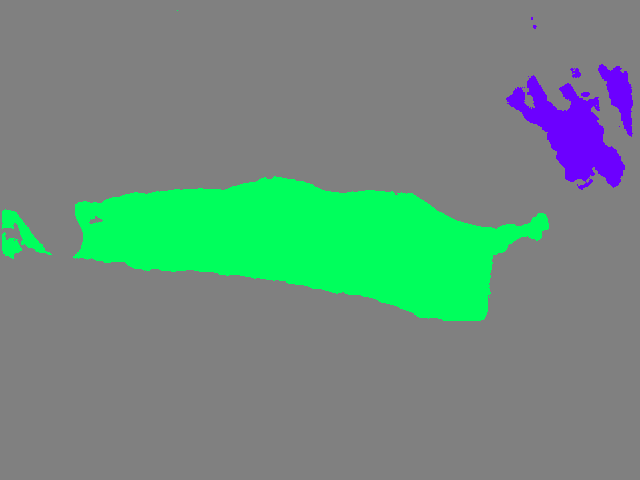} &
\includegraphics[width=.2\linewidth,valign=m]{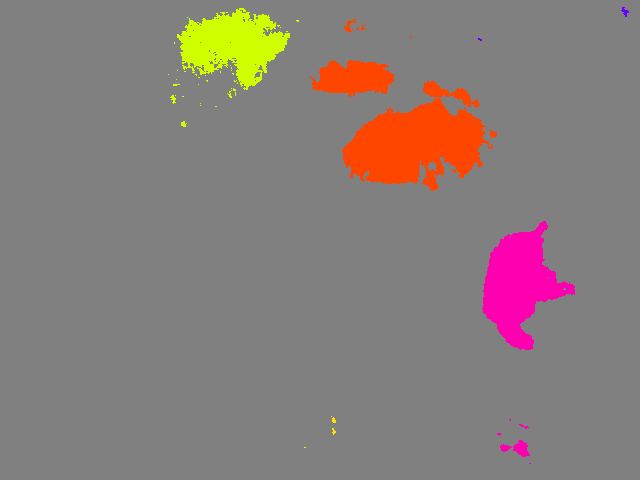} &
\includegraphics[width=.2\linewidth,valign=m]{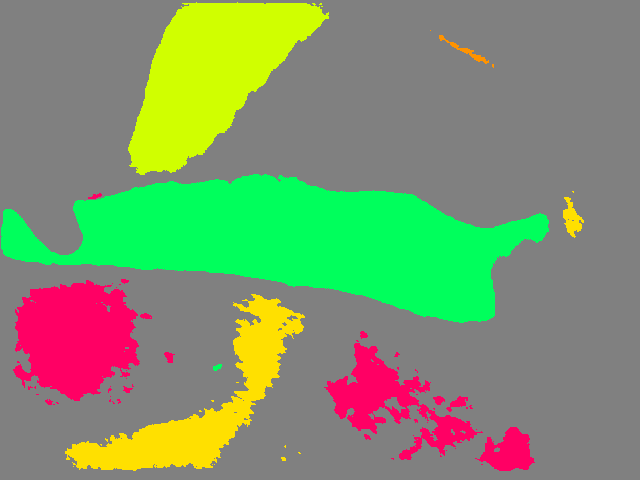} &
\includegraphics[width=.2\linewidth,valign=m]{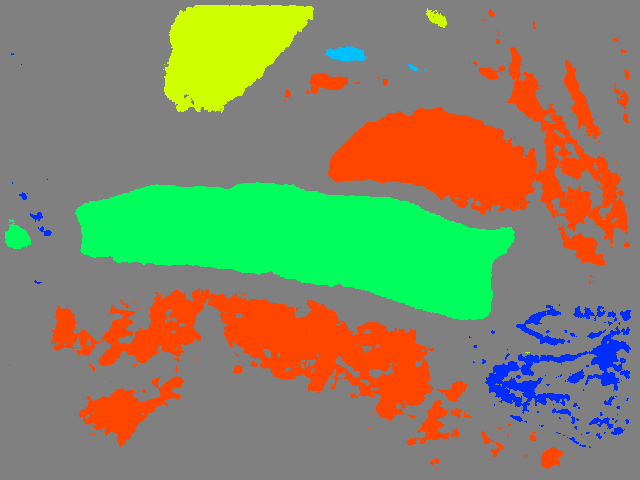}\\
 &
\multicolumn{4}{c}{\includegraphics[width=16.1cm, height=0.5cm]{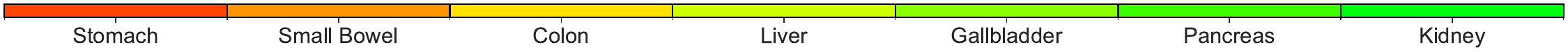}} \\
 &
\multicolumn{4}{c}{\includegraphics[width=16.1cm, height=0.5cm]{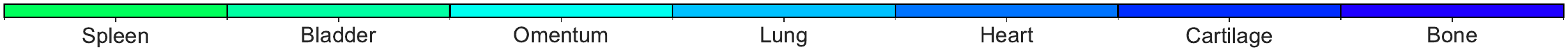}} \\
 &
\multicolumn{4}{c}{\includegraphics[width=16.1cm, height=0.5cm]{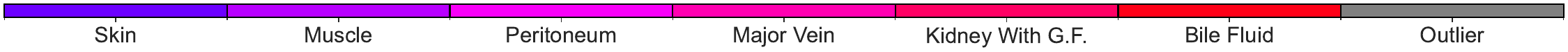}}
\end{tabular}
\caption{Qualitative result of first case from Heiporspectral dataset. We show results of the same image from four class partitions (represented as $\CP_1$ to $\CP_4$). For each $\CP$, classes that are held-out are grouped as an extra outlier class for evaluation. We visualise and compare masks generated using different methods at threshold $\tau_{m}$. Baseline results at $\tau_0=0$ are added to represent result without outlier detection.}
\label{fig:segmentation_result_heiporspectral_1}
\end{figure}

\begin{figure}[p]
\centering\footnotesize
\begin{tabular}{ccccc}
& \boldmath$\CP_1$ & \boldmath$\CP_2$ & \boldmath$\CP_3$ & \boldmath$\CP_4$ \\
\vspace{5pt}
\makecell[c] {\textbf{Sparsely} \\ \textbf{annotated} \\ \textbf{ground truth}} &  
\includegraphics[width=.2\linewidth,valign=m]{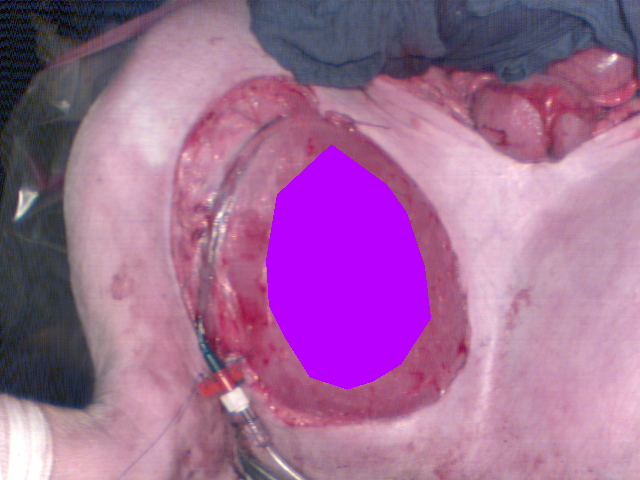} &
\includegraphics[width=.2\linewidth,valign=m]{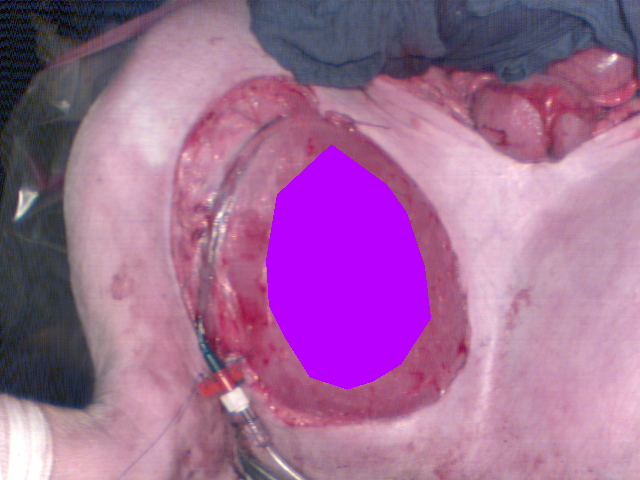} &
\includegraphics[width=.2\linewidth,valign=m]{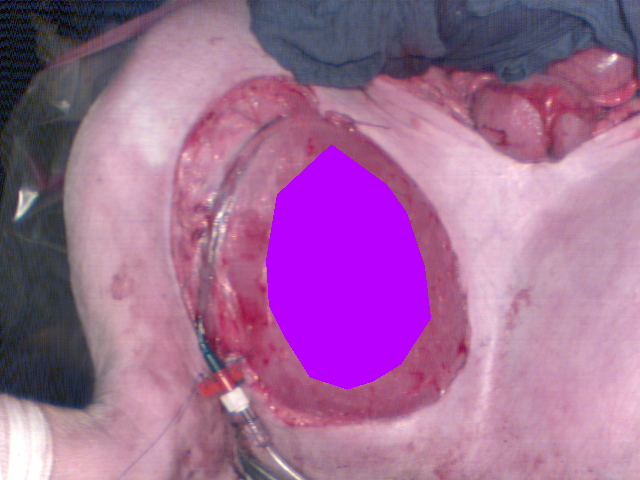} &
\includegraphics[width=.2\linewidth,valign=m]{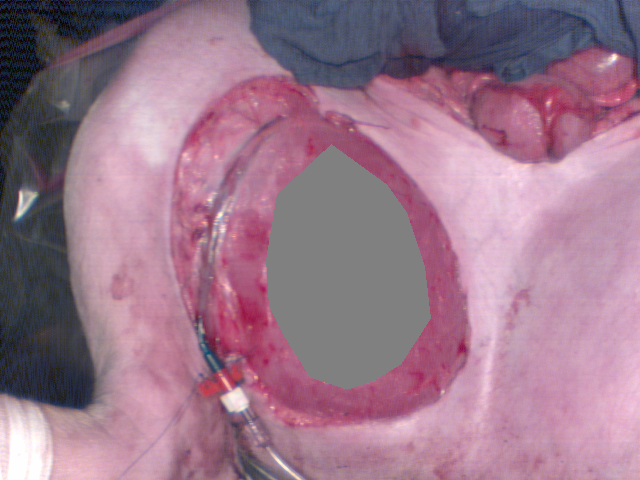}\\
\vspace{5pt}
\textbf{Baseline\ $(\tau_{0}=0$)} & 
\includegraphics[width=.2\linewidth,valign=m]{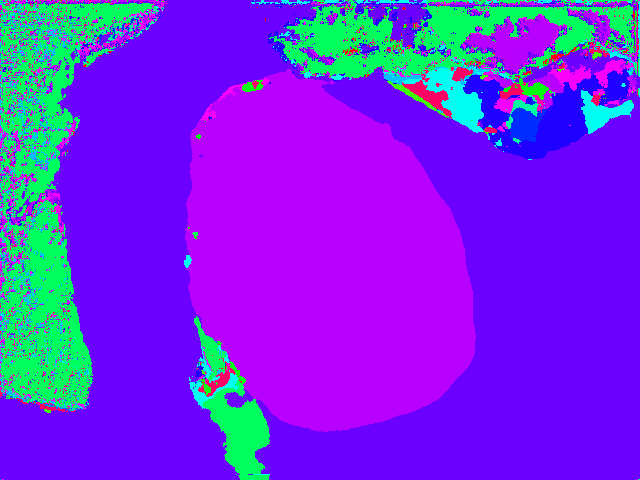} &
\includegraphics[width=.2\linewidth,valign=m]{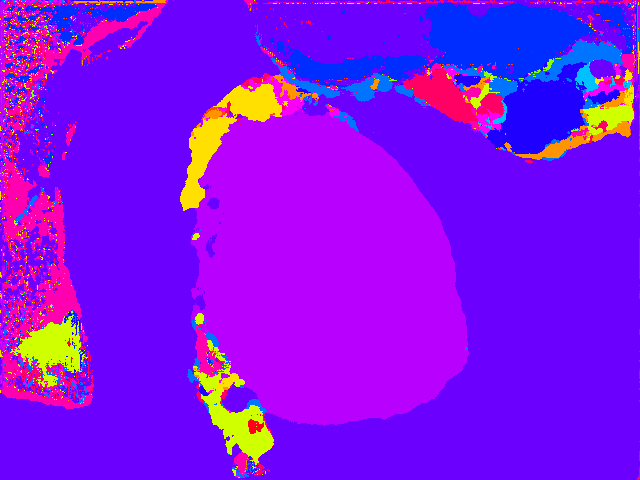} &
\includegraphics[width=.2\linewidth,valign=m]{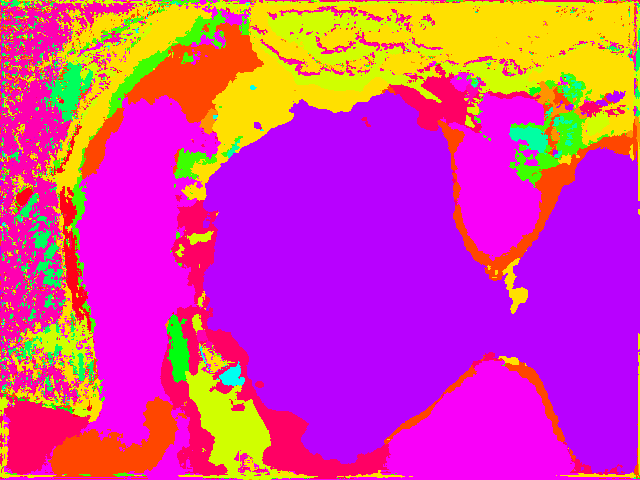} &
\includegraphics[width=.2\linewidth,valign=m]{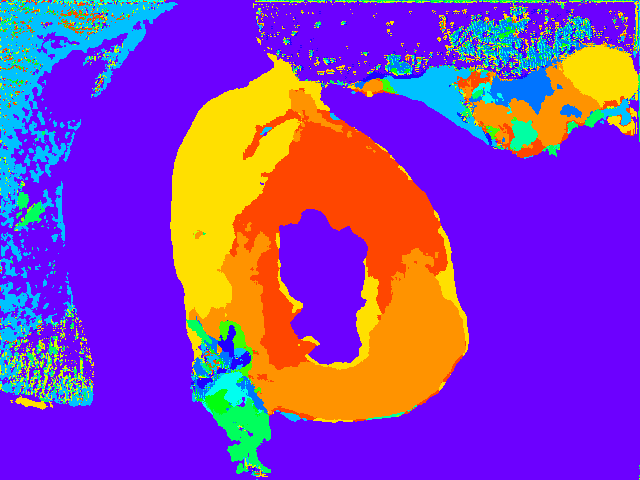}\\
\vspace{5pt}
\textbf{Baseline\ $(\tau_{m}$)} & 
\includegraphics[width=.2\linewidth,valign=m]{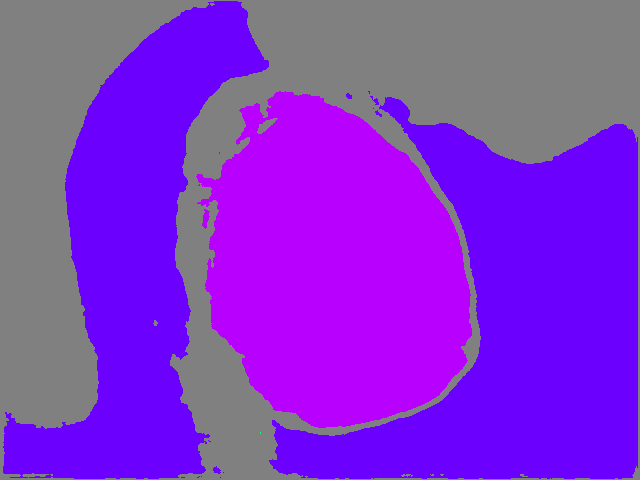} &
\includegraphics[width=.2\linewidth,valign=m]{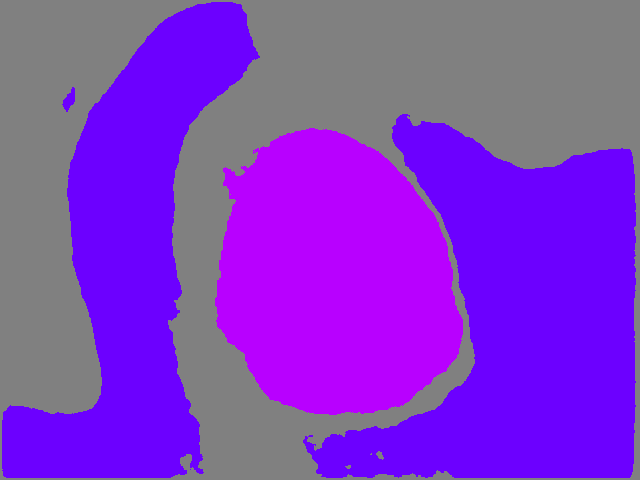} &
\includegraphics[width=.2\linewidth,valign=m]{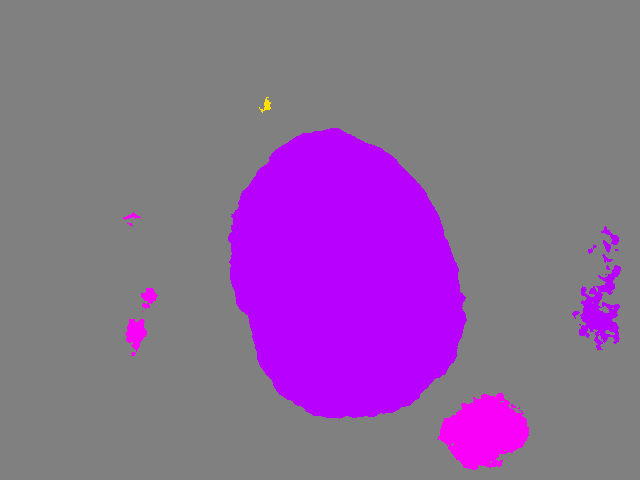} &
\includegraphics[width=.2\linewidth,valign=m]{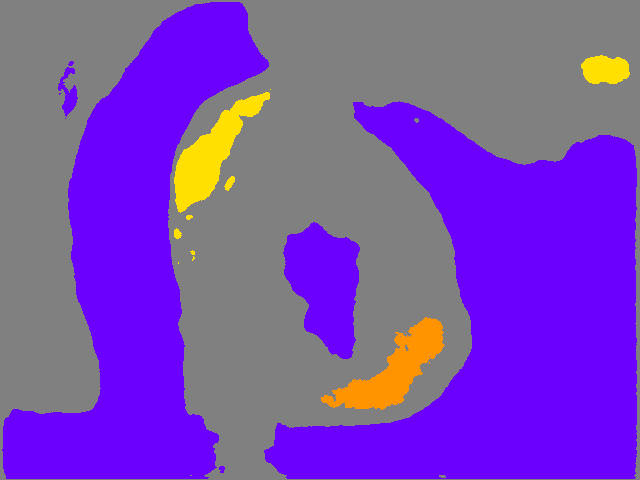}\\
\vspace{5pt}
\textbf{ODIN} & 
\includegraphics[width=.2\linewidth,valign=m]{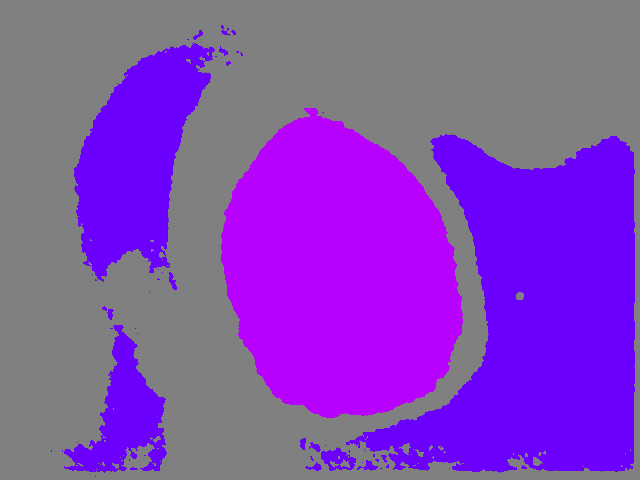} &
\includegraphics[width=.2\linewidth,valign=m]{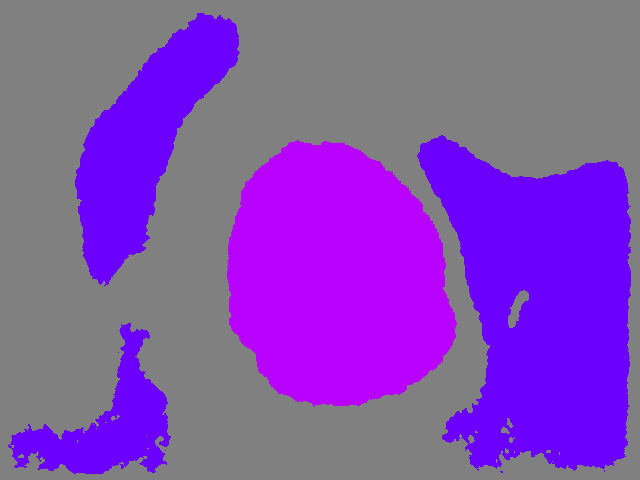} &
\includegraphics[width=.2\linewidth,valign=m]{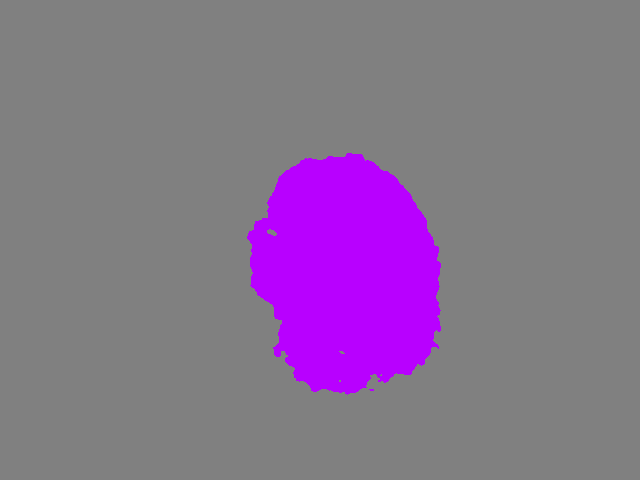} &
\includegraphics[width=.2\linewidth,valign=m]{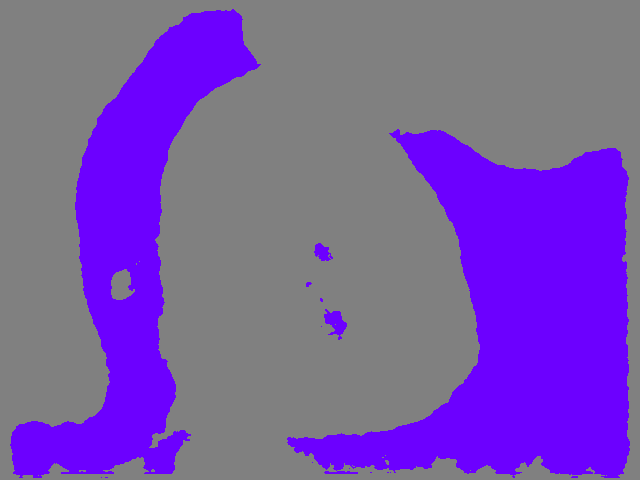}\\
\vspace{5pt}
\textbf{Mahalanobis} & 
\includegraphics[width=.2\linewidth,valign=m]{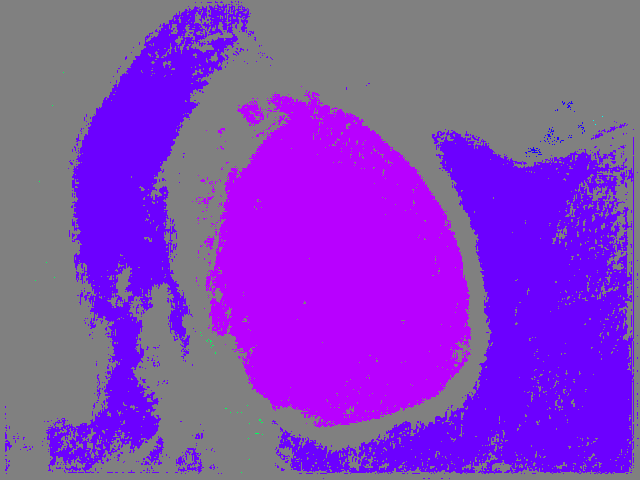} &
\includegraphics[width=.2\linewidth,valign=m]{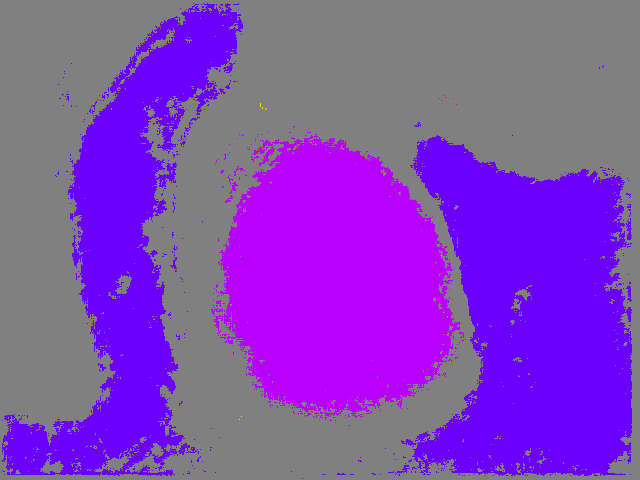} &
\includegraphics[width=.2\linewidth,valign=m]{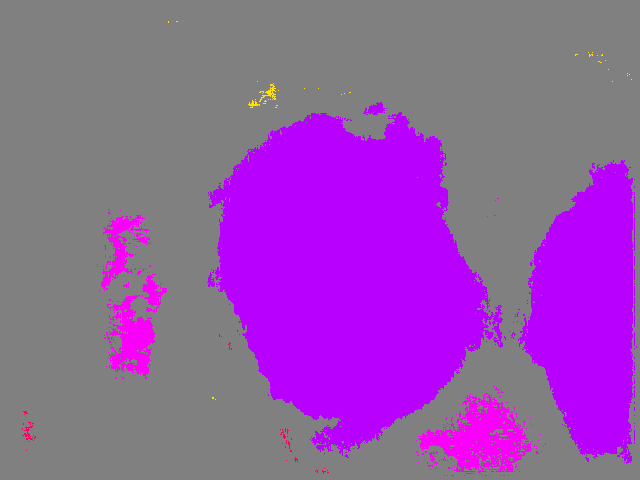} &
\includegraphics[width=.2\linewidth,valign=m]{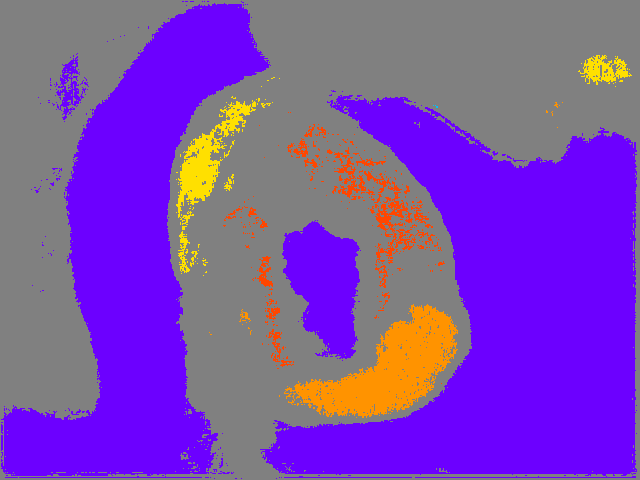}\\
\vspace{5pt}
\textbf{GODIN} & 
\includegraphics[width=.2\linewidth,valign=m]{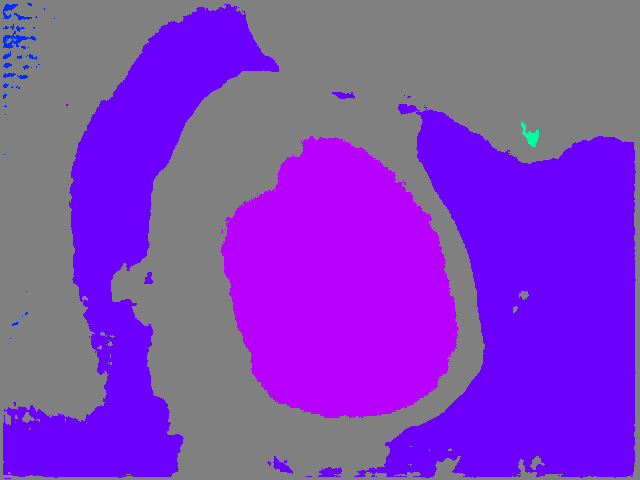} &
\includegraphics[width=.2\linewidth,valign=m]{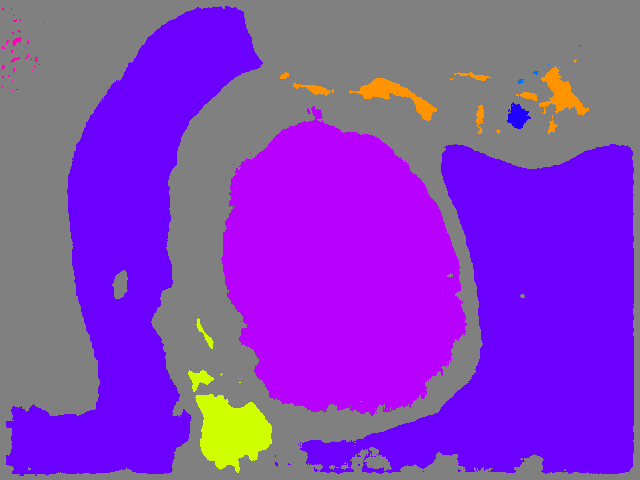} &
\includegraphics[width=.2\linewidth,valign=m]{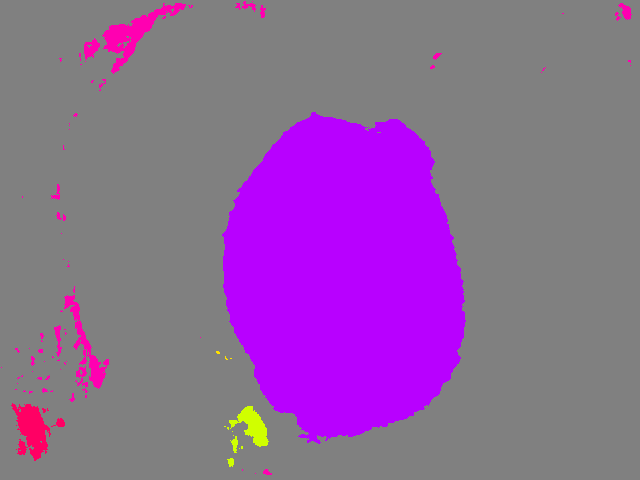} &
\includegraphics[width=.2\linewidth,valign=m]{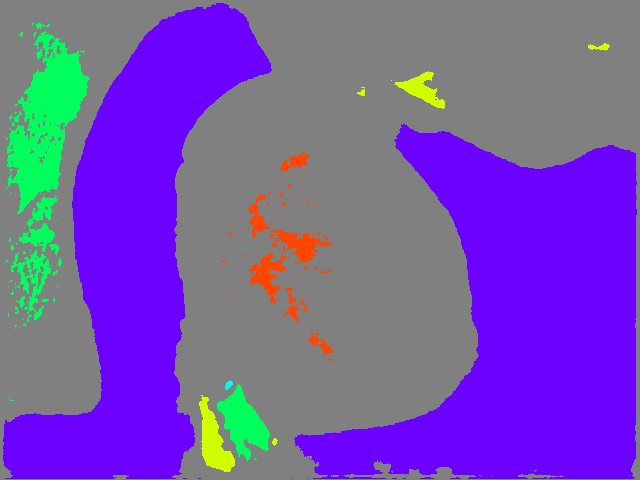}\\
 &
\multicolumn{4}{c}{\includegraphics[width=16.1cm, height=0.5cm]{figures/colorbar_heiporspectral_0.pdf}} \\
 &
\multicolumn{4}{c}{\includegraphics[width=16.1cm, height=0.5cm]{figures/colorbar_heiporspectral_1.pdf}} \\
 &
\multicolumn{4}{c}{\includegraphics[width=16.1cm, height=0.5cm]{figures/colorbar_heiporspectral_2.pdf}}
\end{tabular}
\caption{Qualitative result of second case from Heiporspectral dataset. We show results of the same image from four class partitions (represented as $\CP_1$ to $\CP_4$). For each $\CP$, classes that are held-out are grouped as an extra outlier class for evaluation. We visualise and compare masks generated using different methods at threshold $\tau_{m}$. Baseline results at $\tau_0=0$ are added to represent result without outlier detection.}
\label{fig:segmentation_result_heiporspectral_2}
\end{figure}

\begin{figure}[p]
\centering\footnotesize
\begin{tabular}{ccccc}
& \boldmath$\CP_1$ & \boldmath$\CP_2$ & \boldmath$\CP_3$ & \boldmath$\CP_4$ \\
\vspace{5pt}
\makecell[c] {\textbf{Sparsely} \\ \textbf{annotated} \\ \textbf{ground truth}} &  
\includegraphics[width=.2\linewidth,height=.15\linewidth,valign=m]{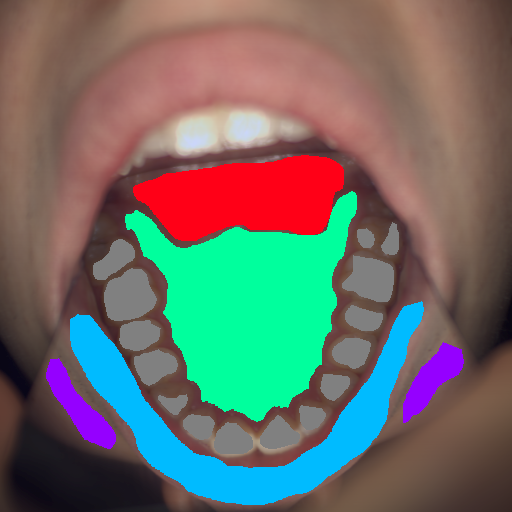} &
\includegraphics[width=.2\linewidth,height=.15\linewidth,valign=m]{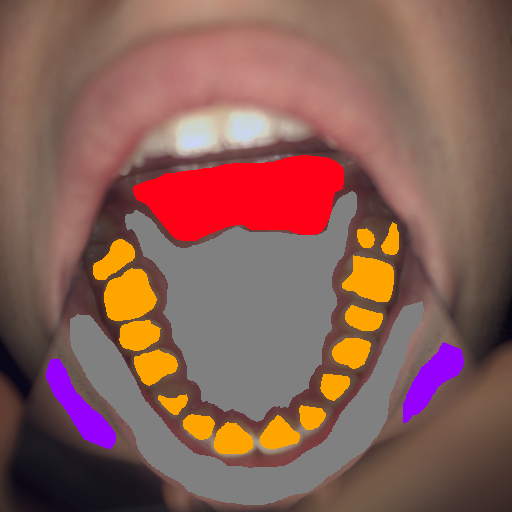} &
\includegraphics[width=.2\linewidth,height=.15\linewidth,valign=m]{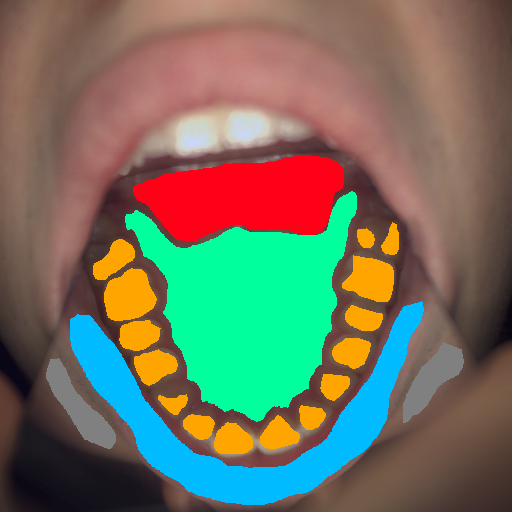} &
\includegraphics[width=.2\linewidth,height=.15\linewidth,valign=m]{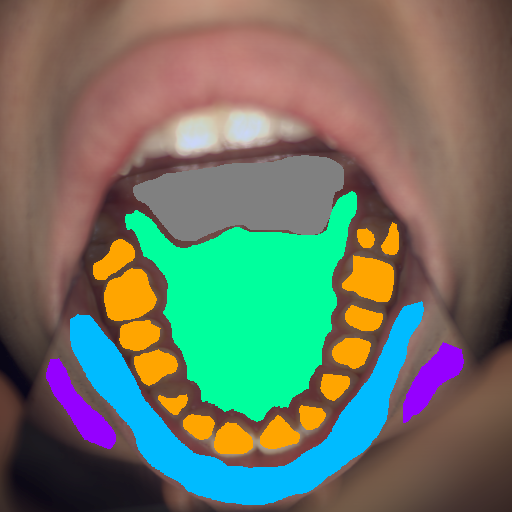}\\
\vspace{5pt}
\textbf{Baseline\ $(\tau_{0}=0$)} & 
\includegraphics[width=.2\linewidth,height=.15\linewidth,valign=m]{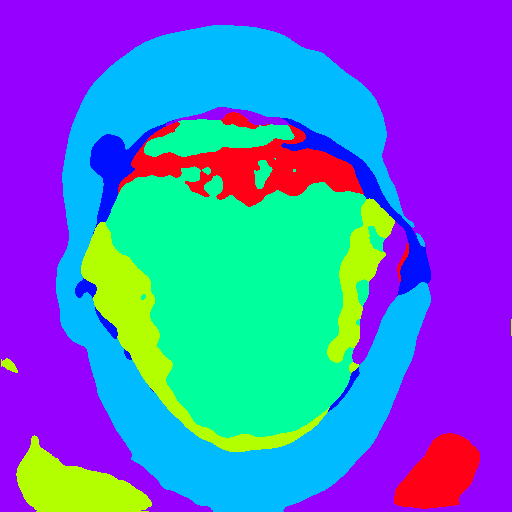} &
\includegraphics[width=.2\linewidth,height=.15\linewidth,valign=m]{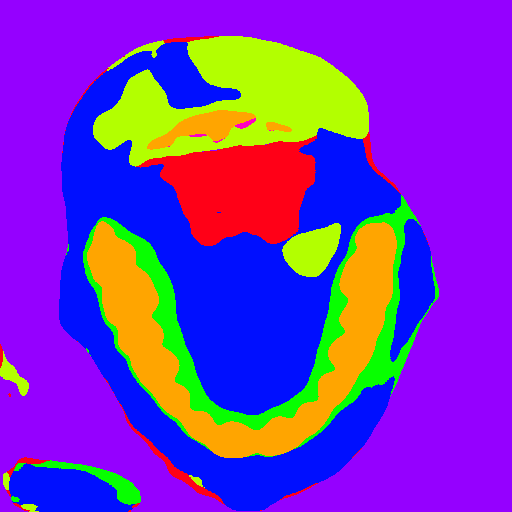} &
\includegraphics[width=.2\linewidth,height=.15\linewidth,valign=m]{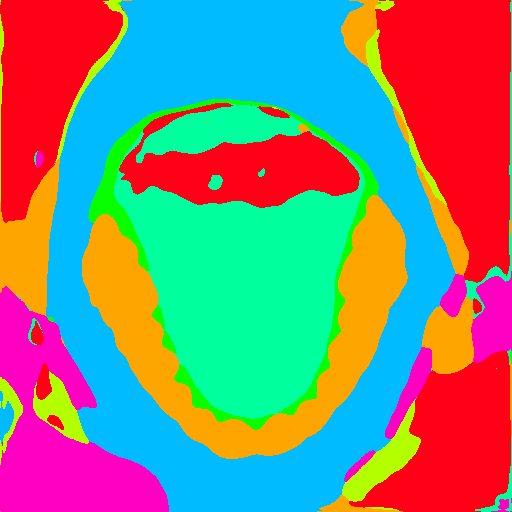} &
\includegraphics[width=.2\linewidth,height=.15\linewidth,valign=m]{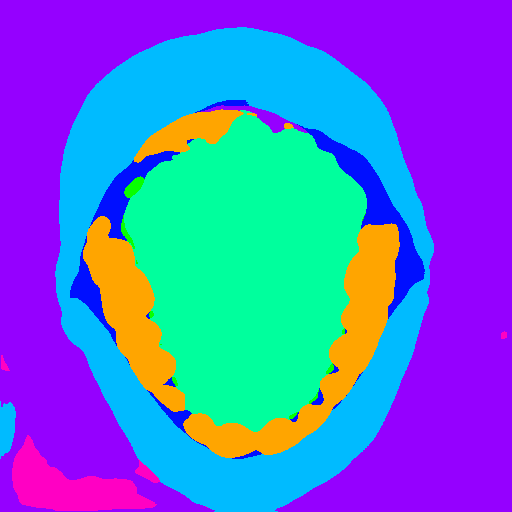}\\
\vspace{5pt}
\textbf{Baseline\ $(\tau_{m}$)} & 
\includegraphics[width=.2\linewidth,height=.15\linewidth,valign=m]{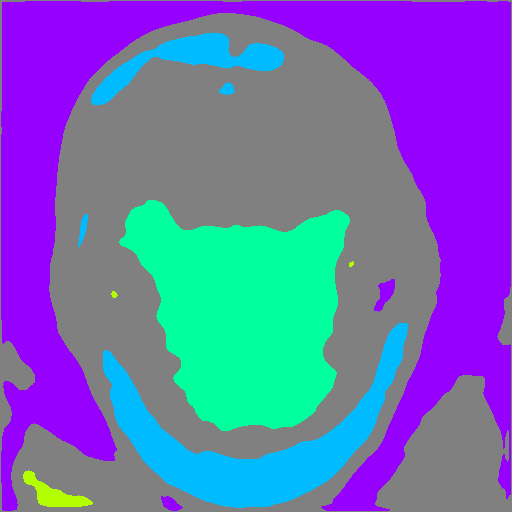} &
\includegraphics[width=.2\linewidth,height=.15\linewidth,valign=m]{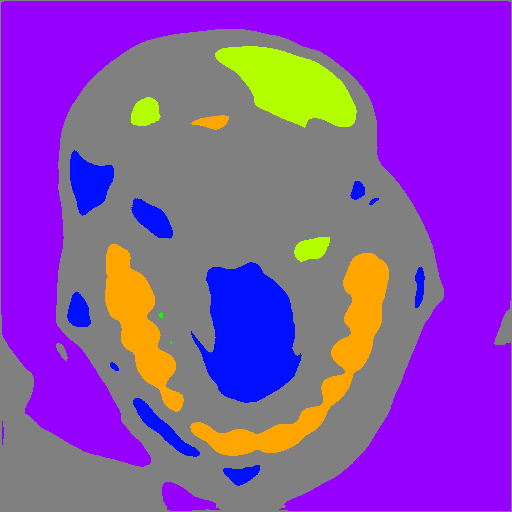} &
\includegraphics[width=.2\linewidth,height=.15\linewidth,valign=m]{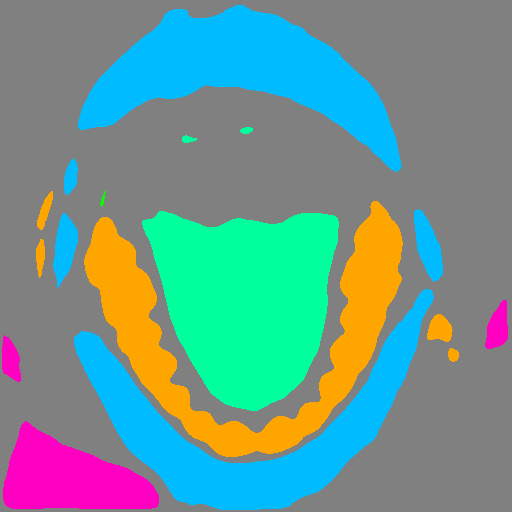} &
\includegraphics[width=.2\linewidth,height=.15\linewidth,valign=m]{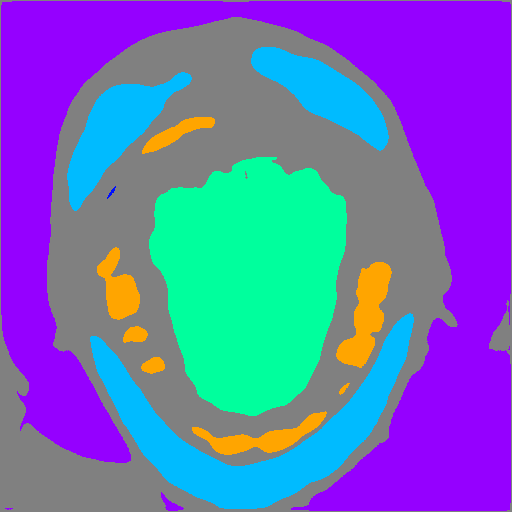}\\
\vspace{5pt}
\textbf{ODIN} & 
\includegraphics[width=.2\linewidth,height=.15\linewidth,valign=m]{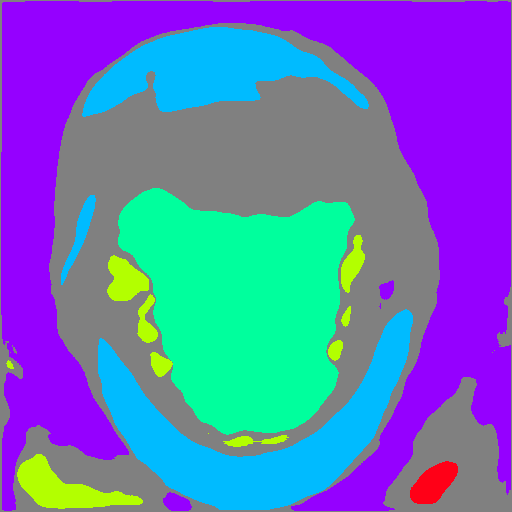} &
\includegraphics[width=.2\linewidth,height=.15\linewidth,valign=m]{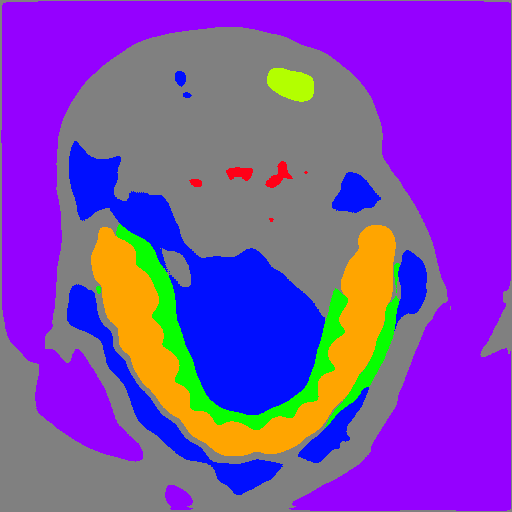} &
\includegraphics[width=.2\linewidth,height=.15\linewidth,valign=m]{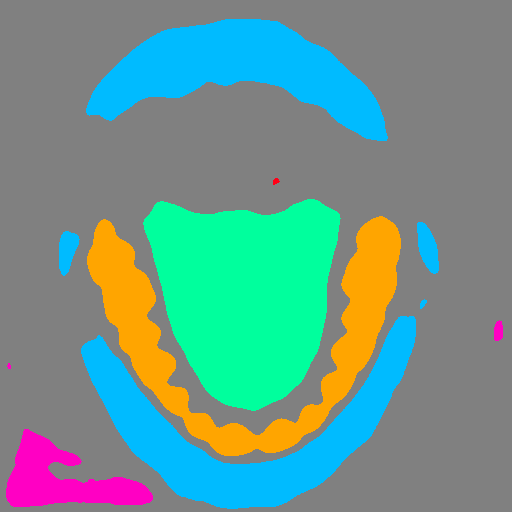} &
\includegraphics[width=.2\linewidth,height=.15\linewidth,valign=m]{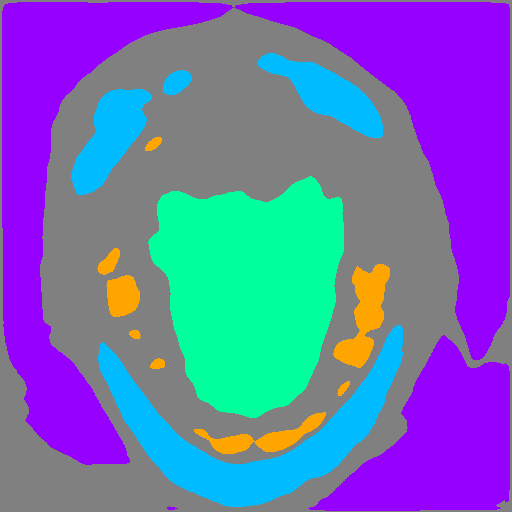}\\
\vspace{5pt}
\textbf{Mahalanobis} & 
\includegraphics[width=.2\linewidth,height=.15\linewidth,valign=m]{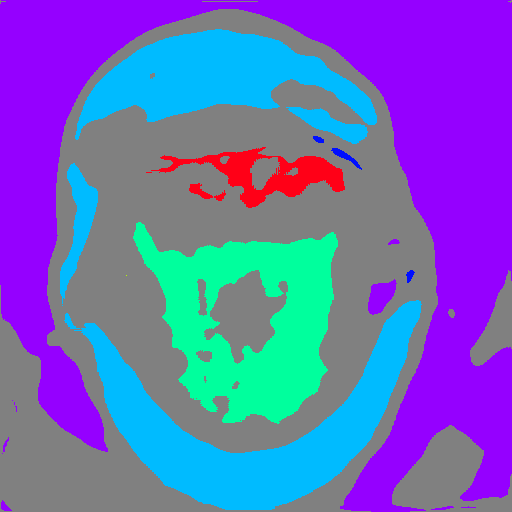} &
\includegraphics[width=.2\linewidth,height=.15\linewidth,valign=m]{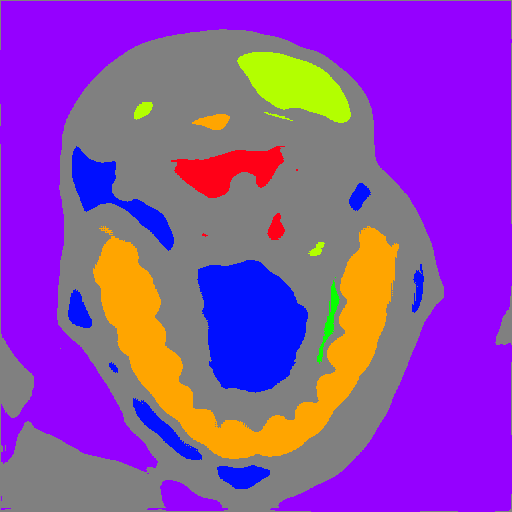} &
\includegraphics[width=.2\linewidth,height=.15\linewidth,valign=m]{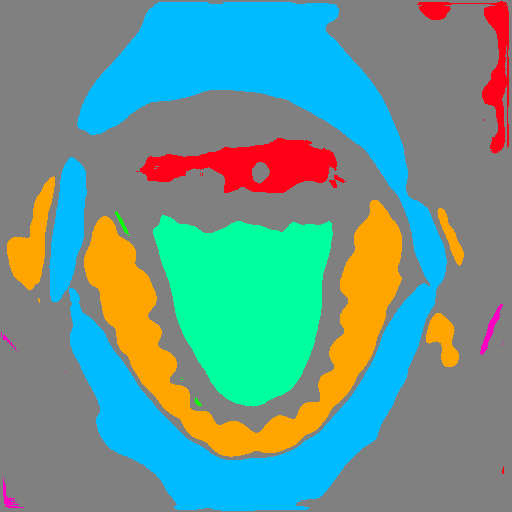} &
\includegraphics[width=.2\linewidth,height=.15\linewidth,valign=m]{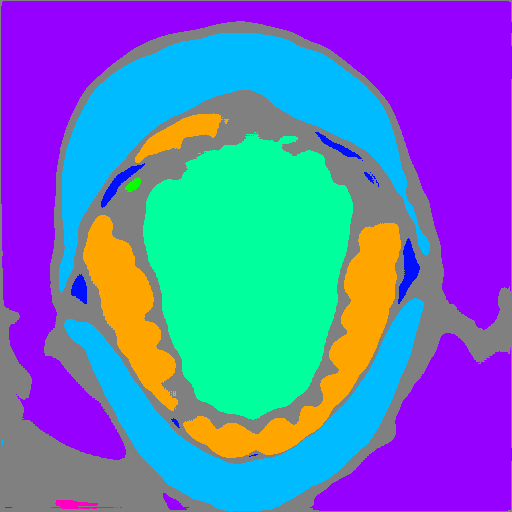}\\
\vspace{5pt}
\textbf{GODIN} & 
\includegraphics[width=.2\linewidth,height=.15\linewidth,valign=m]{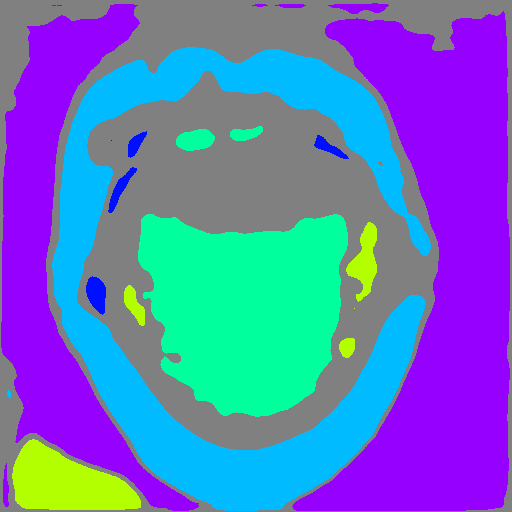} &
\includegraphics[width=.2\linewidth,height=.15\linewidth,valign=m]{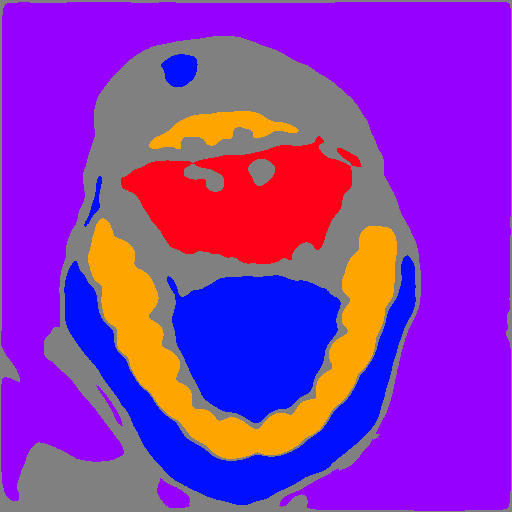} &
\includegraphics[width=.2\linewidth,height=.15\linewidth,valign=m]{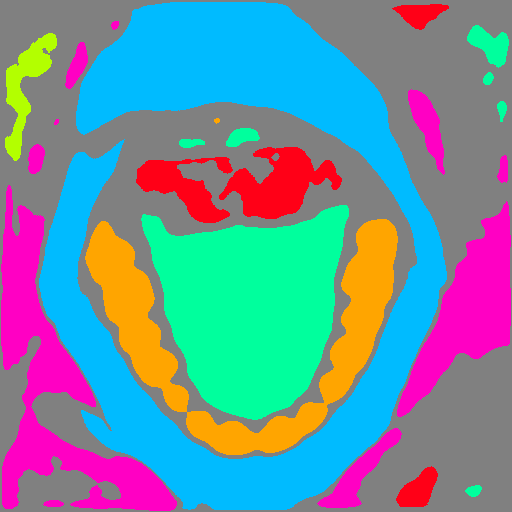} &
\includegraphics[width=.2\linewidth,height=.15\linewidth,valign=m]{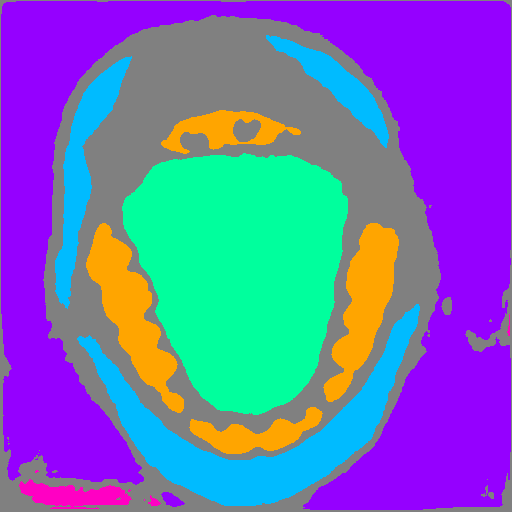}\\
&
\multicolumn{4}{c}{\includegraphics[width=16.1cm, height=0.5cm]{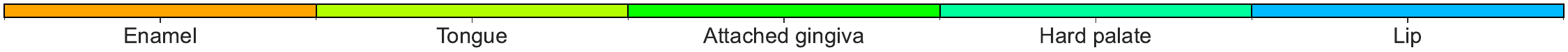}} \\
&
\multicolumn{4}{c}{\includegraphics[width=16.1cm, height=0.5cm]{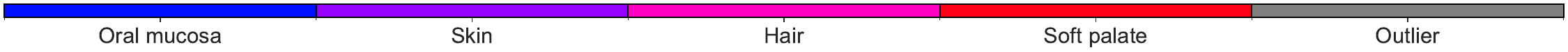}} \\
\end{tabular}
\caption{Qualitative result of first case from ODSI-DB dataset. We show results of the same image from four class partitions (represented as $\CP_1$ to $\CP_4$). For each $\CP$, classes that are held-out are grouped as an extra outlier class for evaluation. We visualise and compare masks generated using different methods at threshold $\tau_{m}$. Baseline results at $\tau_0=0$ are added to represent result without outlier detection.}
\label{fig:segmentation_result_odsi_1}
\end{figure}

\begin{figure}[p]
\centering\footnotesize
\begin{tabular}{ccccc}
& \boldmath$\CP_1$ & \boldmath$\CP_2$ & \boldmath$\CP_3$ & \boldmath$\CP_4$ \\
\vspace{5pt}
\makecell[c] {\textbf{Sparsely} \\ \textbf{annotated} \\ \textbf{ground truth}} &  
\includegraphics[width=.2\linewidth,height=.15\linewidth,valign=m]{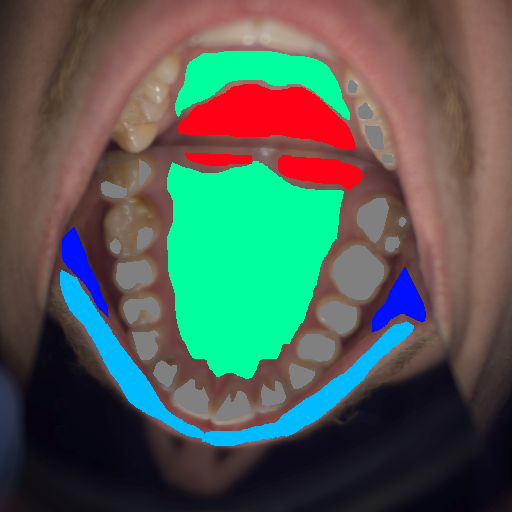} &
\includegraphics[width=.2\linewidth,height=.15\linewidth,valign=m]{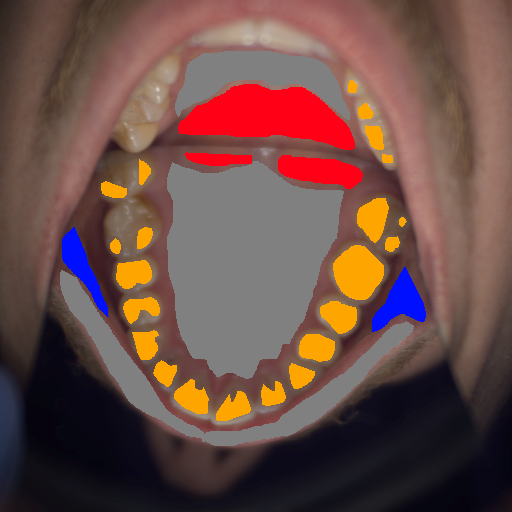} &
\includegraphics[width=.2\linewidth,height=.15\linewidth,valign=m]{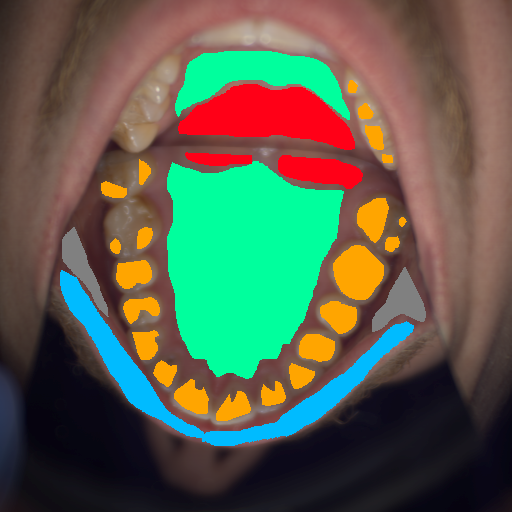} &
\includegraphics[width=.2\linewidth,height=.15\linewidth,valign=m]{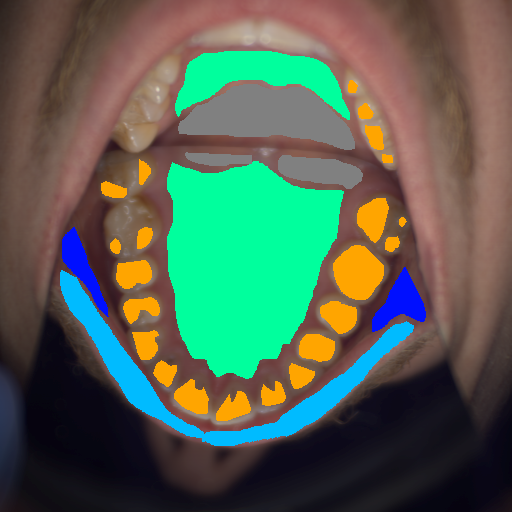}\\
\vspace{5pt}
\textbf{Baseline\ $(\tau_{0}=0$)} & 
\includegraphics[width=.2\linewidth,height=.15\linewidth,valign=m]{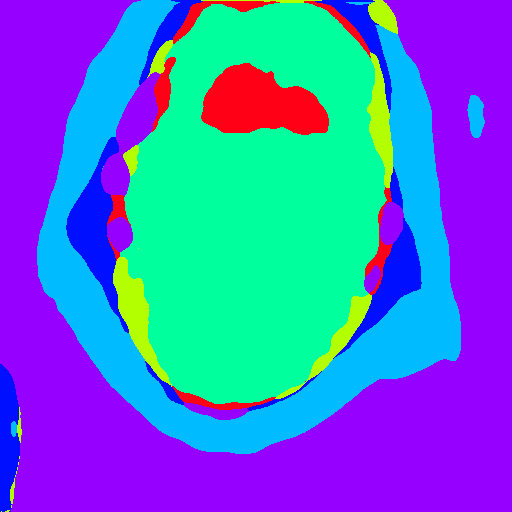} &
\includegraphics[width=.2\linewidth,height=.15\linewidth,valign=m]{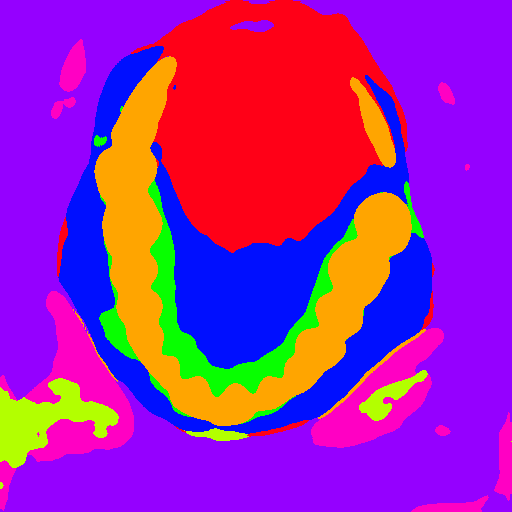} &
\includegraphics[width=.2\linewidth,height=.15\linewidth,valign=m]{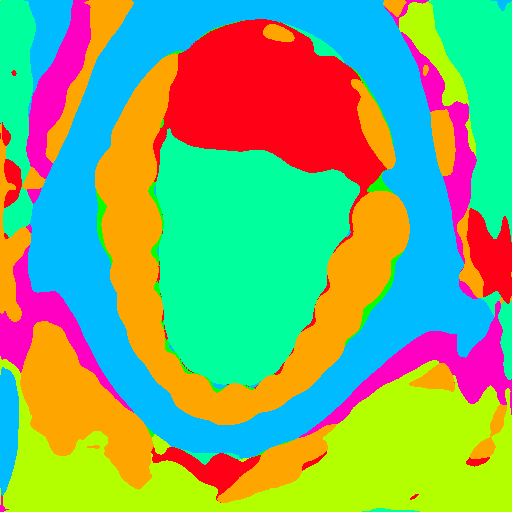} &
\includegraphics[width=.2\linewidth,height=.15\linewidth,valign=m]{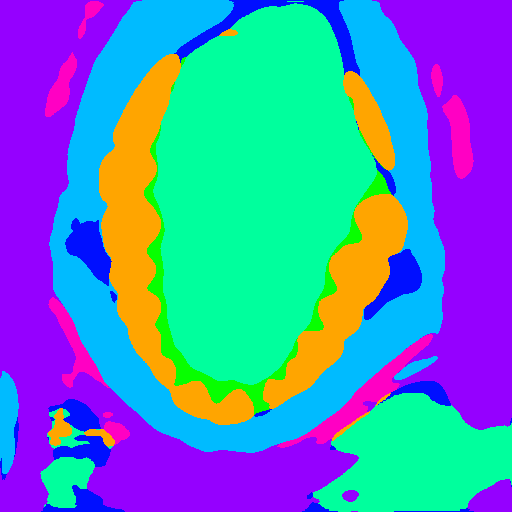}\\
\vspace{5pt}
\textbf{Baseline\ $(\tau_{m}$)} & 
\includegraphics[width=.2\linewidth,height=.15\linewidth,valign=m]{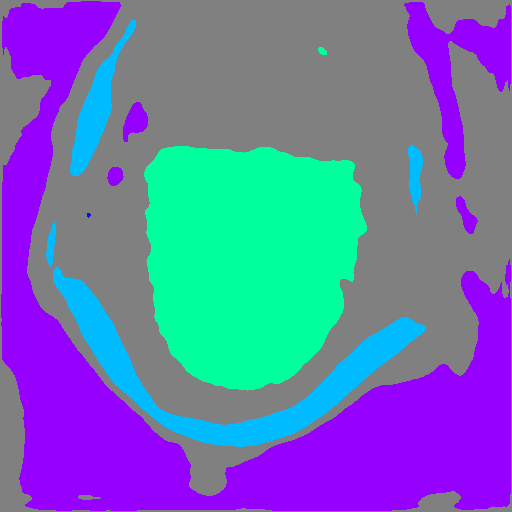} &
\includegraphics[width=.2\linewidth,height=.15\linewidth,valign=m]{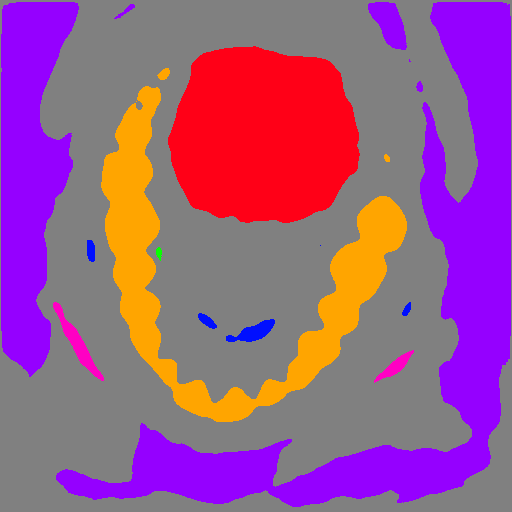} &
\includegraphics[width=.2\linewidth,height=.15\linewidth,valign=m]{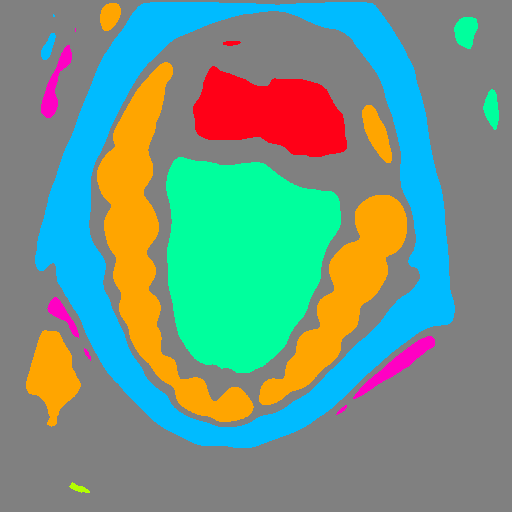} &
\includegraphics[width=.2\linewidth,height=.15\linewidth,valign=m]{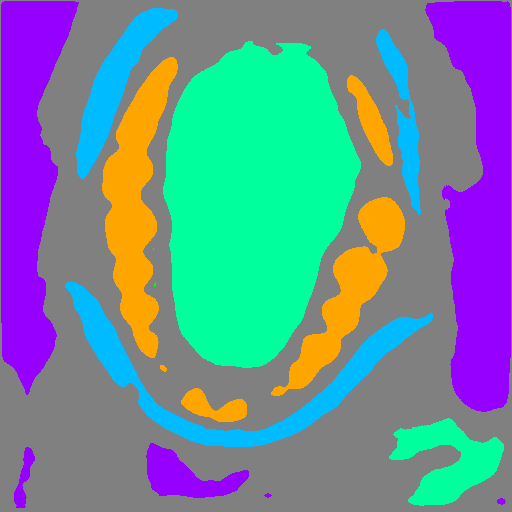}\\
\vspace{5pt}
\textbf{ODIN} & 
\includegraphics[width=.2\linewidth,height=.15\linewidth,valign=m]{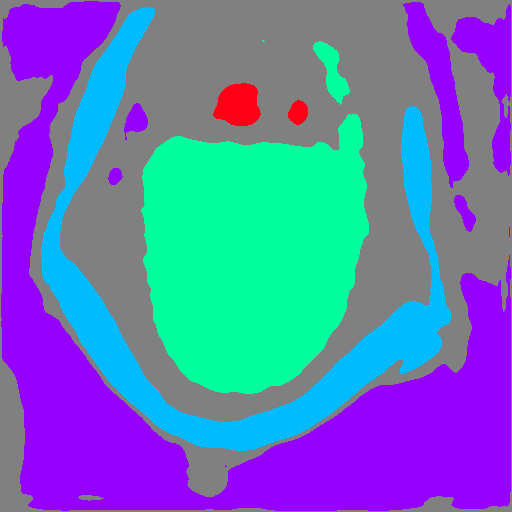} &
\includegraphics[width=.2\linewidth,height=.15\linewidth,valign=m]{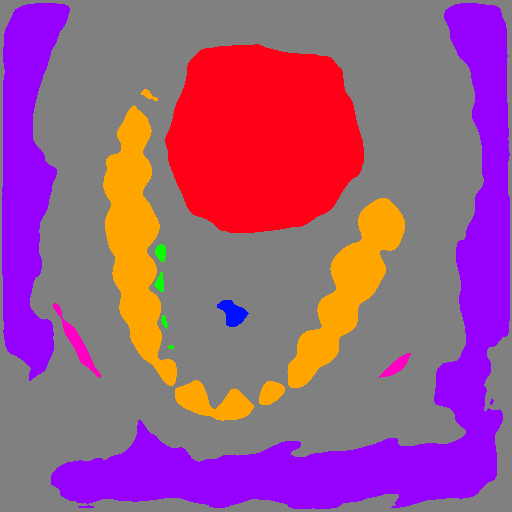} &
\includegraphics[width=.2\linewidth,height=.15\linewidth,valign=m]{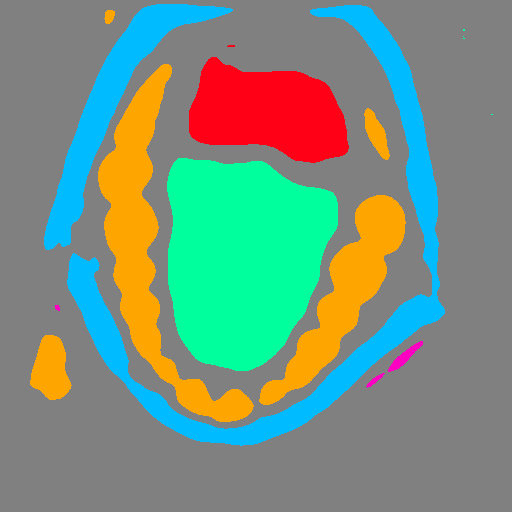} &
\includegraphics[width=.2\linewidth,height=.15\linewidth,valign=m]{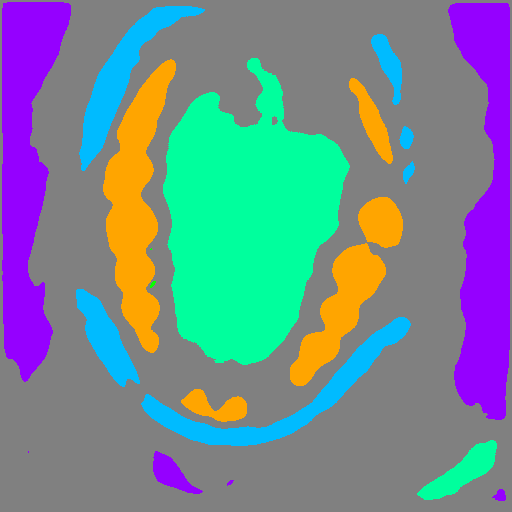}\\
\vspace{5pt}
\textbf{Mahalanobis} & 
\includegraphics[width=.2\linewidth,height=.15\linewidth,valign=m]{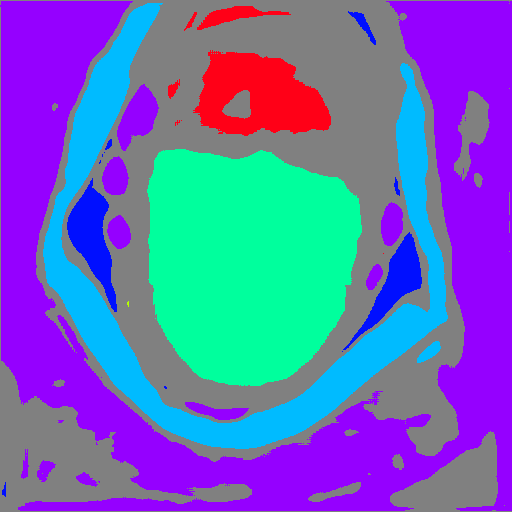} &
\includegraphics[width=.2\linewidth,height=.15\linewidth,valign=m]{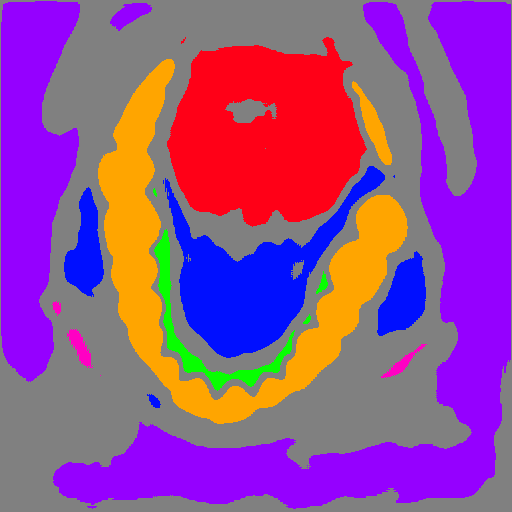} &
\includegraphics[width=.2\linewidth,height=.15\linewidth,valign=m]{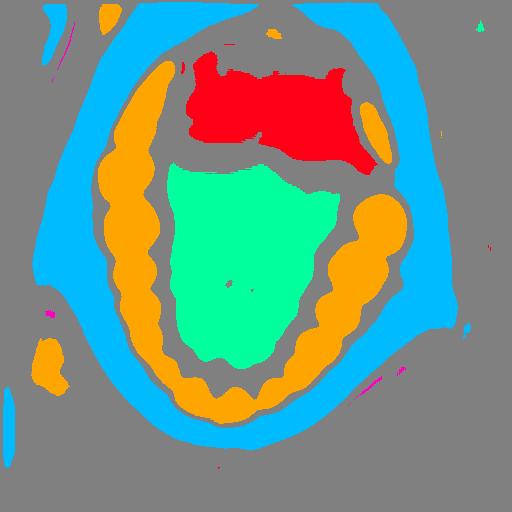} &
\includegraphics[width=.2\linewidth,height=.15\linewidth,valign=m]{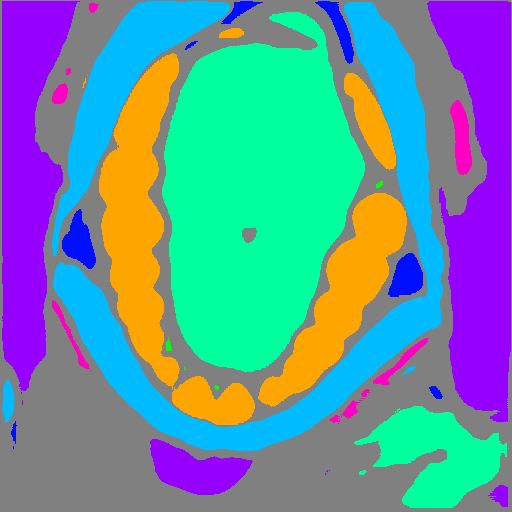}\\
\vspace{5pt}
\textbf{GODIN} & 
\includegraphics[width=.2\linewidth,height=.15\linewidth,valign=m]{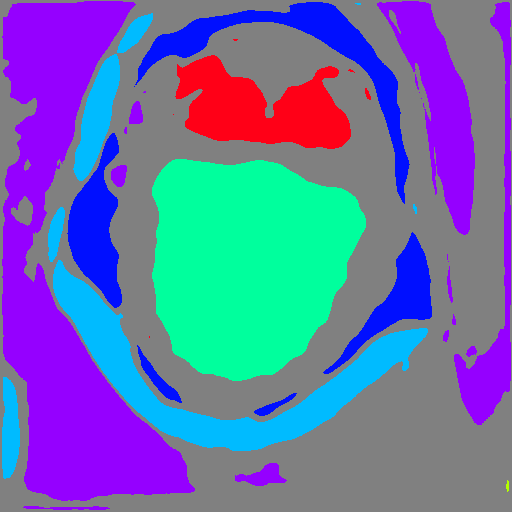} &
\includegraphics[width=.2\linewidth,height=.15\linewidth,valign=m]{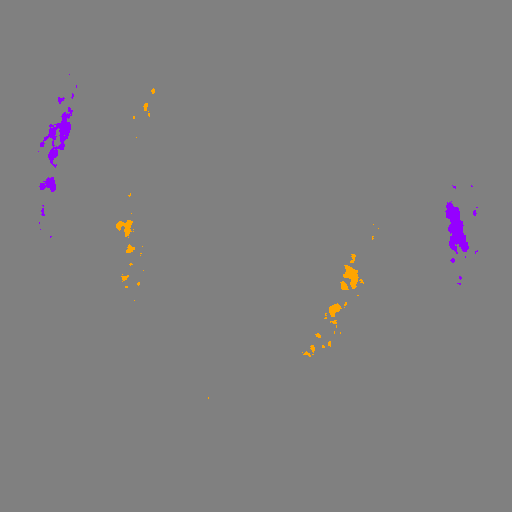} &
\includegraphics[width=.2\linewidth,height=.15\linewidth,valign=m]{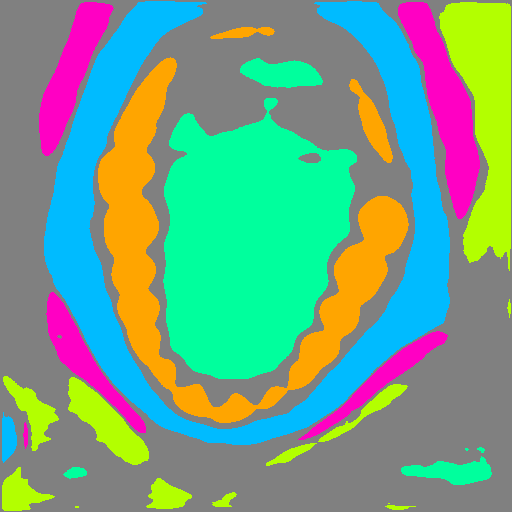} &
\includegraphics[width=.2\linewidth,height=.15\linewidth,valign=m]{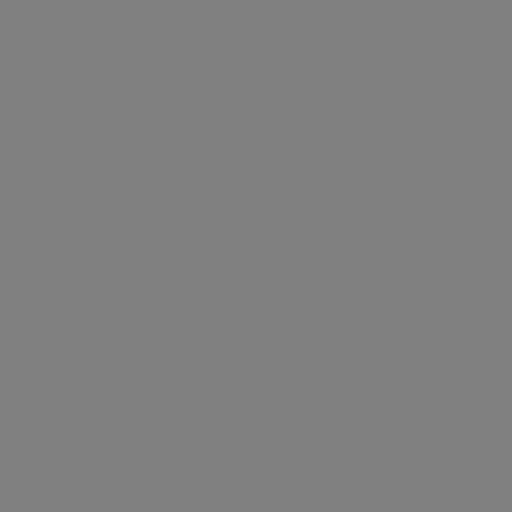}\\
&
\multicolumn{4}{c}{\includegraphics[width=16.1cm, height=0.5cm]{figures/colorbar_odsi_1.pdf}} \\
&
\multicolumn{4}{c}{\includegraphics[width=16.1cm, height=0.5cm]{figures/colorbar_odsi_2.pdf}} \\
\end{tabular}
\caption{Qualitative result of second case from ODSI-DB dataset. We show results of the same image from four class partitions (represented as $\CP_1$ to $\CP_4$). For each $\CP$, classes that are held-out are grouped as an extra outlier class for evaluation. We visualise and compare masks generated using different methods at threshold $\tau_{m}$. Baseline results at $\tau_0=0$ are added to represent result without outlier detection.}
\label{fig:segmentation_result_odsi_2}
\end{figure}

\begin{figure}[p]
\centering\footnotesize
\begin{tabular}{ccccc}
& \boldmath$\CP_1$ & \boldmath$\CP_2$ & \boldmath$\CP_3$ & \boldmath$\CP_4$ \\
\vspace{5pt}
\makecell[c] {\textbf{Sparsely} \\ \textbf{annotated} \\ \textbf{ground truth}} &  
\includegraphics[width=.2\linewidth,valign=m]{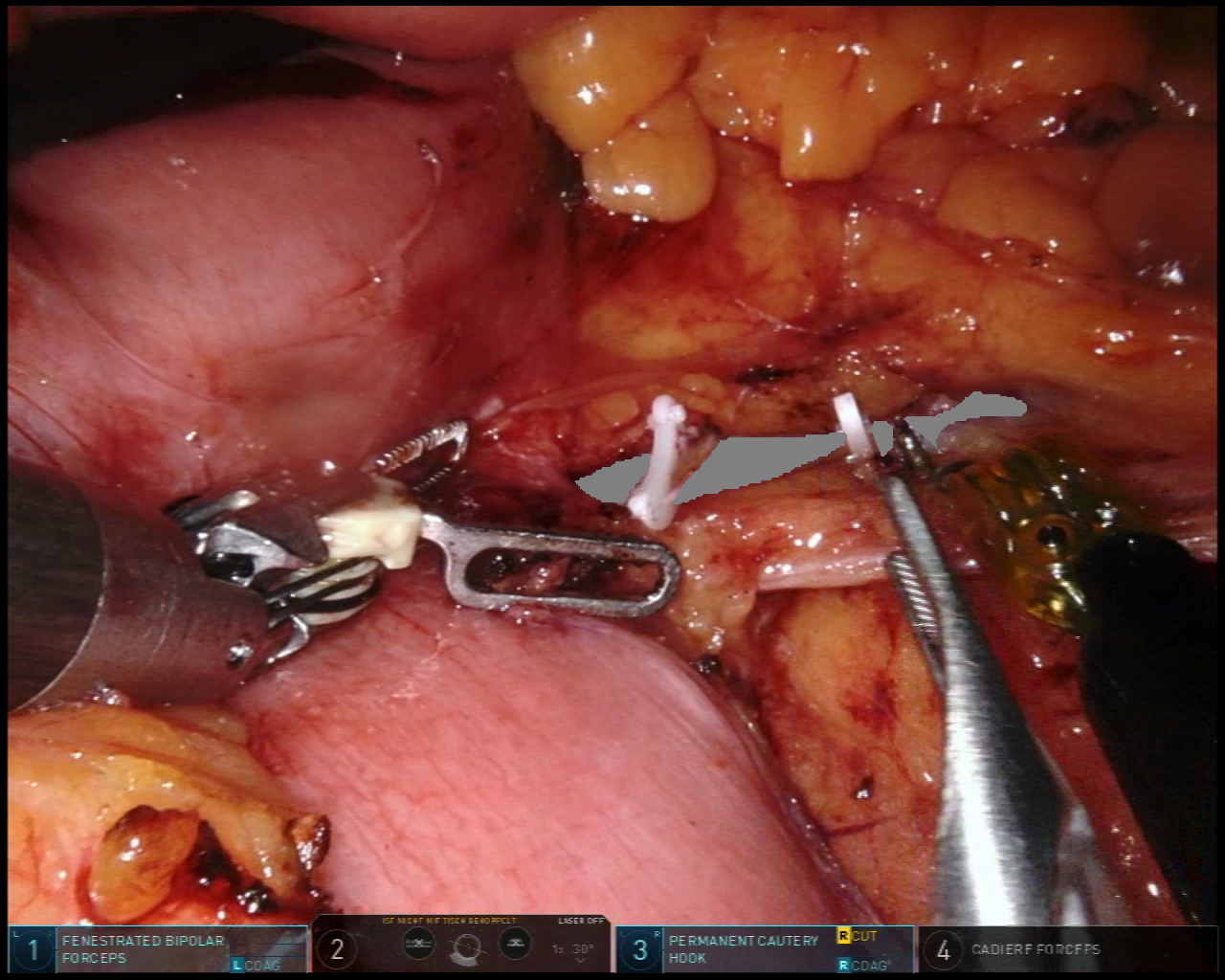} &
\includegraphics[width=.2\linewidth,valign=m]{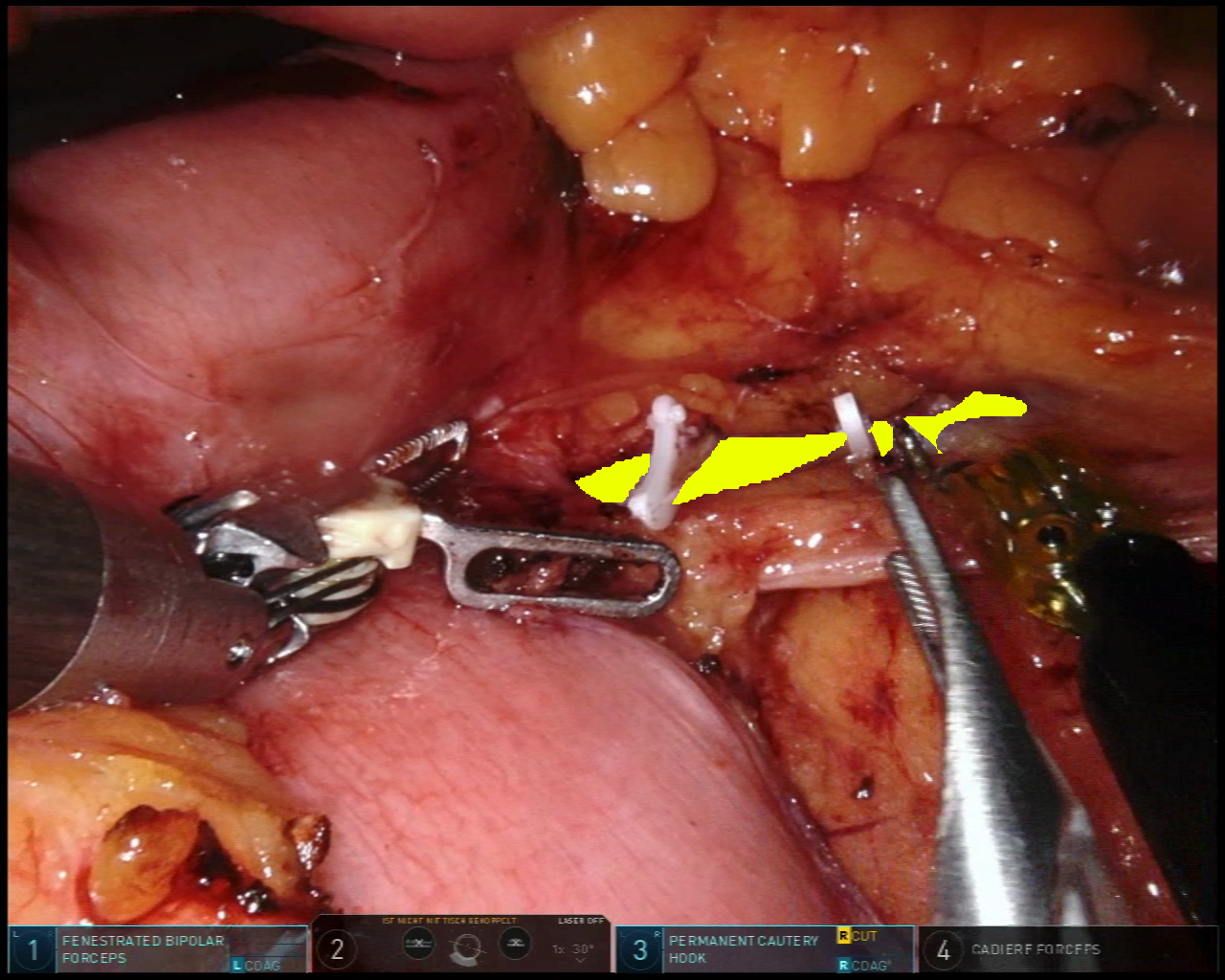} &
\includegraphics[width=.2\linewidth,valign=m]{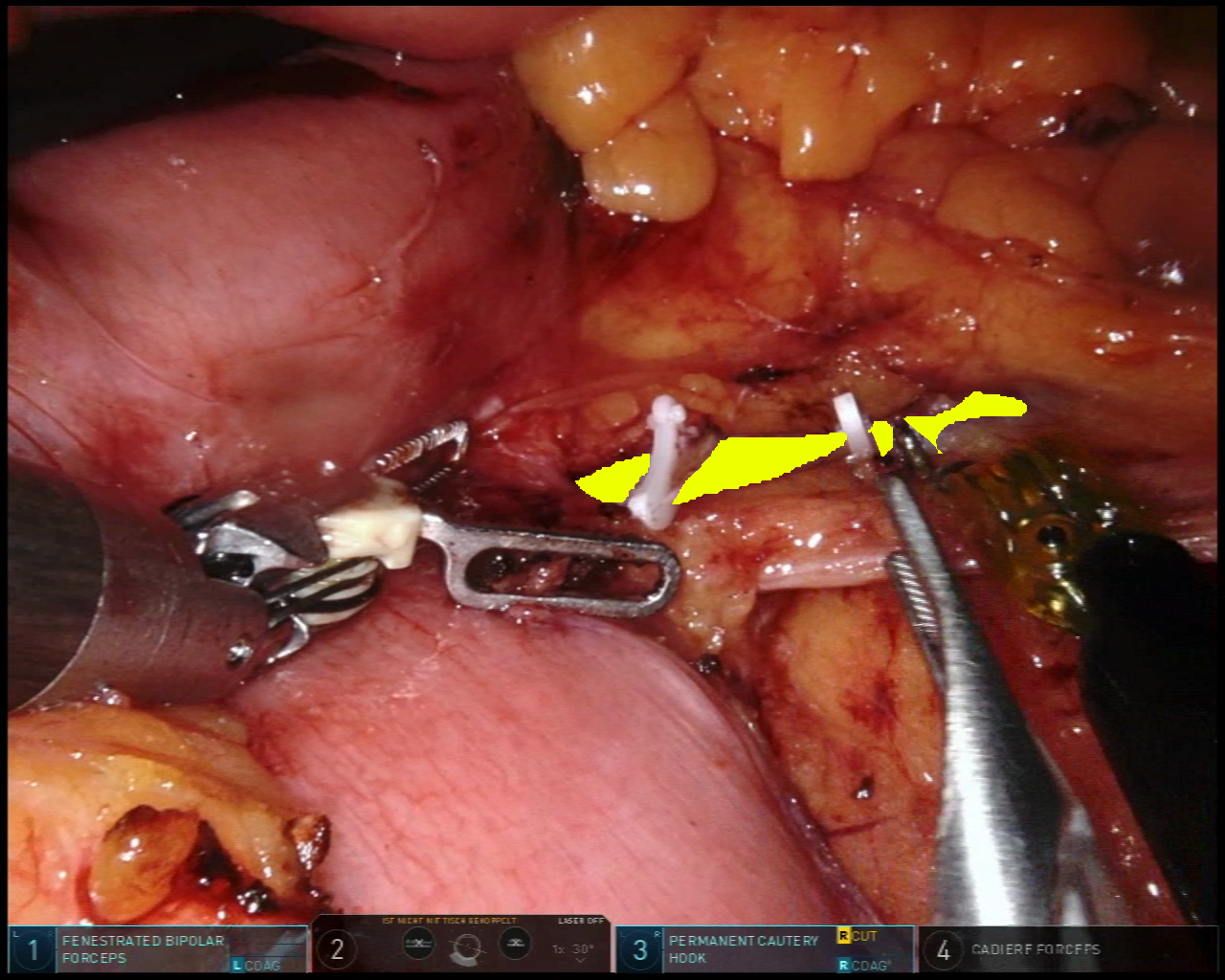} &
\includegraphics[width=.2\linewidth,valign=m]{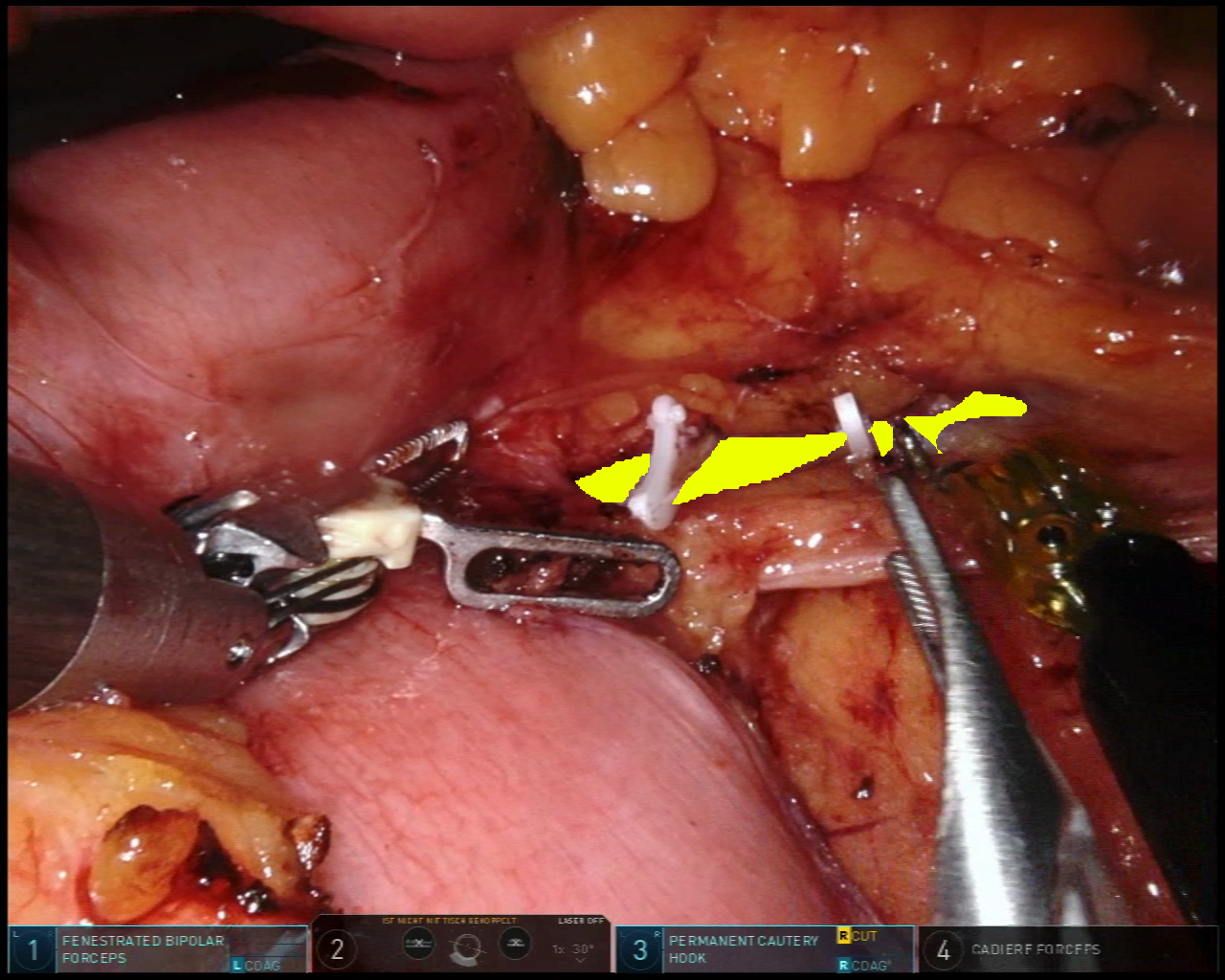}\\
\vspace{5pt}
\textbf{Baseline\ $(\tau_{0}=0$)} & 
\includegraphics[width=.2\linewidth,valign=m]{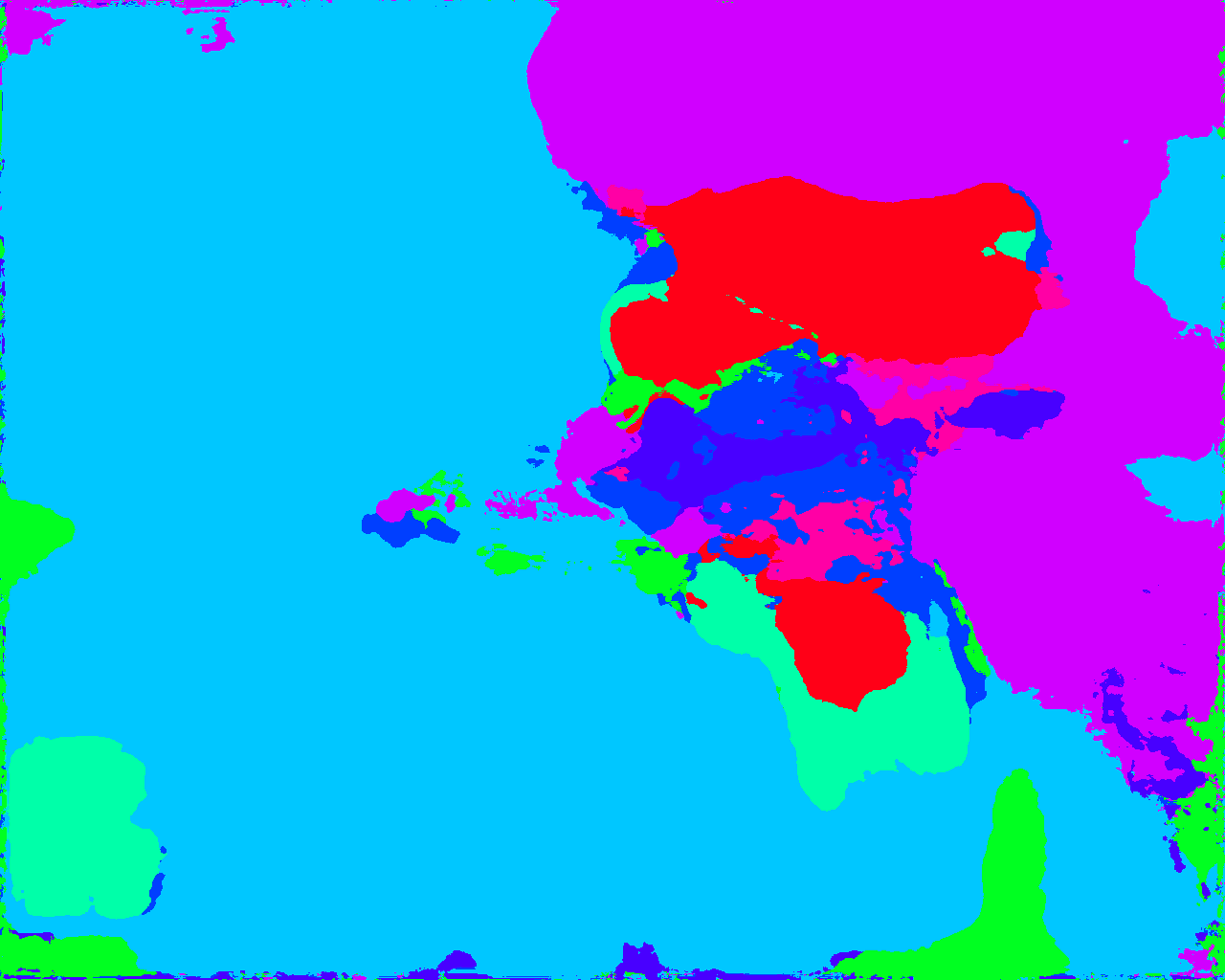} &
\includegraphics[width=.2\linewidth,valign=m]{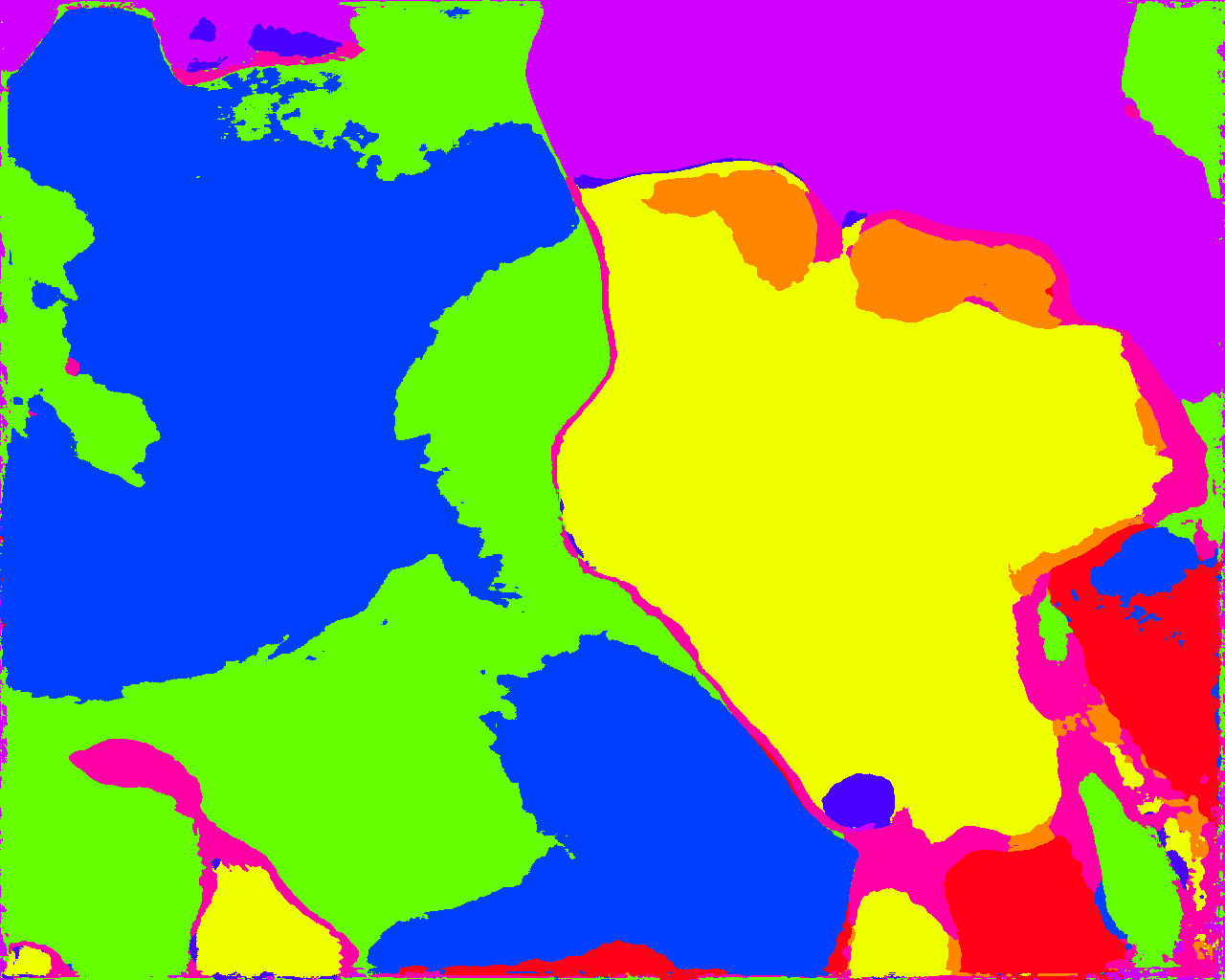} &
\includegraphics[width=.2\linewidth,valign=m]{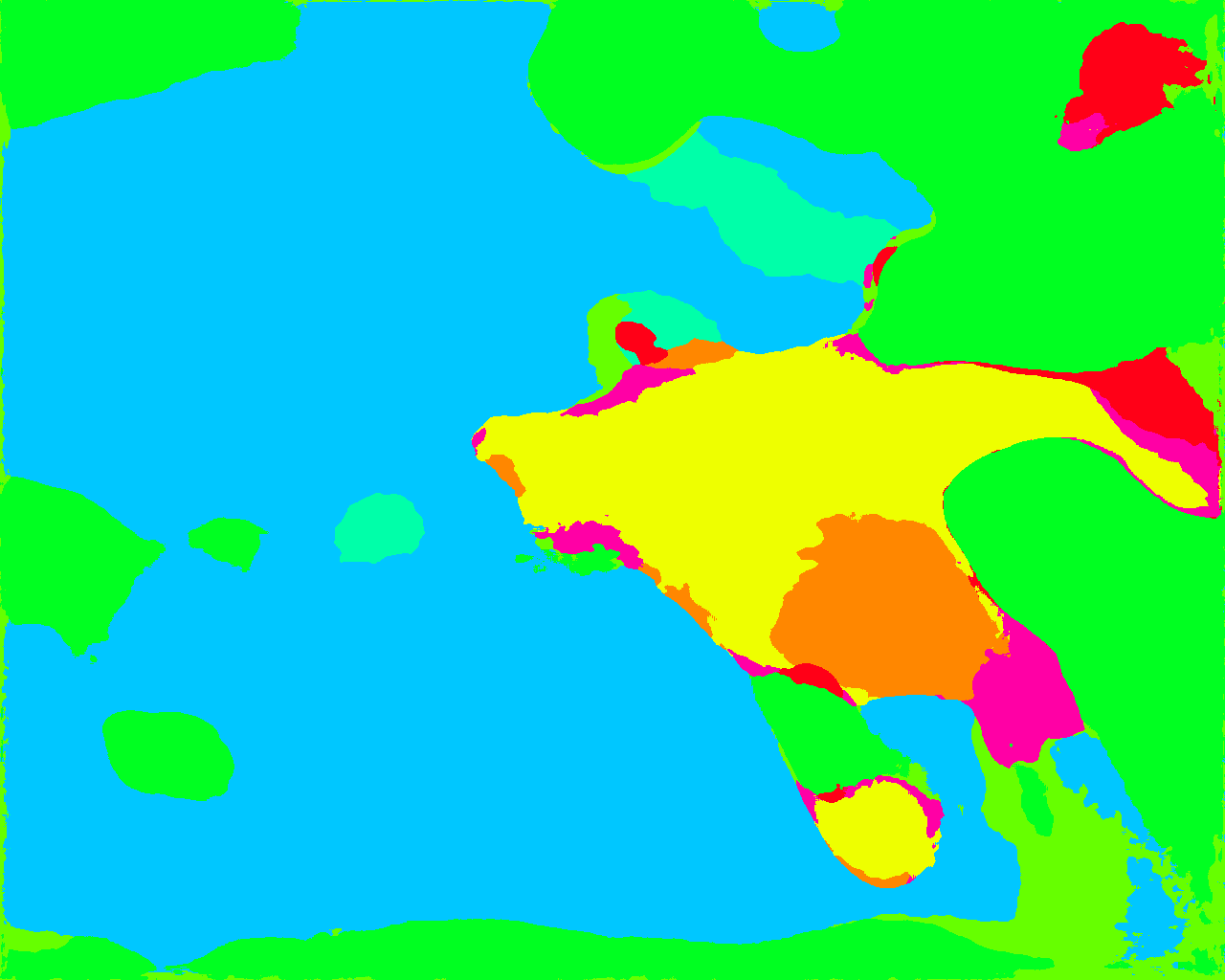} &
\includegraphics[width=.2\linewidth,valign=m]{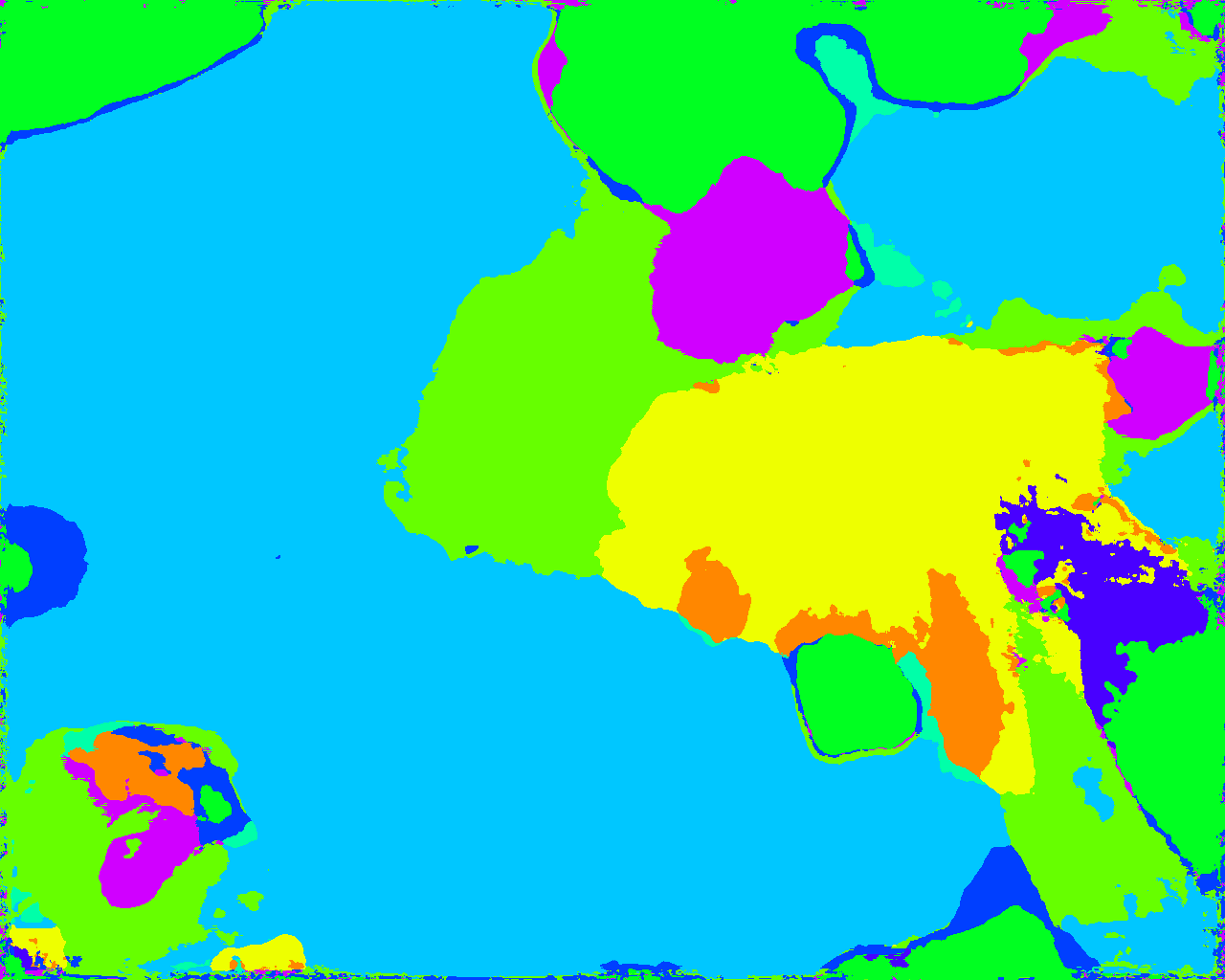}\\
\vspace{5pt}
\textbf{Baseline\ $(\tau_{m}$)} & 
\includegraphics[width=.2\linewidth,valign=m]{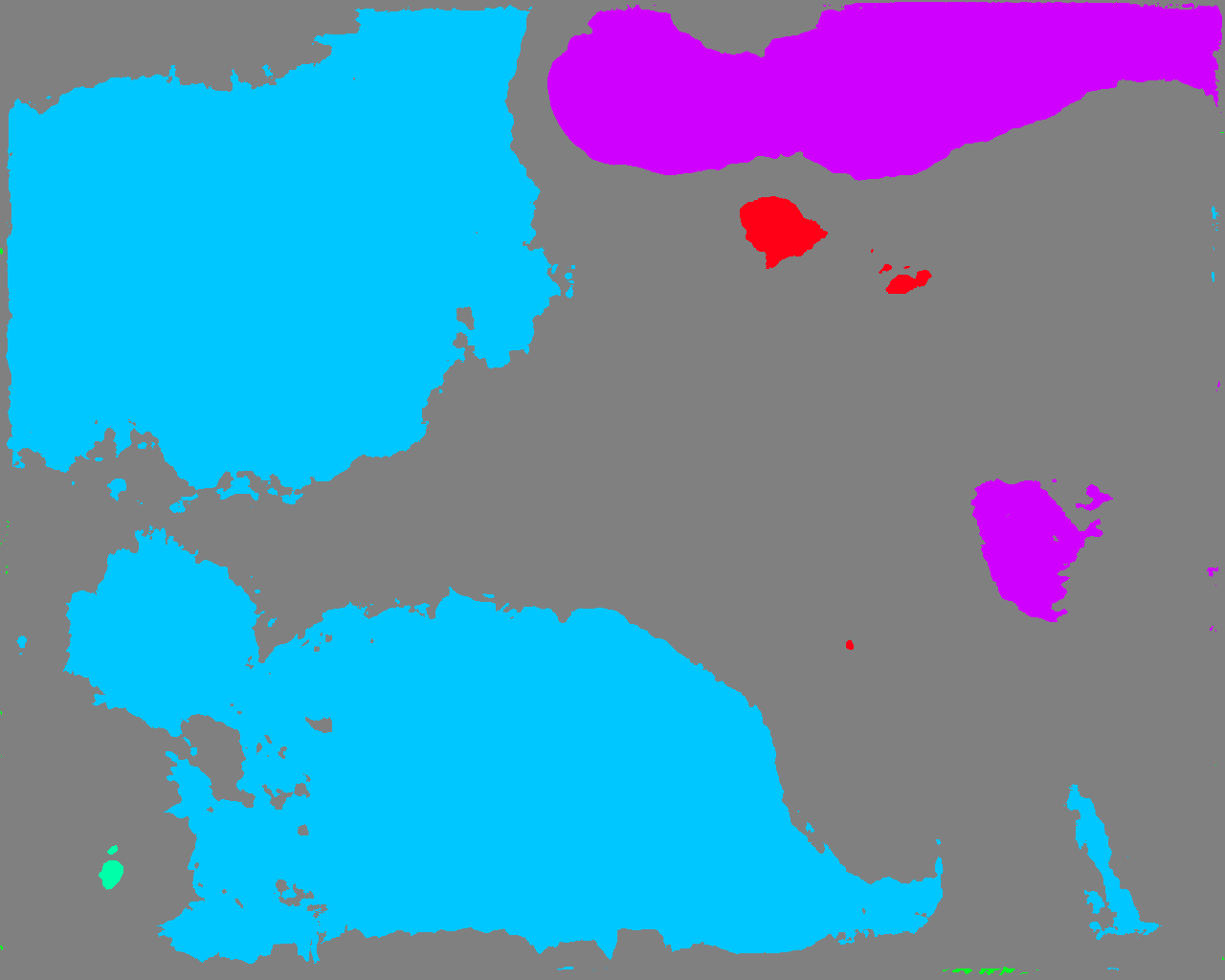} &
\includegraphics[width=.2\linewidth,valign=m]{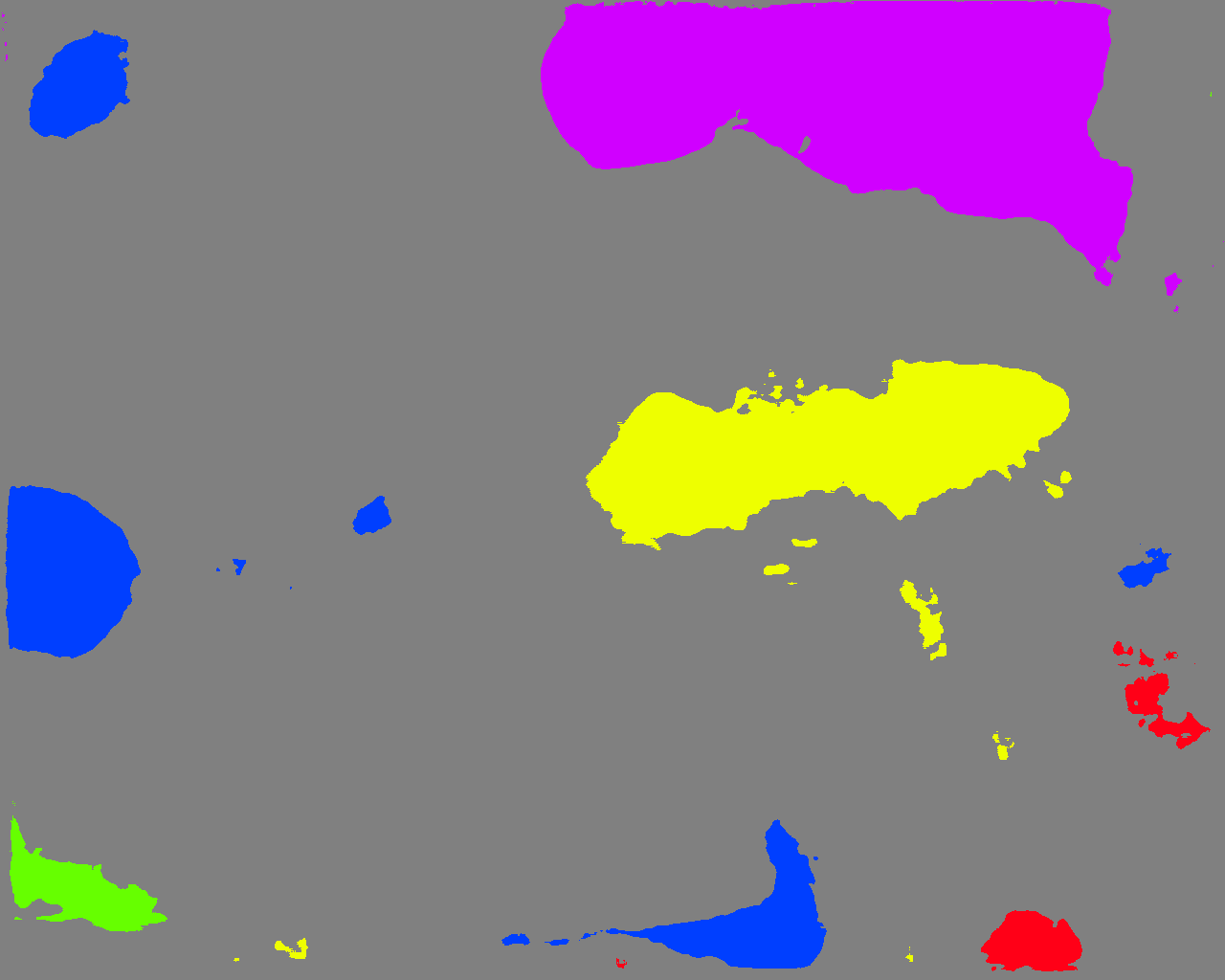} &
\includegraphics[width=.2\linewidth,valign=m]{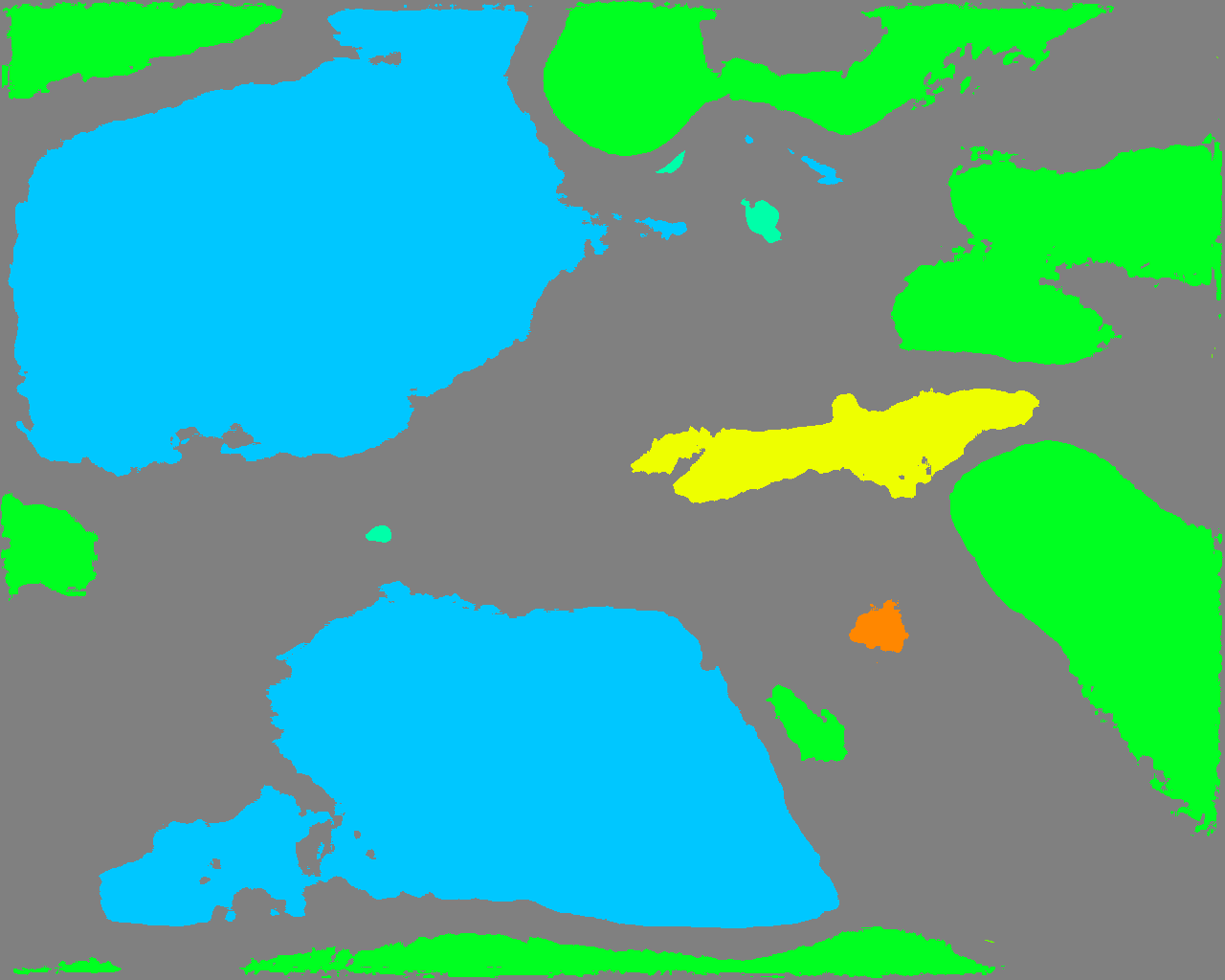} &
\includegraphics[width=.2\linewidth,valign=m]{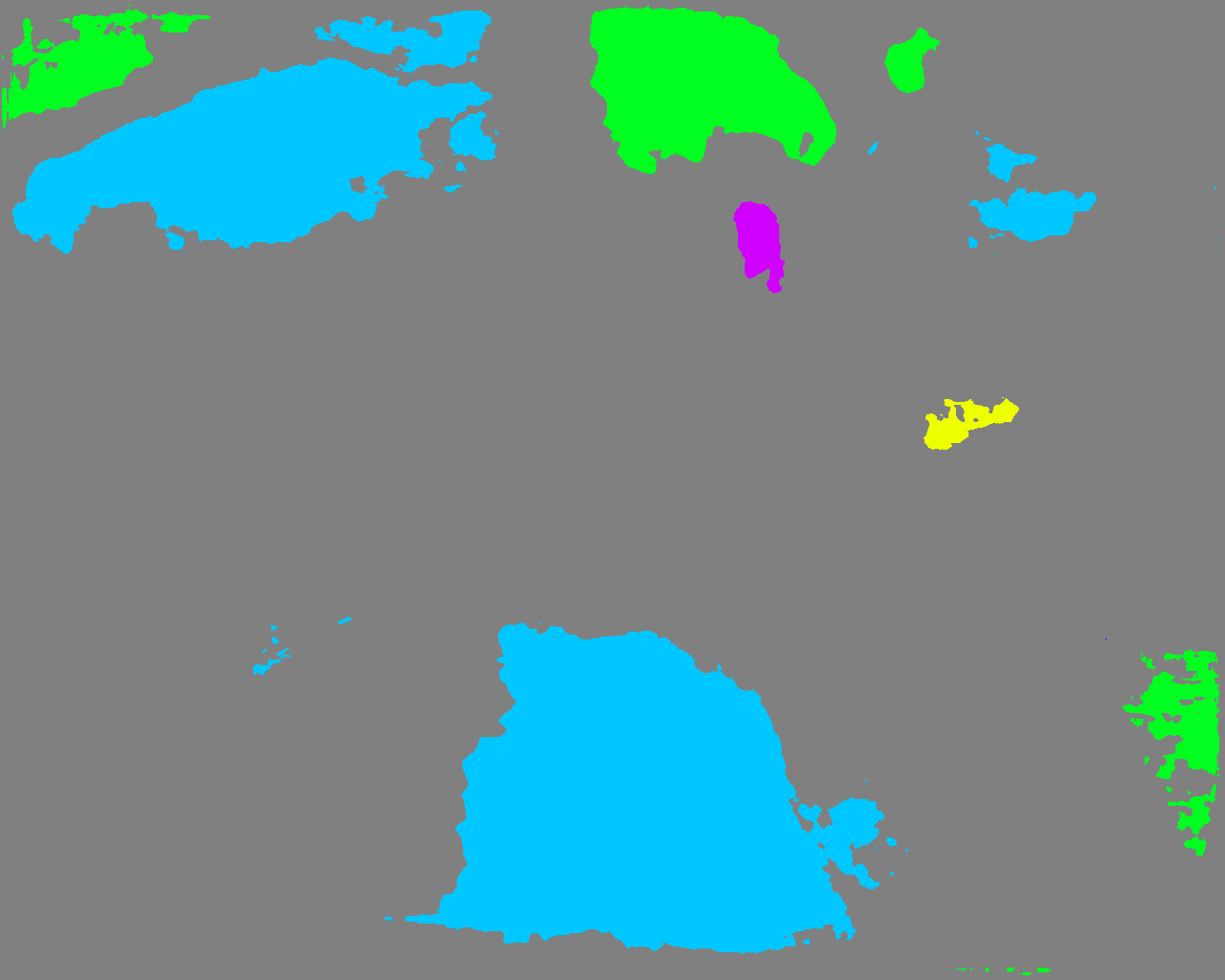}\\
\vspace{5pt}
\textbf{ODIN} & 
\includegraphics[width=.2\linewidth,valign=m]{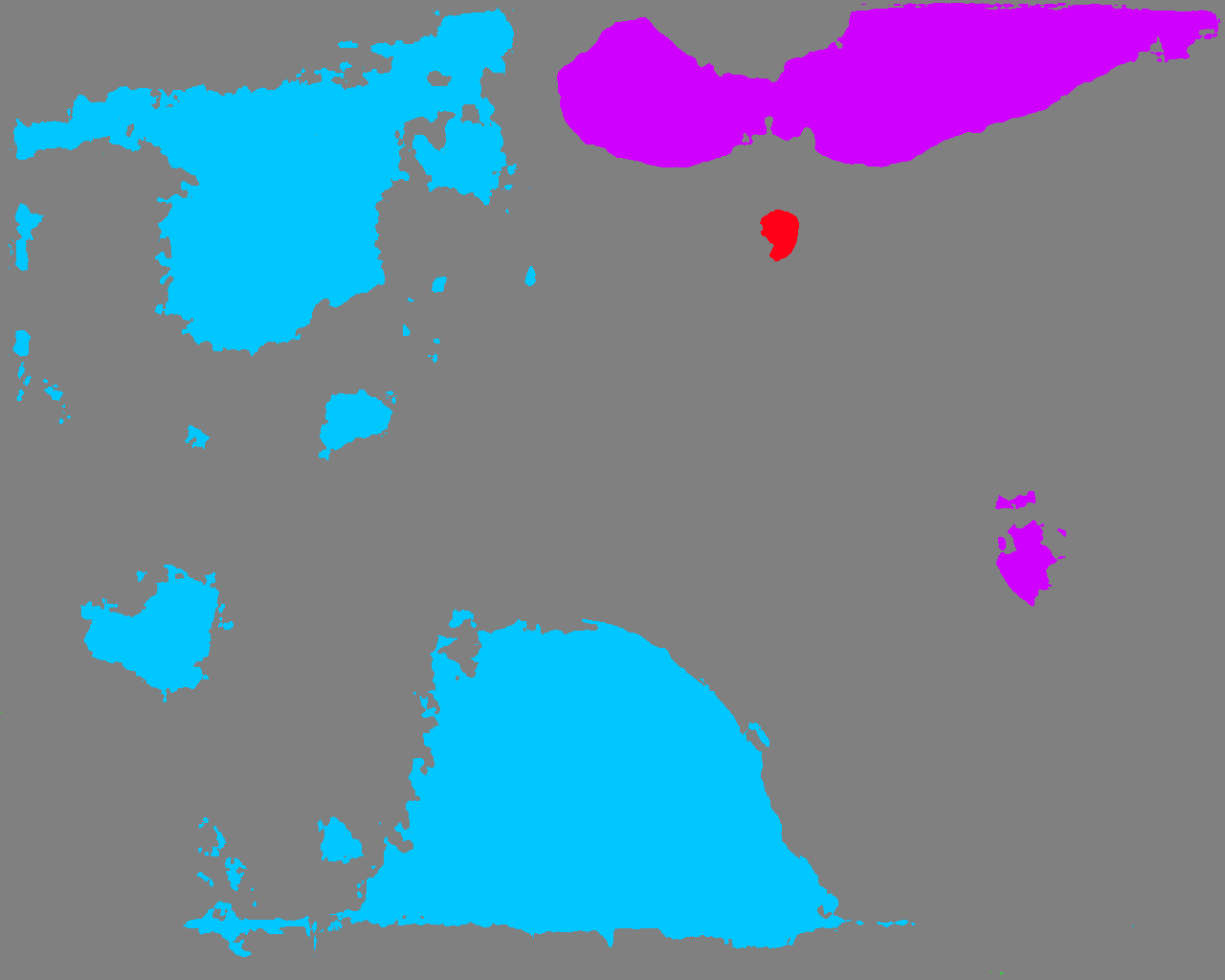} &
\includegraphics[width=.2\linewidth,valign=m]{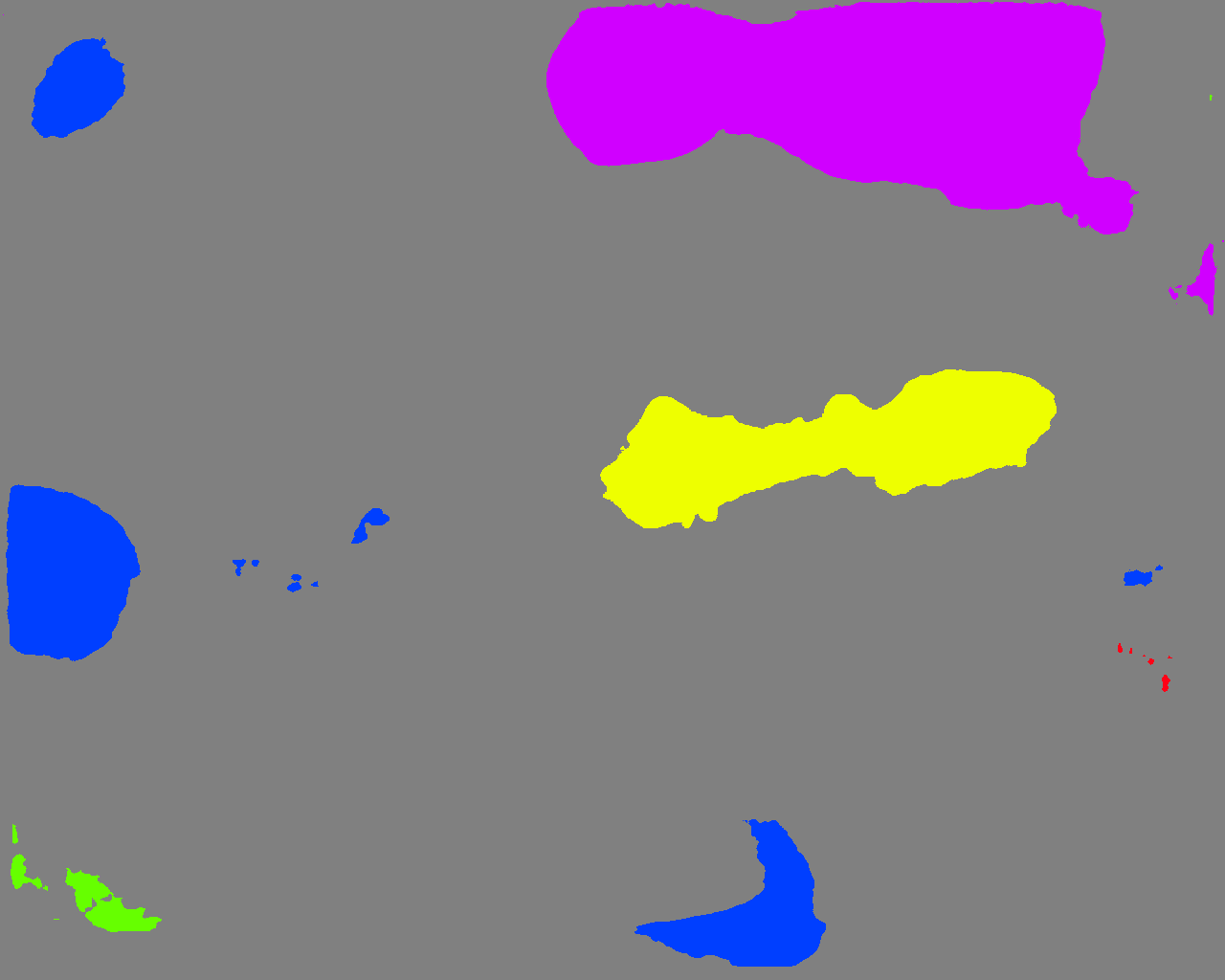} &
\includegraphics[width=.2\linewidth,valign=m]{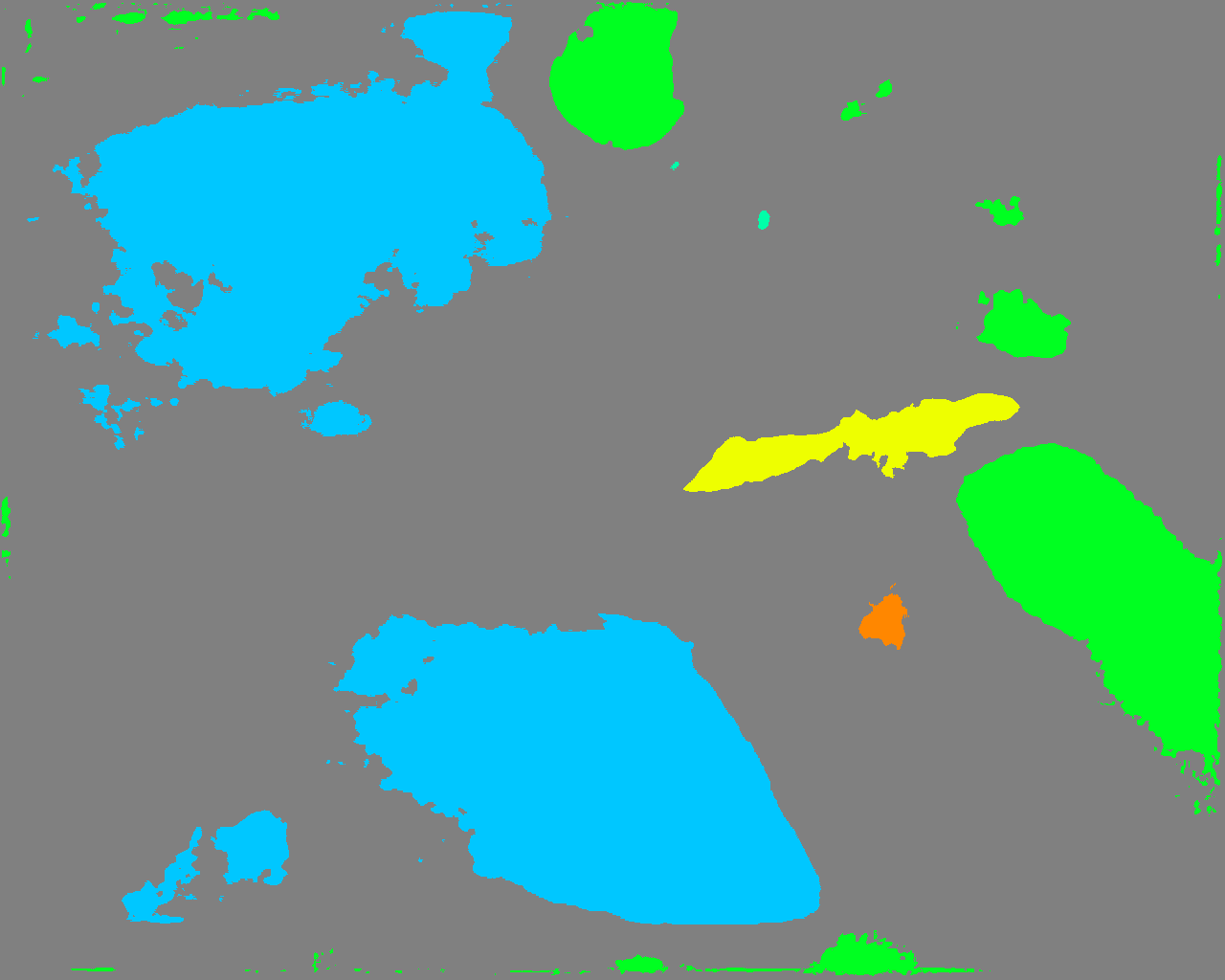} &
\includegraphics[width=.2\linewidth,valign=m]{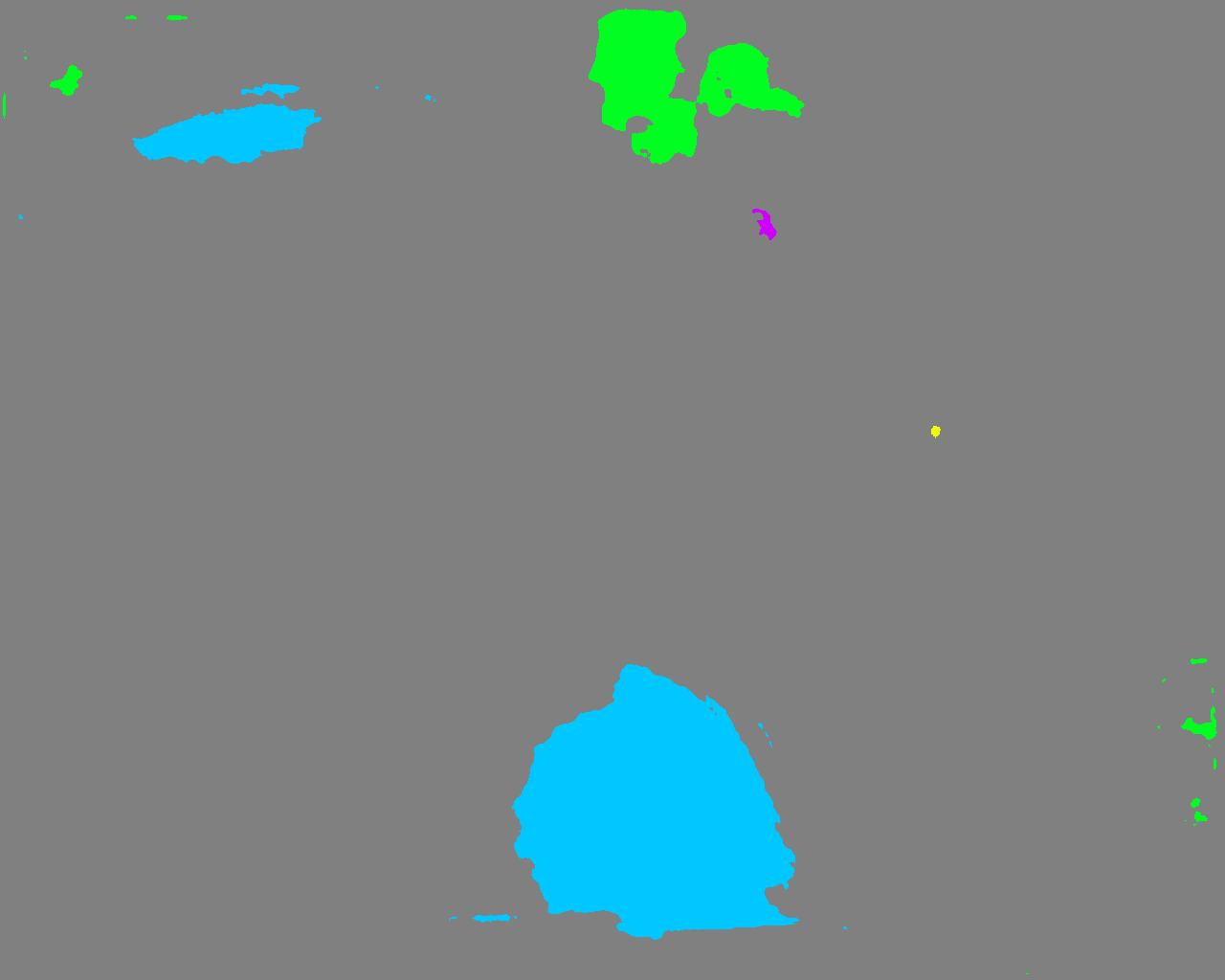}\\
\vspace{5pt}
\textbf{Mahalanobis} & 
\includegraphics[width=.2\linewidth,valign=m]{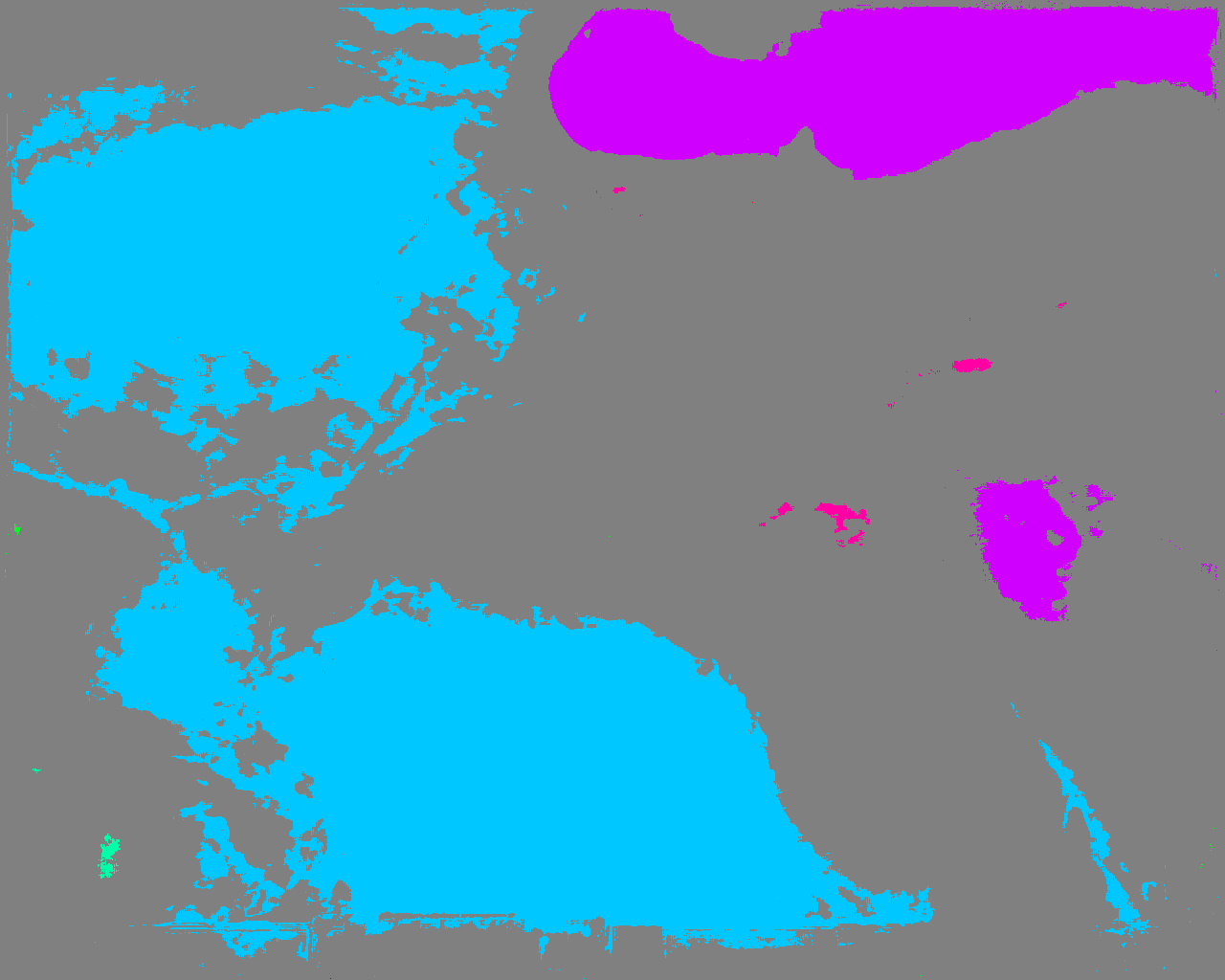} &
\includegraphics[width=.2\linewidth,valign=m]{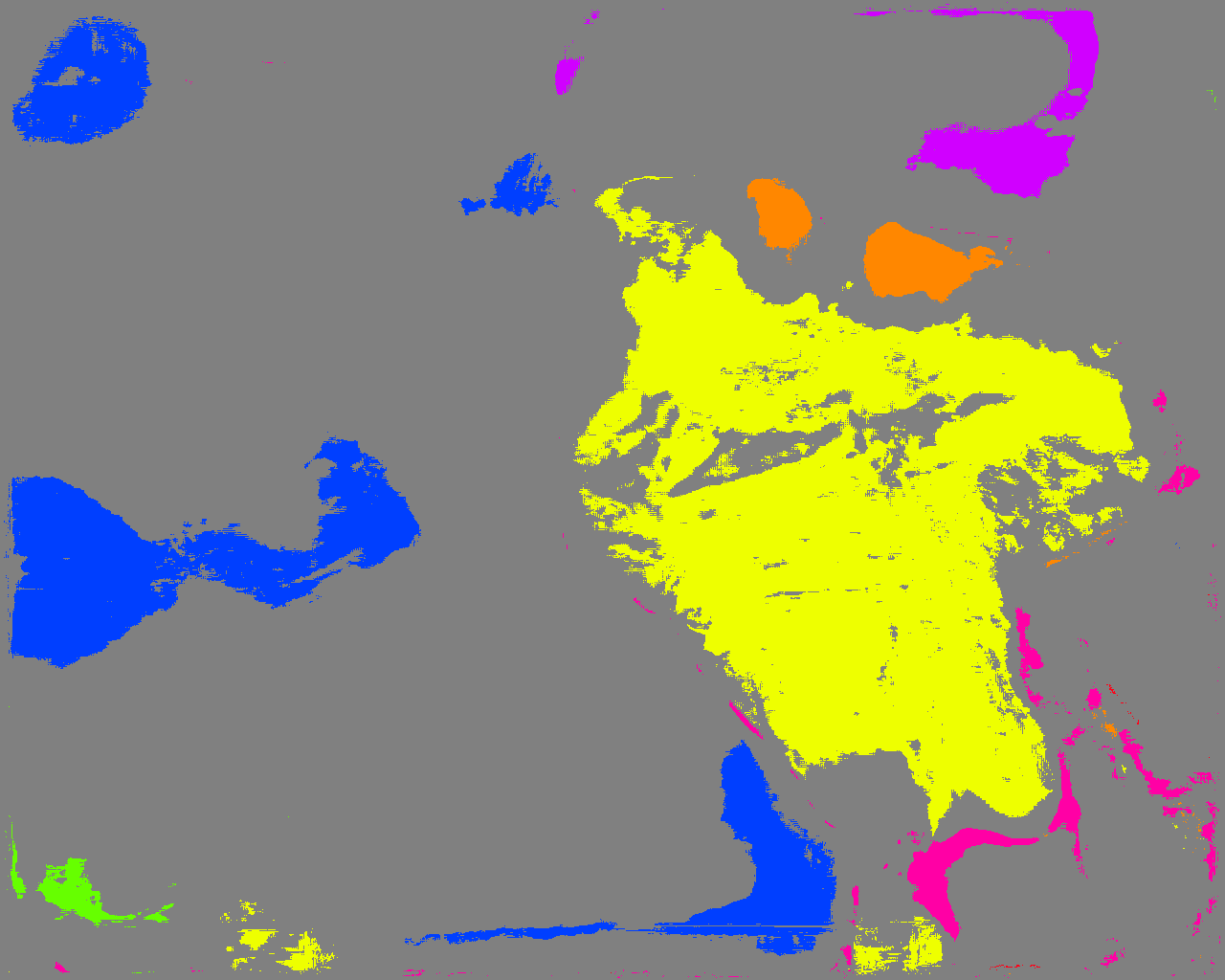} &
\includegraphics[width=.2\linewidth,valign=m]{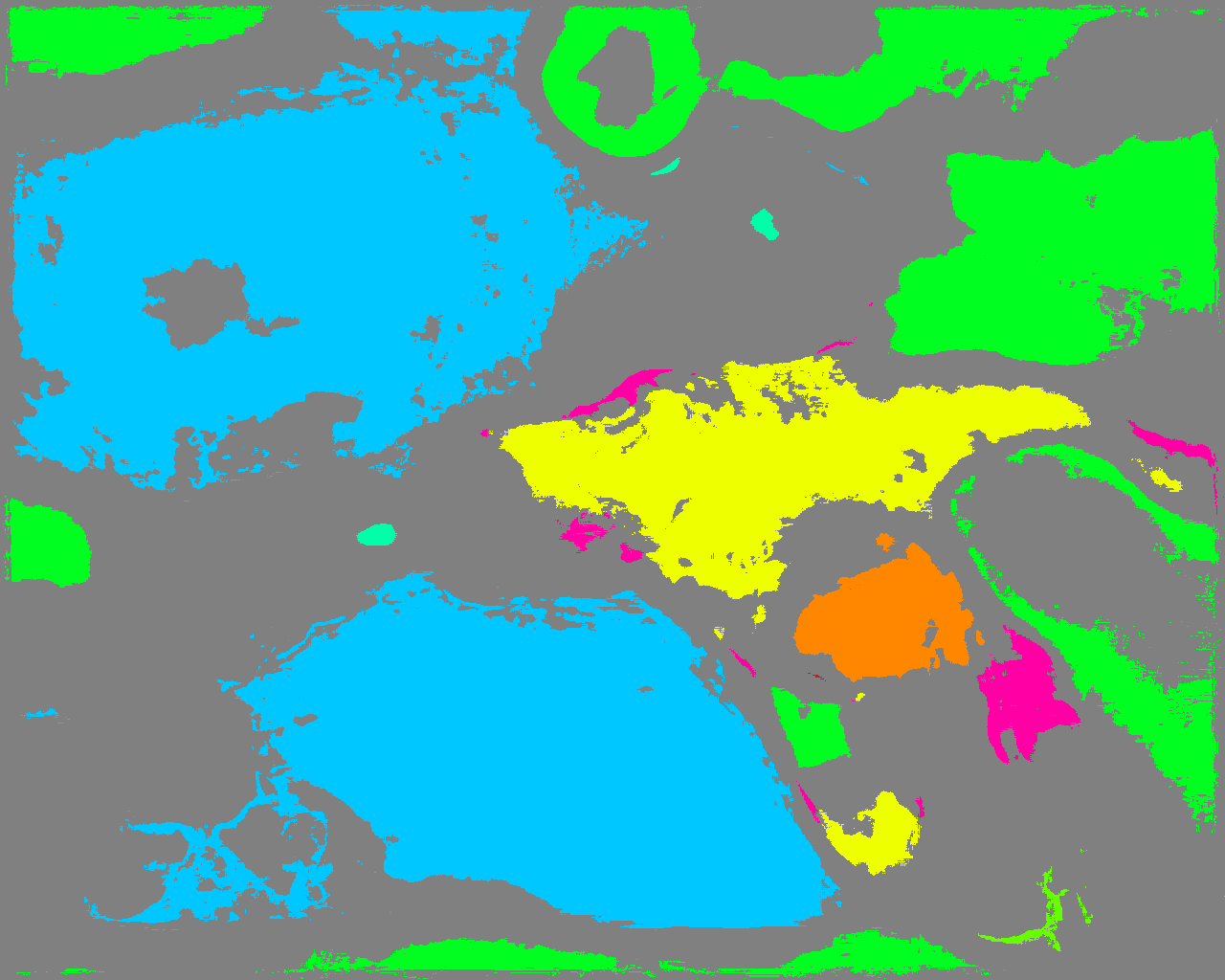} &
\includegraphics[width=.2\linewidth,valign=m]{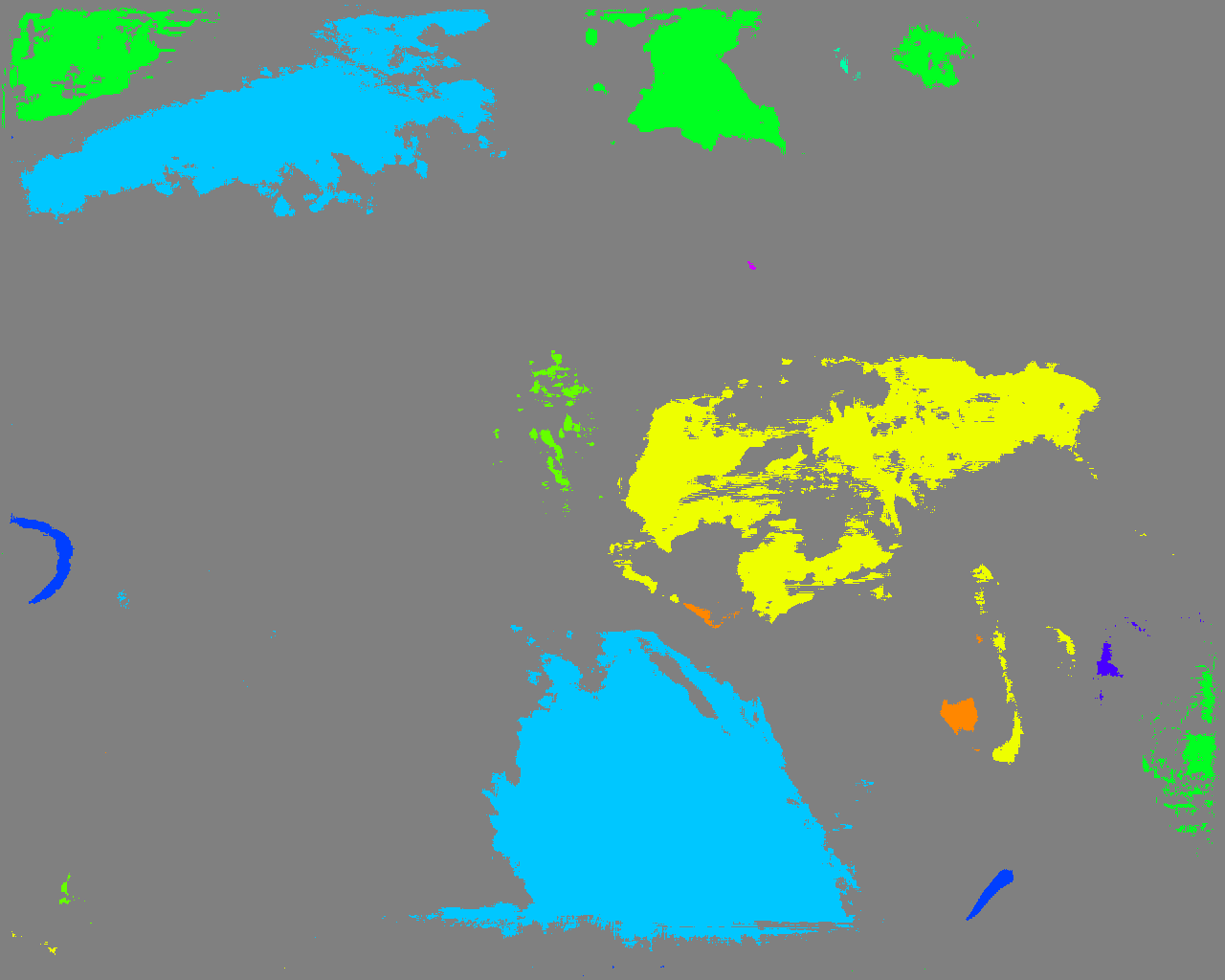}\\
\vspace{5pt}
\textbf{GODIN} & 
\includegraphics[width=.2\linewidth,valign=m]{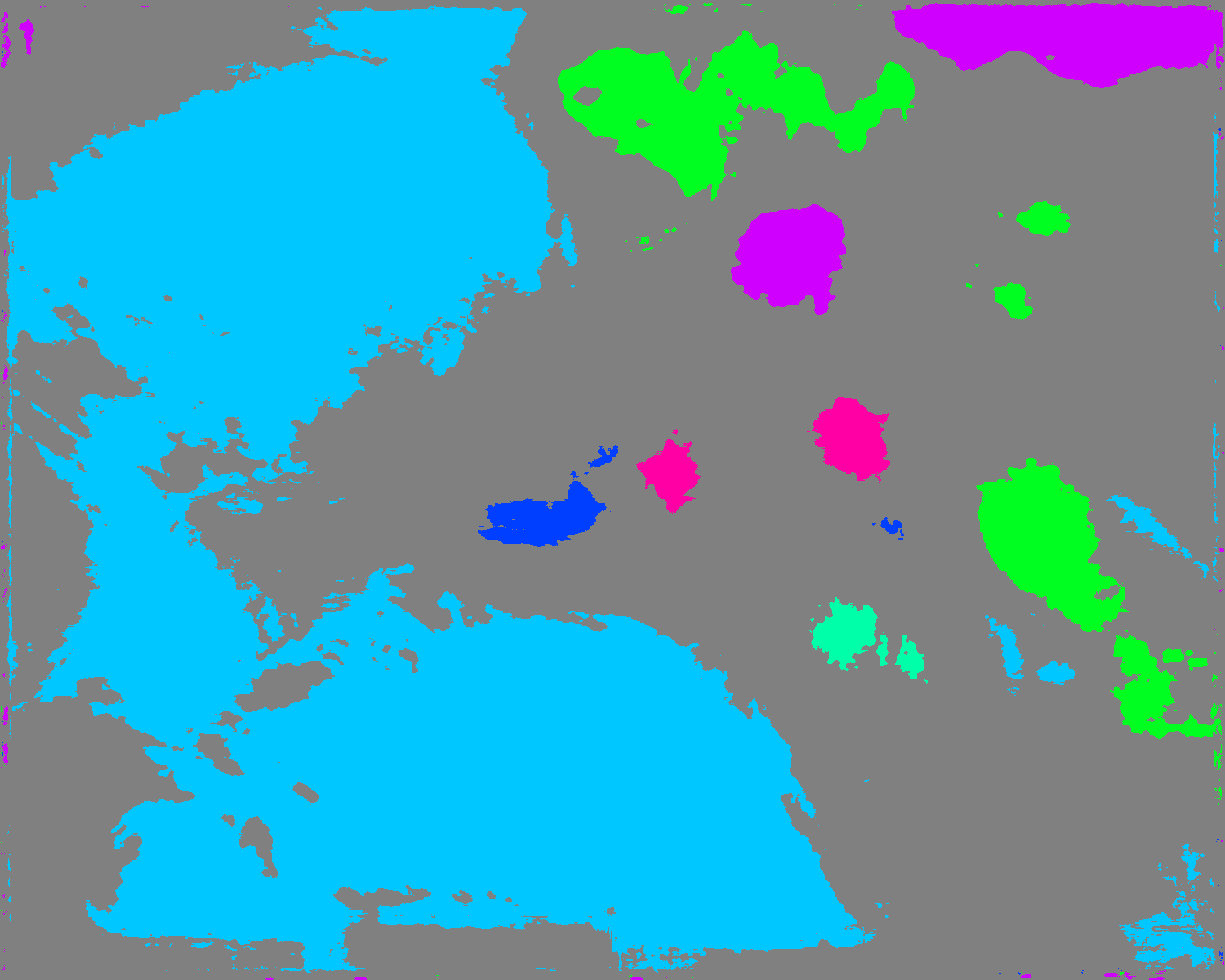} &
\includegraphics[width=.2\linewidth,valign=m]{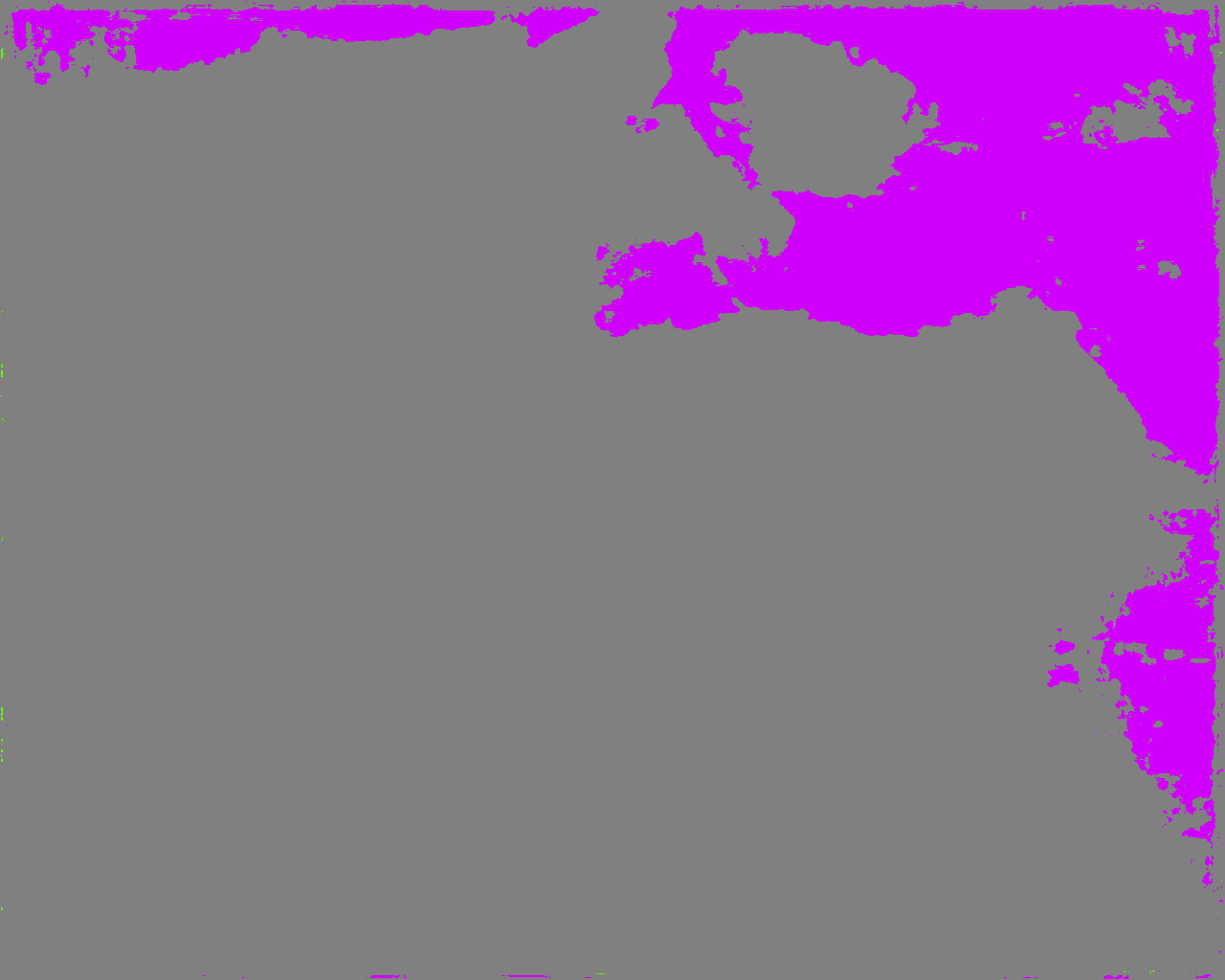} &
\includegraphics[width=.2\linewidth,valign=m]{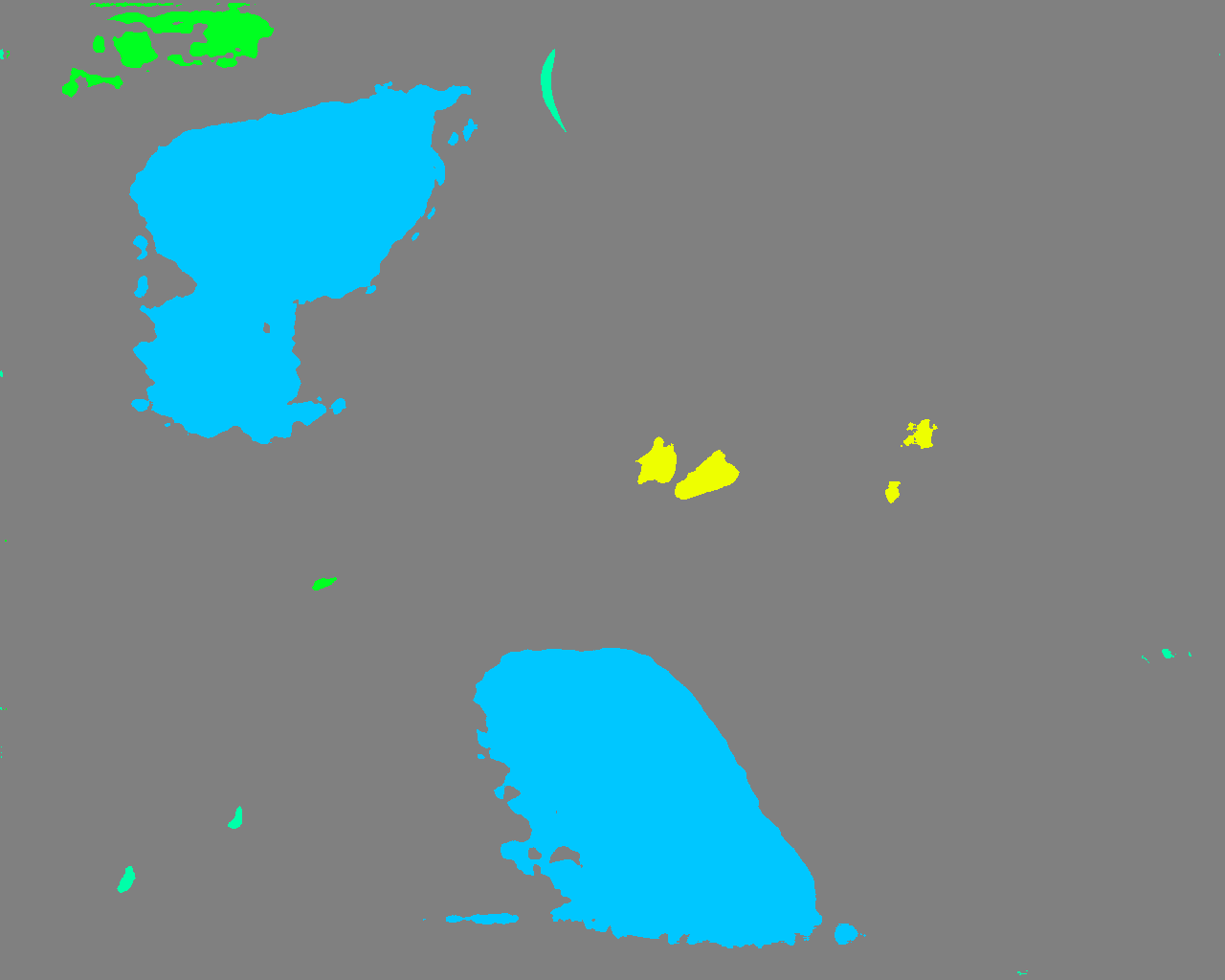} &
\includegraphics[width=.2\linewidth,valign=m]{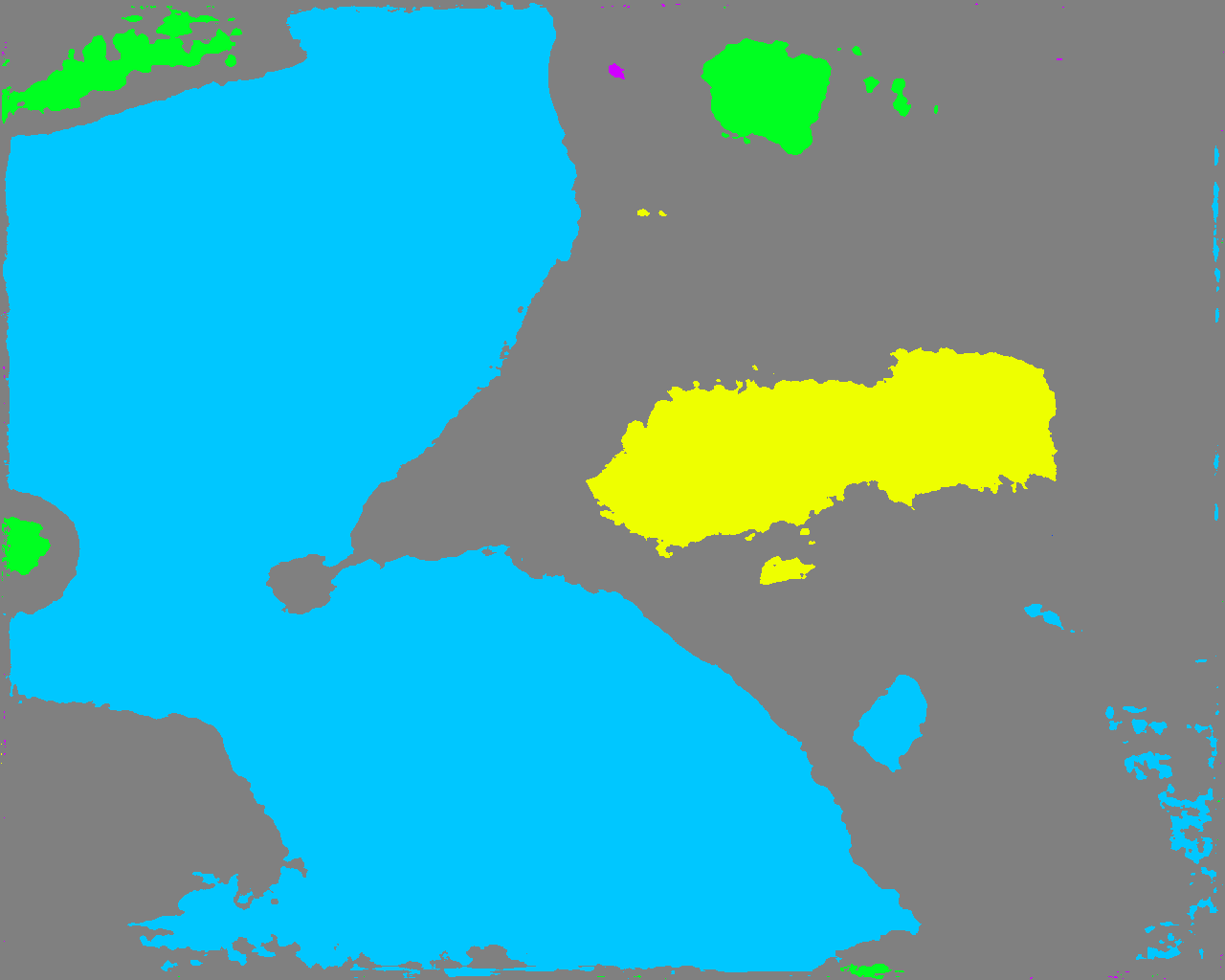}\\
&
\multicolumn{4}{c}{\includegraphics[width=16.1cm, height=0.5cm]{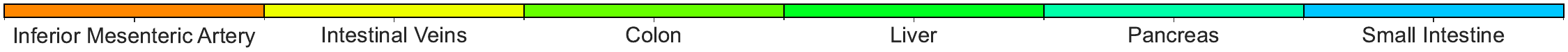}} \\
&
\multicolumn{4}{c}{\includegraphics[width=16.1cm, height=0.5cm]{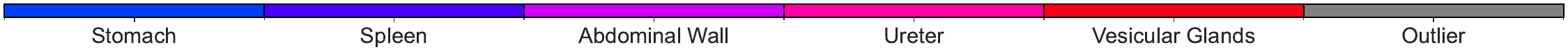}} \\
\end{tabular}
\caption{Qualitative result of first case from DSAD dataset. We show results of the same image from four class partitions (represented as $\CP_1$ to $\CP_4$). For each $\CP$, classes that are held-out are grouped as an extra outlier class for evaluation. We visualise and compare masks generated using different methods at threshold $\tau_{m}$. Baseline results at $\tau_0=0$ are added to represent result without outlier detection.}
\label{fig:segmentation_result_dsad_1}
\end{figure}

\begin{figure}[p]
\centering\footnotesize
\begin{tabular}{ccccc}
& \boldmath$\CP_1$ & \boldmath$\CP_2$ & \boldmath$\CP_3$ & \boldmath$\CP_4$ \\
\vspace{5pt}
\makecell[c]
{\textbf{Sparsely} \\ \textbf{annotated} \\ \textbf{ground truth}} & 
\includegraphics[width=.2\linewidth,valign=m]{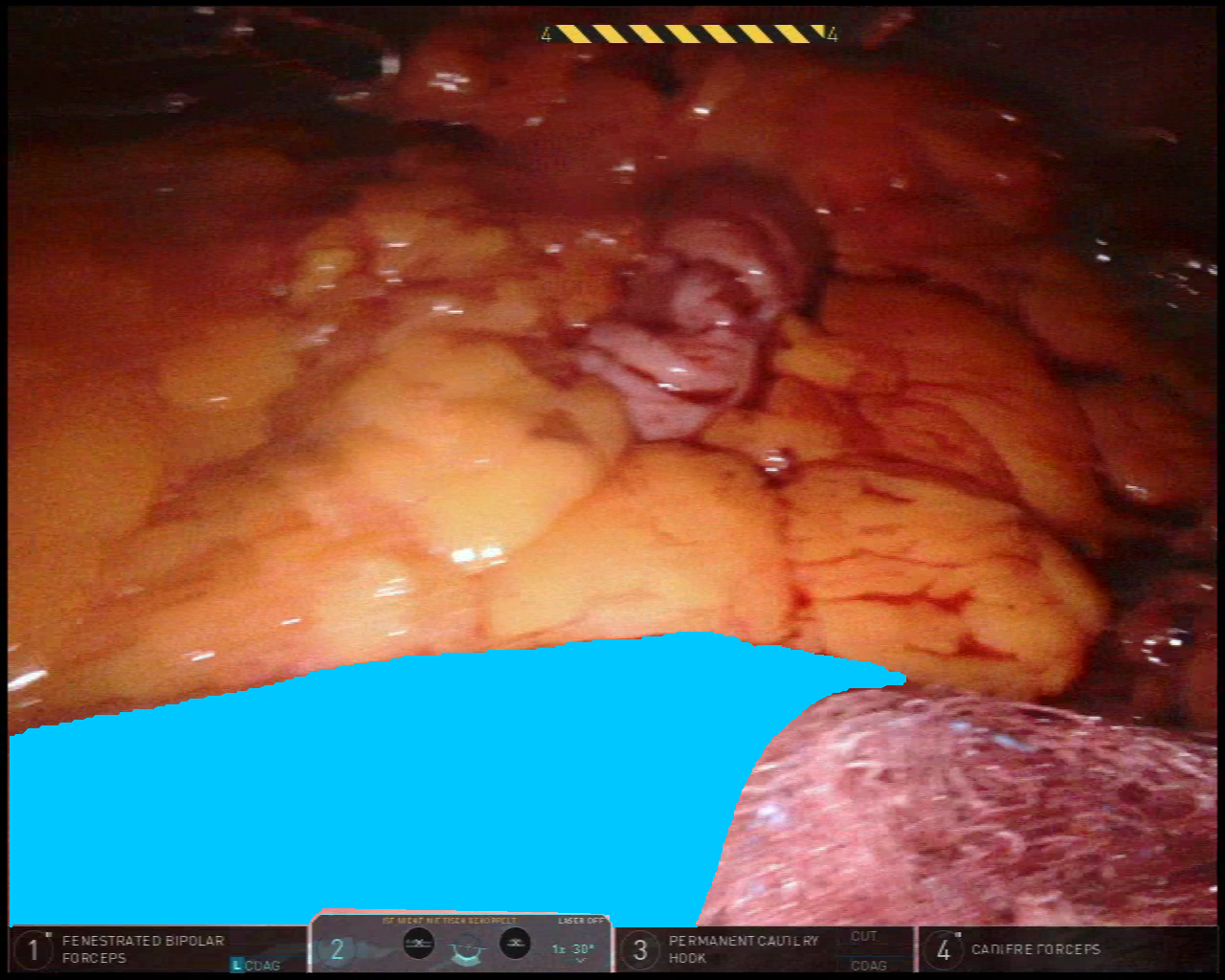} &
\includegraphics[width=.2\linewidth,valign=m]{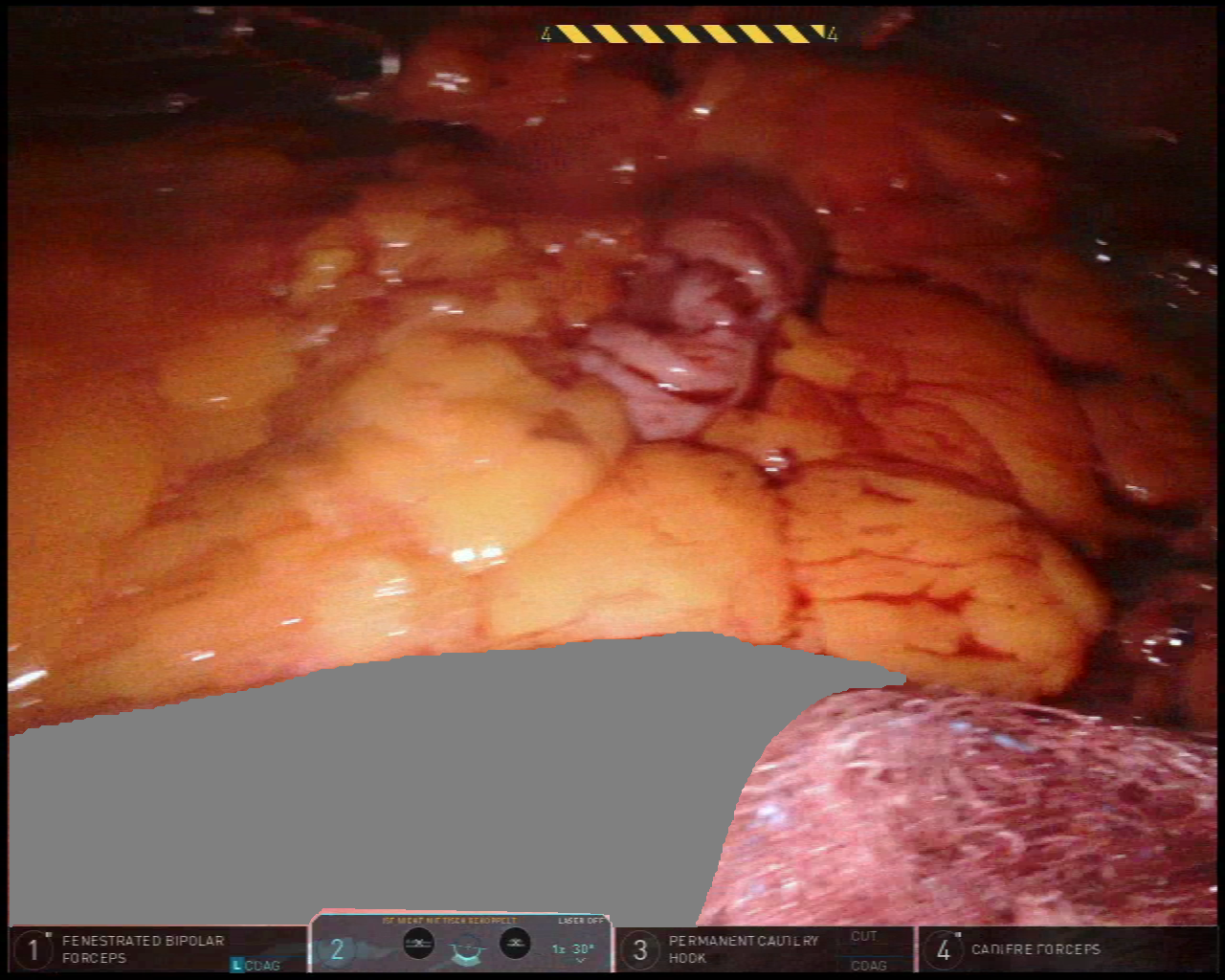} &
\includegraphics[width=.2\linewidth,valign=m]{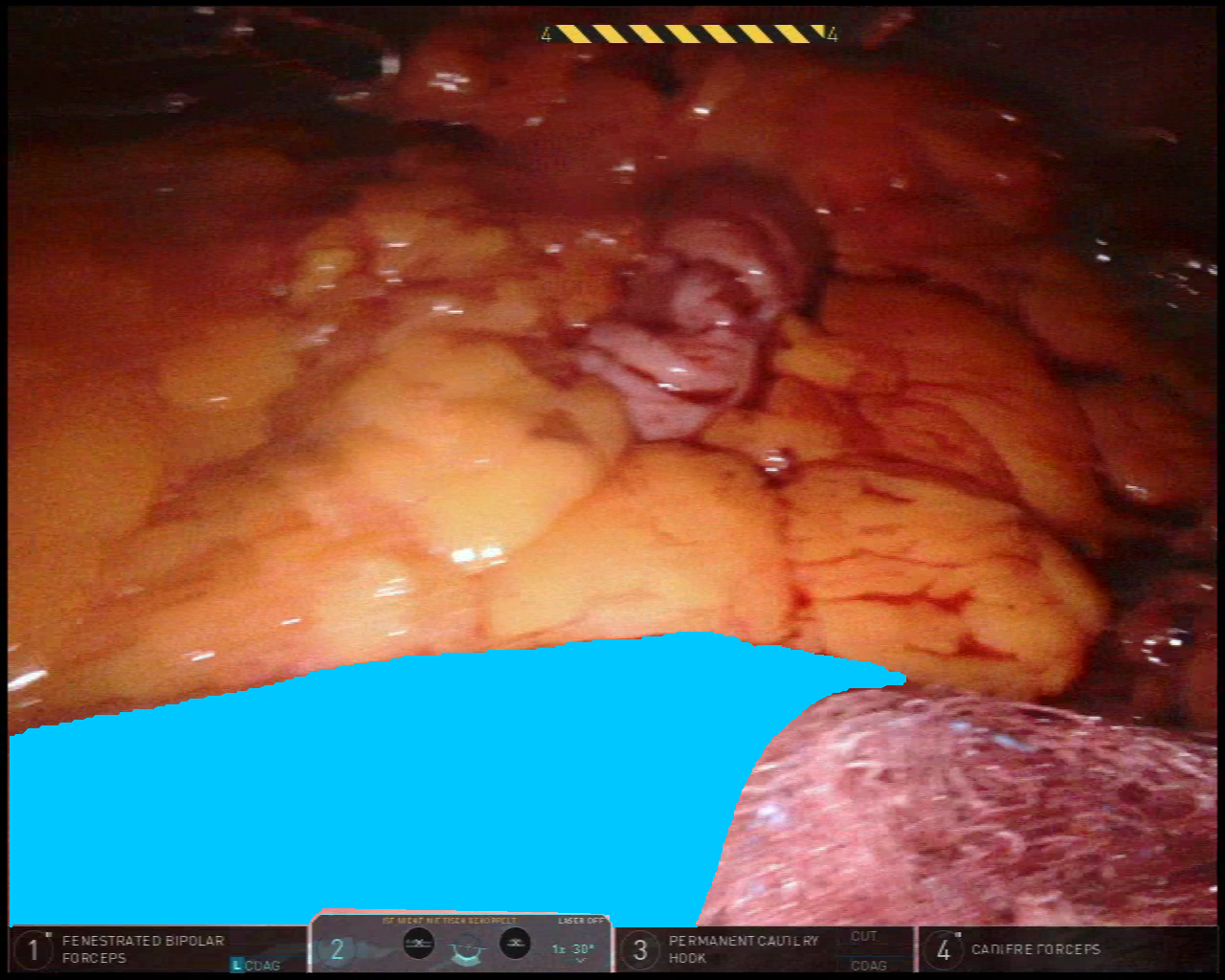} &
\includegraphics[width=.2\linewidth,valign=m]{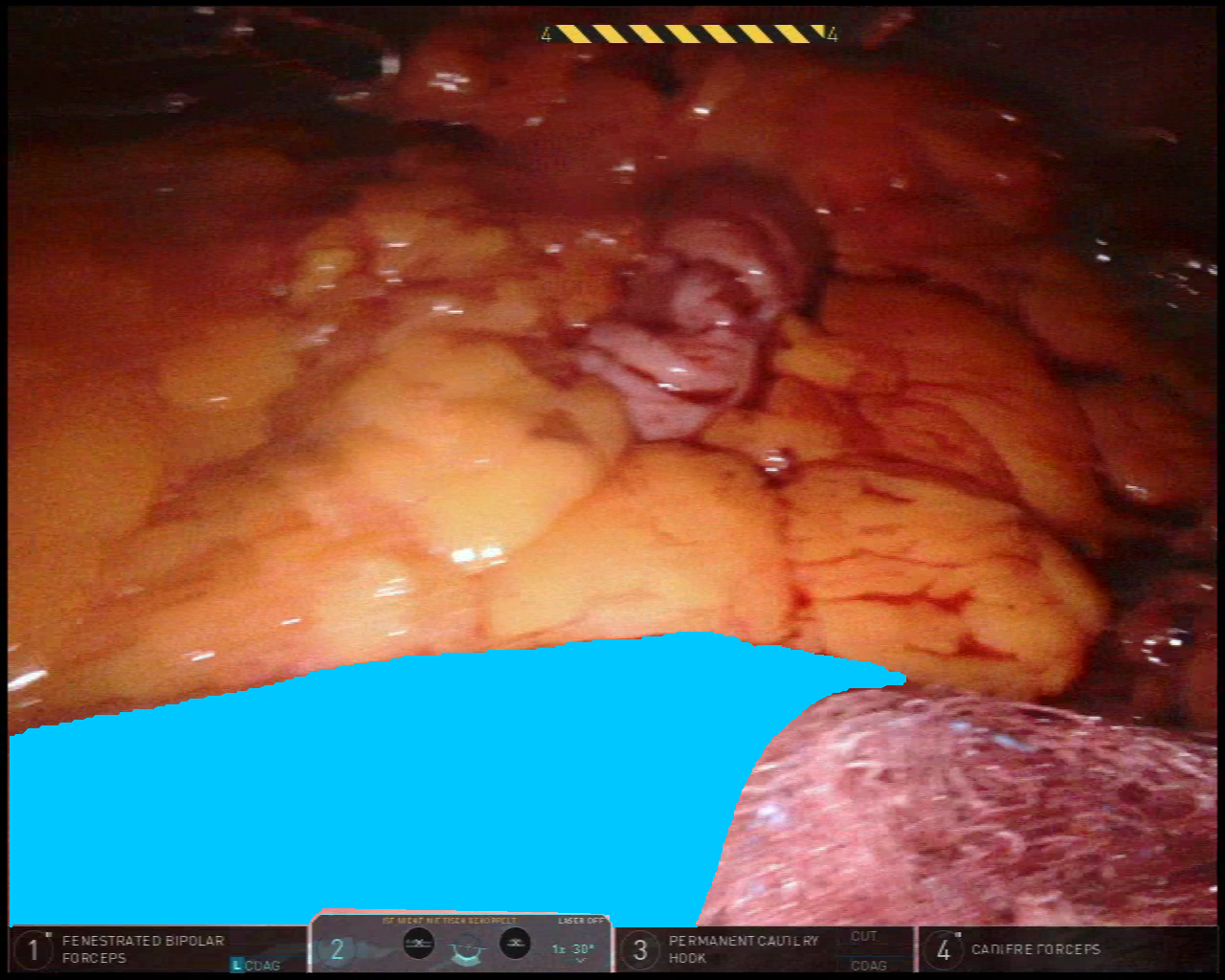}\\
\vspace{5pt}
\textbf{Baseline\ $(\tau_{0}=0$)} & 
\includegraphics[width=.2\linewidth,valign=m]{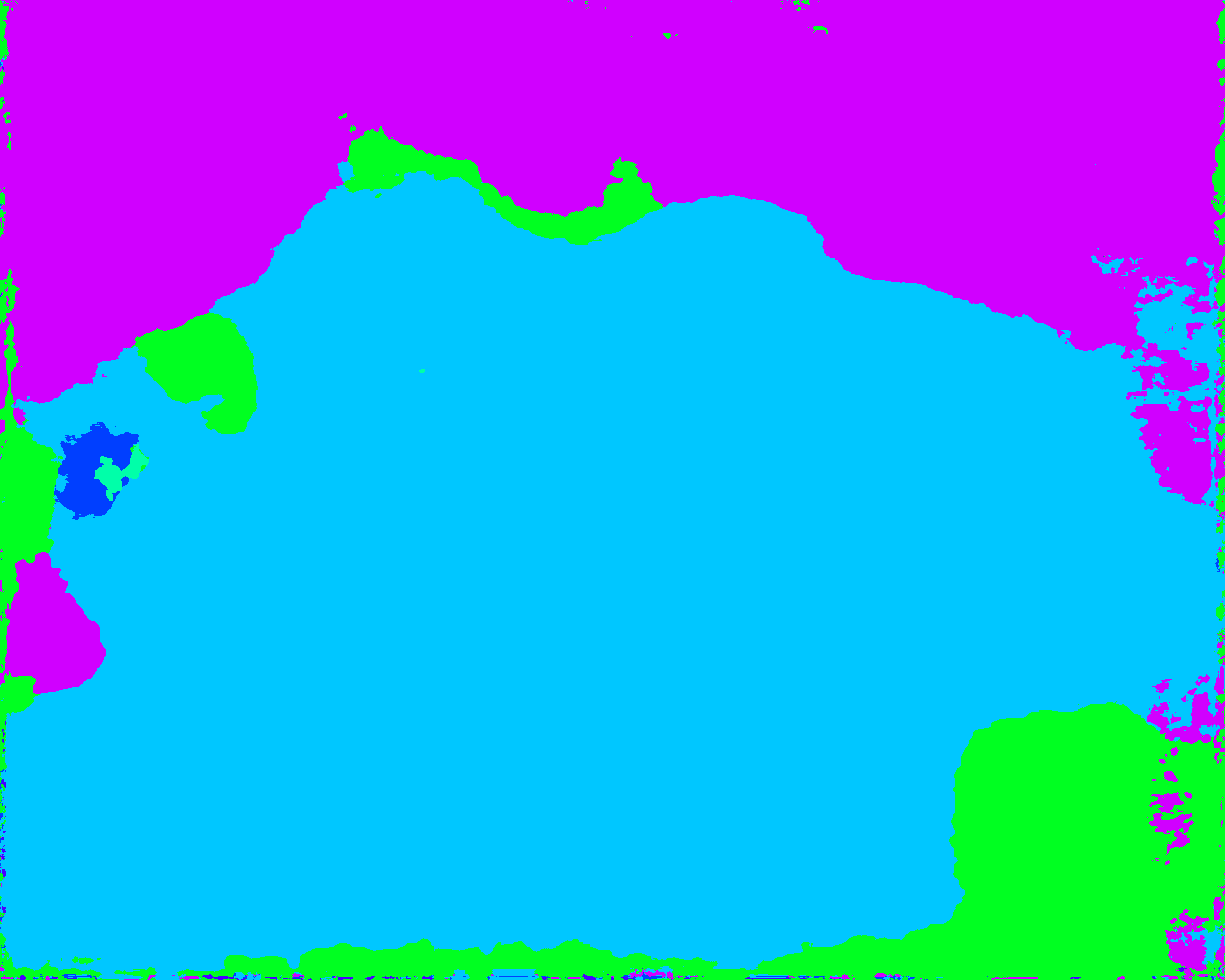} &
\includegraphics[width=.2\linewidth,valign=m]{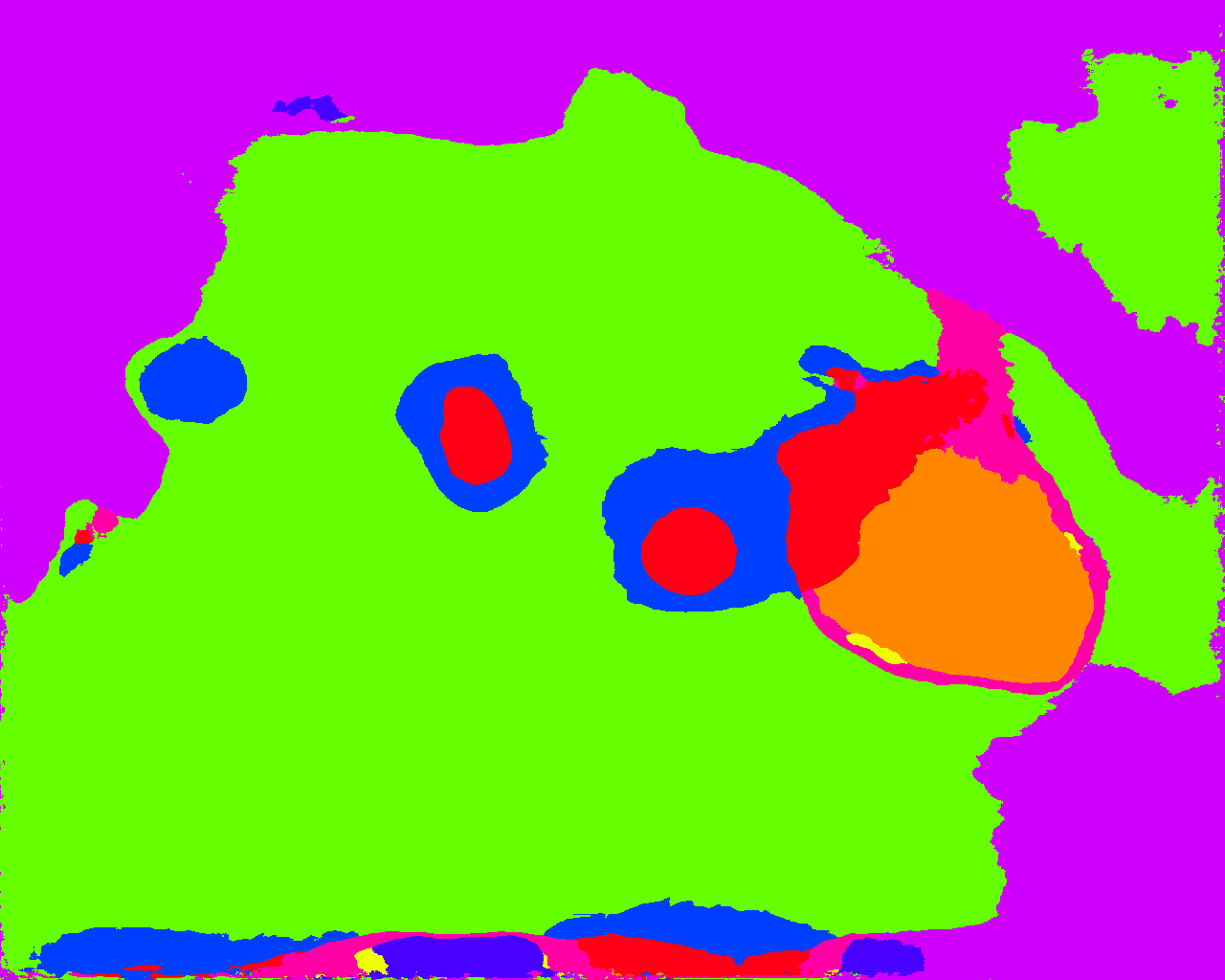} &
\includegraphics[width=.2\linewidth,valign=m]{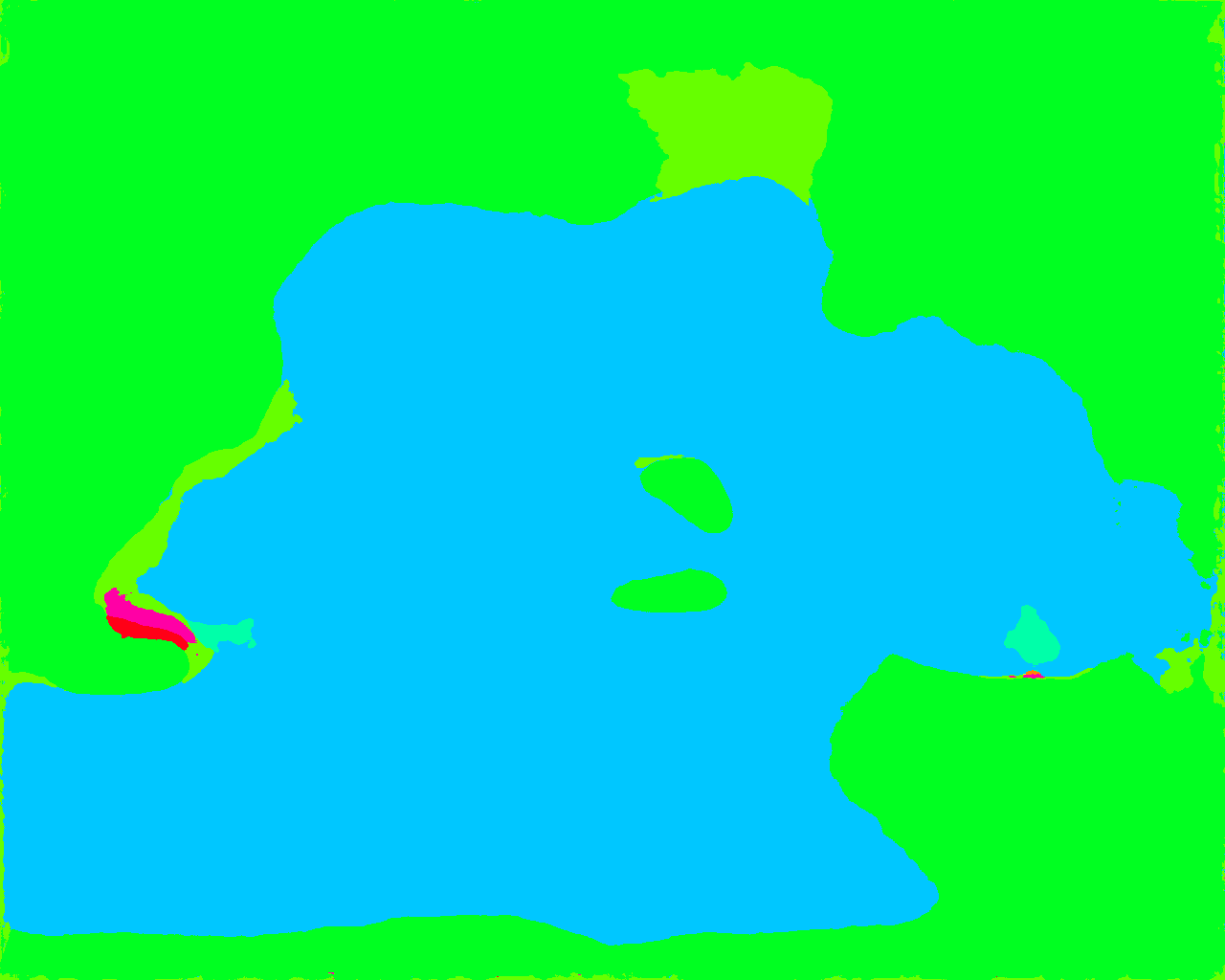} &
\includegraphics[width=.2\linewidth,valign=m]{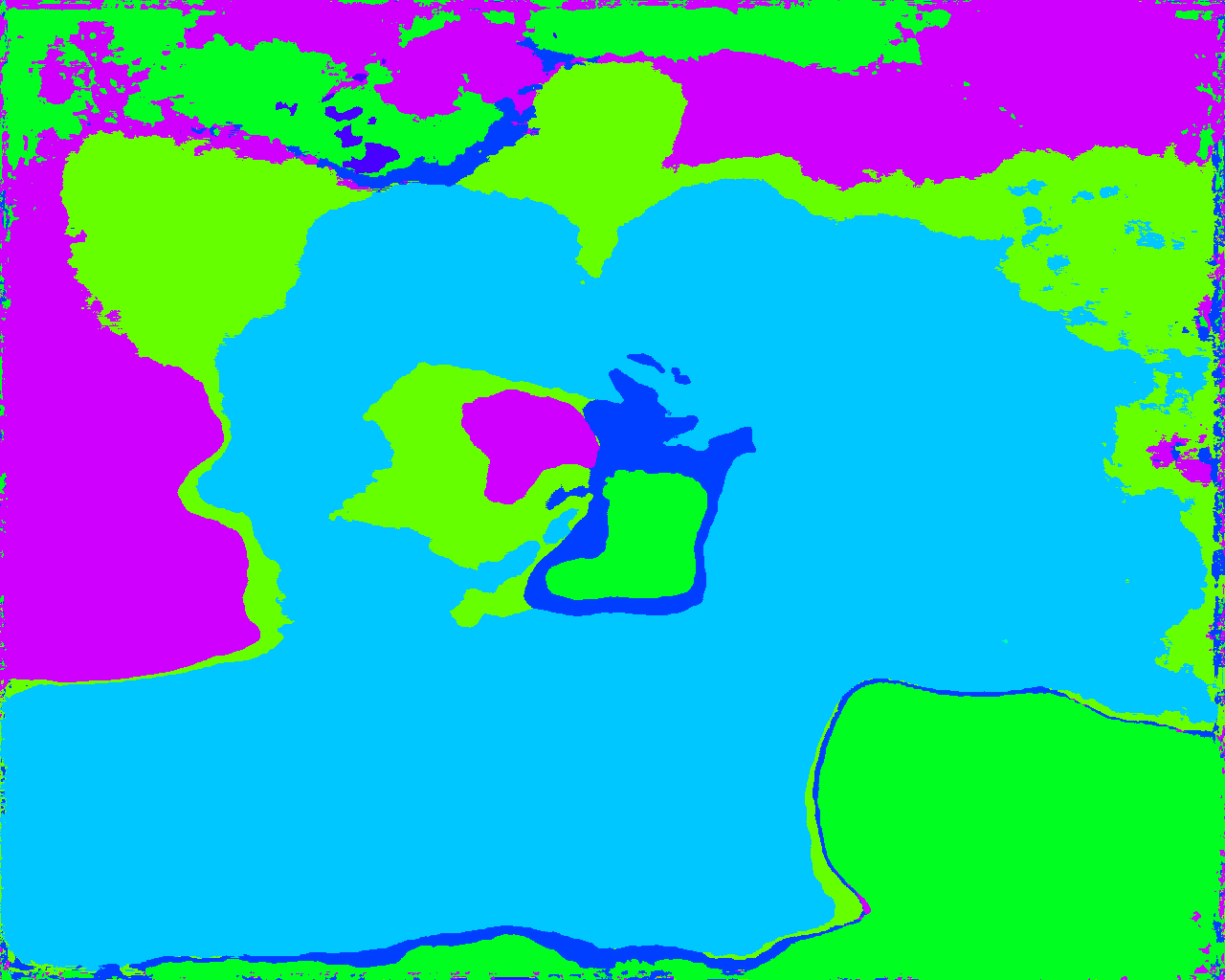}\\
\vspace{5pt}
\textbf{Baseline\ $(\tau_{m}$)} & 
\includegraphics[width=.2\linewidth,valign=m]{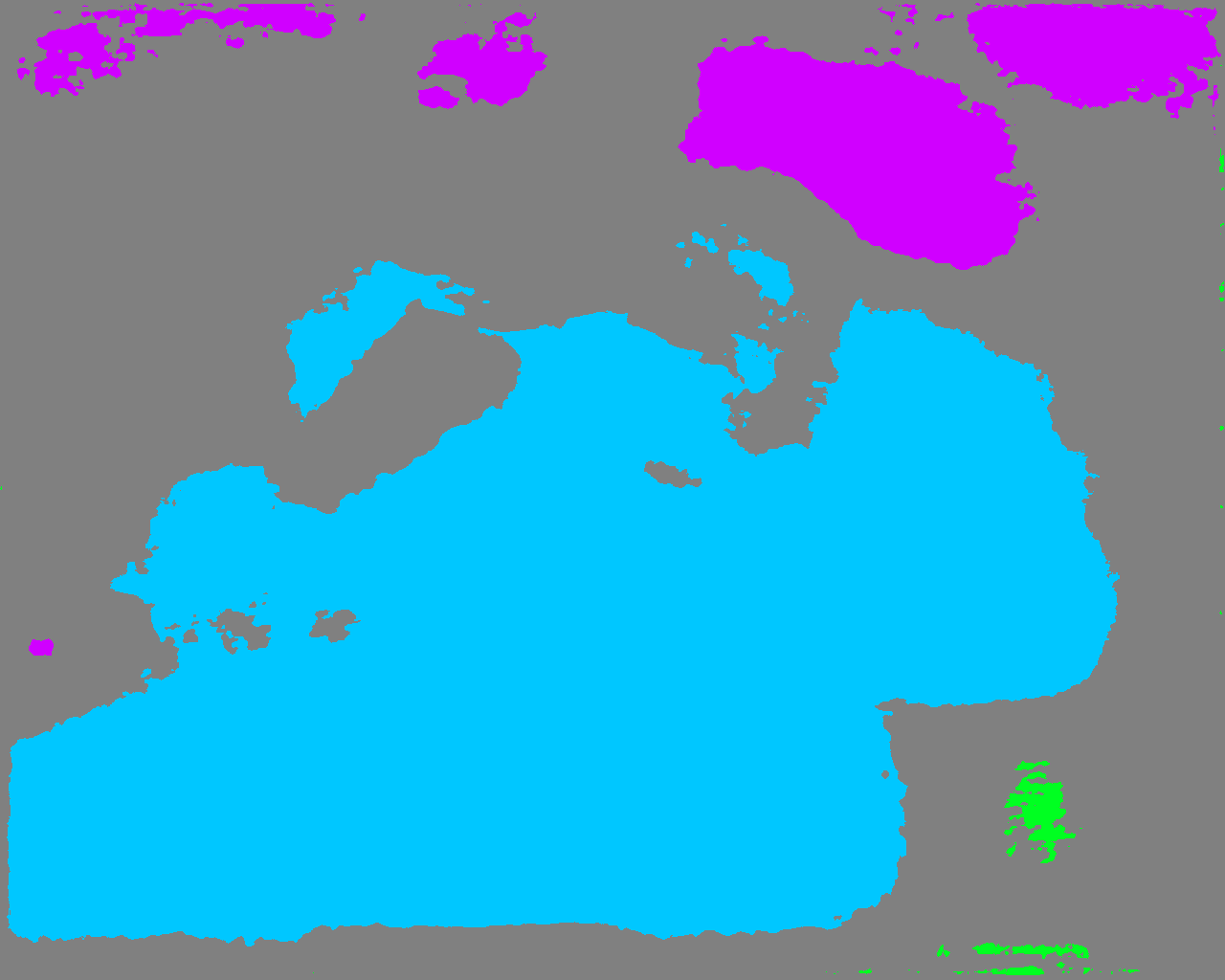} &
\includegraphics[width=.2\linewidth,valign=m]{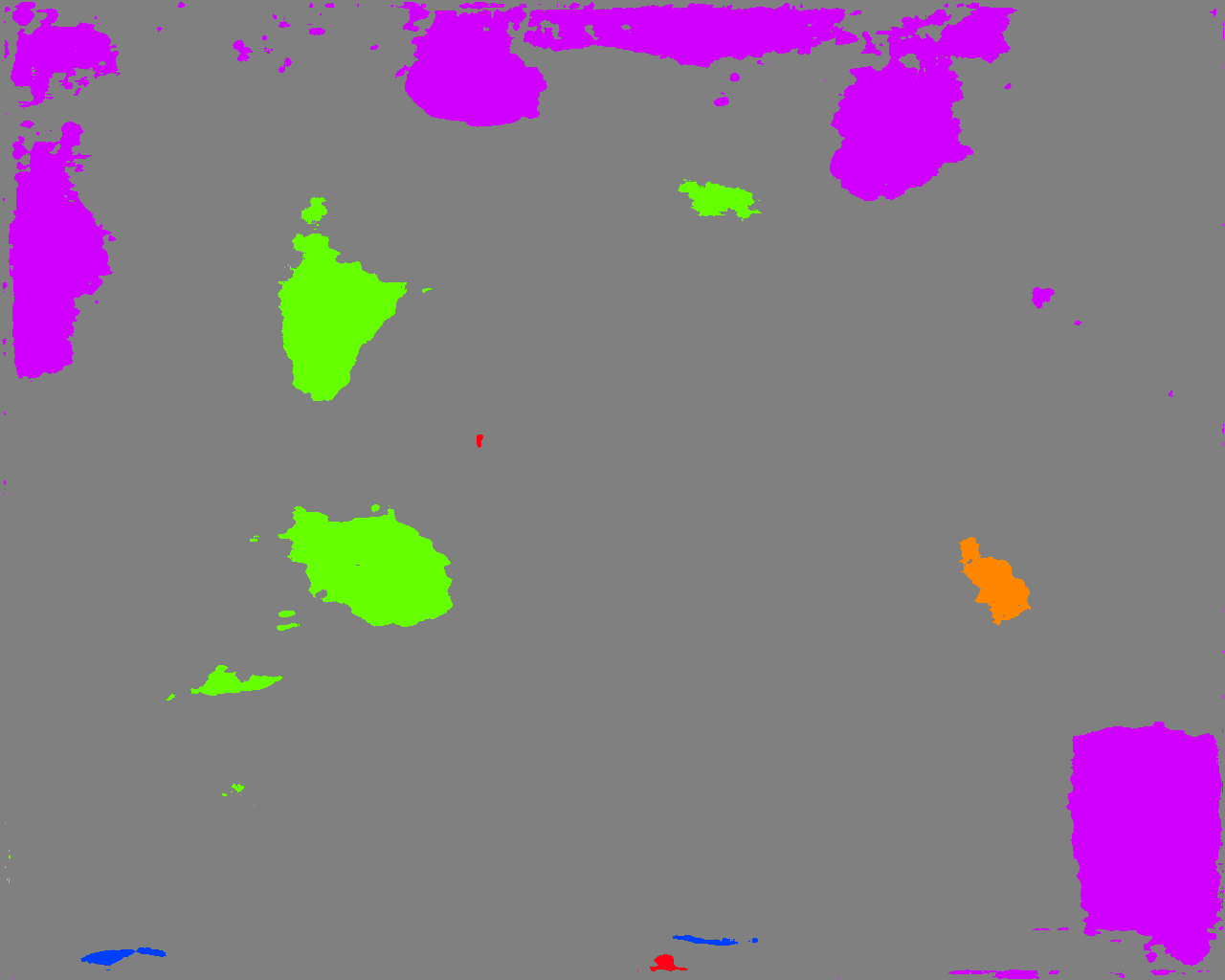} &
\includegraphics[width=.2\linewidth,valign=m]{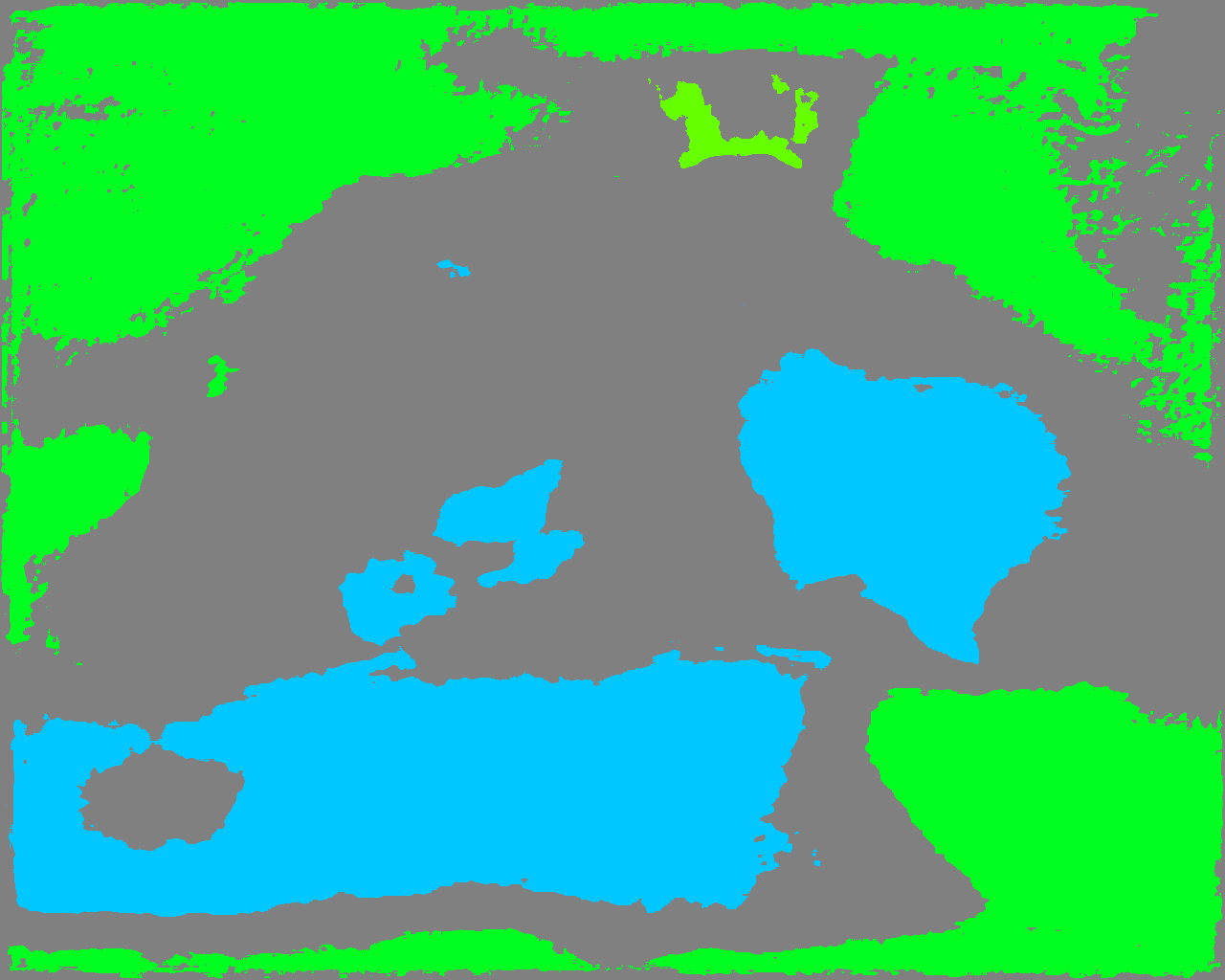} &
\includegraphics[width=.2\linewidth,valign=m]{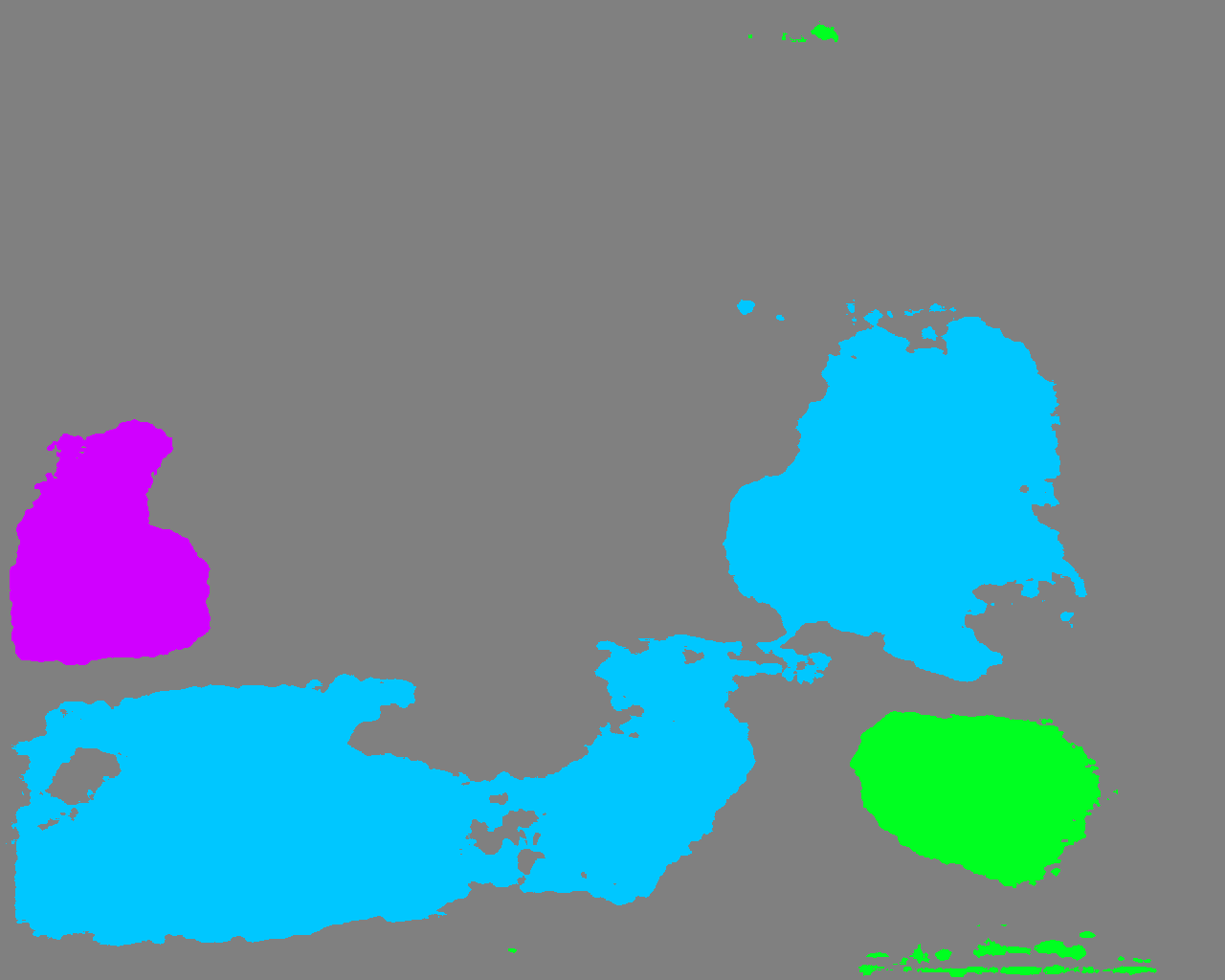}\\
\vspace{5pt}
\textbf{ODIN} & 
\includegraphics[width=.2\linewidth,valign=m]{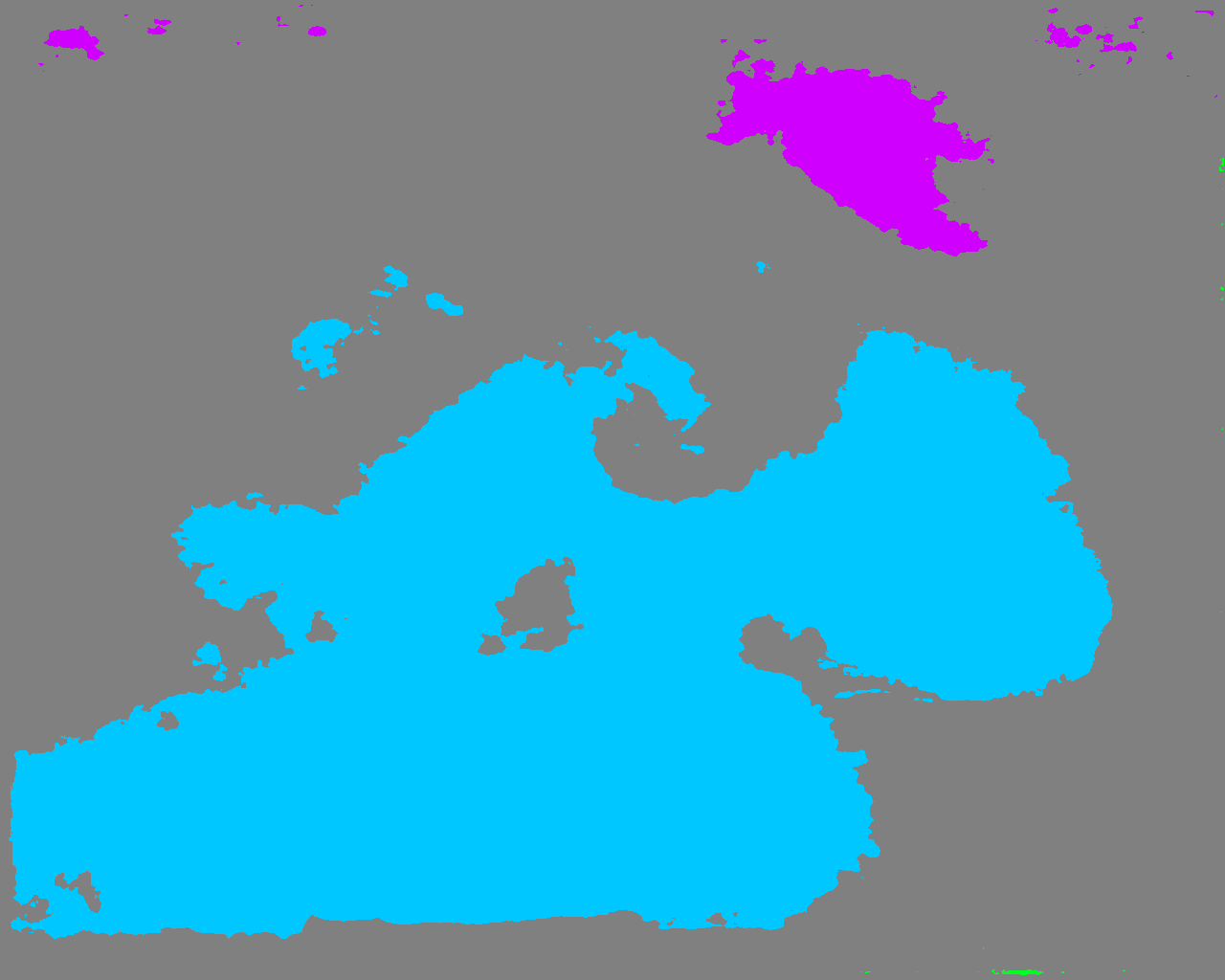} &
\includegraphics[width=.2\linewidth,valign=m]{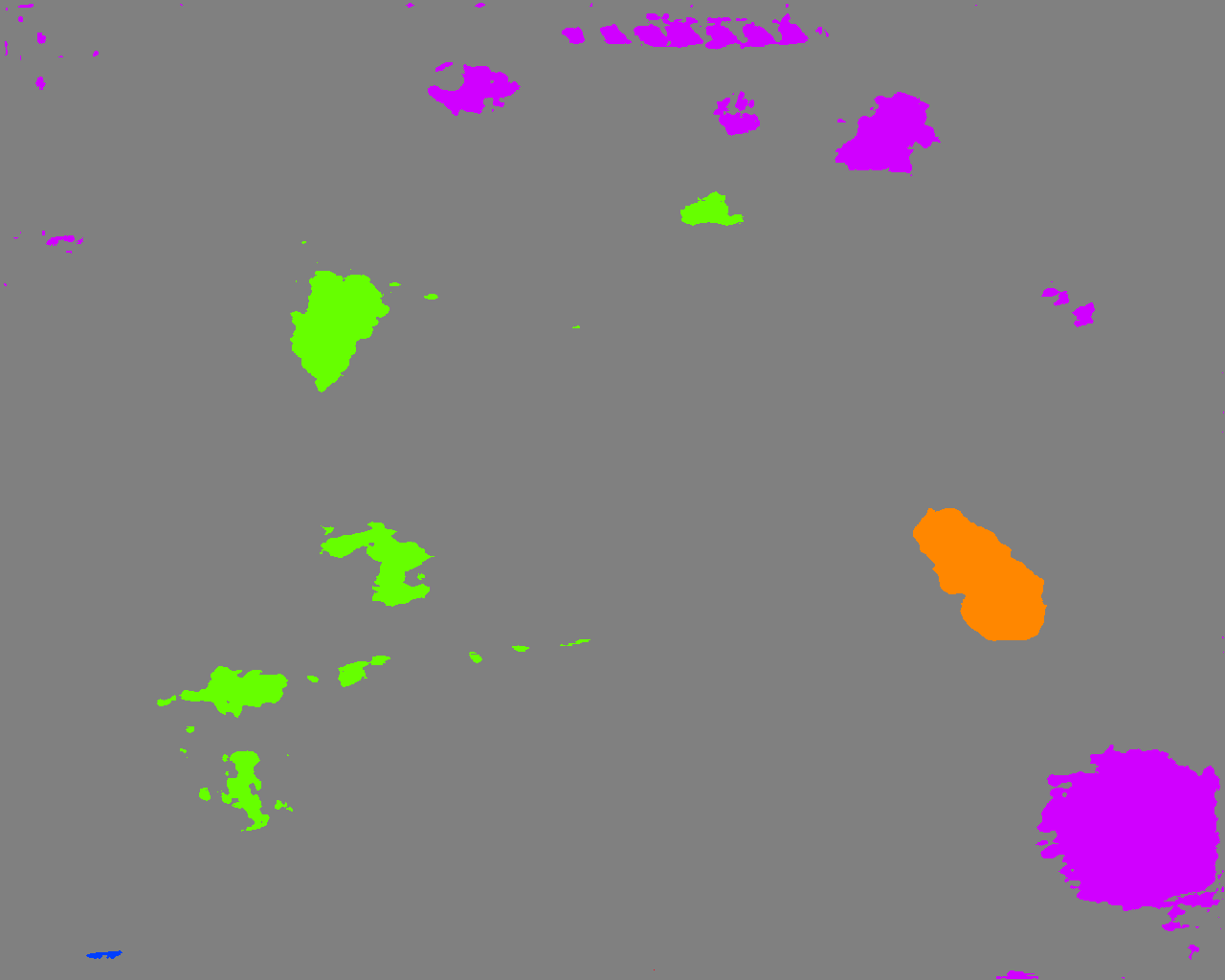} &
\includegraphics[width=.2\linewidth,valign=m]{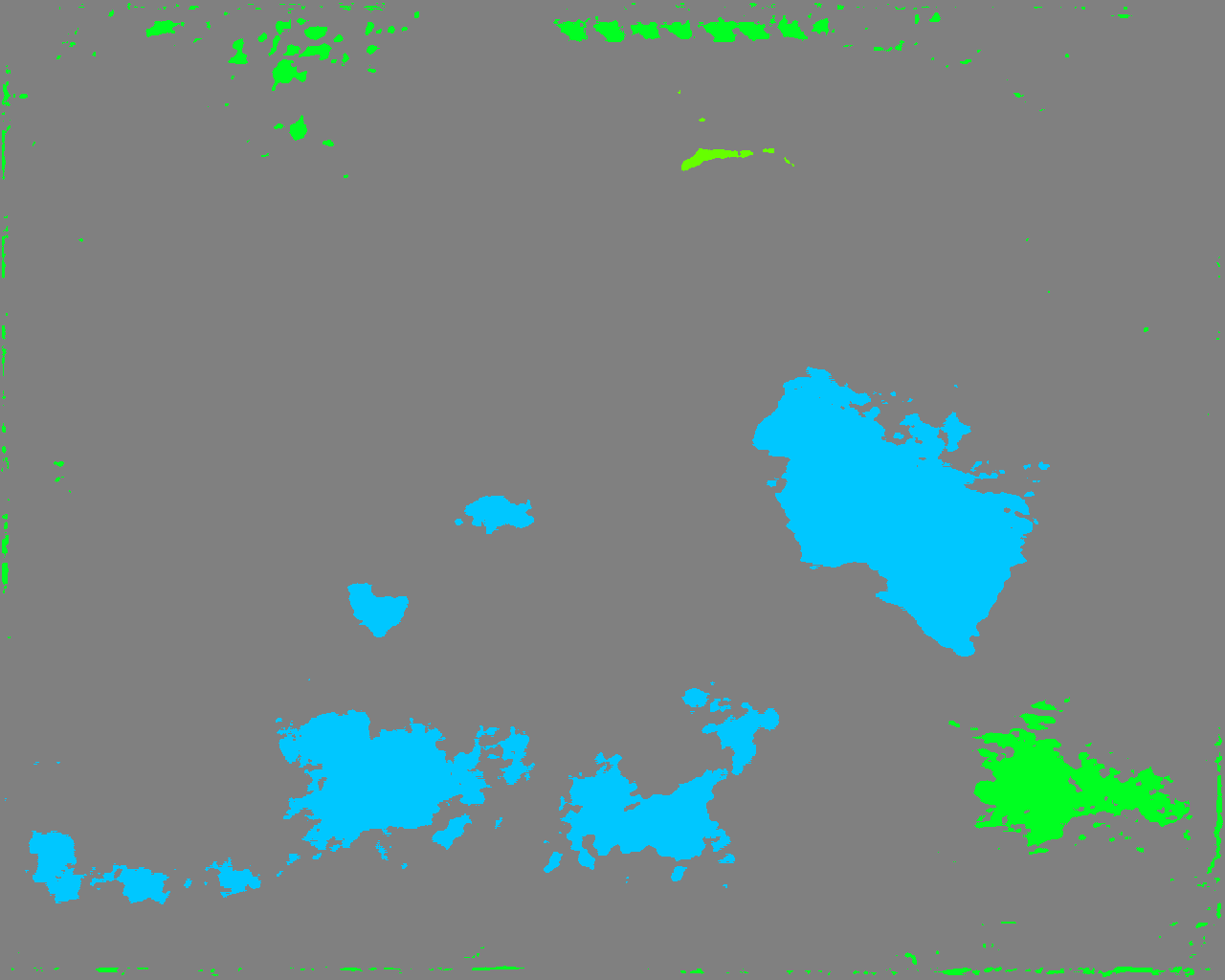} &
\includegraphics[width=.2\linewidth,valign=m]{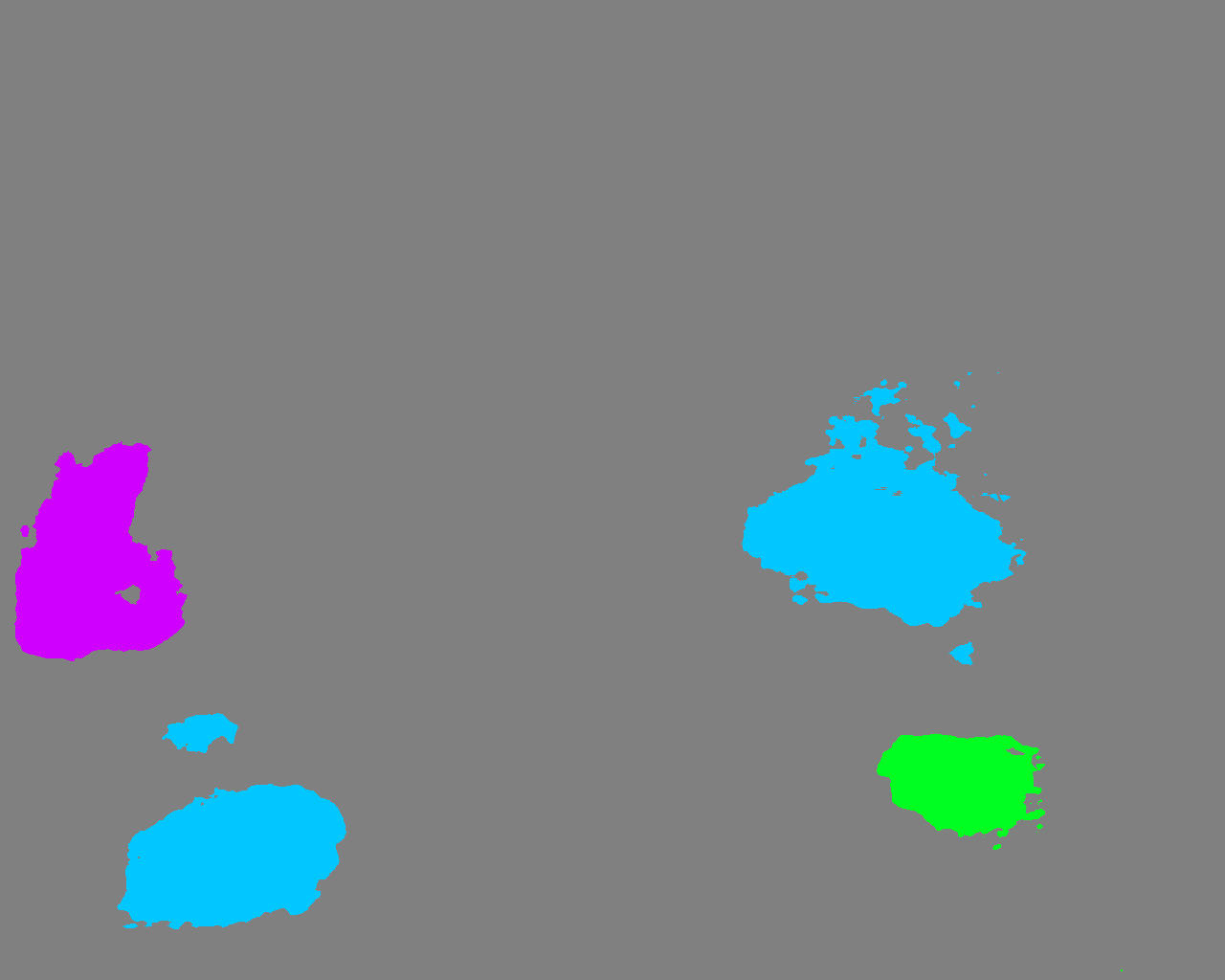}\\
\vspace{5pt}
\textbf{Mahalanobis} & 
\includegraphics[width=.2\linewidth,valign=m]{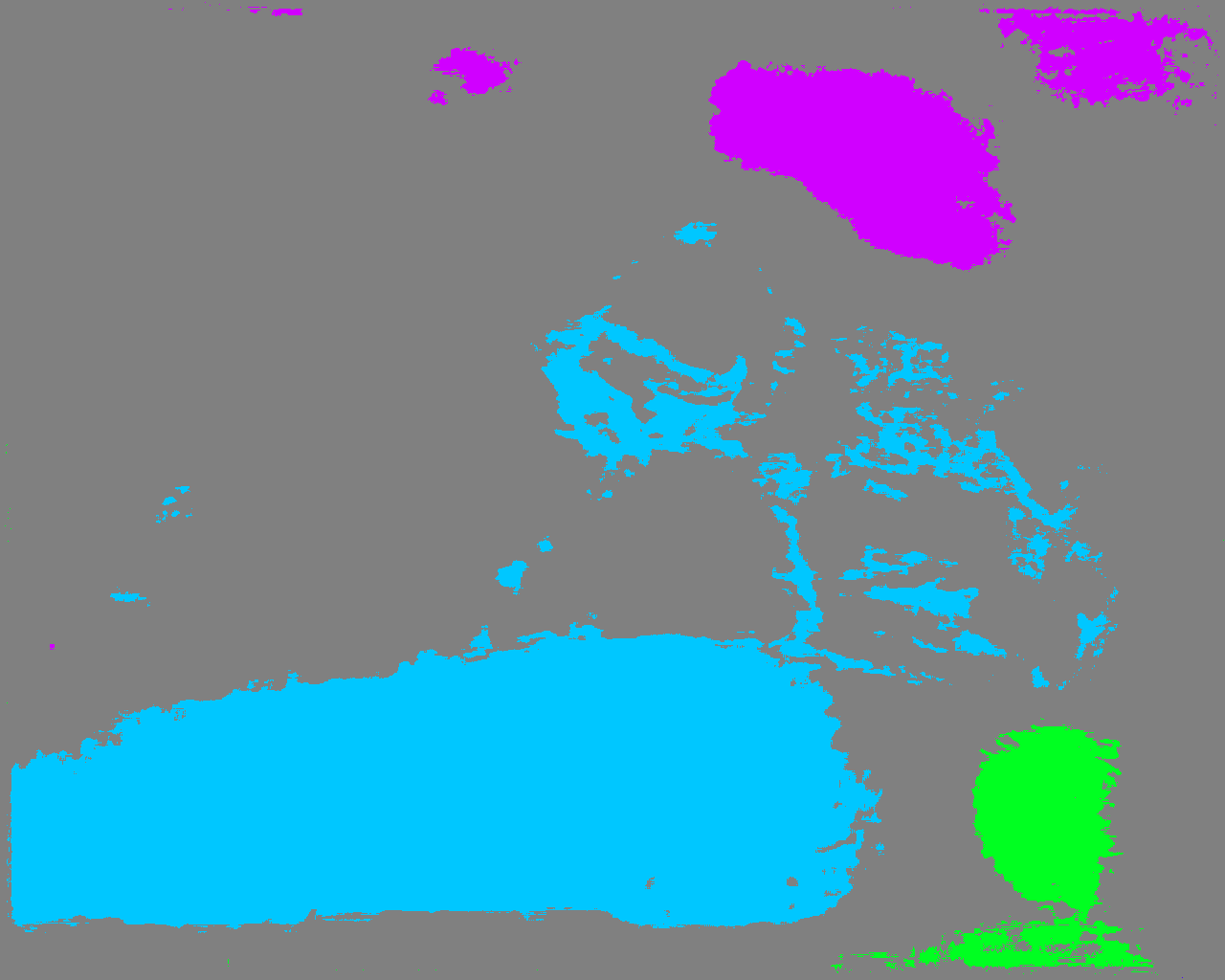} &
\includegraphics[width=.2\linewidth,valign=m]{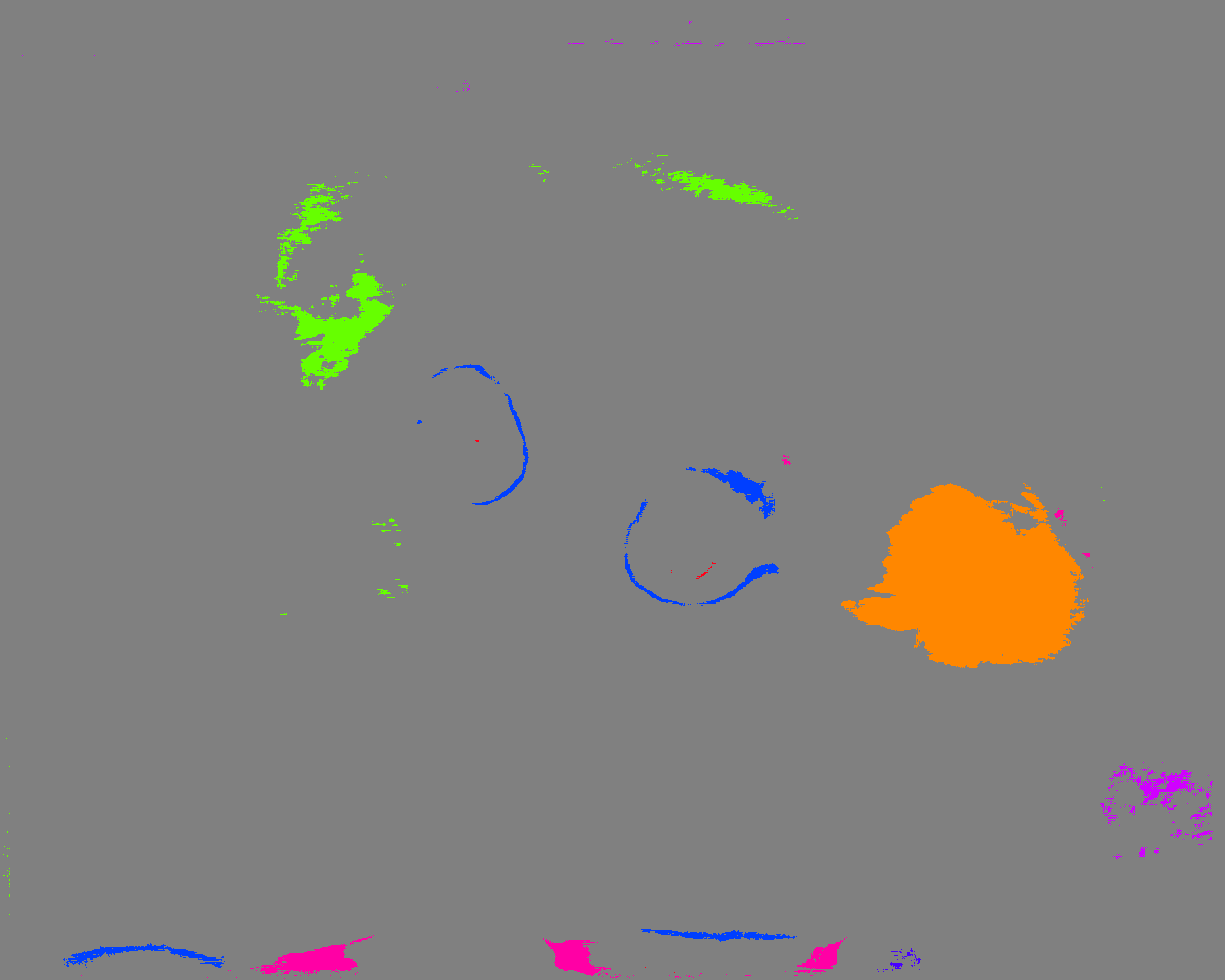} &
\includegraphics[width=.2\linewidth,valign=m]{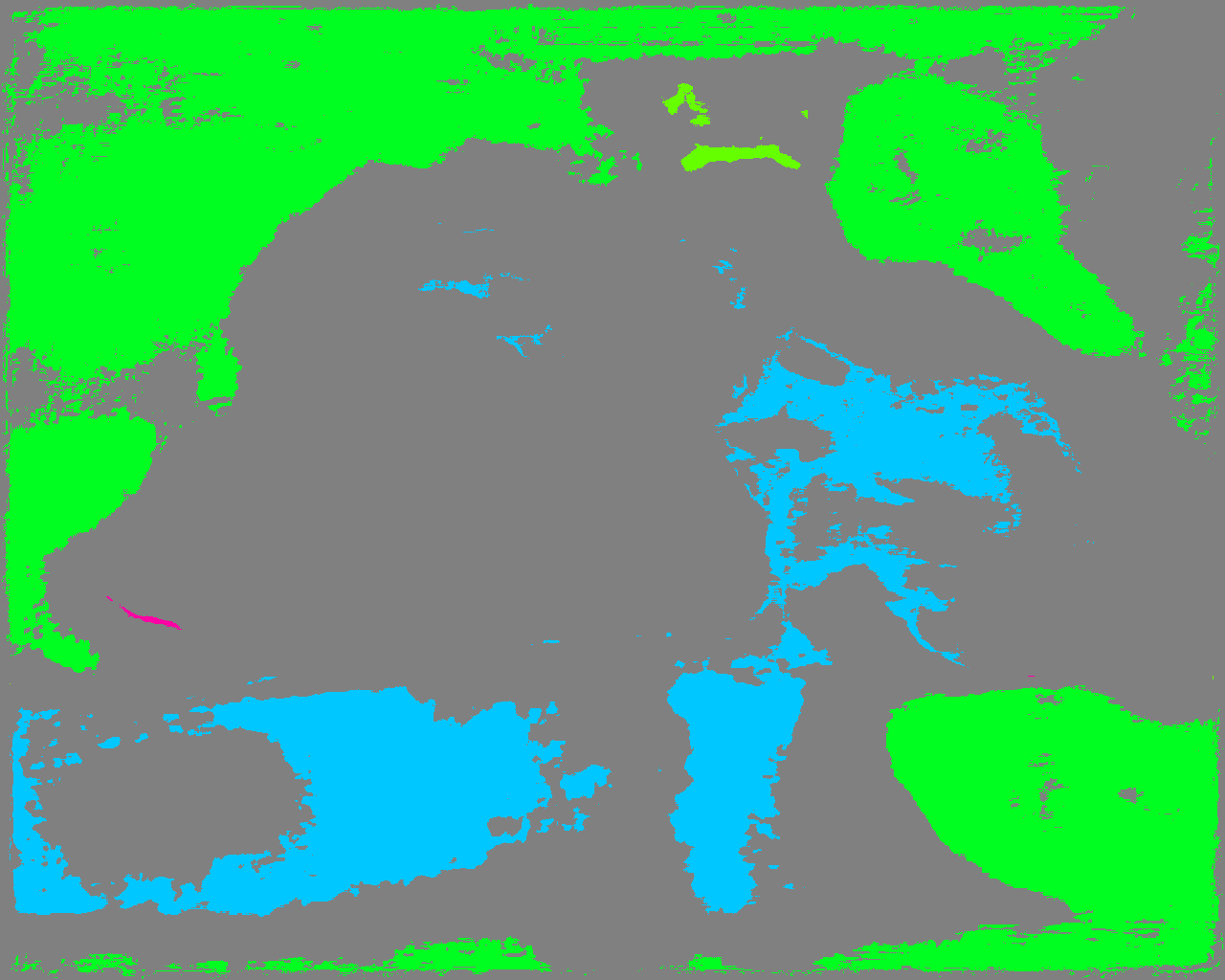} &
\includegraphics[width=.2\linewidth,valign=m]{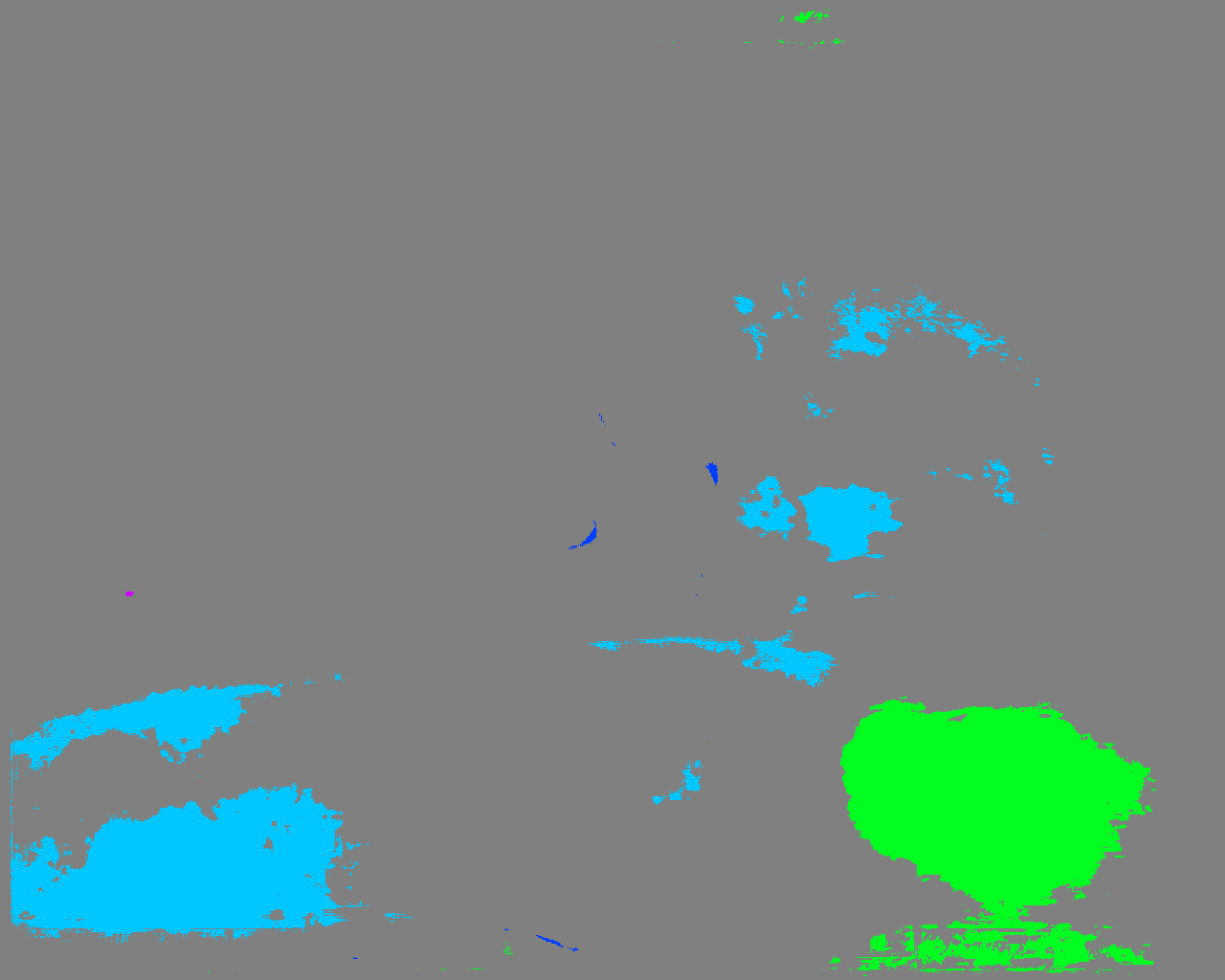}\\
\vspace{5pt}
\textbf{GODIN} & 
\includegraphics[width=.2\linewidth,valign=m]{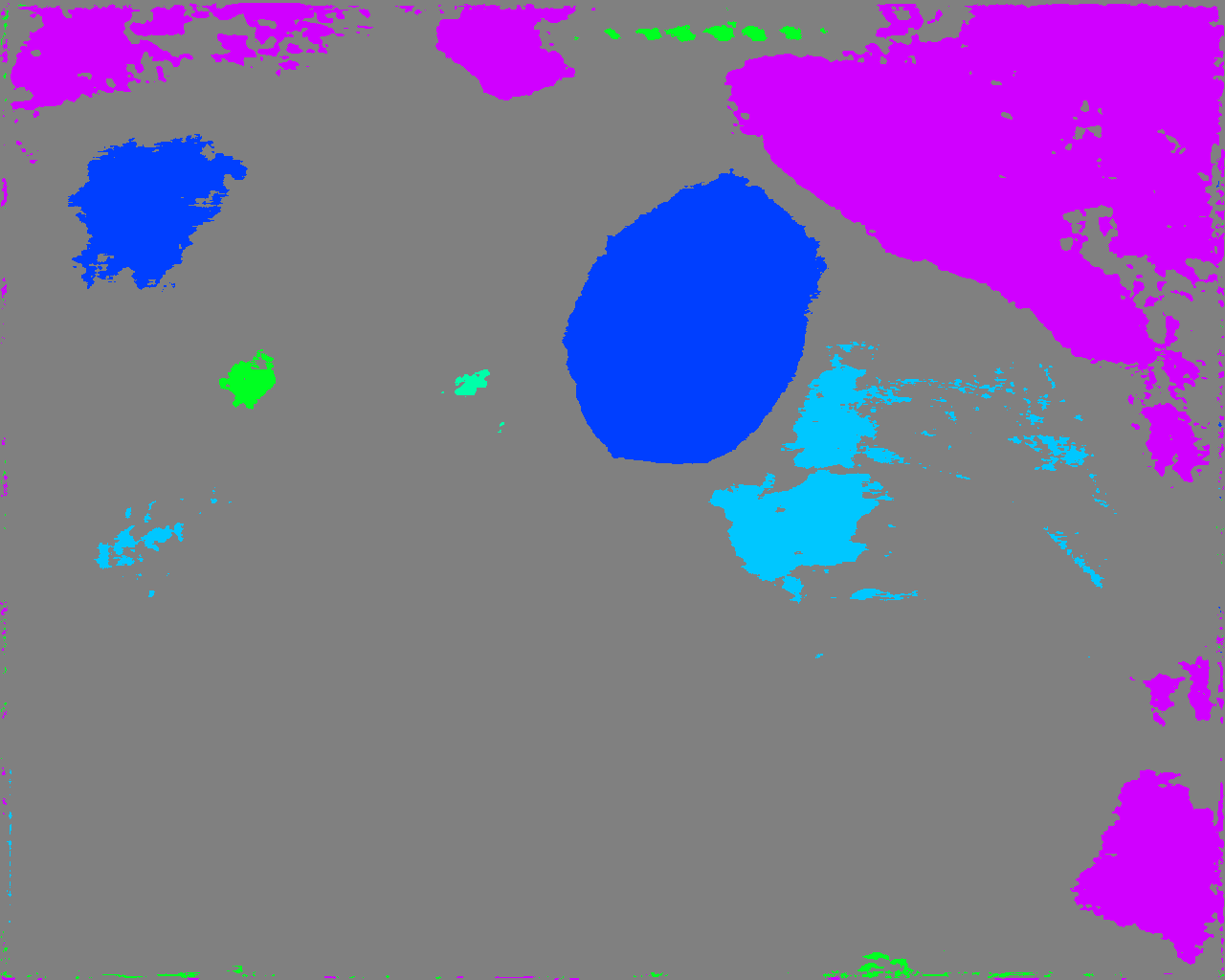} &
\includegraphics[width=.2\linewidth,valign=m]{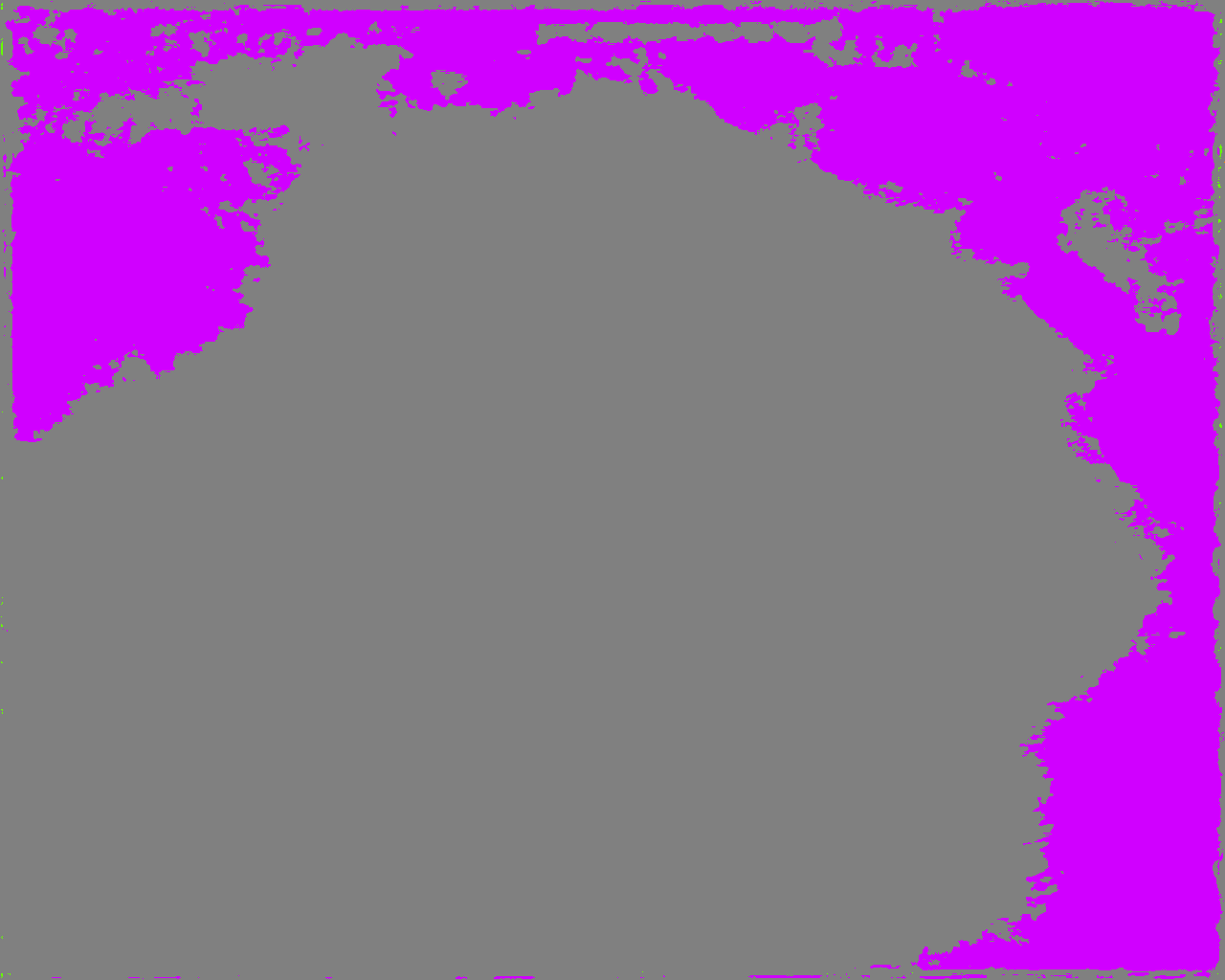} &
\includegraphics[width=.2\linewidth,valign=m]{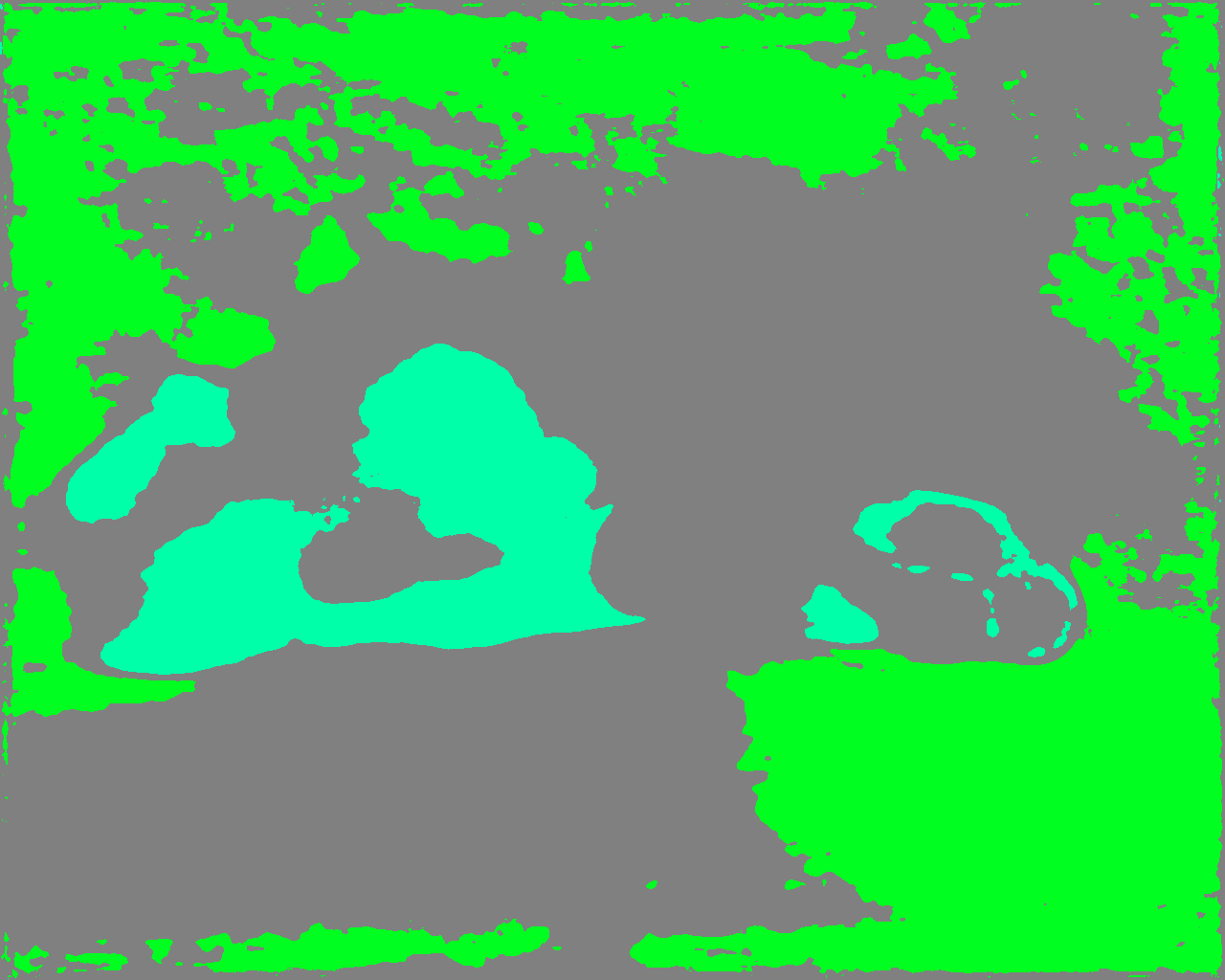} &
\includegraphics[width=.2\linewidth,valign=m]{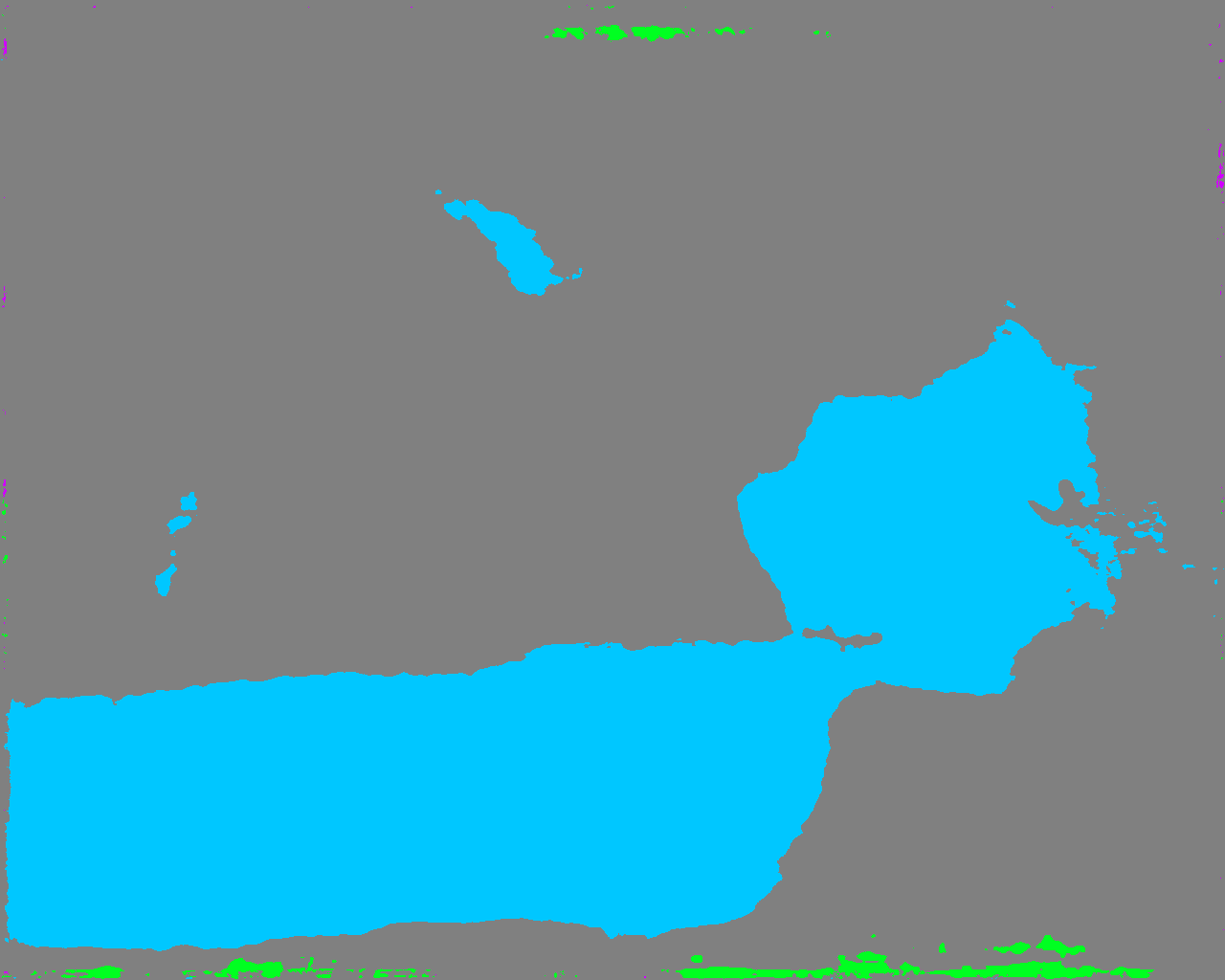}\\
 &
\multicolumn{4}{c}{\includegraphics[width=16.1cm, height=0.5cm]{figures/colorbar_dsad_1.pdf}} \\
 &
\multicolumn{4}{c}{\includegraphics[width=16.1cm, height=0.5cm]{figures/colorbar_dsad_2.pdf}} \\
\end{tabular}
\caption{Qualitative result of second case from DSAD dataset. We show results of the same image from four class partitions (represented as $\CP_1$ to $\CP_4$). For each $\CP$, classes that are held-out are grouped as an extra outlier class for evaluation. We visualise and compare masks generated using different methods at threshold $\tau_{m}$. Baseline results at $\tau_0=0$ are added to represent result without outlier detection.}
\label{fig:segmentation_result_dsad_2}
\end{figure}

\end{document}